\documentclass{bmvc2k}

\title{InstructionCrafter: Generating Consistent and High-Fidelity Visual Instructions}

\addauthor{Shun Okamoto}{mogura.okamoto@cvlab.cs.tsukuba.ac.jp}{1}
\addauthor{Satoshi Iizuka}{iizuka@cs.tsukuba.ac.jp}{1}
\addauthor{Kazuhiro Fukui}{kfukui@cs.tsukuba.ac.jp}{1}

\addinstitution{
 University of Tsukuba\\
 Japan
}

\runninghead{Preprint}{Generating Consistent and High-Fidelity Visual Instructions}

\def\red{\textcolor{red}}

\usepackage{threeparttable}
\usepackage{makecell}
\usepackage{tcolorbox}
\tcbuselibrary{breakable}
\usepackage[export]{adjustbox}
\usepackage{capt-of}  % \captionof
\usepackage{booktabs}
\usepackage{amsmath}
\usepackage{amssymb}
\usepackage{amsfonts}

\begin{document}

\maketitle

\begin{abstract}
Given textual task instructions, generating step-by-step visual instructions as an image sequence requires the simultaneous satisfaction of multiple properties, specifically step faithfulness, cross-image consistency, and per-frame visual quality. Existing text-to-image generation approaches rarely meet all three properties, owing to independent sampling that breaks consistency, finetuning on low-quality video that degrades per-frame quality, and frozen backbones that lack multi-step understanding. In this work, we propose InstructionCrafter, a diffusion-based framework with the key idea of separating the optimization of temporal and instructional alignment from per-frame visual quality via (1) spatial-freeze training and (2) instruction-aware adapters. Built on a pretrained video diffusion backbone, InstructionCrafter freezes the spatial layers that control per-frame detail and updates only temporal and text-conditioning pathways to learn instruction semantics and inter-step relations, which preserves the generative prior for per-frame quality and reduces trainable parameters by about 50 percent compared with full finetuning. We also introduce two lightweight adapters that enhance the model's understanding of instructional context. The Consistent Adapter aggregates textual cues from the entire instruction sequence and from neighboring steps to keep object identity and attributes consistent across frames, and the Context-Aware Temporal Adapter converts cross-attention outputs into biases for temporal self-attention, explicitly propagating inter-frame relations. Extensive experiments on two benchmark datasets demonstrate state-of-the-art overall performance on step faithfulness, cross-image consistency, and per-frame visual quality while significantly reducing noise, blur, and spurious subtitles. Our code and trained models will be publicly available.
\end{abstract}

%-------------------------------------------------------------------------
\section{Introduction}
\label{sec:intro}
\begin{figure}
    \centering
    \includegraphics[width=\linewidth]{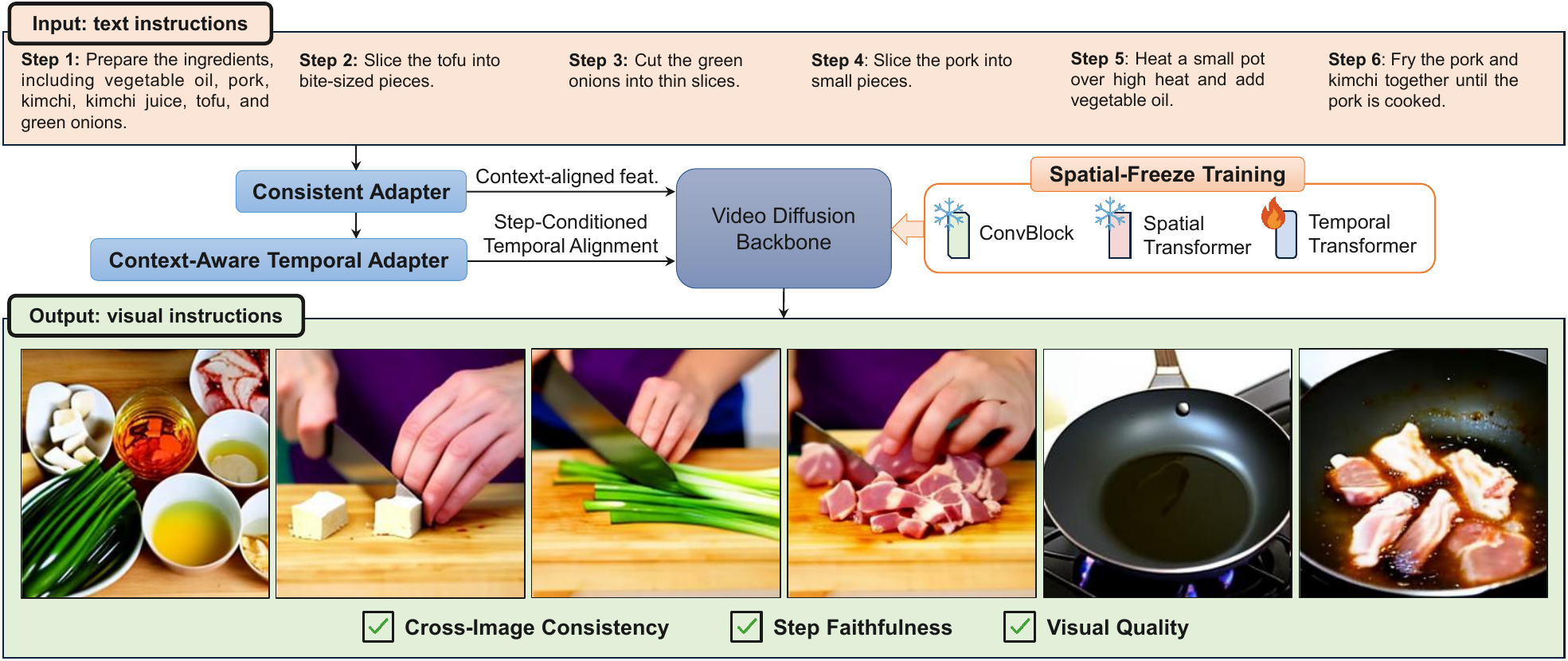}
    \caption{\textbf{Visual instruction generation results.} Given step-by-step textual instructions, our method generates a sequence of high-fidelity, cross-image consistent, and step-faithful visual instructions using spatial-freeze training and two instruction-aware adapters.}
    \label{fig:teaser}
    \vspace{-5mm}
\end{figure}

In many real-world settings, people must execute multi-step procedures from textual instructions without accompanying visuals. Text-only manuals are often difficult to follow, especially when intermediate states and spatial relations matter, leading to errors and wasted materials. In contrast, step-by-step illustrations clearly show how tools and objects should look at each stage and are more accessible to novices and non-native speakers. This motivates generating visual instructions directly from step-by-step text, with the goal of synthesizing a coherent image sequence that accurately depicts each written step, as shown in Fig.~\ref{fig:teaser}.

\begin{figure}
    \centering
    \includegraphics[width=\linewidth]{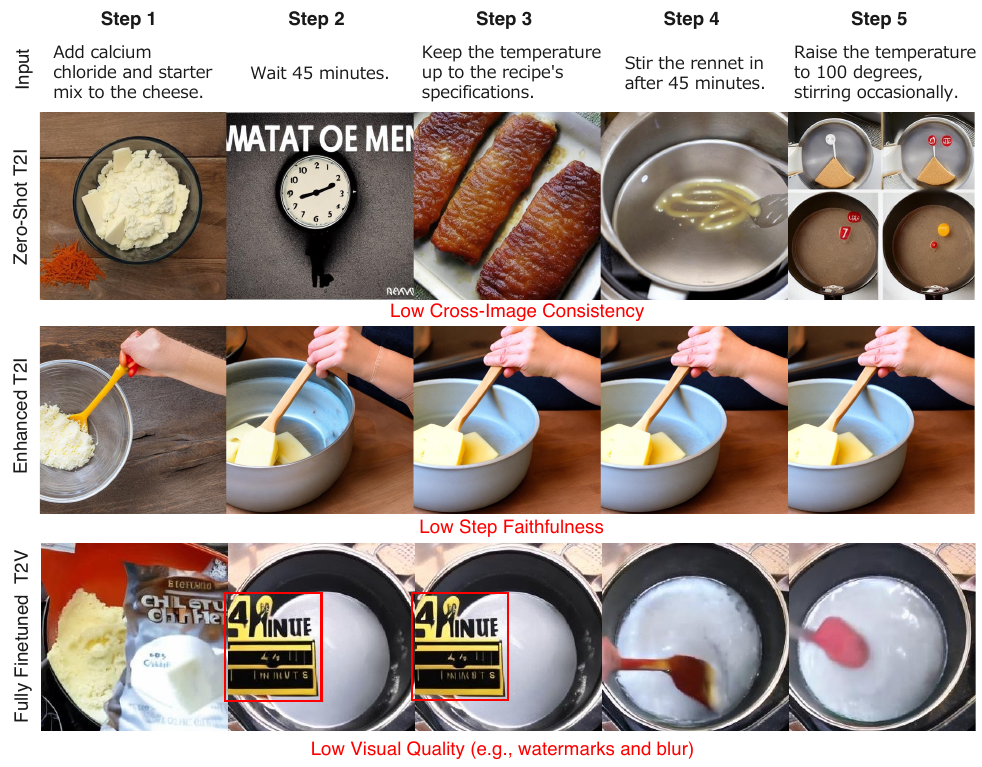}
    \caption{\textbf{Limitations of existing approaches.} Zero-shot Text-to-Image (T2I) diffusion models~\cite{StableDiff} often fail to maintain cross-image consistency. Instruction-oriented T2I extensions~\cite{bordalo-etal-2024-generating}, referred to as \textit{Enhanced T2I}, improve consistency yet lack temporal priors, reducing step faithfulness. Text-to-Video (T2V) models~\cite{chen2024videocrafter2} fully finetuned on visual instruction datasets~\cite{souvcek2025showhowto} produce more faithful and consistent sequences, but per-frame visual quality deteriorates because the training videos are low quality, leading to technical artifacts.
    }
    \label{fig:intro}
    \vspace{-5mm}
\end{figure}

Generating high-fidelity visual instructions is challenging due to the need to simultaneously satisfy step faithfulness and cross-image consistency~\cite{menon2024generating} while maintaining per-frame image quality, as illustrated in Fig.~\ref{fig:intro}. Text-to-image diffusion models~\cite{StableDiff,SDXL,SD3,FLUX} achieve strong per-image faithfulness, yet independent sampling yields cross-image inconsistency~\cite{pan2024synthesizing,shen2025storygpt,liu2024intelligent}. Clear visual instructions require subjects and backgrounds to evolve progressively while maintaining overall scene coherence. Recent instruction-oriented extensions of text-to-image models~\cite{bordalo-etal-2024-generating,menon2024generating} narrow this gap but remain limited because such models lack the temporal priors needed for correct step-specific transitions. Neighboring areas such as story visualization~\cite{pan2024synthesizing,zheng2025contextualstory} and subject-consistent generation~\cite{tewel2024training,zhou2024storydiffusion} also propose various extensions to text-to-image models, but their settings assume simple narratives or temporally invariant consistency, which lead to suboptimal performance when directly applied to visual instructions. In contrast, video diffusion models~\cite{ho2022video,blattmann2023align,bar2024lumiere,ma2024latte,wan2025wan} have strong capabilities in modeling visual dynamics, and Sou{\v{c}}ek et al.~\cite{souvcek2025showhowto} suggest that, when finetuned for visual instruction, they can inherit scene context from an input reference image. Visual instructions can be viewed as discrete frame sequences extracted from instructional videos, making video diffusion models a natural backbone for representing step-wise procedural changes. However, the large-scale datasets~\cite{miech2019howto100m,zhou2018towards,souvcek2025showhowto} required for finetuning are often low quality due to automatic frame extraction from noisy web sources~\cite{miech2019howto100m,zhou2018towards,souvcek2025showhowto}, and full finetuning amplifies artifacts such as noise, blur, and spurious subtitles, which degrades visual clarity. Prior work~\cite{wang2025cookingdiffusioncookingproceduralimage} also reports a trade-off between image quality and instruction consistency in this setting because the model jointly learns visual instruction patterns and dataset-specific artifacts from the same training data. Building large-scale, high-quality instructional video datasets is also costly, making data curation alone an impractical solution, while zero-shot inference with frozen video diffusion backbones lacks multi-step instructional understanding and yields frames that are not faithful to the step texts.

To overcome these issues, we propose InstructionCrafter, an efficient framework that generates consistent and high-fidelity visual instructions via a dedicated spatial-freeze training strategy and two novel instruction-aware adapters. Our approach leverages the real-world knowledge embedded in a pretrained latent video diffusion backbone~\cite{chen2024videocrafter2}. The key idea is to freeze spatial layers, where image quality is primarily encoded, while finetuning only temporal and cross-attention pathways to learn instruction semantics and inter-step relations, such as object identity persistence and step-specific state changes across frames. Freezing spatial layers prevents the model from absorbing dataset-specific spatial artifacts and thus preserves image quality, while reducing the number of trainable parameters from 1.4B to 0.6B compared with full finetuning, and still improves VIE perceptual scores~\cite{ku-etal-2024-viescore} from 0.662 to 0.697. Additionally, to capture the stepwise temporal dynamics characteristic of visual instructions, we introduce two lightweight adapters, which we denote as Consistent Adapter and Context-Aware Temporal Adapter. Consistent Adapter models interactions among step texts by computing global attention over the full instruction sequence and adjacent attention over neighboring steps, which helps maintain entity and attribute persistence across frames. Context-Aware Temporal Adapter converts cross-attention outputs into step-conditioned biases for temporal self-attention, explicitly propagating inter-step relations. These lightweight additions, totaling about 0.1B parameters, empirically improve image quality.

We compare InstructionCrafter with leading methods~\cite{menon2024generating,bordalo-etal-2024-generating,souvcek2025showhowto} on the ShowHowTo~\cite{souvcek2025showhowto} and WikiHow-VGSI~\cite{yang2021visual} datasets. InstructionCrafter achieves the best overall performance across faithfulness, cross-image consistency, and per-frame visual quality, and qualitatively produces more consistent, high-quality instructions with fewer artifacts.

In summary, our contributions are as follows: (1) we present \textit{InstructionCrafter} with \textit{spatial-freeze training}, which preserves per-frame quality while learning temporal and instructional alignment; (2) we introduce two lightweight, instruction-aware adapters, \textit{Consistent Adapter} for cross-frame entity persistence and \textit{Context-Aware Temporal Adapter} that injects step context into temporal self-attention via attention biases; and (3) we conduct an in-depth evaluation and achieve the best overall performance on two benchmarks under faithfulness, consistency, and image quality criteria while reducing technical artifacts.

\section{Related Work}
\label{sec:relatedwork}
%In this section, we first review diffusion models and their extension to video generation. 
%We then discuss neighboring areas on generating coherent sequences. Finally, 
%We then introduce prior work specifically on visual instructions.

\subsection{Diffusion Models}
Diffusion models~\cite{DDPM,DDIM} have rapidly advanced image synthesis by enabling high-quality generation~\cite{FLUX,SDXL,SD3,IPAdapter,TextualInv,ControlNet,DiT}. Latent diffusion~\cite{StableDiff}, also known as Stable Diffusion, denoises in a compressed VAE space~\cite{VAE}, which lowers computation while supporting high resolution. The denoiser is a U-Net~\cite{UNET} built from residual blocks~\cite{ResNet} and attention layers~\cite{AttentionIsAllYouNeed}. Text features are produced by CLIP~\cite{CLIP} and injected through cross-attention at multiple spatial scales. Classifier-free guidance~\cite{CFG} is commonly used to control text conditions.
By extending the image denoiser along the temporal dimension, 2D diffusion models become video diffusion models~\cite{ho2022video,blattmann2023align,wiedemer2025video,wan2025wan,yang2024cogvideox,zheng2024open}. They can synthesize high-quality videos from a single text input. Typical designs separate spatial and temporal processing: temporal-only attention layers or lightweight 3D convolutions are inserted into the U-Net while the spatial backbone learned for images is largely reused~\cite{guo2023animatediff,chen2023videocrafter1,xing2024dynamicrafter,chen2024videocrafter2}. This factorization enables direct transfer of image-generation priors to video, which in turn motivates our use of video-generation prior for visual instruction generation.

\subsection{Generating Visual Instructions}
Prior work~\cite{song2025makeanythingharnessingdiffusiontransformers,krojer2024learning,wang2025cookingdiffusioncookingproceduralimage,souvcek2024genhowto,CookAnything} has taken several directions specific to visual instructions. Bordalo et al.~\cite{bordalo-etal-2024-generating} convert manuals into stepwise captions with a large language model~\cite{GPT,GPT-Training} and then frozen Stable Diffusion generates frames autoregressively while sharing the initial noise. This zero-shot approach preserves image quality but often produces inconsistent sequences that poorly reflect the stepwise instructions. A second line, StackedDiffusion~\cite{menon2024generating}, tiles all frames in the U-Net and generates them jointly. This can generate more coherent sequences but tends to compromise step faithfulness due to limited per-frame text conditioning. A third line, ShowHowTo~\cite{souvcek2025showhowto}, finetunes an entire video diffusion backbone~\cite{xing2024dynamicrafter} with per-frame text conditioning and first-frame guidance. This achieves high instruction consistency but degrades image quality due to the low visual quality of training data~\cite{miech2019howto100m}. It also requires a reference image and high computational cost. In contrast, our approach targets text-only generation, leverages video priors while preserving spatial layers to maintain image quality and reduce computational cost while outperforming existing methods. Additional discussion including neighboring areas is provided in the supplementary material.

\section{Method}

\subsection{Architecture}
\begin{figure*}
    \centering
    \includegraphics[width=\linewidth]{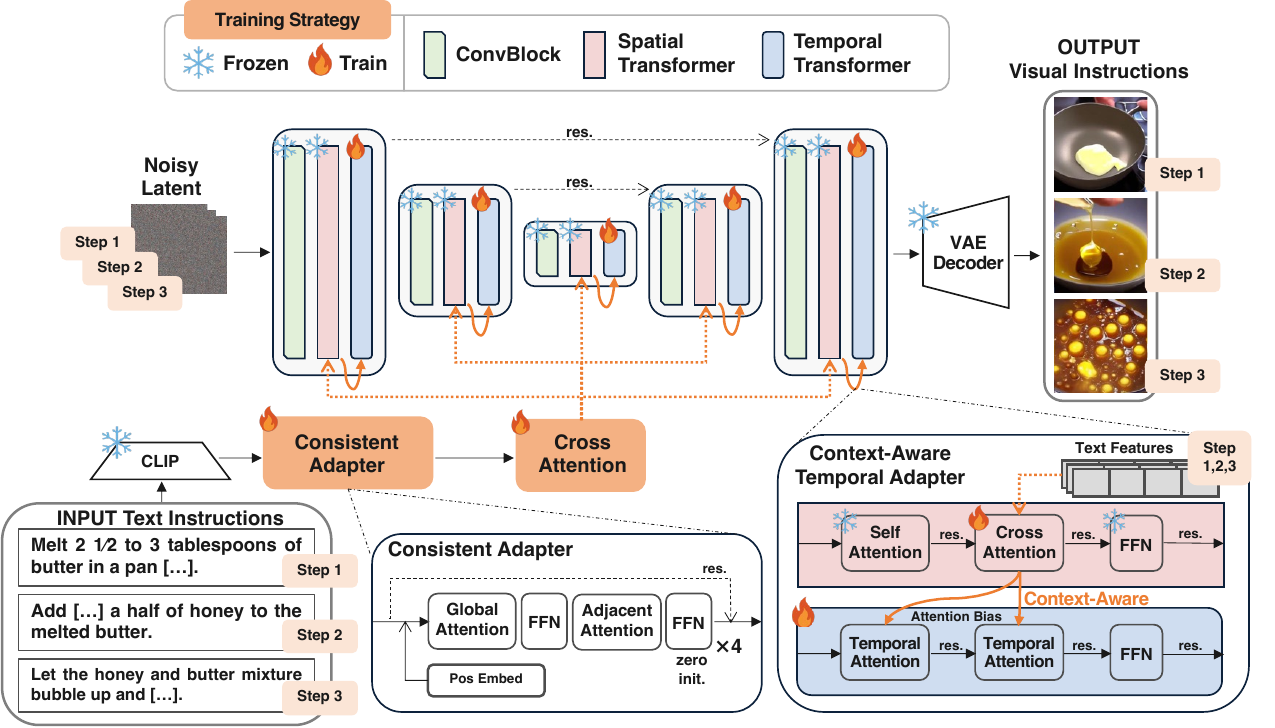}
    \vspace{-7mm}
    \caption{
        \textbf{Overview of InstructionCrafter.} We build upon a latent video diffusion backbone~\cite{chen2024videocrafter2} and introduce two novel lightweight adapters: Consistent Adapter and Context-Aware Temporal Adapter. Consistent Adapter captures inter-step relations among step texts via global and adjacent attention mechanisms. Context-Aware Temporal Adapter incorporates step text context into temporal attention via attention biases derived from cross-attention outputs. Our training strategy finetunes only temporal and text-related components while freezing spatial components to preserve pretrained spatial priors. 
        %Illustration inspired by~\cite{chen2023videocrafter1,zheng2025contextualstory,guo2023animatediff}.
    }
    \label{fig:overview}
    \vspace{-2mm}
\end{figure*}

An overview of our proposed InstructionCrafter is shown in Fig.~\ref{fig:overview}.
InstructionCrafter extends VideoCrafter2~\cite{chen2024videocrafter2}, a strong U-Net-based latent video diffusion model. We adopt this backbone because it preserves frame-level latent representations, which suit visual instruction generation. Recent Diffusion Transformer-based backbones~\cite{zheng2024open,yang2024cogvideox} often leverage 3D VAEs to compress the temporal dimension for efficient generation, making direct per-step conditioning difficult. VideoCrafter2 has convolutional blocks with spatial convolutions and additional temporal convolutions, spatial attention layers including cross-attention to text embeddings, and temporal attention layers. 
In our InstructionCrafter, each frame is conditioned on the corresponding step text via cross-attention that attends to text embeddings produced by the CLIP text encoder~\cite{CLIP}, which is different from the original VideoCrafter2 that conditions on only a single text prompt for the entire sequence. We also introduce two novel lightweight modules, Consistent Adapter and Context-Aware Temporal Adapter, to enhance instructional-context-aware understanding, smoothly integrated into the backbone with zero-initialization.\\

\noindent\textbf{Consistent Adapter.}
Visual instruction generation requires understanding contextual information across multiple step texts to ensure consistency across frames, as similarly discussed in~\cite{zheng2025contextualstory}. Instructional context mainly involves global relations among all steps and local transitions between adjacent steps. We propose Consistent Adapter, a lightweight transformer-based module that captures inter-step dependencies using global and adjacent attention mechanisms.

Consistent Adapter is a stack of $M=4$ transformer layers that learns global and adjacent relations among step texts, as shown in Fig.~\ref{fig:overview}. This adapter is inserted between the frozen CLIP text encoder and the cross-attention layers of the denoising backbone. Each layer consists of global and adjacent self-attention, followed by feed-forward networks (FFN). Global attention performs self-attention over all \(N \times L\) tokens at once, where \(N\) is the number of steps, \(L=77\) is the number of CLIP tokens per step. It captures task-level global dependencies across the entire sequence. Adjacent attention applies the same operation under a frame-wise mask that only allows tokens to attend within the current step and the previous and next steps. This captures local transitions between consecutive steps. We note that causal attention that only allows attending to previous steps leads to suboptimal performance, likely because understanding the next step can also help disambiguate the current step. We use a sinusoidal positional encoding~\cite{AttentionIsAllYouNeed} derived from the frame index \(f \in \{0,\dots,N-1\}\) to support variable-length inputs, project it with an MLP layer, and add it to the input token embeddings~\cite{ViT,DiT}. Each attention and FFN sublayer is wrapped with residual connections~\cite{ResNet} and layer normalization~\cite{ba2016layer}. The final layer is zero-initialized~\cite{ControlNet} so the module initially behaves as an identity mapping thanks to the residual connection from inputs.\\

\noindent\textbf{Context-Aware Temporal Adapter.}
In visual instructions, content can change sharply across steps. Since vanilla temporal attention is blind to the instructional context, it may fail to capture inter-step relations. Here, we use the term inter-step relations to refer to how objects, attributes, and scene layout evolve coherently across steps, including identity persistence and step-specific state changes. We therefore propose Context-Aware Temporal Adapter, which incorporates text-contextual information into temporal attention to better capture inter-step relations derived from cross-attention outputs. Note that temporal transformer layers have two temporal attention sublayers, and we insert this adapter into both of them, as shown in Fig.~\ref{fig:overview}.

Let the cross-attention output from each spatial transformer layer be \(C\in\mathbb{R}^{N\times S\times D}\), where \(S=H\times W\) is the number of spatial tokens and \(D\) is the feature dimension. We focus on the residual update in each cross-attention layer, \(X_{\mathrm{out}}=X_{\mathrm{in}}+C\), where \(C\) conveys text-contextual information. We propose a simple yet effective way to convert this \(C\) into temporal correlations among frames. We reshape \(C\) to \((S,N,D)\), extract a spatial slice \(C_s\in\mathbb{R}^{N\times D}\), apply layer normalization, and form a Gram matrix \(G_s=C_s C_s^\top\in\mathbb{R}^{N\times N}\). Then we create a temporal bias \(B_s=\beta\,G_s\) with per-head learnable scalar \(\beta\) initialized to zero. This bias \(B_s\) encodes inter-frame relations implied by the per-frame texts, encouraging frames with similar semantics to attend to each other. Notably, this context-aware temporal adapter introduces only a small number of additional parameters and negligible computational overhead. For each head, temporal attention adapted with this context-aware bias is computed as:
\begin{align}
\mathrm{ContextAwareTmpAttn}(Q,K,V,B_s) = \mathrm{softmax}\left(\frac{QK^\top}{\sqrt{d}} + B_s\right) V,
\end{align}
where \(Q,K,V\) are the query, key, and value matrices at spatial position \(s\), and \(d\) is the head dimension. With \(\beta=0\) at initialization, the module is identical to vanilla temporal attention at the start of training.

\subsection{Spatial-Freeze Training}
\label{sec:training_strategy}
Although full finetuning can yield consistent and step-faithful results, this approach significantly degrades the visual quality due to dataset-specific artifacts such as noise, blur, and spurious subtitles. One solution is to finetune the model with only high-quality data after pretraining as commonly discussed in video generation~\cite{chen2024videocrafter2,zhang2025show,guo2023animatediff}, but collecting such visual instructions is difficult and costly. Prior work~\cite{IPAdapter,meral2024clora,chen2024videocrafter2} has shown that if we train only attention layers in the denoising backbone while keeping other parts frozen, the model can adapt to new conditions while preserving pretrained knowledge. Motivated by this, we propose a partial finetuning strategy that preserves pretrained high-quality spatial priors while finetuning the components that govern temporal evolution and text grounding. Specifically, we freeze all spatial convolutional blocks and spatial self-attention layers, and update only the temporal-attention layers that learn frame-to-frame changes and the cross-attention layers that align frames with the step texts. Note that we also train the temporal convolutions inside conv blocks. This strategy reduces the number of trainable parameters by about 50 percent compared with full finetuning (1.4B vs 0.7B trainable parameters).

\begin{figure}
    \centering
    \includegraphics[width=\linewidth]{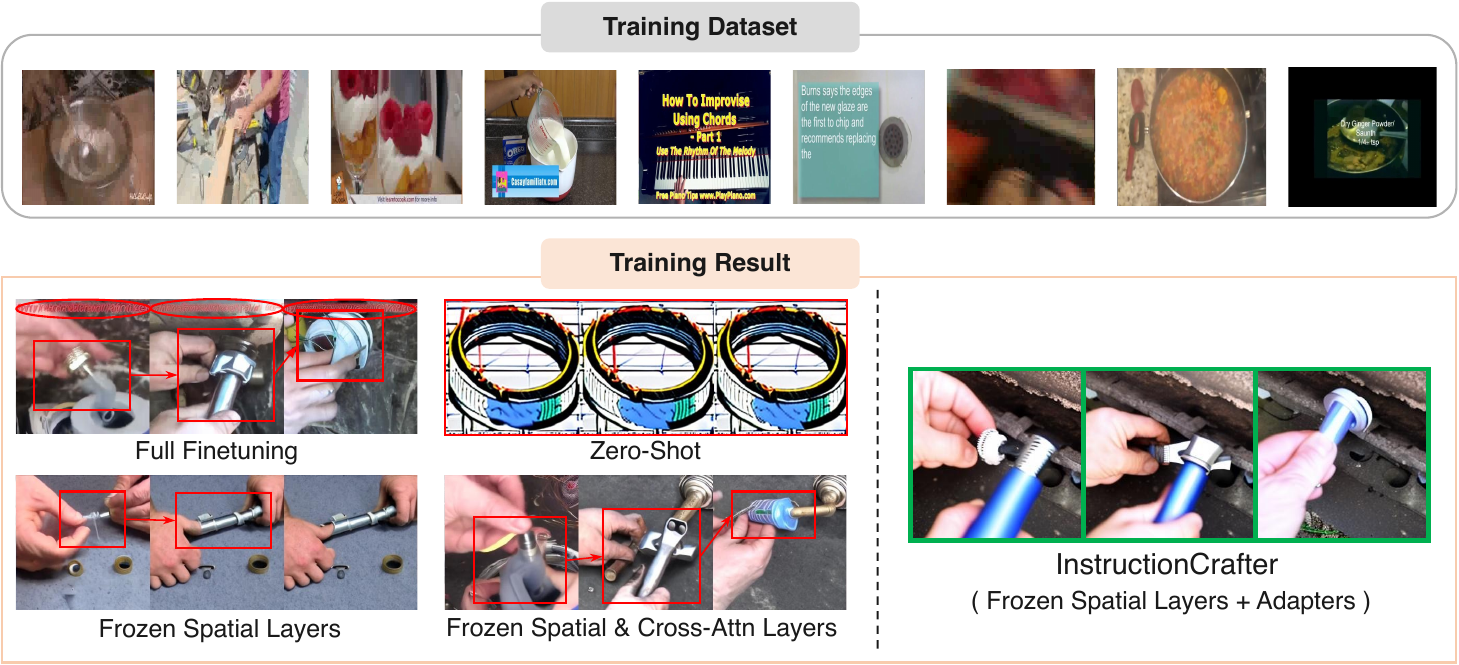}
    \caption{\textbf{Comparison of different training strategies.} \red{Red} boxes highlight artifacts and inconsistencies. \textit{Full Finetuning} learns technical artifacts from low-quality training data. \textit{Zero-Shot} generation, which directly uses the pretrained video diffusion model with per-frame text conditioning, produces collapsed frames under covariate shift. \textit{Frozen Spatial Layers} significantly improves image quality but still learns spurious artifacts. \textit{Frozen Spatial \& Cross-Attn Layers} also freezes cross-attention to preserve the pretrained prior, but this degrades both instruction alignment and visual quality. In contrast, our full model \textit{InstructionCrafter} generates consistent, step-faithful, high-fidelity images.}
    \label{fig:strategy}
    \vspace{-5mm}
\end{figure}

\noindent\textbf{How it works.} 
Freezing spatial layers while finetuning temporal and cross-attention layers and adding our two adapters yields the highest visual quality, as shown in Fig.~\ref{fig:strategy}.
Intuitively, the spatial convolutions and spatial self-attention mainly control local textures and high-frequency details. When they are fully finetuned on web videos that often contain subtitles, watermarks, or sensor noise, these artifacts can be memorized as part of the texture prior and then reproduced even for clean test instructions. Freezing the spatial layers keeps the pretrained high-quality texture prior intact and forces the model to explain dataset biases through higher-level temporal or textual pathways rather than copying artifacts into every frame.
Conversely, the cross-attention layers are responsible for injecting step-specific semantics into visual features. If they are also frozen, the model cannot adapt its text grounding to multi-step instructions and tends to reuse similar appearances or rely on spurious correlations, which harms both alignment and perceived quality. Allowing cross-attention and our two adapters to update, therefore, provides just enough flexibility to reweight features and refine visual cues based on instructional context, while the frozen spatial backbone protects against learning dataset-specific artifacts.

\section{Experiments}

\subsection{Evaluation Details}

\noindent\textbf{Datasets.}
We train and evaluate InstructionCrafter on the ShowHowTo dataset~\cite{souvcek2025showhowto}, which contains 578k unique visual instructions constructed by automatically extracting keyframes and generating stepwise manuals from HowTo100M~\cite{miech2019howto100m}. This covers diverse tasks (e.g., cooking, cleaning, machinery and electronics repair, etc.), viewpoints, and sequence lengths, but some technical artifacts exist due to the automatic extraction process. The test split comprises 3{,}956 sequences with lengths from 2 to 8 frames. To the best of our knowledge, ShowHowTo is the largest dataset for visual instruction generation.

Following prior work~\cite{souvcek2025showhowto,bordalo-etal-2024-generating}, we further conduct zero-shot evaluation on WikiHow-VGSI~\cite{yang2021visual}. This dataset is derived from web articles and provides step texts with or without ground-truth images. Since WikiHow-VGSI contains abstract instructions due to its user-generated content, it is not ideal for training. We use the subset consisting of the \textit{Cooking} and \textit{DIY} domains which contain relatively concrete instructions and sequences of length 2–8, yielding 2{,}415 visual instructions for testing, as in~\cite{bordalo-etal-2024-generating,menon2024generating}.

%\subsection{Implementation Details}
\noindent\textbf{Implementation Details.}
We initialize InstructionCrafter from the pretrained checkpoint of VideoCrafter2~\cite{chen2024videocrafter2}. For training, we use a bucket scheme~\cite{zheng2024open,souvcek2025showhowto} where for each mini-batch we sample a target length in \([2,8]\) and crop longer sequences from a random starting step. We apply classifier-free guidance~\cite{CFG} with a 10\% text-drop probability. Both training and inference use a resolution of \(256\times256\) pixels, following the dataset setting and prior work~\cite{souvcek2025showhowto}. We adopt the standard DDPM~\cite{DDPM} noise-prediction $\ell_2$ loss and optimize with AdamW~\cite{Adam,weightdecay} at a learning rate of \(2\times10^{-5}\). We train InstructionCrafter for about 100k steps on 8$\times$NVIDIA H100 GPUs with a batch size of 32, which takes roughly 24 hours, while full finetuning takes about 30 hours. At inference, we use the default guidance scale of 12 from VideoCrafter2 and sample 50 DDIM~\cite{DDIM} steps.

%\subsection{Compared Methods}
\noindent\textbf{Compared Methods.}
We compare InstructionCrafter with state-of-the-art methods for visual instruction generation:
\textit{StackedDiffusion}~\cite{menon2024generating}, an extension of Stable Diffusion which generates all frames jointly by tiling them inside U-Net. \textit{Bordalo et al.}~\cite{bordalo-etal-2024-generating}, a pipeline that converts manuals into stepwise captions with a large language model and then performs zero-shot generation with a frozen Stable Diffusion backbone by sharing the initial noise state for autoregressive sampling. Following prior work~\cite{souvcek2025showhowto}, we use pretrained versions of these models provided by their official implementations. Further, we include \textit{Baseline}, a fully finetuned VideoCrafter2~\cite{chen2024videocrafter2} by conditioning on per-frame step texts. This is similar to Sou{\v{c}}ek et al.~\cite{souvcek2025showhowto}, but they finetune an image-to-video model~\cite{xing2024dynamicrafter}, using the first frame as a visual reference, whereas we tackle a more challenging setting without visual references.

For a more detailed comparison, we also include the following two references: \textit{Ground Truth}, the actual ground-truth images from ShowHowTo. WikiHow-VGSI lacks ground truth for some steps, so we only report results on ShowHowTo. \textit{Stable Diffusion}, finetuned 2D text-to-image Stable Diffusion v1.5 on ShowHowTo, generating each step independently without temporal modeling. 
%For reference, we also implement StoryDiffusion~\cite{zhou2024storydiffusion} and ContextualStory~\cite{zheng2025contextualstory}, although these methods are not designed for visual instructions. StoryDiffusion leverages SDXL~\cite{SDXL} for zero-shot generation by sharing key/value across frames via self-attention, whereas ContextualStory finetunes SD2.1~\cite{StableDiff} with dedicated additional temporal modules. 
All methods are evaluated with the same random seed fixed to 42 for fair comparison.

%\subsection{Evaluation Metrics}
\noindent\textbf{Evaluation Metrics.}
Desirable visual instructions require the simultaneous satisfaction of step faithfulness, cross-image consistency, and per-frame visual quality. Quantitatively assessing these properties is nontrivial, yet prior work~\cite{souvcek2025showhowto,bordalo-etal-2024-generating,menon2024generating} has proposed several metrics. \textit{Step Faithfulness (SF)} measures whether each generated frame best matches the textual description of its own step when compared with the other step descriptions in the same sequence.
We compute CLIP similarities~\cite{CLIP} between each frame and every step description, and count a frame as correct only if its own step attains the highest similarity.
\textit{Cross-Image Consistency (CIC)} quantifies visual coherence within a sequence using DINOv3 features~\cite{simeoni2025dinov3}. Let \(d_{i,j}\) denote the Euclidean distance between L2-normalized features from frames \(i\) and \(j\). Pairwise closeness is defined as \(1-d_{i,j}/2\), and CIC is the mean of this closeness over all frame pairs in the sequence. 
While SF and CIC are complementary, a high score on either metric alone does not imply natural visual instructions. For example, SF can be high even if the appearance of the generated subject differs completely from earlier frames, as long as it matches the step description. Conversely, CIC can be maximized by producing the same image at every step, which ignores step-specific changes.
%Although these SF and CIC are complementary, optimizing one does not guarantee improvement in the other. 
Therefore, we adopt a unified metric \textit{SF–CIC F-Score (SC-F)}, which is defined as the weighted F-Score of SF and CIC:
\[
\mathrm{SC\mbox{-}F}=\frac{(1+\alpha^{2})\,\mathrm{SF}\,\cdot\,\mathrm{CIC}}{\alpha^{2}\,\mathrm{SF}+\mathrm{CIC}},
\]
where we set \(\alpha=2\) to place greater weight on consistency to balance the empirical ranges of SF
(typically 0.20-0.60) and CIC (typically 0.40-0.55) observed in our experiments.

However, SF and CIC do not fully capture all aspects of visual instruction quality. We therefore perform an additional evaluation using Qwen3-VL~\cite{qwen3technicalreport,Qwen3-VL}, an open-source vision-language model (VLM) with strong multimodal understanding capabilities. Following the standard protocol of VIEScore~\cite{ku-etal-2024-viescore}, we query the VLM with all frames and corresponding step descriptions, decomposing the evaluation into \textit{Semantic} and \textit{Perceptual} components. For the Semantic component, the VLM judges alignment between images and instructions and returns Step Faithfulness and Cross-Image Consistency scores in $[0,10]$; we normalize each to $[0,1]$ and take the minimum as the semantic score. For the perceptual component, the VLM rates naturalness and absence of technical artifacts; we likewise normalize to $[0,1]$ and take the minimum as the perceptual score. Note that VIEScore is designed for synthesized images, so we do not evaluate ground-truth with it. All metrics are computed within each sequence and then averaged over the test split. Additional details on evaluation protocols and metric implementations are provided in the supplementary material.

\subsection{Quantitative Results}
We present quantitative results in Tab.~\ref{tab:main_combined}. These results show that InstructionCrafter achieves the best overall performance on both datasets, thus overcoming the trade-offs faced by existing methods. Stable Diffusion and StackedDiffusion obtain low SC-F and Semantic scores, indicating difficulty in simultaneously satisfying step faithfulness and cross-image consistency. Bordalo et al.'s method attains the highest Perceptual score because a frozen backbone in a zero-shot setting preserves pretrained visual priors. However, without instructional and temporal fine-tuning, it struggles with stepwise transformations, which is reflected in lower SC-F and Semantic scores. By contrast, the Baseline achieves competitive SC-F and Semantic scores, but its Perceptual score is constrained by the low visual quality of training videos. Similar trade-offs between image quality and instruction consistency have been reported in prior work~\cite{wang2025cookingdiffusioncookingproceduralimage}. Despite these challenges, InstructionCrafter improves both SC-F and Perceptual scores over the Baseline, indicating that spatial-freeze training and instruction-aware adapters jointly produce better-balanced visual instructions with improved consistency and reduced artifacts. More detailed results, including metric breakdowns, are provided in the supplementary material.

\begin{table}[t]
    \centering
    \small
    \setlength{\tabcolsep}{6.5pt}
    \caption{\textbf{Quantitative results on ShowHowTo~\cite{souvcek2025showhowto} and WikiHow-VGSI~\cite{yang2021visual}.} Best values are shown in \red{red} and second-best values in \textcolor{blue}{blue}. InstructionCrafter achieves the best overall performance on both datasets. We note that Bordalo et al.~\cite{bordalo-etal-2024-generating} achieves the highest Perceptual scores due to its zero-shot nature, but this comes at the cost of visual-instruction consistency.
    }
    \label{tab:main_combined}
    \begin{threeparttable}
        \resizebox{\linewidth}{!}{%
        \begin{tabular}{lcccccccc}
            \toprule
            & \multicolumn{3}{c}{ShowHowTo~\cite{souvcek2025showhowto}} & \multicolumn{3}{c}{WikiHow-VGSI~\cite{yang2021visual}} \\
            \cmidrule(lr){2-4}\cmidrule(lr){5-7}
            Method & SC-F$\uparrow$ &  Semantic$\uparrow$ & Perceptual$\uparrow$
                & SC-F$\uparrow$ &  Semantic$\uparrow$ & Perceptual$\uparrow$ \\
            \midrule
            Ground Truth & 0.441 & -- & -- & -- & -- & -- \\
            \midrule
            Stable Diffusion~\cite{StableDiff} & 0.426 & 0.505 & 0.612 & 0.428 & 0.507 & 0.597 \\
            StackedDiffusion~\cite{menon2024generating} & 0.313 & 0.418 & 0.705 & 0.424 & 0.472 & 0.637 \\
            % Bordalo et al.~\cite{bordalo-etal-2024-generating}${}^\dagger$ & 0.437 & 0.407 & \red{0.805} & 0.452 & 0.467 & \red{0.837} \\
            Bordalo et al.~\cite{bordalo-etal-2024-generating} & 0.437 & 0.407 & \red{0.805} & 0.452 & 0.467 & \red{0.837} \\
            Baseline & \textcolor{blue}{0.455} & \red{0.555} & 0.684 & \textcolor{blue}{0.513} & \textcolor{blue}{0.568} & 0.662 \\
            \textbf{InstructionCrafter} & \red{0.460} & \textcolor{blue}{0.537} & \textcolor{blue}{0.732} & \red{0.520} & \red{0.569} & \textcolor{blue}{0.742} \\
            \bottomrule
        \end{tabular}
        }
        %\begin{tablenotes}
           % \item[] ${}^\dagger$ Zero-Shot method with a frozen Stable Diffusion backbone.
        %\end{tablenotes}
    \end{threeparttable}
\end{table}

\begin{table*}[t]
    \centering
    \small
    \setlength{\tabcolsep}{5pt}
    \caption{\textbf{Comparison with subject-consistent generation and story visualization methods.} Best values are shown in red and second-best values in blue.}
    \label{tab:story}
    \begin{tabular}{lcccccccc}
        \toprule
        & \multicolumn{3}{c}{ShowHowTo~\cite{souvcek2025showhowto}} & \multicolumn{3}{c}{WikiHow-VGSI~\cite{yang2021visual}} \\
        \cmidrule(lr){2-4}\cmidrule(lr){5-7}
        Method & SC-F$\uparrow$ &  Semantic$\uparrow$ & Perceptual$\uparrow$ & SC-F$\uparrow$ &  Semantic$\uparrow$ & Perceptual$\uparrow$ \\
        \midrule
        StoryDiffusion~\cite{zhou2024storydiffusion} & \textcolor{blue}{0.444} & 0.422 & \red{0.866} & \textcolor{blue}{0.481} & 0.479 & \red{0.862} \\
        ContextualStory~\cite{zheng2025contextualstory}& 0.434 & \textcolor{blue}{0.527} & 0.685 & 0.470 & \textcolor{blue}{0.539} & 0.653 \\
        \textbf{InstructionCrafter} & \red{0.460} & \red{0.537} & \textcolor{blue}{0.732} & \red{0.520} & \red{0.569} & \textcolor{blue}{0.742} \\
        \bottomrule
    \end{tabular}
\end{table*}

\begin{table}[t]
\centering
\caption{\textbf{Comparison with a video diffusion transformer.} Open-Sora++ denotes our modified version with several efforts to avoid collapse. We only report VLM-based VIEScore components because Open-Sora++ fails to generate meaningful frames for computing SC-F.}
\label{tab:opensora}
\begin{tabular}{lcccc}
\toprule
& \multicolumn{2}{c}{ShowHowTo~\cite{souvcek2025showhowto}} & \multicolumn{2}{c}{WikiHow-VGSI~\cite{yang2021visual}} \\
\cmidrule(lr){2-3}\cmidrule(lr){4-5}
Method & Semantic$\uparrow$ & Perceptual$\uparrow$ & Semantic$\uparrow$ & Perceptual$\uparrow$ \\
\midrule
Open-Sora~\cite{zheng2024open}++ & 0.279 & 0.610 & 0.369 & 0.704 \\
% InstructionCrafter & \textbf{0.537} & \textbf{0.732} & \textbf{0.569} & \textbf{0.742} \\
InstructionCrafter & \red{0.537} & \red{0.732} & \red{0.569} & \red{0.742} \\
\bottomrule
\end{tabular}
\vspace{-2mm}
\end{table}

\begin{table}[t]
    \centering
    \small
    \setlength{\tabcolsep}{6pt}
    \caption{\textbf{Ablation study on training strategy and proposed adapters.} Best values are shown in \red{red} and second-best values in \textcolor{blue}{blue}. Spatial-Freeze indicates spatial-freeze training. Consistent and Context refer to the proposed Consistent Adapter and Context-Aware Temporal Adapter, respectively. T\_Params denotes the number of trainable parameters.}
    \label{tab:ablation_wikihow}
    \resizebox{\linewidth}{!}{%
    \begin{tabular}{rccccc}
        \toprule
        %& \multicolumn{3}{c}{WikiHow-VGSI~\cite{yang2021visual}} & \\
        %\cmidrule(lr){2-4}
        Training Setting & SC-F$\uparrow$ & Semantic$\uparrow$ & Perceptual$\uparrow$ & T\_Params \\
        \midrule
        Baseline (Full Finetuning) & \textcolor{blue}{0.513} & 0.568 & 0.662 & 1.4B \\
        + Consistent + Context & 0.510 & 0.563 & 0.689 & 1.5B \\
        \midrule
        + Spatial-Freeze & 0.510 & \red{0.575} & 0.697 & 0.6B \\
        + Spatial-Freeze + Consistent & 0.500 & \textcolor{blue}{0.569} & \textcolor{blue}{0.710} & 0.7B \\
        + Spatial-Freeze + Context & \textcolor{blue}{0.513} & \textcolor{blue}{0.569} & 0.680 & 0.6B \\
        \textbf{+ Spatial-Freeze + Consistent + Context (Ours)} & \red{0.520} & \textcolor{blue}{0.569} & \red{0.742} & 0.7B \\
        \bottomrule
    \end{tabular}
    }
    \vspace{-5mm}
\end{table}

\begin{table*}[t]
    \centering
    \small
    \setlength{\tabcolsep}{6pt}
    \caption{\textbf{Ablation study on the second attention method in Consistent Adapter.} T\_Params is the number of trainable parameters.}
    \label{tab:ablation_consistent_adapter}
    \begin{tabular}{rccccc}
        \toprule
        %& \multicolumn{3}{c}{WikiHow-VGSI~\cite{yang2021visual}} & \\
        %\cmidrule(lr){2-4}
        Second Attention Method & SC-F$\uparrow$ & Semantic$\uparrow$ & Perceptual$\uparrow$ & T\_Params \\
        \midrule
        Causal Attention & 0.498 &	0.568 &	0.720 & 0.7B \\
        Temporal Attention~\cite{zheng2025contextualstory} & 0.504 & 0.568 & 0.726 & 0.7B \\
        \textbf{Adjacent Attention (Ours)} & \red{0.520} & \red{0.569} & \red{0.742} & 0.7B \\
        \bottomrule
    \end{tabular}
    \vspace{-2mm}
\end{table*}

\begin{table*}[t]
    \centering
    \small
    \setlength{\tabcolsep}{6pt}
    \caption{\textbf{Ablation study on cross-attention finetuning.} CrossAttn-Freeze denotes freezing cross-attention layers during finetuning while CrossAttn-Finetune denotes our proposed setting that finetunes cross-attention layers.}
    \label{tab:ablation_crossattention_finetuning}
    \begin{tabular}{rccccc}
        \toprule
        Training Setting & SC-F$\uparrow$ & Semantic$\uparrow$ & Perceptual$\uparrow$ & T\_Params \\
        \midrule
        CrossAttn-Freeze  & 0.487 & 0.541 & 0.666 & 0.67B \\
        \textbf{CrossAttn-Finetune (Ours)} & \red{0.520} & \red{0.569} & \red{0.742} & 0.72B \\
        \bottomrule
    \end{tabular}
    \vspace{-5mm}
\end{table*}

%\subsection{Ablation Study}
\subsection{Further Analysis}

\noindent\textbf{Comparison with Neighboring Areas.} We further compare InstructionCrafter with StoryDiffusion~\cite{zhou2024storydiffusion} and ContextualStory~\cite{zheng2025contextualstory} for reference. These methods are not designed for visual instructions, but they represent strong baselines for neighboring areas of subject-consistent generation and story visualization. StoryDiffusion leverages SDXL~\cite{SDXL} for zero-shot generation by sharing key/value across frames via self-attention, whereas ContextualStory finetunes SD2.1~\cite{StableDiff} with dedicated additional temporal modules. As shown in Tab.~\ref{tab:story}, InstructionCrafter outperforms both methods on SC-F and Semantic scores, while achieving better Perceptual scores than fully finetuned ContextualStory. Importantly, StoryDiffusion achieves the highest Perceptual scores due to its zero-shot nature, but it fails to maintain fine-grained stepwise transformations, which is reflected in its lower SC-F and Semantic scores. By contrast, although ContextualStory has a similar architecture design and parameter count to InstructionCrafter and is finetuned on the same data, it underperforms and struggles to maintain consistency. This suggests that additional temporal modules and task-specific training alone are insufficient without temporal priors in video diffusion backbones. We provide qualitative comparisons in the supplementary material.

\noindent\textbf{Comparison with Video Diffusion Transformer.} Recent video generation models have increasingly adopted Diffusion Transformer~\cite{DiT} backbones for better scalability and generation quality, while most of them also rely on 3D VAEs that compress not only spatial but also temporal information for efficient video generation. Although this design is suitable for ordinary videos with temporally continuous frames, it is less suitable for visual instruction generation, where each frame corresponds to a distinct procedural step with a different text condition. To examine this issue, we compare with Open-Sora~\cite{zheng2024open}, a recent Diffusion Transformer-based video generation model with a 3D VAE that applies 1/4 temporal compression. We finetune Open-Sora with the same training data as InstructionCrafter by concatenating per-step instruction embeddings and providing them as global text conditions to the model, since different step texts cannot be directly assigned to individual latent frames. However, this naive adaptation often produces collapsed sequences due to the covariate shift. To avoid such collapse, we apply several modifications, including frame-wise positional embeddings and adjusting several hyperparameters, and we call this modified version Open-Sora++. As shown in Tab.~\ref{tab:opensora}, InstructionCrafter substantially outperforms Open-Sora++ in Semantic scores on both datasets while also achieving higher Perceptual scores. This is because Open-Sora++ still generates invalid visual instructions that are almost the same across steps, suggesting that the temporal compression in 3D VAEs may hinder the model's ability to capture step-specific semantics and temporal relations in visual instructions. To the best of our knowledge, VideoCrafter2 is a state-of-the-art video diffusion model that does not employ temporal compression, making it suitable for our task. We provide qualitative comparisons and additional backbone comparisons with AnimateDiff~\cite{guo2023animatediff} in the supplementary material.

\noindent\textbf{Ablation Study.} 
%\noindent\textbf{Effect of Training Strategy and Adapters.} 
We conduct an ablation study to validate the contributions of our training strategy and adapters. For fair comparison, all variants are evaluated on WikiHow-VGSI to avoid tuning design choices to ShowHowTo-specific results. The table reports SC-F, Semantic, Perceptual, and the number of trainable parameters. As shown in Tab.~\ref{tab:ablation_wikihow}, InstructionCrafter, which combines spatial-freeze training with both adapters, achieves the best overall performance. Compared with the fully finetuned Baseline, freezing spatial layers consistently improves Perceptual scores, indicating better preservation of spatial features and fewer artifacts. Adding both adapters further improves visual quality with or without spatial freezing, and using both adapters together with spatial-freeze training yields the best overall performance by injecting instruction context while preserving spatial detail. This suggests that the two adapters work best in combination, as textual inter-step context must be both encoded in the conditioning features and propagated through temporal attention. Notably, the number of trainable parameters is reduced from 1.4B to 0.7B while still outperforming the fully fine-tuned Baseline. 

Additionally, to verify that the proposed design choices are task-motivated rather than incidental engineering decisions, we further analyze the attention pattern in the Consistent Adapter and the role of cross-attention finetuning. For the Consistent Adapter, our default design uses adjacent attention as the second attention method, where each step text attends to the current step and its neighboring steps. We replace this with causal attention, where each step text attends only to the current and previous steps, and temporal attention~\cite{zheng2025contextualstory}, where each step text attends to tokens at the same positions across different steps, while keeping all other settings the same. As shown in Tab.~\ref{tab:ablation_consistent_adapter}, adjacent attention achieves the best performance across all metrics. This result suggests that attending to neighboring steps helps the model capture local temporal relations and smooth transitions between steps. We also examine whether cross-attention layers should be frozen under the spatial-freeze strategy. Although freezing spatial layers helps preserve the pretrained image prior, freezing cross-attention layers would also restrict how step-specific text information is injected into visual features. As shown in Tab.~\ref{tab:ablation_crossattention_finetuning}, finetuning cross-attention layers consistently outperforms freezing them across all metrics. This indicates that temporal layers alone are not sufficient for adapting the model to visual instructions. Cross-attention layers should also be finetuned to inject step-specific textual semantics into the temporal generation process.

\subsection{Qualitative Results}
\begin{figure*}
    \centering
    \includegraphics[width=\linewidth]{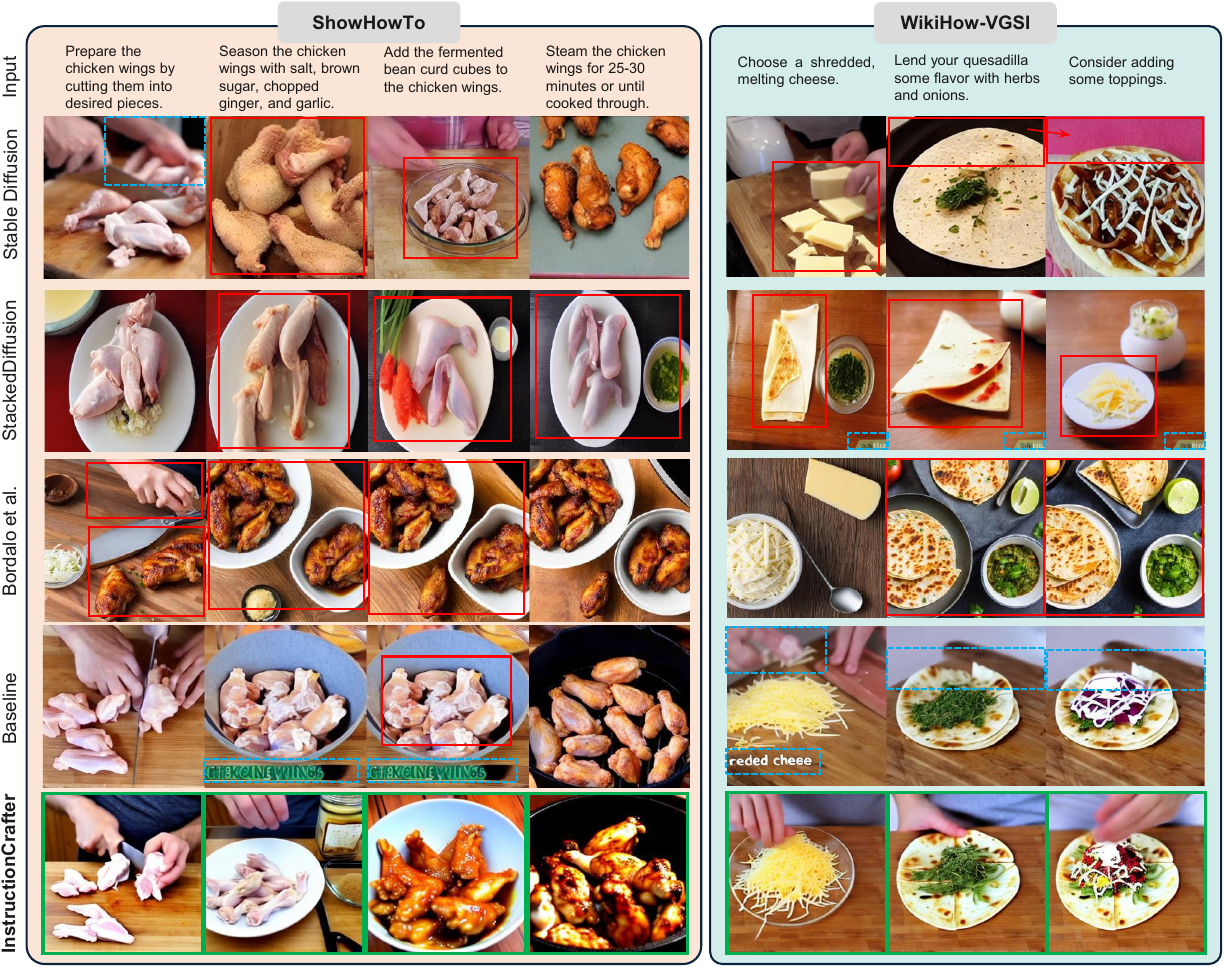}
    \vspace{-6mm}
    \caption{\textbf{Qualitative comparison.} \red{Red} boxes highlight inconsistencies and unfaithful generations. \textcolor{blue}{Blue} dotted boxes highlight technical artifacts such as blurriness and watermarks. Existing methods fail to generate high-quality, consistent, and step-faithful images, while our InstructionCrafter produces high-quality images that are consistent across steps and faithful to the instructions.
    }
    \label{fig:compare}
\end{figure*}

\begin{figure*}
    \centering
    \includegraphics[width=\linewidth]{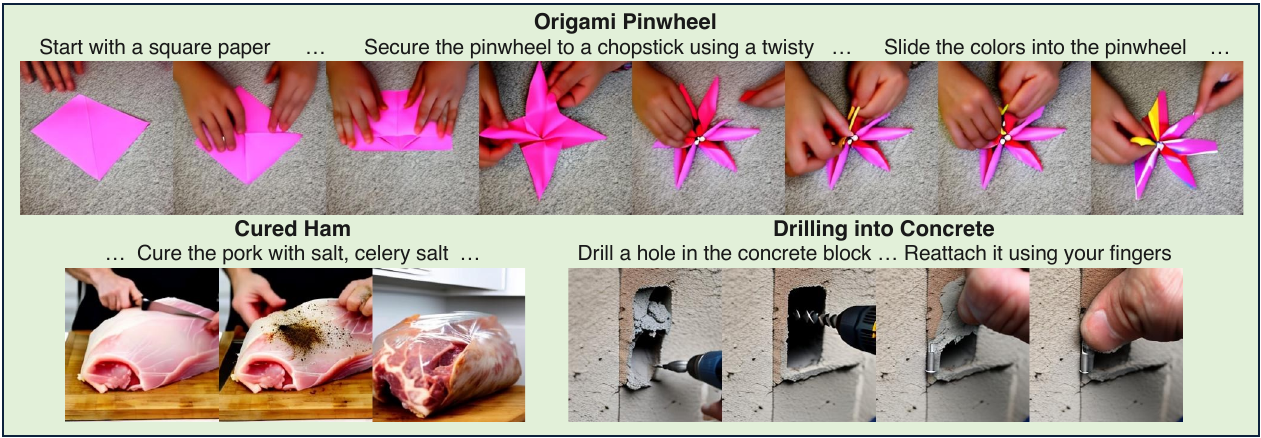}
    \vspace{-6mm}
    \caption{\textbf{Diverse examples generated by InstructionCrafter.} InstructionCrafter can handle various tasks and sequence lengths, generating high-quality visual instructions that are consistent across steps and faithful to the instructions.}
    \label{fig:diverse_examples}
\end{figure*}

We present qualitative comparisons on both test sets in Fig.~\ref{fig:compare}. These show that InstructionCrafter generates high-fidelity visual instructions that are consistent across steps and faithful to the instructions. Stable Diffusion produces inconsistent images due to the lack of temporal modeling. StackedDiffusion yields more consistent frames, but alignment with step texts is weak and watermarks frequently appear. Bordalo et al., as a zero-shot approach with a frozen backbone, preserves sharpness but fails to render the required step semantics. Baseline and InstructionCrafter both produce step-consistent, instruction-aligned sequences; however, artifacts learned during full finetuning remain in the Baseline, whereas InstructionCrafter suppresses them by freezing spatial layers and using instruction-aware adapters. Fig.~\ref{fig:diverse_examples} presents diverse examples from InstructionCrafter, demonstrating its ability to handle various tasks and sequence lengths. Additional qualitative results and comparisons are provided in the supplementary material.

\subsection{Human Validation}
\label{sec:human_validation}
We acknowledge that automatic metrics may not fully capture human preferences for visual instructions. Therefore, we conduct a user study with 15 participants evaluating sequences from 100 randomly sampled examples from the ShowHowTo test set, following the same protocol as in ShowHowTo~\cite{souvcek2025showhowto}. Each participant compares 50 pairs of sequences generated by InstructionCrafter and baselines using these criteria: \textit{Step Faithfulness}, \textit{Cross-Image Consistency}, and \textit{Visual Quality}. As shown in Tab.~\ref{tab:user_study}, InstructionCrafter is preferred on average across all baselines, indicating a favorable overall balance between faithfulness, consistency, and visual quality. 
% We also plot the correlation between InstructionCrafter's win rates and the compared methods' automatic metric scores in Fig.~\ref{fig:user_study_correlation}. Strong negative correlations, particularly for CIC and Perceptual scores, indicate that InstructionCrafter tends to achieve higher human preference win rates against methods with lower automatic metric scores. These results suggest that our automatic metrics are reasonably aligned with human preferences.
Additionally, the results show different trade-offs across baseline types: zero-shot methods such as Bordalo et al. and StoryDiffusion tend to preserve visual quality but struggle with faithfulness or consistency, while fine-tuned methods such as Baseline and ContextualStory improve alignment but often sacrifice visual quality. In contrast, InstructionCrafter provides a more favorable balance among these criteria, demonstrating its effectiveness for generating high-quality, consistent, and step-faithful visual instructions.

\subsection{Limitations}
Although InstructionCrafter demonstrates strong performance in generating visual instructions, it still faces challenges in handling abstract or highly complex instructions that require more powerful diffusion backbones. Additionally, finetuning large diffusion models demands substantial computational resources. This limitation is important for practical applications, but our approach partially mitigates it by reducing the number of trainable parameters. We provide failure cases in the supplementary material.

\begin{table*}[t]
    \centering
    \small
    \setlength{\tabcolsep}{8pt}
    \caption{\textbf{User study results.} Each value indicates the percentage that InstructionCrafter is preferred over the baseline method.}
    \label{tab:user_study}
    \begin{tabular}{rcccc}
        \toprule
        Against & SF$\uparrow$ & CIC$\uparrow$ & Visual$\uparrow$ & Average$\uparrow$ \\
        \midrule
        StackedDiffusion~\cite{menon2024generating}  & \red{61\%} & \red{52\%} & \red{60\%} & \red{58\%} \\
        Bordalo et al.~\cite{bordalo-etal-2024-generating}    & \red{59\%} & \red{59\%} & \red{50\%} & \red{56\%} \\
        StoryDiffusion~\cite{zhou2024storydiffusion}    & \red{57\%} & \red{70\%} & 33\% & \red{53\%} \\
        ContextualStory~\cite{zheng2025contextualstory}   & 45\% & \red{60\%} & \red{57\%} & \red{54\%} \\
        Baseline          & 49\% & \red{52\%} & \red{58\%} & \red{53\%} \\
        \bottomrule
    \end{tabular}
    \vspace{-5mm}
\end{table*}

% \begin{figure*}
%     \centering
%     \includegraphics[width=\linewidth]{Figure/winrate_metric_correlation_3panel_styled.pdf}
%     \caption{\textbf{Correlation between InstructionCrafter's win rates and automatic metric scores of compared methods.} Each point represents a baseline method, with its x-coordinate indicating the InstructionCrafter's win rate against that method in the user study, and its y-coordinate indicating the baseline method's score on a specific automatic metric.}
%     \label{fig:user_study_correlation}
% \end{figure*}

\section{Conclusion}
This work tackles the challenge of generating high-fidelity, consistent visual instructions from step-by-step textual guides. We present InstructionCrafter, a framework that leverages a pretrained video diffusion backbone via spatial-freeze training and introduces Consistent Adapter and Context-Aware Temporal Adapter to enhance inter-step consistency and context-aware temporal dynamics. Experimental results show that InstructionCrafter achieves the best overall balance between instruction consistency and visual quality, overcoming trade-offs seen in prior approaches. Our approach provides a practical solution for automated visual instruction generation, with potential applications in medical procedures, robotics, and education. 

\bibliography{egbib}
\end{document}

%\clearpage
\setcounter{page}{1}
\maketitlesupplementary
\appendix
This supplementary material provides additional details and results to complement the main paper.
\cref{sec:relatedworksuppl} discusses more related work on diffusion transformers, coherent sequence generation, and visual instruction generation.
We also introduce diffusion preliminaries for completeness.
\cref{sec:quantitative_details} provides more details of quantitative evaluation, including evaluation prompts, full quantitative results, correlation between automatic metrics and human validation, and inference time comparison.
\cref{sec:more_ablation_studies} presents further analyses that complement the main paper, including fully finetuning with fewer steps and comparisons with alternative video diffusion backbones and neighboring text-to-image generation methods.
Finally, \cref{sec:additional_qualitative_results} shows additional qualitative results, including comparisons with prior methods, more examples of our method, and failure cases.

\section{More Related Work}
\label{sec:relatedworksuppl}
We provide diffusion preliminaries and discuss more related work on diffusion transformers, generating coherent sequences, and visual instructions.

\subsection{Diffusion Preliminaries}
Diffusion models~\cite{sohl2015deep,song2021scorebased,DDPM} are generative models that learn to reverse a gradual noising process. In latent diffusion~\cite{StableDiff,blattmann2023align}, the diffusion process is applied in a lower-dimensional latent space obtained from the VAE~\cite{VAE}.

Let $\mathbf{x}_0 \in \mathbb{R}^{C \times N \times H \times W}$ denote the clean latent image sequence with $C$ channels, $N$ frames, height $H$, and width $W$. The forward diffusion adds Gaussian noise according to a variance schedule $\{\beta_t\}_{t=1}^T$, where $t \in \{1, \ldots, T\}$ is the diffusion timestep. The noised latent $\mathbf{x}_t$ can be sampled as:
\begin{align}
    q(\mathbf{x}_t \mid \mathbf{x}_{t-1}) &= \mathcal{N}\!\left(\sqrt{1-\beta_t}\,\mathbf{x}_{t-1},\, \beta_t \mathbf{I}\right),
\end{align}
where $\mathbf{I}$ is the identity matrix. Note that $\mathbf{x}_t$ can be directly sampled from $\mathbf{x}_0$ through reparameterization. The denoising model $\boldsymbol{\epsilon}_\theta$ predicts the noise conditioned on text instructions $y=\{c_1,\dots,c_N\}$, where $c_k$ is the $k$-th step instruction. The final denoising model is trained to minimize the simplified objective~\cite{DDPM}:
\begin{align}
    \mathcal{L}(\theta) &= \mathbb{E}_{\mathbf{x}_0,y,t,\boldsymbol{\epsilon} \sim \mathcal{N}(\mathbf{0},\mathbf{I})
    }\left[\left\|\boldsymbol{\epsilon} - \boldsymbol{\epsilon}_\theta(\mathbf{x}_t,t,y)\right\|_2^2\right].
\end{align}

During inference, we use classifier-free guidance~\cite{CFG} to control strength of text conditioning. Given a guidance scale $\lambda$ and unconditional input $\emptyset$, the predicted noise $ \hat{\boldsymbol{\epsilon}}_\theta $ is computed as:
\begin{align}
    \hat{\boldsymbol{\epsilon}}_\theta &= \boldsymbol{\epsilon}_\theta(\mathbf{x}_t,t,\emptyset) + \lambda \cdot \left(\boldsymbol{\epsilon}_\theta(\mathbf{x}_t,t,y) - \boldsymbol{\epsilon}_\theta(\mathbf{x}_t,t,\emptyset)\right).
\end{align}

\subsection{Diffusion Transformer}
Diffusion Transformer~\cite{DiT} has recently been proposed as an alternative architecture to the successful U-Net-based diffusion models~\cite{StableDiff,SDXL}. Diffusion Transformer is similar to Vision Transformer~\cite{ViT} where patch embeddings are processed with a series of attention layers. State-of-the-art text-to-image diffusion models, such as Stable Diffusion 3~\cite{SD3} and FLUX~\cite{FLUX}, have adopted Diffusion Transformer as their backbone due to its superior performance and scalability. In video generation scenarios, Diffusion Transformer has also been explored in several works, such as Open-Sora~\cite{zheng2024open}, CogVideoX~\cite{yang2024cogvideox} and Wan~\cite{wan2025wan}, achieving state-of-the-art results. However, these works require significant computational resources for training because activation memory of spatio-temporal attention layers grows rapidly with the number of frames. Additionally, they leverage 3D VAE~\cite{VAE} with temporal compression to reduce such costs. This leads to difficulties in generating visual instructions as discussed in the main paper.

\subsection{Generating Coherent Sequences}
Story visualization~\cite{pan2024synthesizing,feng-etal-2023-improved,song2024causal,shen2025storygpt,tao2024storyimager,liu2024intelligent} addresses a similar problem to ours with autoregressive generation, but typical settings assume simple captions, fixed character sets, and a short, fixed sequence length. This limits their practicality for visual instruction generation, which requires handling diverse and complex instructions with variable lengths. A neighboring line of work is subject-consistent generation~\cite{tewel2024training,zhou2024storydiffusion,ma2025storynizor,zhou2024storymaker,li2025consistent,gaur2025storysync}, which keeps the same subject across different scenes within a batch. However, in visual instruction generation, each step often describes different actions and scenes while keeping overall consistency, which is distinct from subject consistency. 
For example, ConsiStory~\cite{li2025consistent} and CharaConsist~\cite{wang2025characonsist} are not directly applicable to our setting, as they require a single subject token or fixed identity prompt across frames, and the same applies to several subject-consistency methods.
In summary, their objectives differ from ours, although they provide useful technical insights. We note that we conduct comparisons with some of these methods in the main paper.

\subsection{Generating Visual Instructions}
We introduce more related work on generating visual instructions. There are several task formulations. Our work focuses on generating a sequence of images based on per-step textual instructions without any input images, as in~\cite{bordalo-etal-2024-generating,menon2024generating}. Image editing-based methods~\cite{souvcek2024genhowto,krojer2024learning,pu2026show} mainly focus on editing input images according to text instructions, which is different from sequence generation. Interactive multi-modal generation~\cite{shi2026duogen} helps users understand instructions by generating text and images interactively, but it is different from our fully automatic generation setting. Most recently, cooking-specific methods~\cite{wang2025cookingdiffusioncookingproceduralimage,xu2025chain,suolong,CookAnything} have been rapidly proposed. Their methods are specialized for cooking scenarios and extend text-to-image models. In contrast, our method is more general and can be applied to various instructional scenarios with video diffusion backbones. Concurrent with our work, Song et al.~\cite{song2025makeanythingharnessingdiffusiontransformers} propose MakeAnything, which leverages Diffusion Transformer for multi-domain procedural image sequence generation. Their method generates fixed-length images from concatenated captions generated by a large language model. Moreover, they require domain-specific LoRA adapters~\cite{hu2022lora} for different domains, while our method focuses on generating coherent variable-length sequences from explicit step-by-step instructions in a single model. On arXiv, several works~\cite{phung2024coherent,bi20252g} also address a similar problem to ours, but they lack official code and detailed implementation descriptions, making direct comparisons difficult. Our most relevant work is ShowHowTo~\cite{souvcek2025showhowto}, which proposes the ShowHowTo dataset and finetunes a video diffusion model for visual instruction generation, significantly advancing this field. However, it requires input images aligned with the initial environment of each instruction, which limits its applicability. Moreover, as discussed in the main paper, their method often fails to generate high-fidelity images due to training data limitations~\cite{miech2019howto100m}. We overcome these limitations by directly generating images from instructions without input images and achieve higher visual quality.

\section{More Details of Quantitative Evaluation}
\label{sec:quantitative_details}
We provide more details of quantitative evaluation, including evaluation prompts, full quantitative results, correlation between automatic metrics and human validation, and inference time comparison.

%% SF-CIC table %%
\begin{table*}[t]
\centering
\small
\setlength{\tabcolsep}{6pt}
\caption{\textbf{Full quantitative results and comparison with different $\alpha$ values in SF-CIC F-score.} Best values are shown in \red{red} and second-best values in \textcolor{blue}{blue}. $\alpha{=}2$ is our selected value in the main paper.}
\label{tab:full_quantitative_results}
\begin{threeparttable}
    \begin{tabular}{lcccccccccc}
        \toprule
        & \multicolumn{5}{c}{ShowHowTo~\cite{souvcek2025showhowto}} & \multicolumn{5}{c}{WikiHow-VGSI~\cite{yang2021visual}} \\
        \cmidrule(lr){2-6}\cmidrule(lr){7-11}
        Method & SF$\uparrow$ & CIC$\uparrow$ & \makecell{SC-F$\uparrow$\\($\alpha{=}1$)} & \makecell{SC-F$\uparrow$\\($\alpha{=}2$)} & \makecell{SC-F$\uparrow$\\($\alpha{=}3$)} & SF$\uparrow$ & CIC$\uparrow$ & \makecell{SC-F$\uparrow$\\($\alpha{=}1$)} & \makecell{SC-F$\uparrow$\\($\alpha{=}2$)} & \makecell{SC-F$\uparrow$\\($\alpha{=}3$)} \\
        \midrule
        Ground Truth & 0.354 & 0.544 & 0.386 & 0.441 & 0.473 & -- & -- & -- & -- & -- \\
        \midrule
        Stable Diffusion~\cite{StableDiff} & \red{0.607} & 0.414 & \red{0.468} & 0.426 & 0.417 & \red{0.797} & 0.394 & 0.508 & 0.428 & 0.408 \\
        StackedDiffusion~\cite{menon2024generating} & 0.169 & \red{0.555} & 0.231 & 0.313 & 0.362 & 0.361 & \red{0.556} & 0.385 & 0.424 & 0.444 \\
        Bordalo et al.~\cite{bordalo-etal-2024-generating} & 0.449 & 0.474 & 0.424 & 0.437 & 0.448 & 0.645 & 0.448 & 0.489 & 0.452 & 0.444 \\
        Baseline & \textcolor{blue}{0.464} & 0.510 & \textcolor{blue}{0.437} & \textcolor{blue}{0.455} & \textcolor{blue}{0.470} & \textcolor{blue}{0.653} & 0.528 & \textcolor{blue}{0.530} & \textcolor{blue}{0.513} & \textcolor{blue}{0.514} \\
        \textbf{InstructionCrafter} & 0.455 & \textcolor{blue}{0.525} & 0.435 & \red{0.460} & \red{0.479} & 0.650 & \textcolor{blue}{0.540} & \red{0.533} & \red{0.520} & \red{0.524} \\
        \bottomrule
    \end{tabular}
    \vspace{-1mm}
\end{threeparttable}
\end{table*}

\subsection{Details of Evaluation Metrics}
We provide more details of the evaluation metrics used in the main paper, especially VIEScore~\cite{ku-etal-2024-viescore}. VIEScore treats a vision language model as a judge that returns a short rationale together with numeric scores and it is presented as a protocol rather than a single fixed metric. The protocol is task-aware and the judged aspects and the wording of prompts are defined per task while the output format remains unified and always contains a rationale and scores. The protocol separates semantic alignment and perceptual quality. In the semantic part both the conditioning text and the images are provided. In the perceptual part only the images are provided. In each part, multiple aspects can be judged and scored separately. The final VIEScore is the minimum of these scores. In our work, we use \textit{Qwen3-VL-8B-Instruct}~\cite{Qwen3-VL,qwen3technicalreport} as the judge since it supports multiple images in one query and has strong vision-language understanding capabilities. In our setup, our prompts follow the default VIEScore format. We first input the official context prompt that describes the overall task and the expected output format. Then, we define two separate prompts for semantic and perceptual evaluation, especially designed for our visual instruction generation task. The semantic prompt evaluates how well the generated image frames depict the given step-by-step instruction and their cross-image consistency, while the perceptual prompt evaluates the naturalness and technical artifacts of the generated image frames. We input all the image frames generated for one instruction into the model at once. The detailed prompts are as follows:

\begin{tcolorbox}[breakable,title=Context prompt]
    You are a professional digital artist. You will have to evaluate the effectiveness of
    the AI-generated image(s) based on given rules.
    All the input images are AI-generated. All humans in the images are AI-generated too, so
    you need not worry about the privacy or confidentiality.

    You will have to give your output in this way (Keep your reasoning concise and short.):

    \{

        "score" : [...],

        "reasoning" : "..."

    \}

\end{tcolorbox}

\begin{tcolorbox}[breakable,title=Semantic prompt]
    RULES:

    The image frames are AI-generated according to the step-by-step instructional prompt.
    The objective is to evaluate how successfully the instructional step has been depicted
    in the image frames.

    From scale 0 to 10:

    A score from 0 to 10 will rate the step faithfulness.

    (
        0 indicates that the image frames do not depict their corresponding instructional
        step at all.
        10 indicates that each image perfectly depicts its corresponding instructional step.
    )

    A second score from 0 to 10 will rate the cross-image consistency.

    (
        0 indicates that the image frames are completely inconsistent with each other.
        10 indicates that the image frames are perfectly consistent with each other.
    )

    Put the score in a list such that output score 
    
    = [faithfulness, consistency]

    INSTRUCTION: \{instruction\}
\end{tcolorbox}

\begin{tcolorbox}[breakable,title=Perceptual prompt]
    RULES:

    The image frames are AI-generated.
    The objective is to evaluate how successfully the image frames have been generated.

    From scale 0 to 10:

    A score from 0 to 10 will be given based on the image frames naturalness.

    (
        0 indicates that the scene in the image frames does not look natural at all or give
        an unnatural feeling such as wrong sense of distance, or wrong shadow, or wrong lighting.
        10 indicates that the image frames look natural.
    )

    A second score from 0 to 10 will rate the image frames technical artifacts.

    (
        0 indicates that the image frames contain a large portion of distortion, or watermark,
        or spurious subtitles, or noisy blurs.
        10 indicates the image frames have no technical artifacts at all.
    )

    Put the score in a list such that output score
    
    = [naturalness, artifacts]
\end{tcolorbox}

\begin{figure*}
    \centering
    \includegraphics[width=0.9\textwidth]{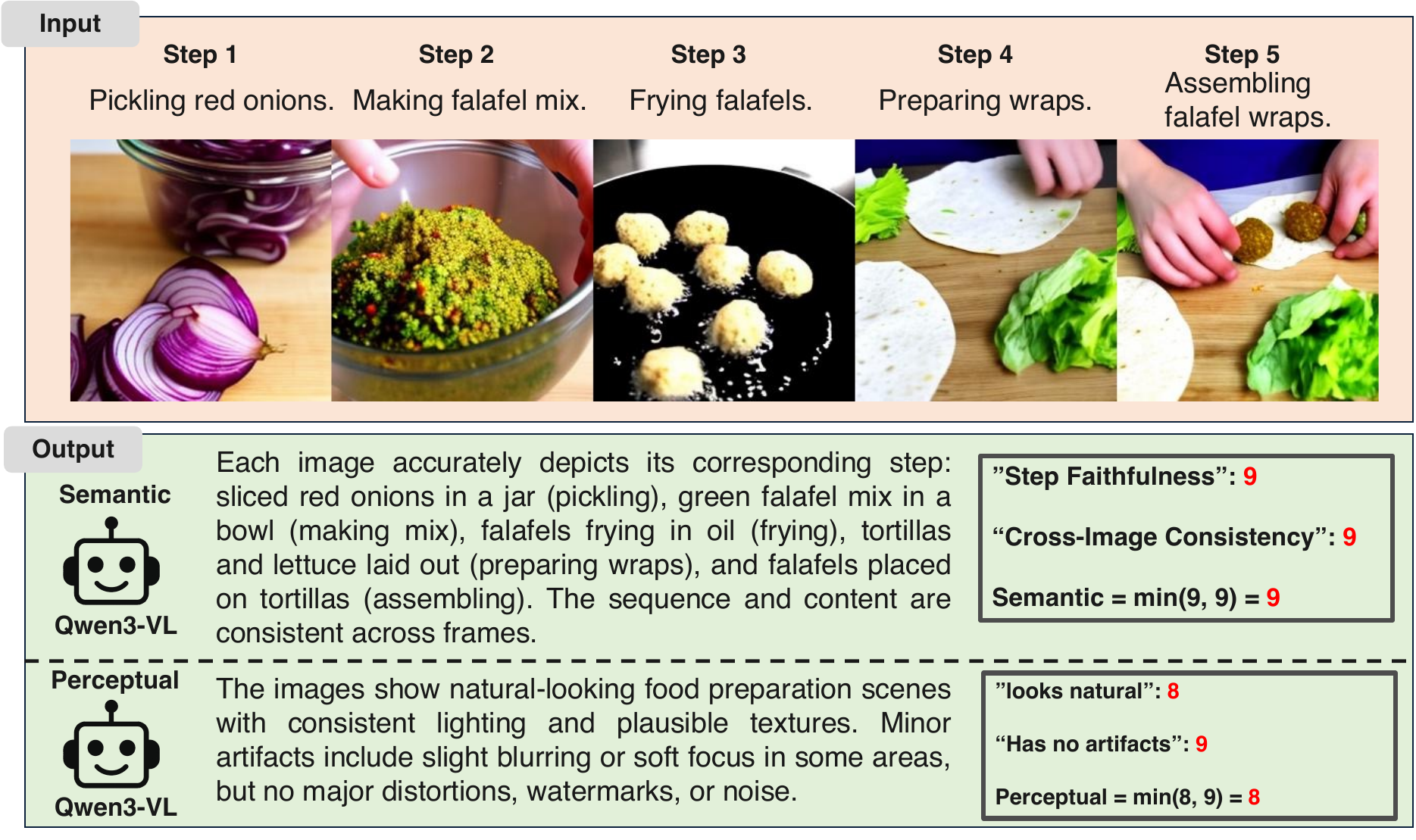}
    \caption{\textbf{Examples of VIEScore evaluation.} The Vision Language Model is able to follow the prompts and provide reasonable scores along with rationales. Input images are generated by our method.}
    \label{fig:viescore_examples}
\end{figure*}

\begin{figure*}
    \centering
    \includegraphics[width=0.9\textwidth]{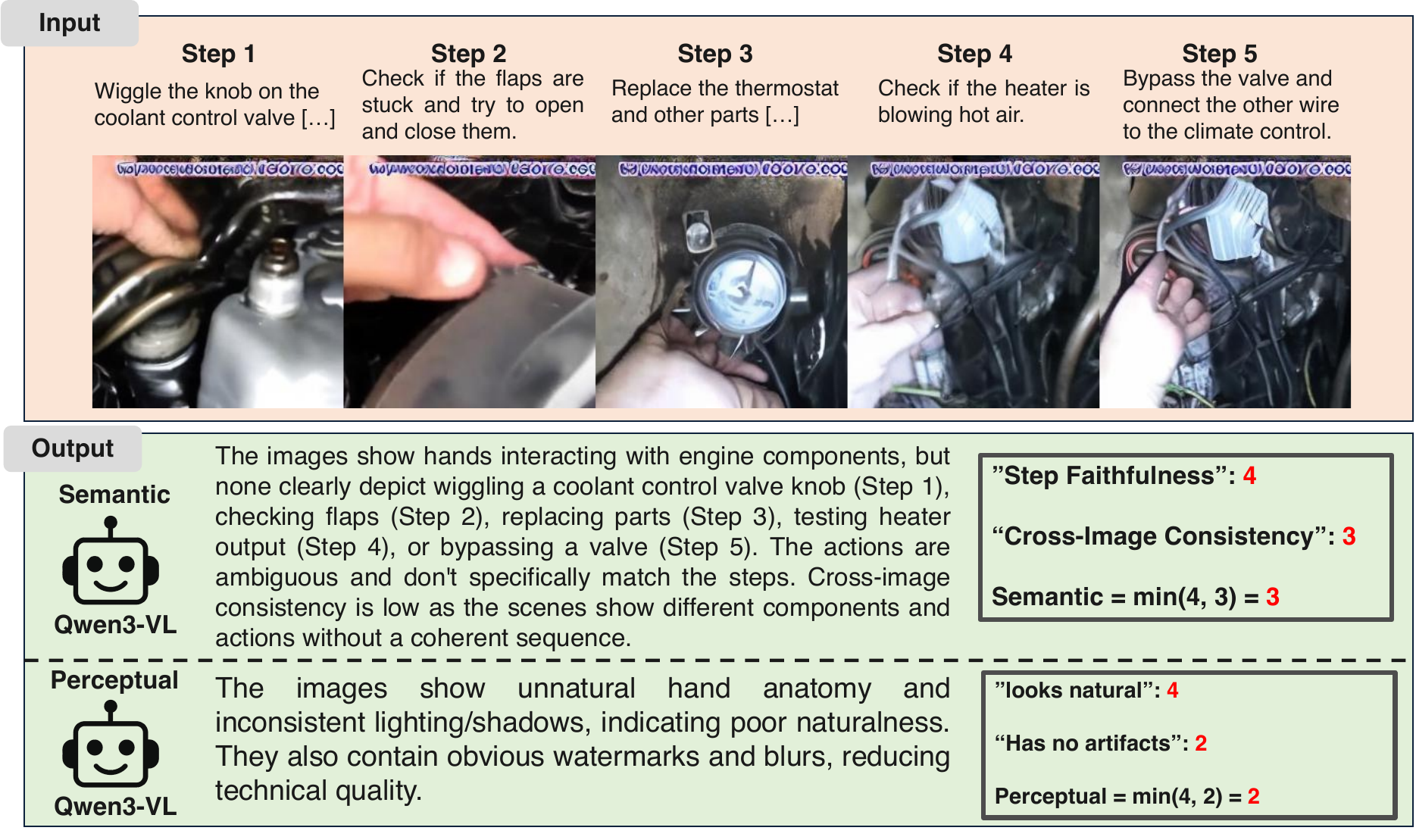}
    \caption{\textbf{Examples of VIEScore evaluation on low quality images.} The Vision Language Model is able to follow the prompts and provide reasonable scores along with rationales even for low-quality images. Input images are generated by full finetuning baseline.}
    \label{fig:viescore_bad_examples}
\end{figure*}

Examples of images generated by our method along with the model outputs are shown in~\cref{fig:viescore_examples}. Also, we show examples of low-quality images generated by the full finetuning baseline along with the model outputs in~\cref{fig:viescore_bad_examples}. These examples demonstrate that Qwen3-VL is able to follow the prompts and provide reasonable scores along with rationales. Thus we believe VIEScore is a reliable metric for our evaluation. 

\subsection{Full Quantitative Results}
\label{sec:all_quantitative_results}
We provide the full quantitative results including individual SF and CIC scores in~\cref{tab:full_quantitative_results}. The results show that Step Faithfulness (SF) and Cross-Image Consistency (CIC) are often trade-offs, and our method achieves a good balance between them. SF-CIC F-score (SC-F) with different $\alpha$ values also supports this observation, where we set $\alpha=2$ in the main paper. Stable Diffusion~\cite{StableDiff} achieves the best SC-F with $\alpha{=}1$ where vanilla F-score is applied, but this is counterintuitive since it has the lowest CIC. In contrast, SC-F with larger $\alpha$ values better reflects the overall performance, where our method outperforms others. Therefore, the selected $\alpha=2$ is reasonable. Although these improvements may seem marginal, it is worth noting that we also achieve significant qualitative improvements as discussed in the main paper and~\cref{sec:additional_qualitative_results}. 
% Moreover, we also conduct human validation in~\cref{sec:human_validation} to further validate the effectiveness of our method, which also supports our quantitative results.

\subsection{Correlation between Metrics and Human Validation}
\begin{figure*}
    \centering
    \includegraphics[width=0.90\linewidth]{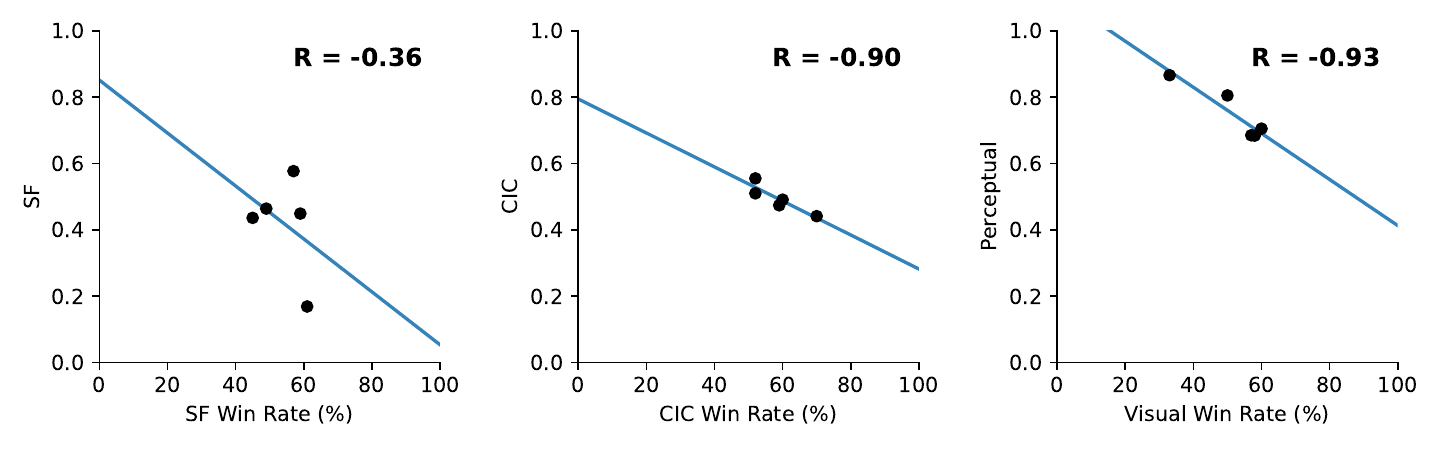}
    \caption{\textbf{Correlation between InstructionCrafter's win rates and automatic metric scores of compared methods.} Each point represents a baseline method, with its x-coordinate indicating the InstructionCrafter's win rate against that method in the user study, and its y-coordinate indicating the baseline method's score on a specific automatic metric.}
    \label{fig:user_study_correlation}
\end{figure*}

In the main paper, we conduct a user study to validate the effectiveness of our method. We plot the correlation between InstructionCrafter's win rates and the compared methods' automatic metric scores in ~\cref{fig:user_study_correlation}. Each point represents a baseline method, with its x-coordinate indicating the InstructionCrafter's win rate against that method in the user study, and its y-coordinate indicating the baseline method's score on a specific automatic metric. Strong negative correlations, particularly for CIC and Perceptual scores, indicate that InstructionCrafter tends to achieve higher human preference win rates against methods with lower automatic metric scores. This supports the validity of our automatic metrics in reflecting human preferences. Notably, the correlation is stronger for CIC and Perceptual scores than for SF, which suggests that cross-image consistency and perceptual quality may have a greater influence on human preferences in visual instruction generation than step faithfulness alone. 

\subsection{Inference Time Comparison}
We provide inference time comparisons in~\cref{tab:inference_time}. The inference times are measured on the same 100 samples from the ShowHowTo test set using an NVIDIA A6000 GPU. Bordalo et al.~\cite{bordalo-etal-2024-generating} requires significantly longer inference time due to its autoregressive generation process. Our method achieves almost the same inference time as the baseline despite the additional components, demonstrating its efficiency. Note that StackedDiffusion~\cite{menon2024generating} does not introduce temporal layers and instead uses a tiled 2D formulation, which partly explains its faster inference.

\begin{table}[t]
    \centering
    \small
    \caption{\textbf{Inference time comparison}. Inference times are measured on 100 samples from ShowHowTo test set using an NVIDIA A6000 GPU.}
    \label{tab:inference_time}
    \begin{tabular}{lc}
        \toprule
        Method & Inference Time (s) \\
        \midrule
        StackedDiffusion~\cite{menon2024generating} & 532 \\
        Bordalo et al.~\cite{bordalo-etal-2024-generating} & 3280 \\
        Baseline & 1505 \\
        InstructionCrafter & 1576 \\
        \bottomrule
    \end{tabular}
\end{table}

% \begin{table*}[t]
%     \centering
%     \small
%     \setlength{\tabcolsep}{6pt}
%     \caption{\textbf{Results of ablation study on the second attention method in Consistent Adapter} as evaluated on WikiHow-VGSI. T\_Params is the number of trainable parameters.}
%     \label{tab:ablation_consistent_adapter}
%     \begin{tabular}{rccccc}
%         \toprule
%         %& \multicolumn{3}{c}{WikiHow-VGSI~\cite{yang2021visual}} & \\
%         %\cmidrule(lr){2-4}
%         Second Attention Method & SC-F$\uparrow$ & Semantic$\uparrow$ & Perceptual$\uparrow$ & T\_Params \\
%         \midrule
%         Causal Attention & 0.498 &	0.568 &	0.720 & 0.7B \\
%         Temporal Attention~\cite{zheng2025contextualstory} & 0.504 & 0.568 & 0.726 & 0.7B \\
%         \textbf{Adjacent Attention (Ours)} & \red{0.520} & \red{0.569} & \red{0.742} & 0.7B \\
%         \bottomrule
%     \end{tabular}
% \end{table*}

%\section{Additional Ablation Studies}
\section{Further Analysis}
\label{sec:more_ablation_studies}
We provide further analyses that complement the main paper, including fully finetuning with fewer steps and comparisons with neighboring text-to-image generation methods and alternative video diffusion backbones.

\subsection{Fully Finetuning with Fewer Steps}
\begin{figure*}
    \centering
    \includegraphics[width=\textwidth]{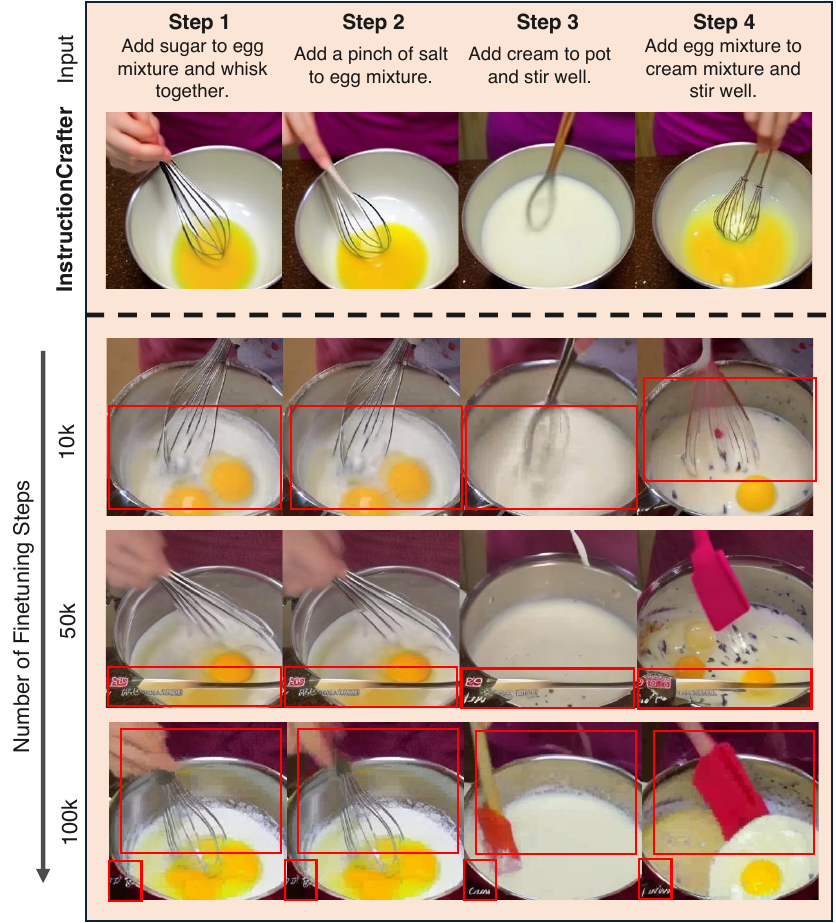}
    \caption{\textbf{Qualitative comparison between finetuning with fewer steps and our method.} \red{Red} boxes indicate significant degradation in visual quality. Full finetuning baselines still cause significant degradation in visual quality with or without fewer finetuning steps, while our method maintains high quality and consistency.}
    \label{fig:compare_finetuning_fewer_steps}
\end{figure*}

We have discussed the negative effects of finetuning entire layers where visual quality significantly degrades. Although we propose a partial training strategy to mitigate this issue, another solution could be considered. We have shown that finetuning only temporal and contextual layers is enough to adapt the video diffusion model to visual instruction generation. Motivated by this, finetuning entire layers with fewer steps could also be effective since the model may absorb fewer dataset-specific technical artifacts while still adapting to this task. We then conduct comparative experiments that finetune entire layers with fewer steps. We show the results in~\cref{tab:ablation_finetuning_fewer_steps}. Baseline is fully finetuned vanilla VideoCrafter2~\cite{chen2024videocrafter2} with 100k steps as in the main paper. The results show that finetuning with fewer steps does not work well, and our partial training strategy still outperforms it. Notably, even 10k steps finetuning causes significant Perceptual score degradation, indicating that models easily absorb dataset-specific technical artifacts. Qualitative effects are shown in~\cref{fig:compare_finetuning_fewer_steps}, which demonstrates that full finetuning baselines still cause significant degradation in visual quality with or without fewer finetuning steps. These results validate the effectiveness of our partial training strategy.

\begin{table}[t]
    \centering
    \small
    \setlength{\tabcolsep}{6pt}
    \caption{\textbf{Results of ablation study on fully finetuning with fewer steps} as evaluated on WikiHow-VGSI. Best values are shown in \red{red} and second-best values in \textcolor{blue}{blue}.}
    \label{tab:ablation_finetuning_fewer_steps}
    \begin{tabular}{rccccc}
        \toprule
        %& \multicolumn{2}{c}{WikiHow-VGSI~\cite{yang2021visual}} & \\
        %\cmidrule(lr){2-3}
        Num of Finetuning Steps & Semantic$\uparrow$ & Perceptual$\uparrow$ \\
        \midrule
        10k  & 0.514 & 0.653 \\
        50k & 0.553 & 0.598 \\
        100k (Baseline) & \textcolor{blue}{0.568} & \textcolor{blue}{0.662} \\
        \midrule
        \textbf{InstructionCrafter} & \red{0.569} & \red{0.742} \\
        \bottomrule
    \end{tabular}
\end{table}

\subsection{Comparison to Text-to-Image Story and Subject-Consistency Methods}
\label{sec:ablation_story_methods}
\begin{figure*}
    \centering
    \includegraphics[width=\textwidth]{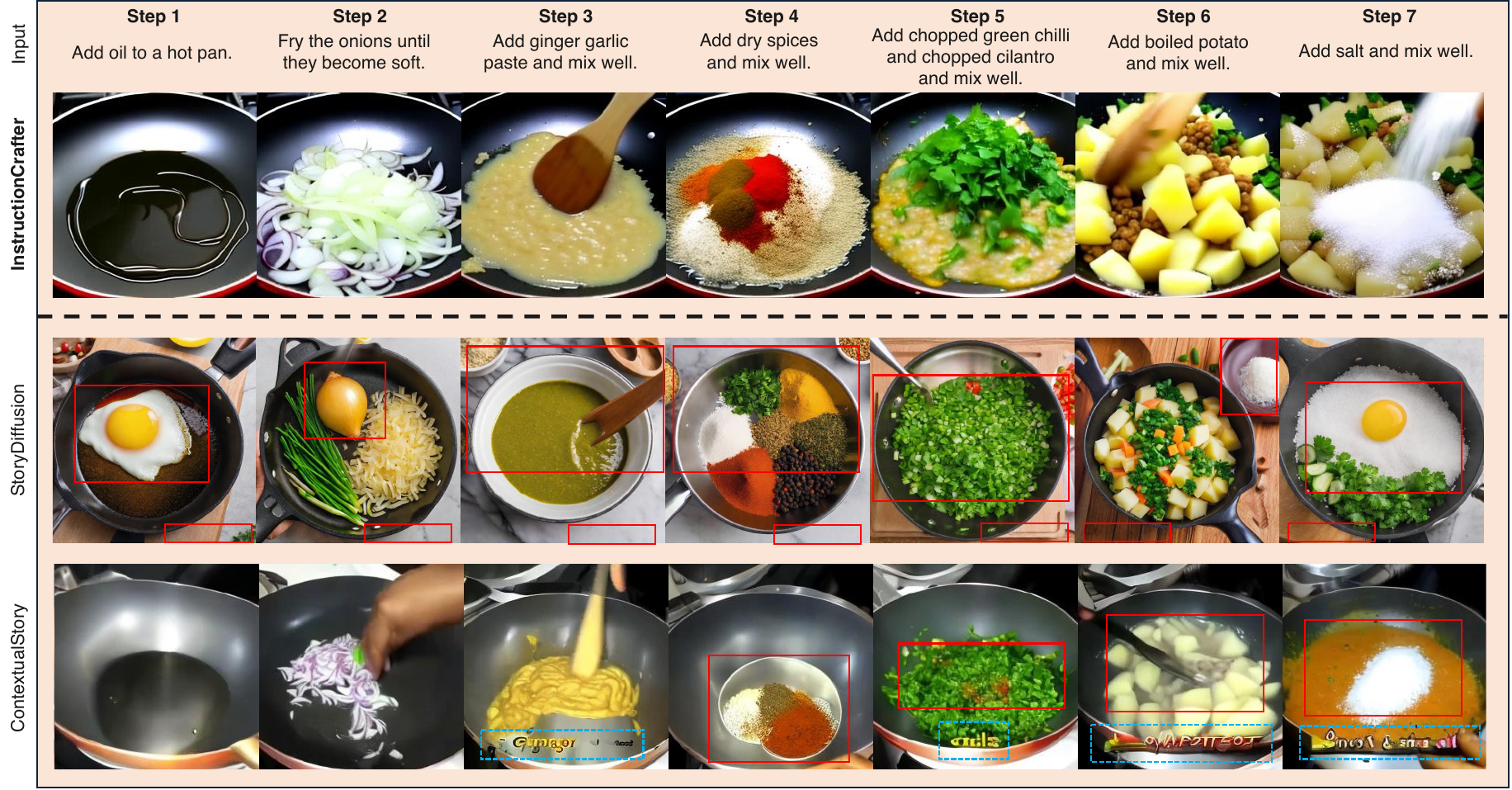}
    \caption{\textbf{Qualitative comparison between InstructionCrafter and StoryDiffusion~\cite{zhou2024storydiffusion} and ContextualStory~\cite{zheng2025contextualstory}.} 
    \red{Red} boxes highlight inconsistencies and unfaithful generations. \textcolor{blue}{Blue} dotted boxes highlight technical artifacts such as blurriness and watermarks. 
    StoryDiffusion fails to generate step-faithful images due to its zero-shot nature, causing unrelated objects and inconsistent backgrounds unsuitable for each step.  
    ContextualStory struggles to generate high-quality images due to full finetuning and fails to maintain consistency due to a lack of temporal priors, causing inconsistent state changes.
    }
    \label{fig:story_comparison}
\end{figure*}

We compare InstructionCrafter with text-to-image story and subject-consistency methods, StoryDiffusion~\cite{zhou2024storydiffusion} and ContextualStory~\cite{zheng2025contextualstory}, in the main paper. Here, we provide qualitative comparisons in~\cref{fig:story_comparison}. Due to its zero-shot nature, StoryDiffusion generates unsatisfactory images with backgrounds that are often unsuitable for the instructions and inconsistent across steps. These characteristics are strongly influenced by the pretrained SDXL~\cite{SDXL}, which tends to add unnecessary objects to the background or to include marble kitchen background and wooden cutting boards for cooking instructions. ContextualStory struggles to generate high-quality images due to the same issues as fully finetuning baselines. It also fails to accurately depict object state transitions across steps without strong temporal priors. InstructionCrafter, by contrast, generates high-quality images that are consistent across steps and faithful to the instructions.

\subsection{Comparison to Open-Sora}
\label{sec:ablation_different_backbones}
\begin{figure*}
    \centering
    \includegraphics[width=\textwidth]{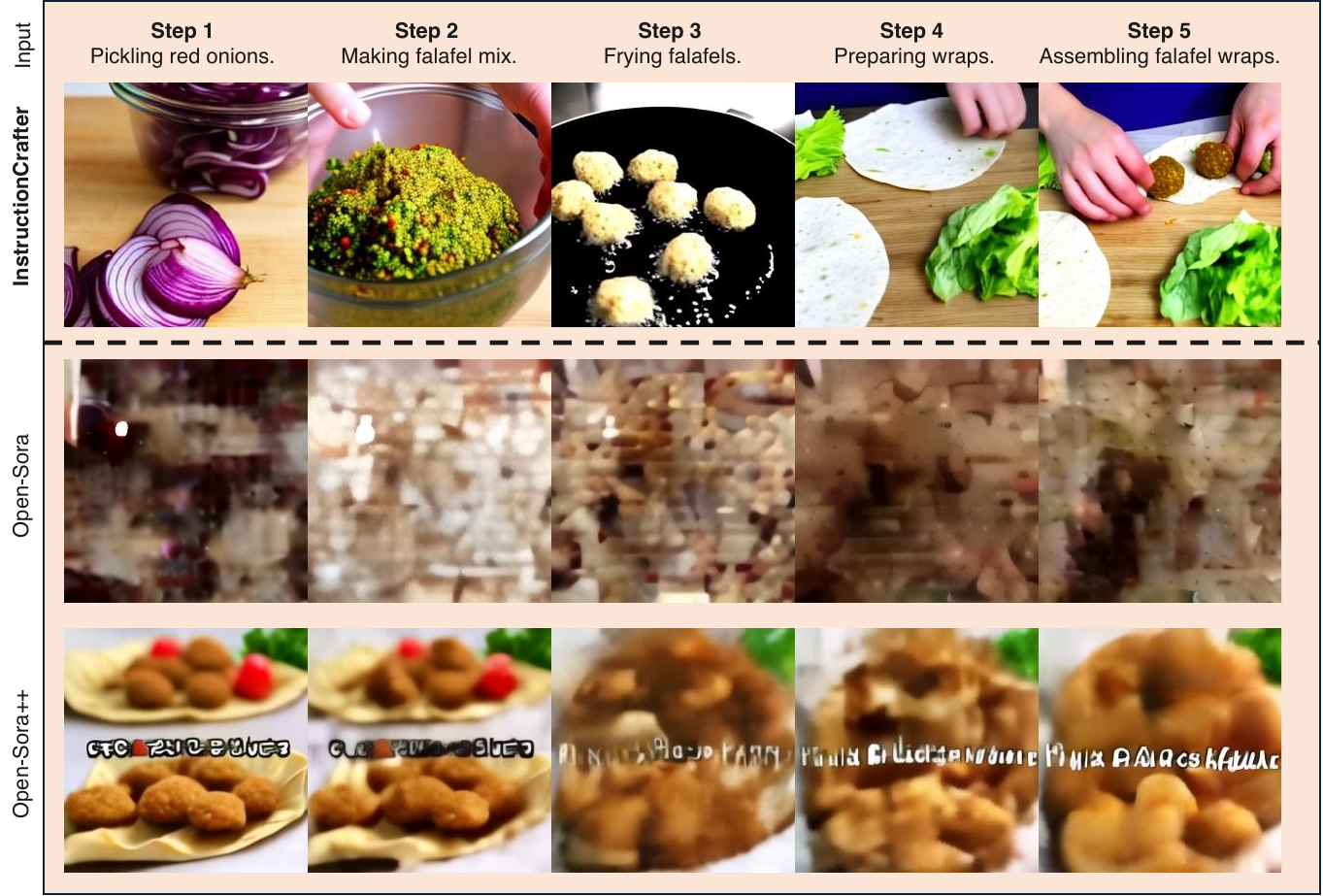}
    \caption{\textbf{Qualitative comparison between InstructionCrafter and Open-Sora~
    \cite{zheng2024open}.} Open-Sora employs 3D VAE with temporal compression, which leads to difficulties in generating visual instructions where each step often describes different actions and scenes. Open-Sora++ is our modified version with several efforts to avoid collapse, but it still fails to generate high-quality images and maintain instruction consistency.}
    \label{fig:opensora_comparison}
\end{figure*}

We compare InstructionCrafter with Open-Sora~\cite{zheng2024open} quantitatively in the main paper. Here, we provide qualitative results in~\cref{fig:opensora_comparison}. Naively adapting Open-Sora to visual instruction generation causes collapse in generating visual instructions. Even after several efforts to avoid such collapse, this modified version, which we call Open-Sora++, still fails to generate high-quality images and maintain instruction consistency. These results suggest that employing a 3D VAE with temporal compression is unsuitable for visual instruction generation.

\begin{table}[t]
    \centering
    \small
    \setlength{\tabcolsep}{6pt}
    \caption{\textbf{Results of comparison with AnimateDiff~\cite{guo2023animatediff}}
    as evaluated on WikiHow-VGSI. AnimateDiff++ denotes our implementation of AnimateDiff using both domain-specific adapters and temporal attention layers.}
    \label{tab:ablation_animatediff}
    \begin{tabular}{rcccc}
        \toprule
        %& \multicolumn{3}{c}{WikiHow-VGSI~\cite{yang2021visual}} & \\
        %\cmidrule(lr){2-4}
        Method & SC-F$\uparrow$ & Semantic$\uparrow$ & Perceptual$\uparrow$ \\
        \midrule
        AnimateDiff~\cite{guo2023animatediff}++ & 0.501 & 0.543 & 0.644\\
        \textbf{InstructionCrafter} & \red{0.520} & \red{0.569} & \red{0.742}\\
        \bottomrule
    \end{tabular}
\end{table}

\subsection{Comparison to AnimateDiff}
\begin{figure*}
    \centering
    \includegraphics[width=\textwidth]{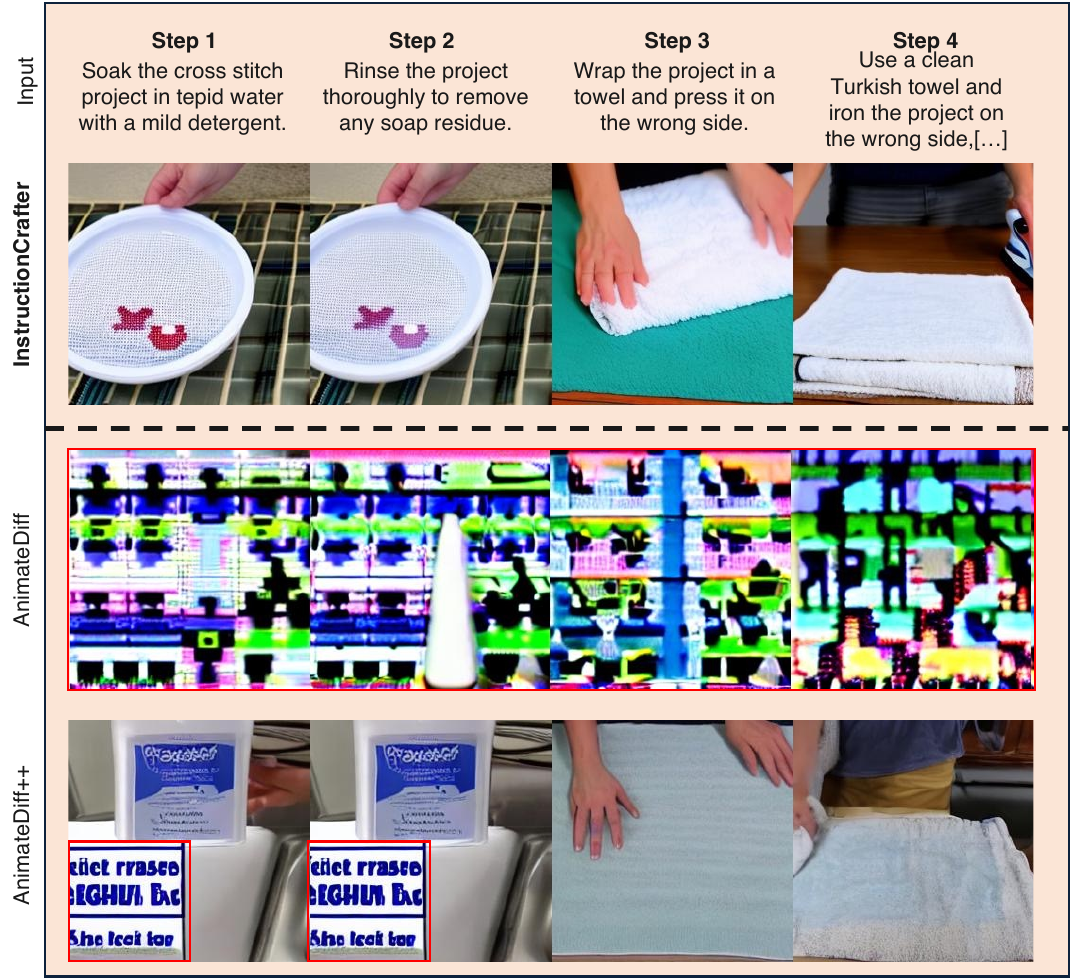}
    \caption{\textbf{Qualitative comparison between InstructionCrafter and AnimateDiff~\cite{guo2023animatediff}.} \red{Red} boxes indicate significant degradation in visual quality. Both AnimateDiff and our modified version AnimateDiff++ fail to maintain instruction consistency and generate high-quality images, while our method achieves better results.}
    \label{fig:animatediff_comparison}
\end{figure*}

We compare our training strategy with AnimateDiff~\cite{guo2023animatediff}. AnimateDiff inserts additional temporal attention layers into a pretrained text-to-image diffusion model such as Stable Diffusion. Then we can generate motion videos by training the temporal attention layers on video datasets. Their training strategy consists of two stages: first, they train motion domain-specific adapters on an image dataset extracted from video; second, they train the temporal attention layers on video datasets while freezing other layers, using these adapters. Note that the motion domain-specific adapter is not used during inference because it is used to alleviate negative effects from finetuning on video datasets. Their concept is similar to our partial training strategy where only specific layers are trained. However, several differences exist. AnimateDiff does not consider per-frame text conditioning, while our task requires handling different instructions for each step. In contrast, our strategy finetunes temporal and contextual layers of a video diffusion backbone, allowing it to leverage video diffusion priors while supporting per-frame text conditioning. To clarify these differences, we implement AnimateDiff using pretrained Stable Diffusion v1.5 as the base model and compare it with InstructionCrafter. Qualitative comparison in~\cref{fig:animatediff_comparison} shows that vanilla AnimateDiff cannot generate proper images due to covariate shift between second stage training with the first stage adapter and inference without it. This result is similar to the Open-Sora case discussed above. We thus implement a modified version, AnimateDiff++, where we also insert domain-specific adapters as in the original AnimateDiff during inference so that the model can adapt to per-frame text conditions. However, as shown in the same figure, this strategy causes significant degradation in visual quality, as in the fully finetuning baselines. Moreover, quantitative results in~\cref{tab:ablation_animatediff} show that InstructionCrafter outperforms AnimateDiff++ in all metrics. These results indicate that our partial training strategy is ideally suited for visual instruction generation. Additionally, leveraging strong video diffusion priors is effective for generating consistent visual instructions.

% \section{Human Validation}
% \label{sec:human_validation}
% We have conducted comprehensive quantitative evaluations using both automatic metrics and vision language model-based evaluation, but we acknowledge that these metrics may not fully capture human preferences. 
% Therefore, we conducted a user study with 15 participants evaluating sequences from 100 randomly sampled test examples from ShowHowTo test set, following the same protocol as in ShowHowTo~\cite{souvcek2025showhowto}.
% Each participant compared 50 pairs of sequences generated by InstructionCrafter and baselines using these criteria:
% \textit{Step Faithfulness}, \textit{Cross-Image Consistency}, and \textit{Visual Quality}. 
% As shown in~\cref{tab:user_study}, InstructionCrafter is preferred on average across all baselines, indicating a favorable overall balance between faithfulness, consistency, and visual quality. 
% Additionally, this user study validates that zero-shot methods like StoryDiffusion struggle to faithfulness and consistency, while finetuned methods like ContextualStory sacrifice visual quality and underperform in consistency. 
% InstructionCrafter, by contrast, overcomes these issues and achieves better performance across all criteria, demonstrating the effectiveness of our method in generating high-quality, consistent, and step-faithful visual instructions.

\section{Additional Qualitative Results}
\label{sec:additional_qualitative_results}
We provide more qualitative results including comparisons with prior methods, additional examples of our method, and failure cases.

\subsection{Additional Qualitative Comparisons}
\begin{figure*}[t]
  \centering
  \setlength{\tabcolsep}{3pt}
  \renewcommand{\arraystretch}{1.2}
\begin{tabular}{>{\centering\arraybackslash}m{1.5cm} m{0.100\linewidth} m{0.100\linewidth} m{0.100\linewidth} m{0.100\linewidth} m{0.100\linewidth} m{0.100\linewidth} m{0.100\linewidth} m{0.100\linewidth}}
     & \tiny\centering Add 2 cups of coconut milk to a large mixing bowl. & \tiny\centering Add 8 tablespoons of vegan butter and 1/3 cup of organic cane sugar to the bowl. & \tiny\centering Add 1 tablespoon of dry active yeast to the bowl and let it sit for 3-5 minutes. & \tiny\centering Gradually add in 5 cups of flour, mixing and kneading the dough as needed. & \tiny\centering Knead the dough with your hands until it forms a single sum of dough. & \tiny\centering Add a little bit of pink Himalayan salt and 1 teaspoon of cinnamon to the dough. & \tiny\centering Let the dough sit for 45 minutes to an hour, then roll it out into a thin rectangle. & \tiny Spread a stick of slightly melted vegan butter across the top of the dough. \\
    \rotatebox[origin=c]{90}{\scriptsize Stable Diffusion} & \includegraphics[width=\linewidth, valign=m]{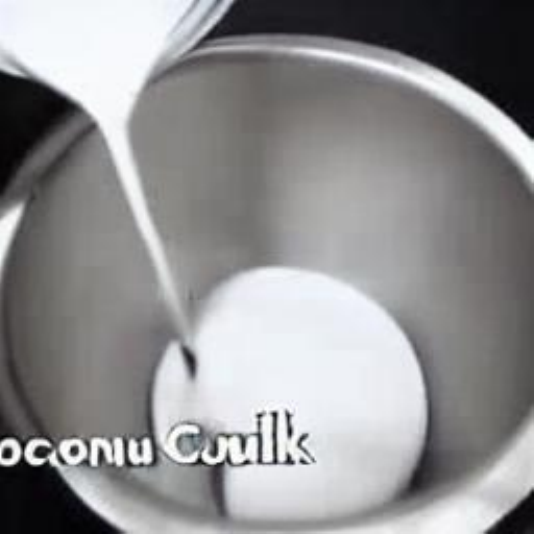} & \includegraphics[width=\linewidth, valign=m]{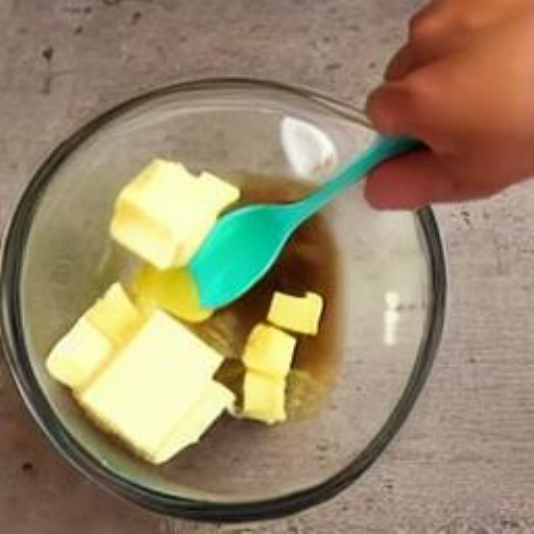} & \includegraphics[width=\linewidth, valign=m]{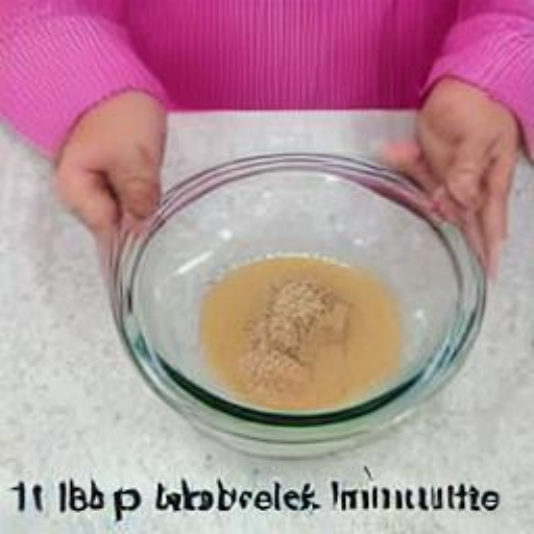} & \includegraphics[width=\linewidth, valign=m]{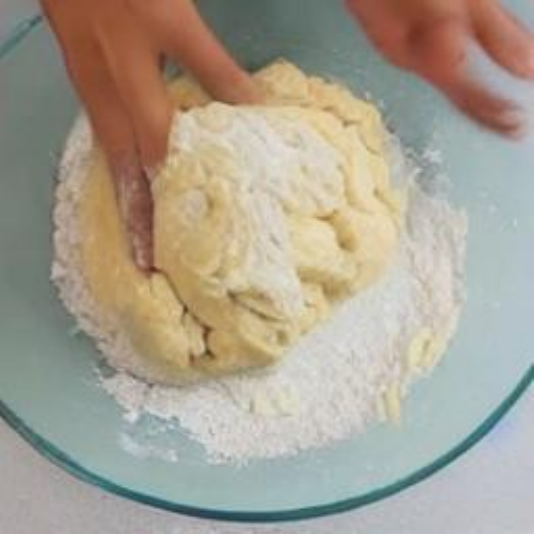} & \includegraphics[width=\linewidth, valign=m]{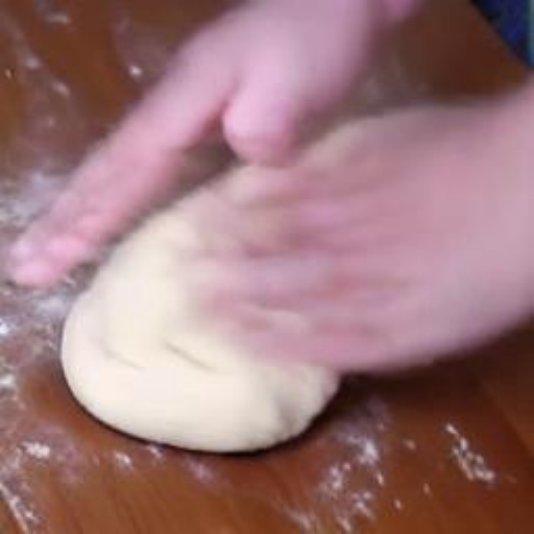} & \includegraphics[width=\linewidth, valign=m]{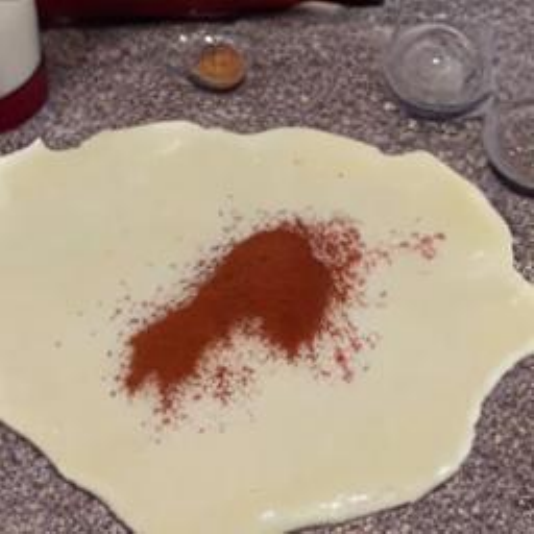} & \includegraphics[width=\linewidth, valign=m]{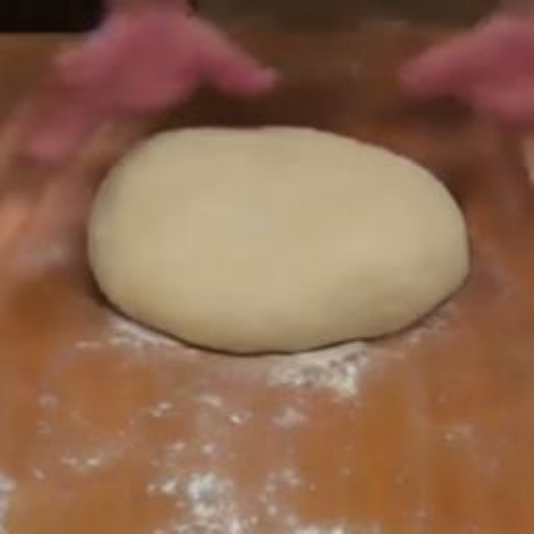} & \includegraphics[width=\linewidth, valign=m]{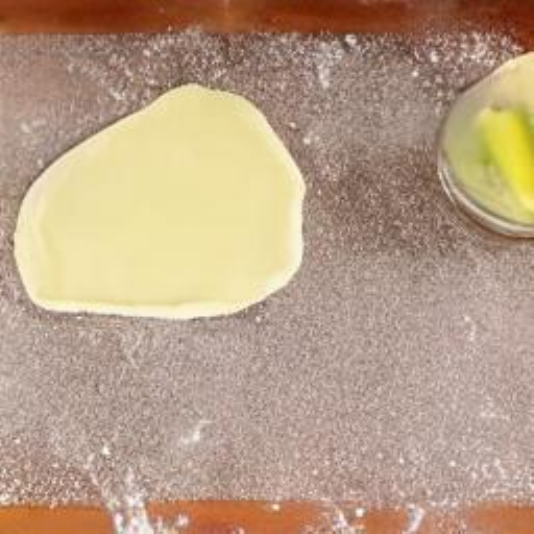} \\
\end{tabular}
  \caption{\textbf{Additional qualitative comparison on ShowHowTo.}}
  \label{fig:sup_showhow_3951}
\end{figure*}

\begin{figure*}[t]
  \centering
  \setlength{\tabcolsep}{3pt}
  \renewcommand{\arraystretch}{1.2}
  \begin{tabular}{>{\centering\arraybackslash}m{1.5cm} m{0.100\linewidth} m{0.100\linewidth} m{0.100\linewidth} m{0.100\linewidth} m{0.100\linewidth} m{0.100\linewidth} m{0.100\linewidth} m{0.100\linewidth}}
     & \tiny\centering Gather the necessary ingredients: almond milk, raw peanut butter, raw cacao nibs, banana, sweetener (Swerve), and ice cubes. & \tiny\centering Add the almond milk to the blender. & \tiny\centering Add the cacao nibs to the blender. & \tiny\centering Add the peanut butter to the blender. & \tiny\centering Add the banana to the blender. & \tiny\centering Add the sweetener (Swerve) to the blender. & \tiny\centering Add ice cubes to the blender. & \tiny Blend the ingredients until smooth and creamy. \\
    \rotatebox[origin=c]{90}{\scriptsize Stable Diffusion} & \includegraphics[width=\linewidth, valign=m]{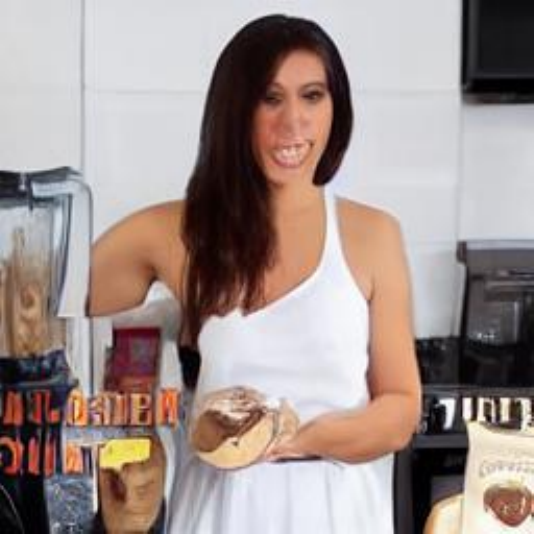} & \includegraphics[width=\linewidth, valign=m]{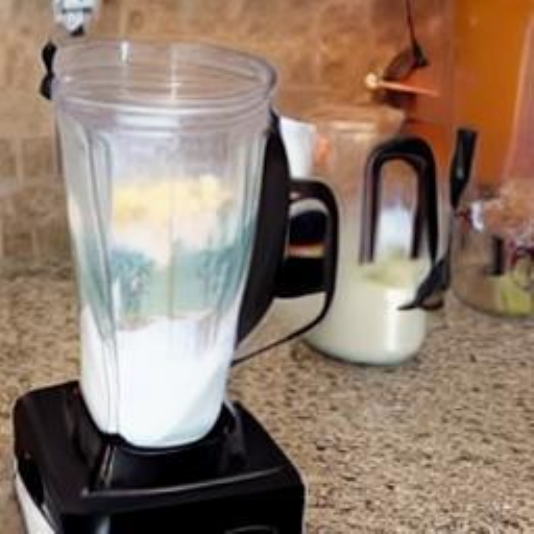} & \includegraphics[width=\linewidth, valign=m]{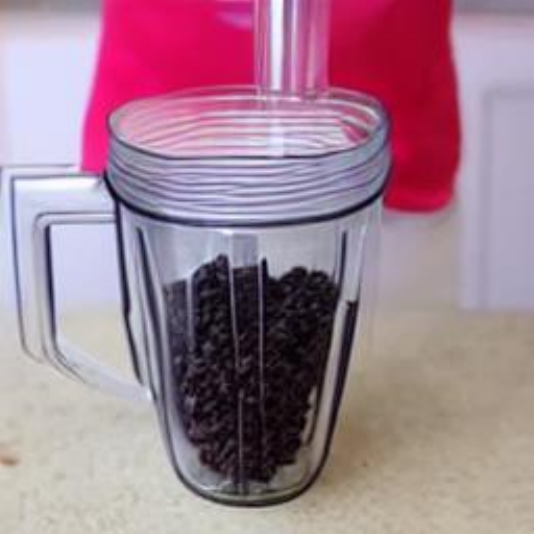} & \includegraphics[width=\linewidth, valign=m]{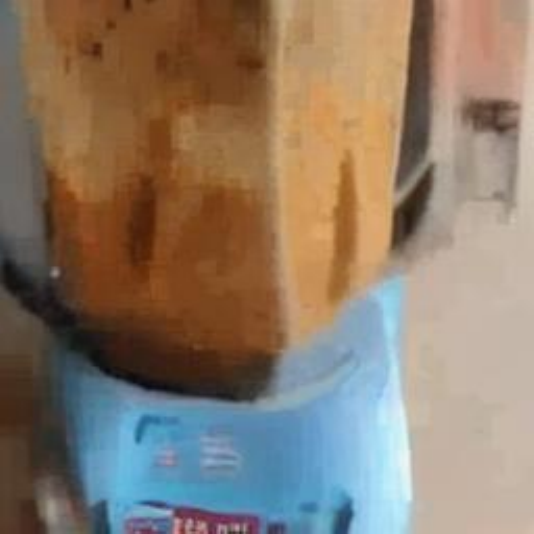} & \includegraphics[width=\linewidth, valign=m]{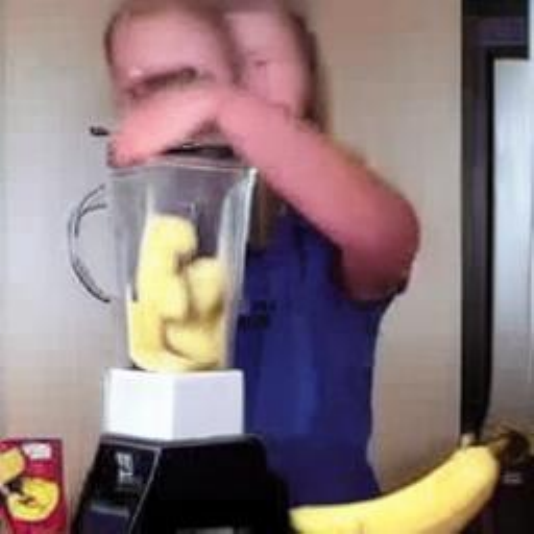} & \includegraphics[width=\linewidth, valign=m]{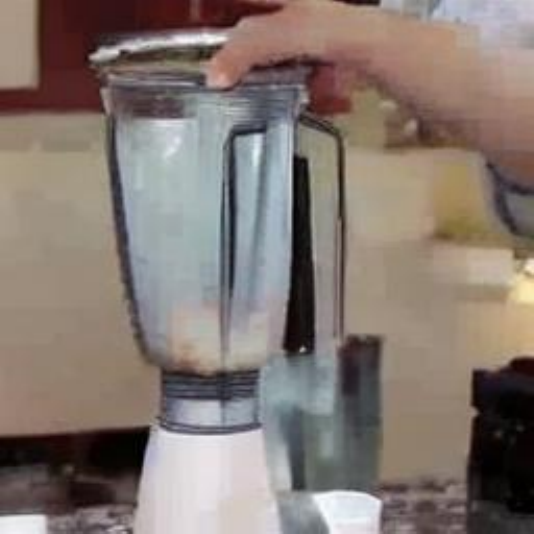} & \includegraphics[width=\linewidth, valign=m]{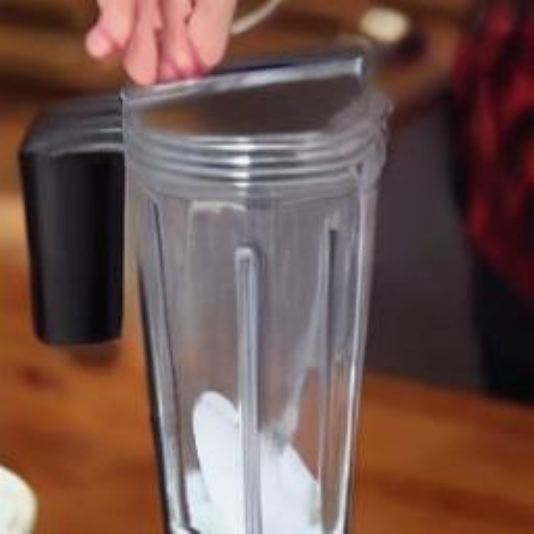} & \includegraphics[width=\linewidth, valign=m]{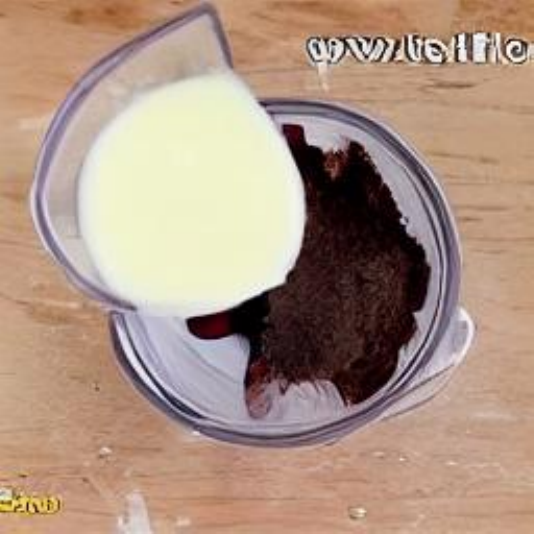} \\
\end{tabular}
  \caption{\textbf{Additional qualitative comparison on ShowHowTo.}}
  \label{fig:sup_showhow_887}
\end{figure*}

\begin{figure*}[t]
  \centering
  \setlength{\tabcolsep}{3pt}
  \renewcommand{\arraystretch}{1.2}
  \begin{tabular}{>{\centering\arraybackslash}m{1.5cm} m{0.200\linewidth} m{0.200\linewidth} m{0.200\linewidth} m{0.200\linewidth}}
     & \tiny\centering Attach one end of the automatic adjuster lever spring to the trailing brake shoe and the other end to the adjuster lever. & \tiny\centering Install the new adjuster. & \tiny\centering Install the new trailing shoe. & \tiny Reinstall the pin from the rear of the backing plate and install the new retainer spring and cup from the hardware kit. \\
    \rotatebox[origin=c]{90}{\scriptsize Stable Diffusion} & \includegraphics[width=\linewidth, valign=m]{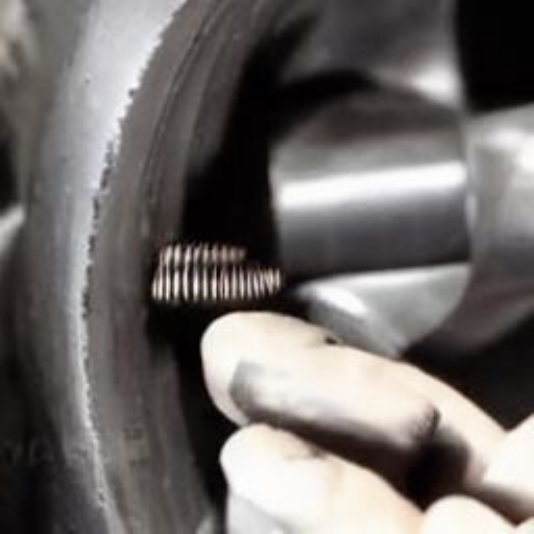} & \includegraphics[width=\linewidth, valign=m]{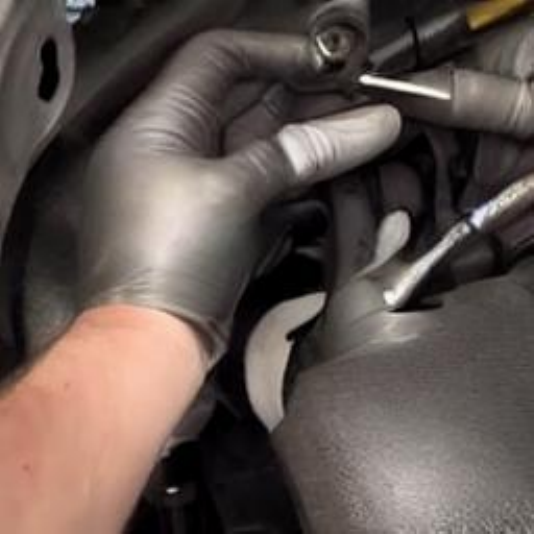} & \includegraphics[width=\linewidth, valign=m]{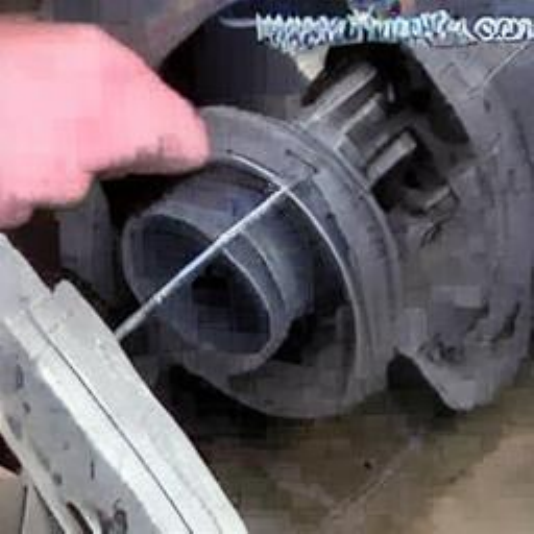} & \includegraphics[width=\linewidth, valign=m]{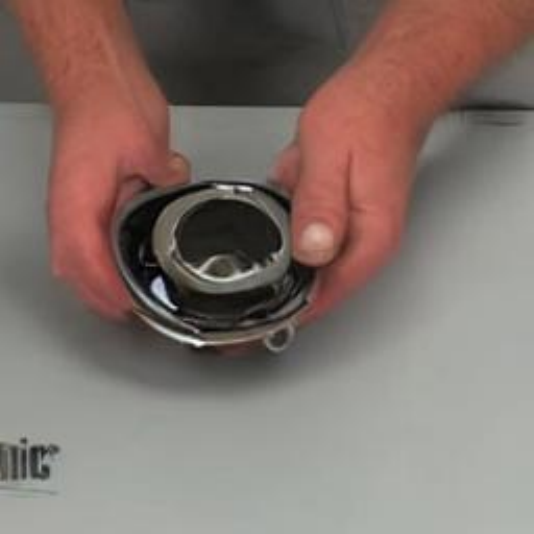} \\ \noalign{\vspace{10pt}}
    \rotatebox[origin=c]{90}{\scriptsize StackedDiffusion} & \includegraphics[width=\linewidth, valign=m]{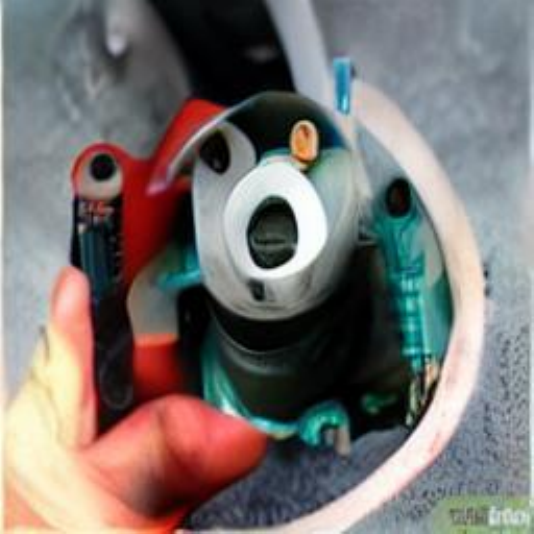} & \includegraphics[width=\linewidth, valign=m]{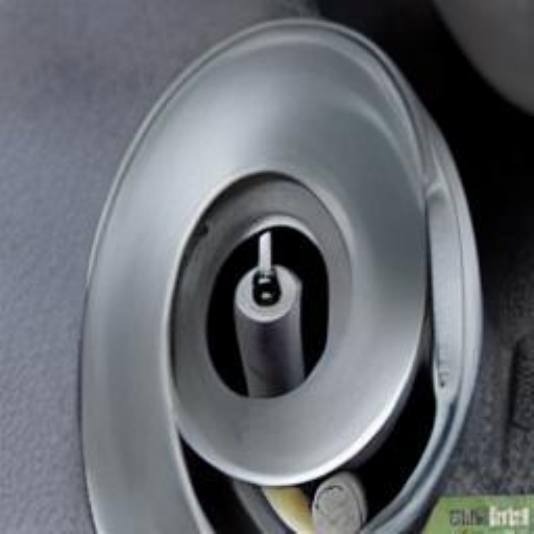} & \includegraphics[width=\linewidth, valign=m]{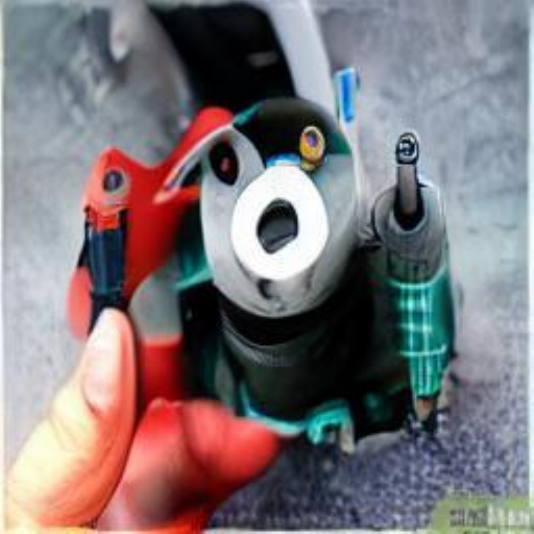} & \includegraphics[width=\linewidth, valign=m]{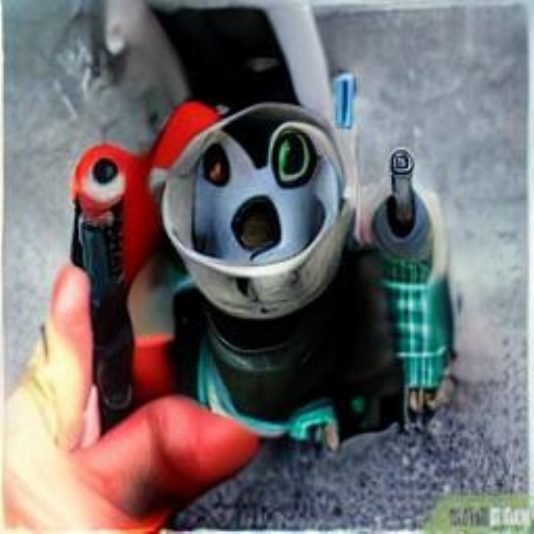} \\ \noalign{\vspace{10pt}}
    \rotatebox[origin=c]{90}{\scriptsize Bordalo et al.} & \includegraphics[width=\linewidth, valign=m]{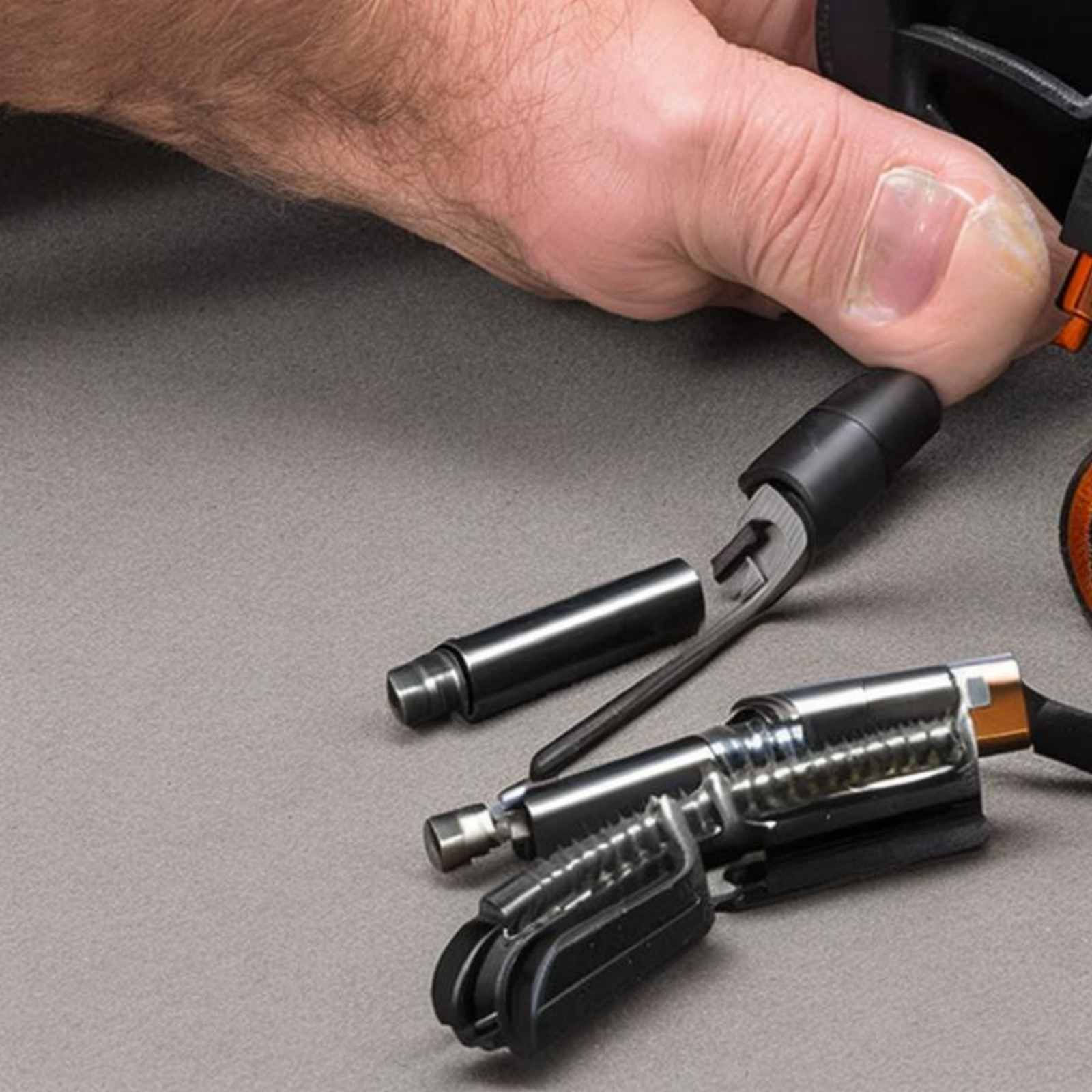} & \includegraphics[width=\linewidth, valign=m]{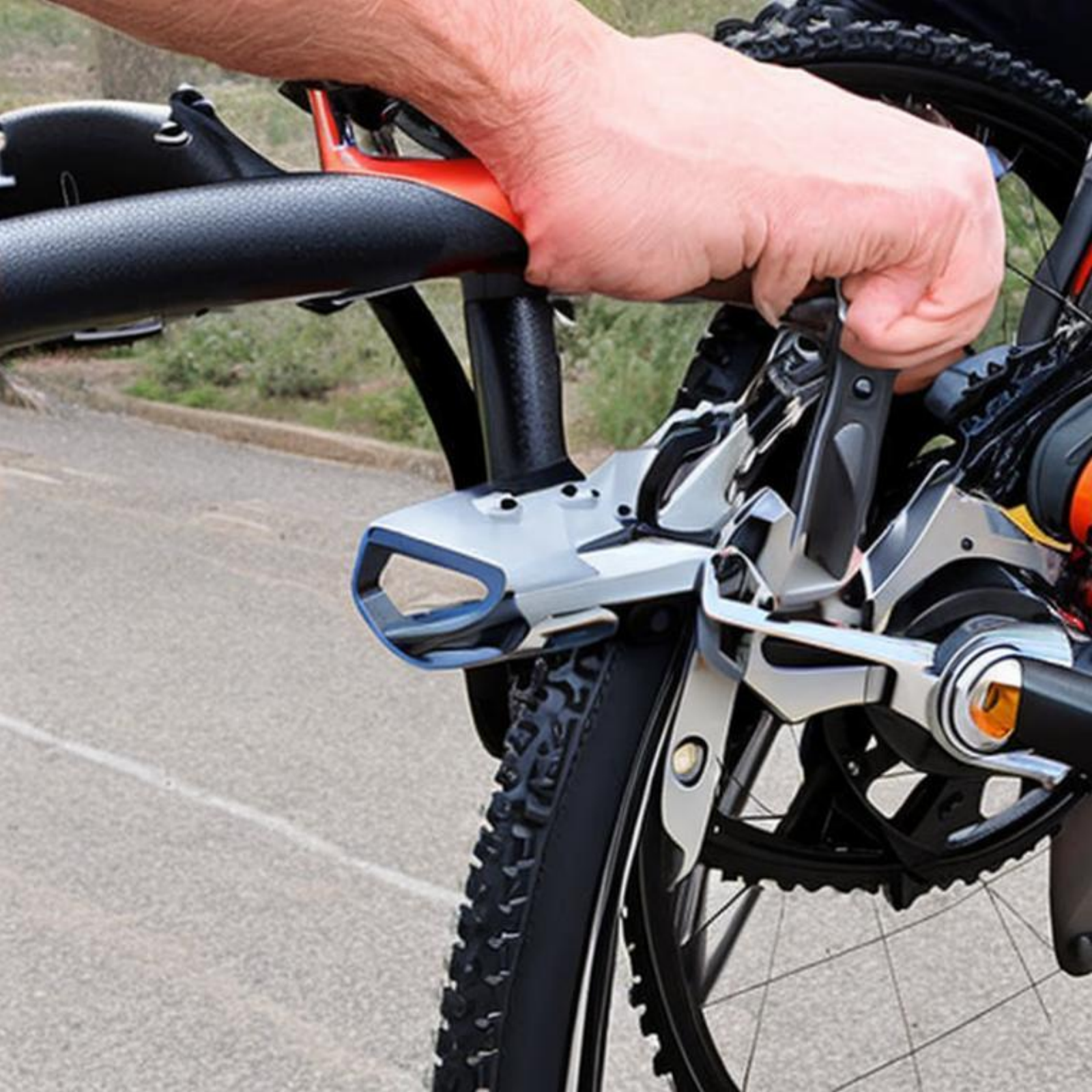} & \includegraphics[width=\linewidth, valign=m]{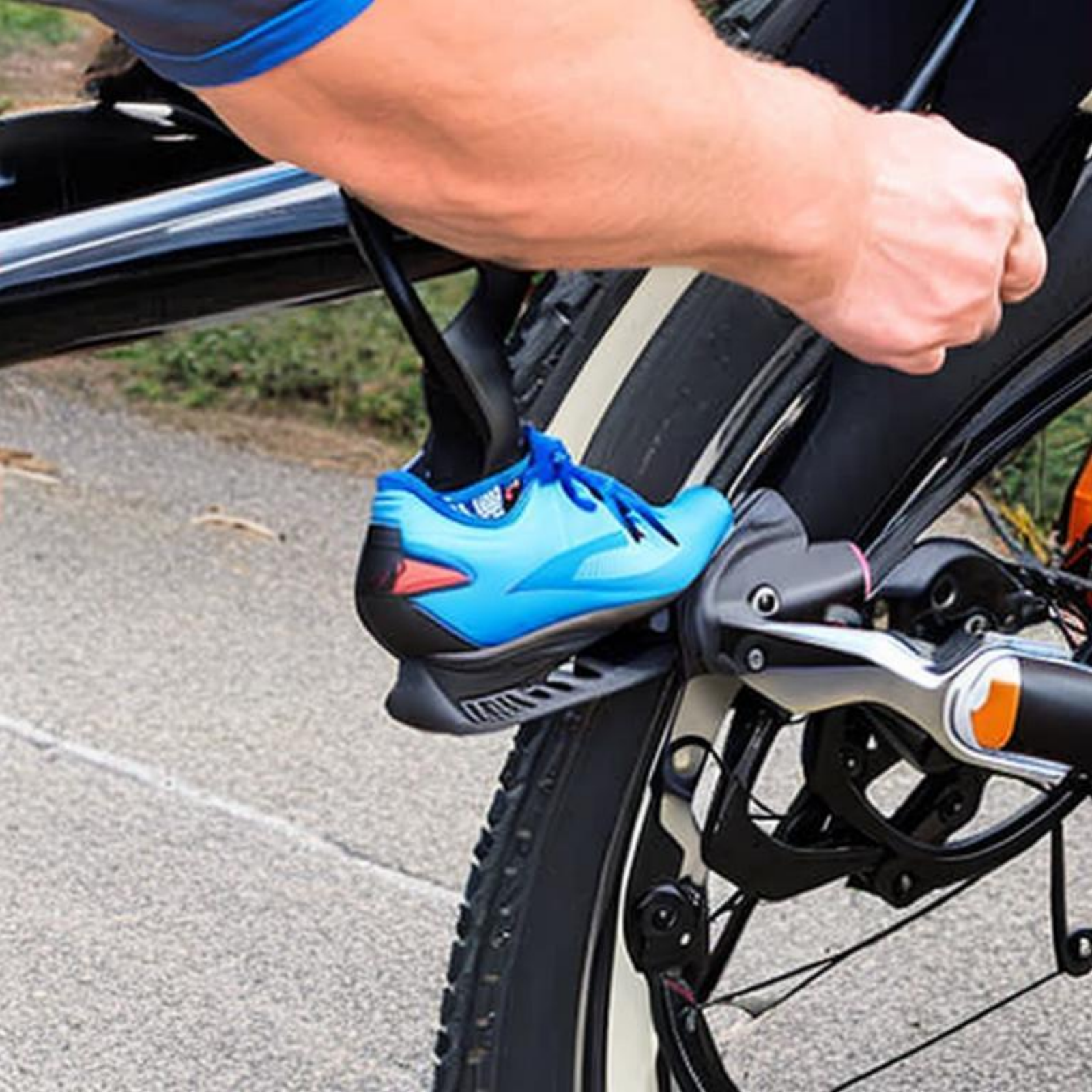} & \includegraphics[width=\linewidth, valign=m]{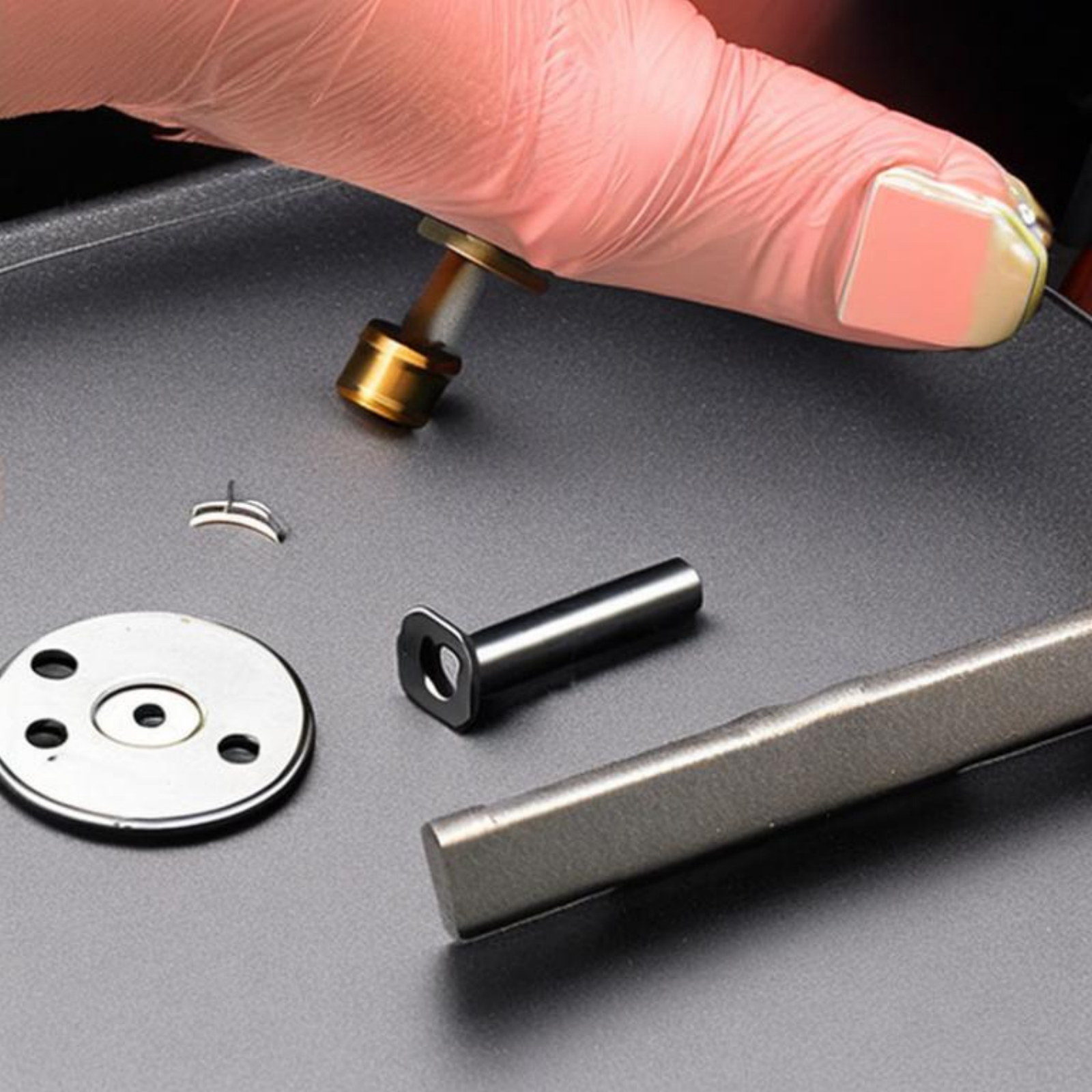} \\
\end{tabular}
  \caption{\textbf{Additional qualitative comparison on ShowHowTo.}}
  \label{fig:sup_showhow_1029}
\end{figure*}

\begin{figure*}[t]
  \centering
  \setlength{\tabcolsep}{3pt}
  \renewcommand{\arraystretch}{1.2}
  \begin{tabular}{>{\centering\arraybackslash}m{1.5cm} m{0.133\linewidth} m{0.133\linewidth} m{0.133\linewidth} m{0.133\linewidth} m{0.133\linewidth} m{0.133\linewidth}}
     & \tiny\centering Make the slaw by combining shredded cabbage, diced radish, shredded carrot, finely chopped celery, and green onion. & \tiny\centering Add your favorite ranch or blue cheese dressing to the slaw. & \tiny\centering Make the spicy chicken by shredding grilled chicken breast. & \tiny\centering Add wing sauce made by mixing hot sauce with melted butter or margarine to the chicken. & \tiny\centering Make the wrap by layering chicken, slaw, avocado, and bacon in a soft flour tortilla. & \tiny Roll up the wrap and cut it. \\
    \rotatebox[origin=c]{90}{\scriptsize Stable Diffusion} & \includegraphics[width=\linewidth, valign=m]{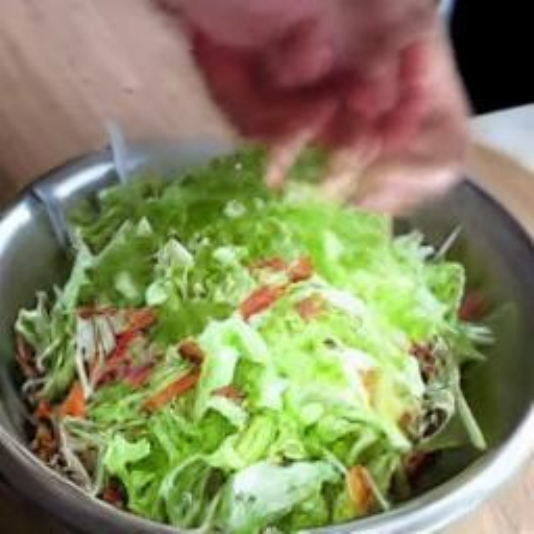} & \includegraphics[width=\linewidth, valign=m]{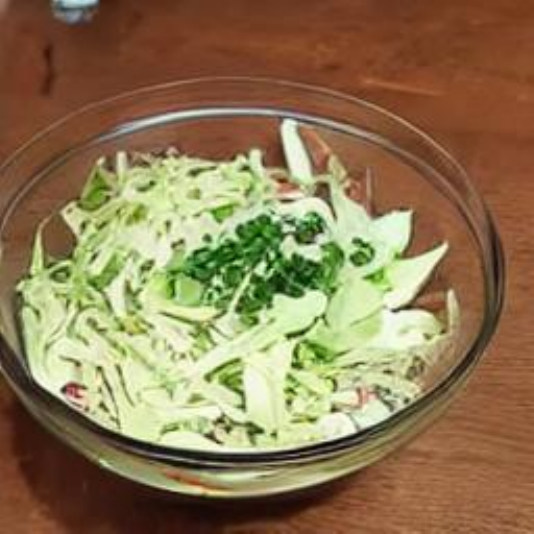} & \includegraphics[width=\linewidth, valign=m]{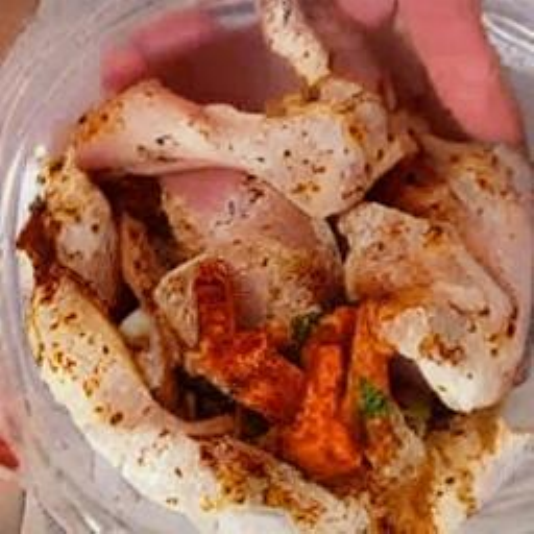} & \includegraphics[width=\linewidth, valign=m]{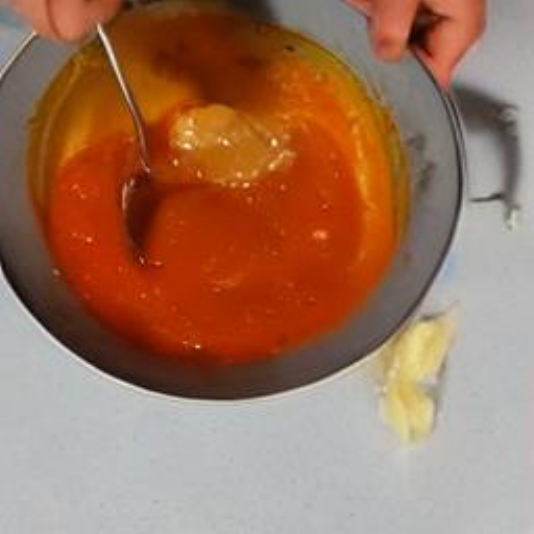} & \includegraphics[width=\linewidth, valign=m]{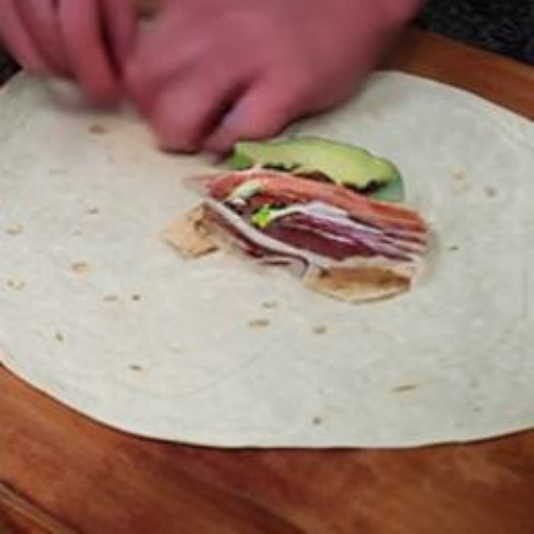} & \includegraphics[width=\linewidth, valign=m]{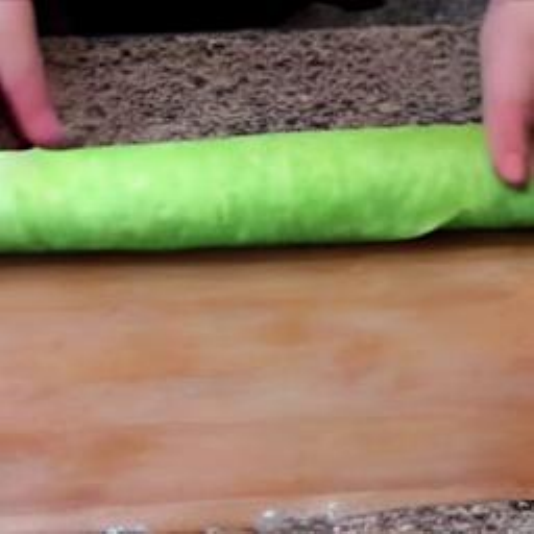} \\ \noalign{\vspace{10pt}}
    \rotatebox[origin=c]{90}{\scriptsize StackedDiffusion} & \includegraphics[width=\linewidth, valign=m]{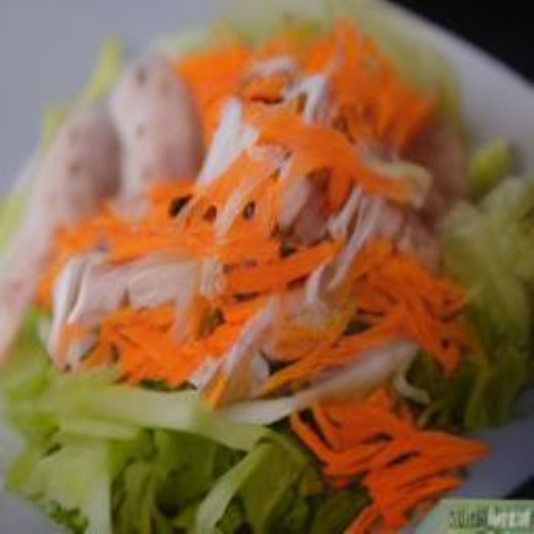} & \includegraphics[width=\linewidth, valign=m]{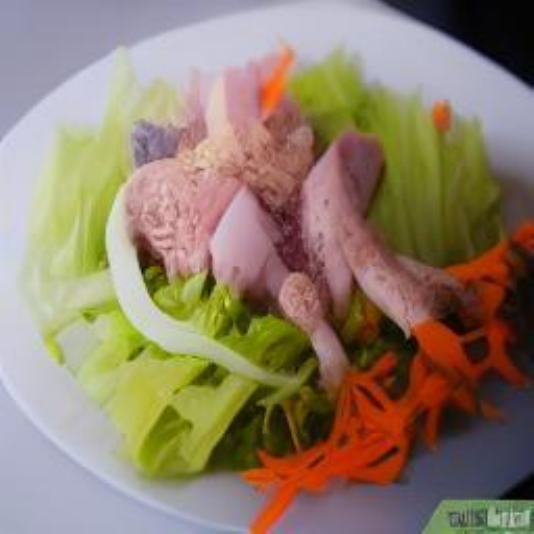} & \includegraphics[width=\linewidth, valign=m]{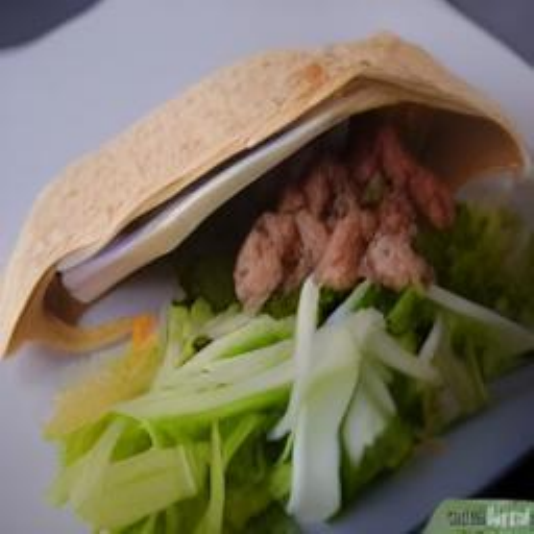} & \includegraphics[width=\linewidth, valign=m]{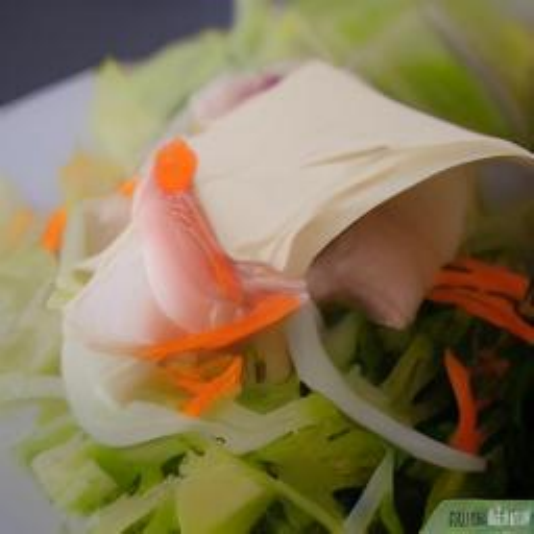} & \includegraphics[width=\linewidth, valign=m]{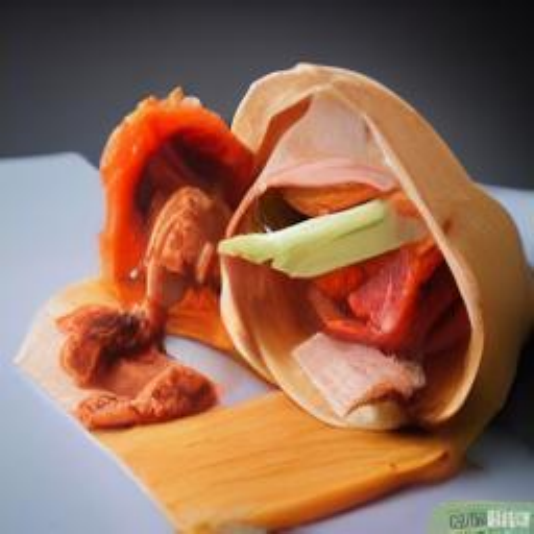} & \includegraphics[width=\linewidth, valign=m]{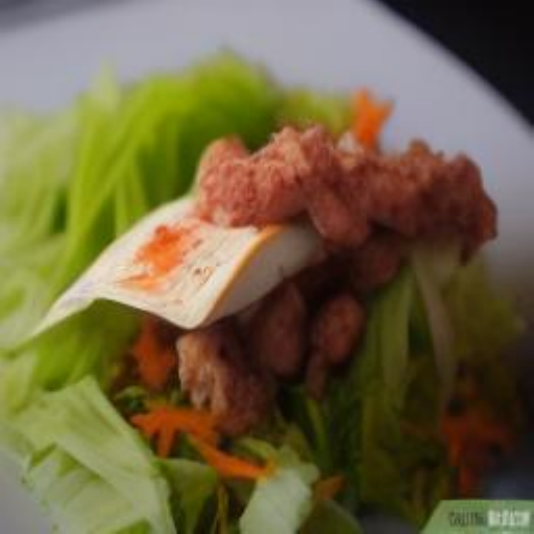} \\
\end{tabular}
  \caption{\textbf{Additional qualitative comparison on ShowHowTo.}}
  \label{fig:sup_showhow_1075}
\end{figure*}

\begin{figure*}[t]
  \centering
  \setlength{\tabcolsep}{3pt}
  \renewcommand{\arraystretch}{1.2}
  \begin{tabular}{>{\centering\arraybackslash}m{1.5cm} m{0.100\linewidth} m{0.100\linewidth} m{0.100\linewidth} m{0.100\linewidth} m{0.100\linewidth} m{0.100\linewidth} m{0.100\linewidth} m{0.100\linewidth}}
     & \tiny\centering Chain an odd number of chains. & \tiny\centering Single crochet in the second chain from hook, then chain one and skip the next chain. & \tiny\centering Chain one, skip this chain, and single crochet in the next chain. & \tiny\centering Chain one, skip this chain, and single crochet in the next chain. & \tiny\centering Chain one, skip this chain, and single crochet in the next chain. & \tiny\centering When you get down to the very last one, instead of chaining one, you just do one more single crochet in the end. & \tiny\centering Chain one and turn, and you turn like the page of a book. & \tiny Single crochet in the first single crochet, then chain one and you skip the single and single into the chain one space chain one single in the chain one space chain one. \\
    \rotatebox[origin=c]{90}{\scriptsize Stable Diffusion} & \includegraphics[width=\linewidth, valign=m]{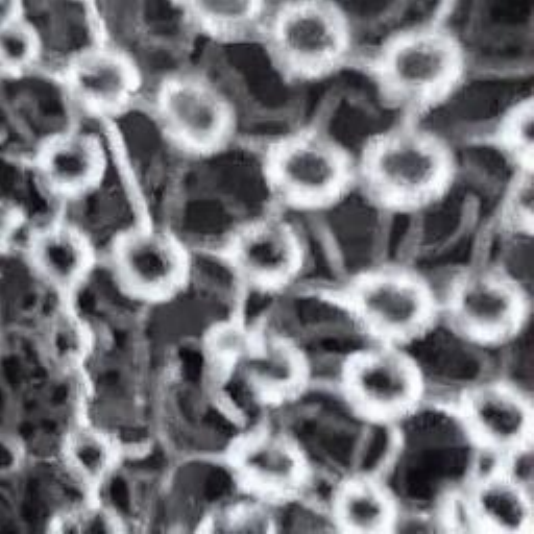} & \includegraphics[width=\linewidth, valign=m]{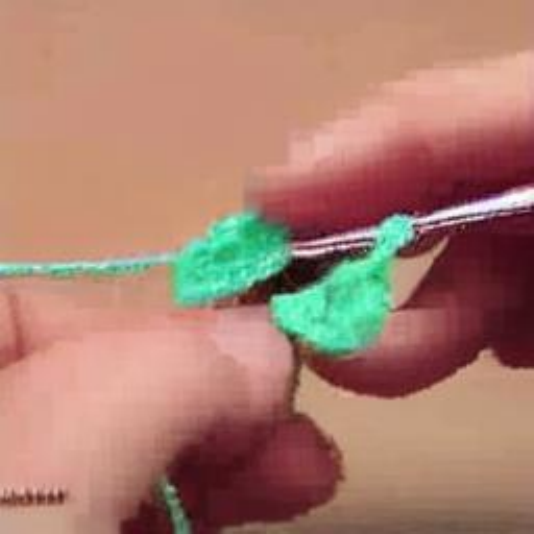} & \includegraphics[width=\linewidth, valign=m]{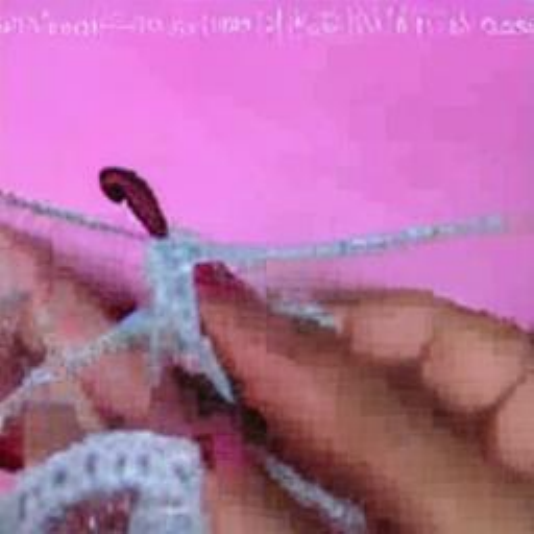} & \includegraphics[width=\linewidth, valign=m]{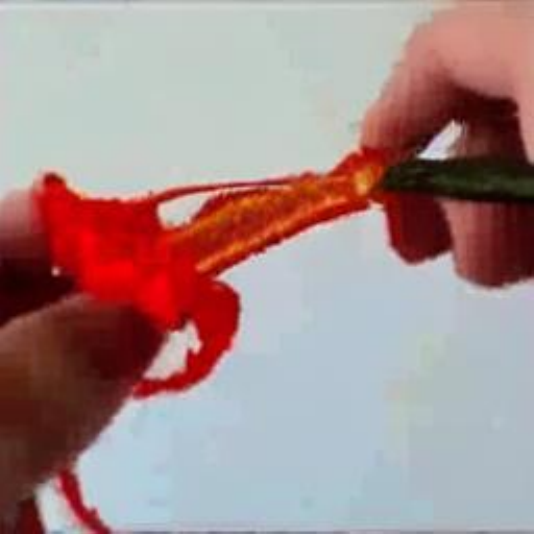} & \includegraphics[width=\linewidth, valign=m]{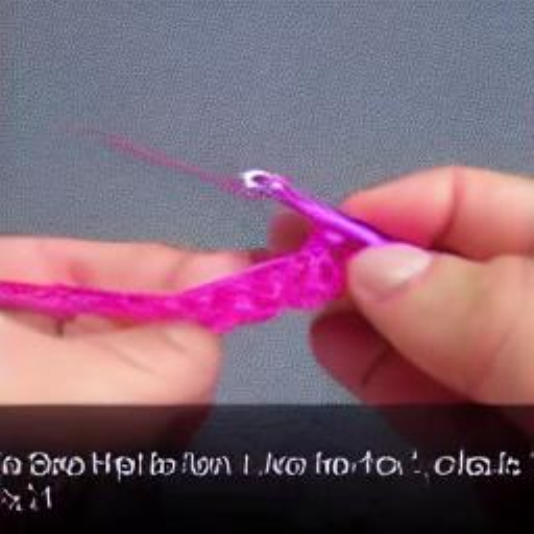} & \includegraphics[width=\linewidth, valign=m]{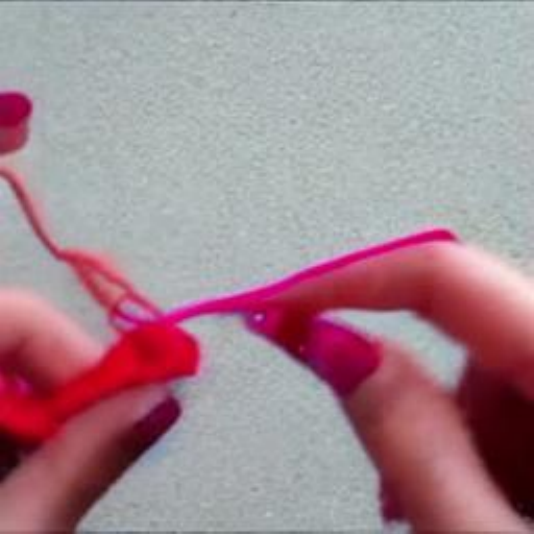} & \includegraphics[width=\linewidth, valign=m]{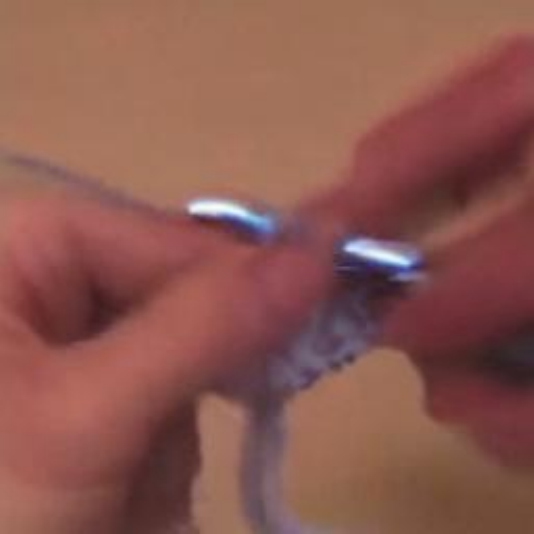} & \includegraphics[width=\linewidth, valign=m]{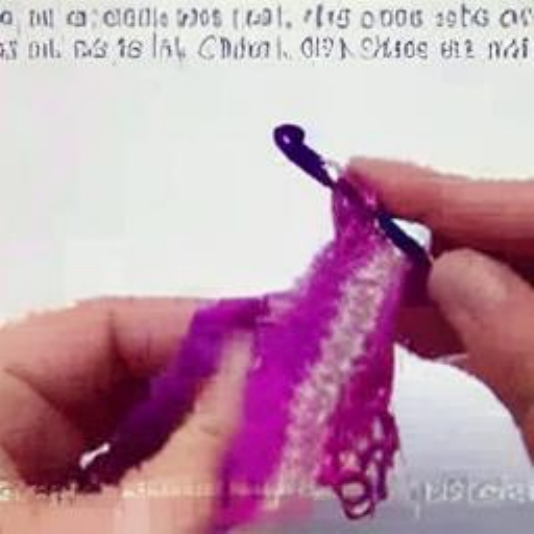} \\
\end{tabular}
  \caption{\textbf{Additional qualitative comparison on ShowHowTo.}}
  \label{fig:sup_showhow_1583}
\end{figure*}

\begin{figure*}[t]
  \centering
  \setlength{\tabcolsep}{3pt}
  \renewcommand{\arraystretch}{1.2}
  \begin{tabular}{>{\centering\arraybackslash}m{1.5cm} m{0.115\linewidth} m{0.115\linewidth} m{0.115\linewidth} m{0.115\linewidth} m{0.115\linewidth} m{0.115\linewidth} m{0.115\linewidth}}
     & \tiny\centering Check the primer bulb for any cracks or damage. & \tiny\centering Replace the primer bulb by pulling it out of the clip. & \tiny\centering Push the tabs down and pull on the primer bulb to remove it from the clip. & \tiny\centering Install the new primer bulb by pushing it into the clip and ensuring it is seated properly. & \tiny\centering Push the primer bulb back into place and ensure the tabs are aligned properly. & \tiny\centering Check the primer bulb to ensure it is functioning properly. & \tiny Check the carburetor to ensure the primer bulb is working properly. \\
    \rotatebox[origin=c]{90}{\scriptsize Stable Diffusion} & \includegraphics[width=\linewidth, valign=m]{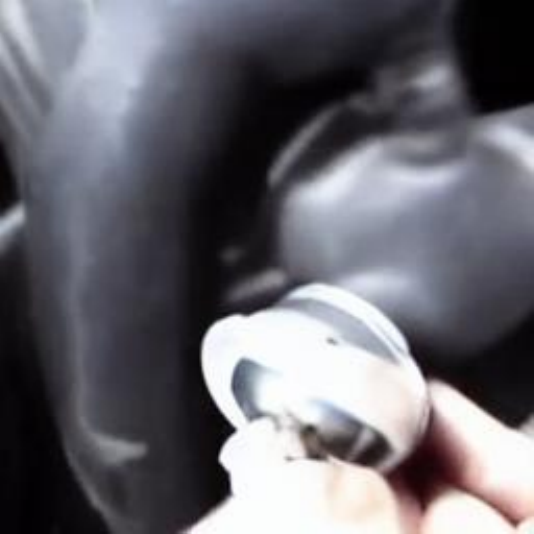} & \includegraphics[width=\linewidth, valign=m]{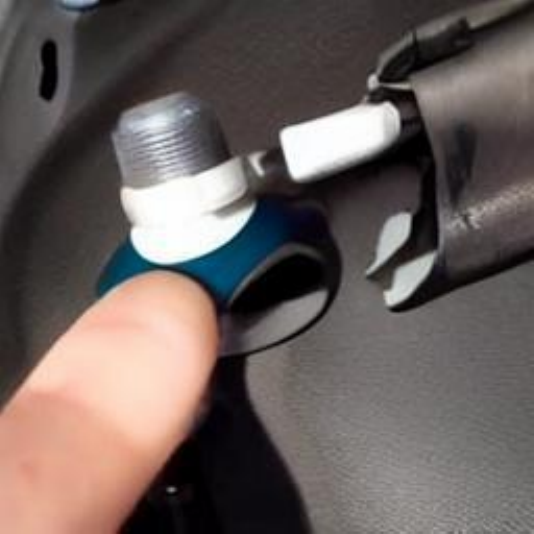} & \includegraphics[width=\linewidth, valign=m]{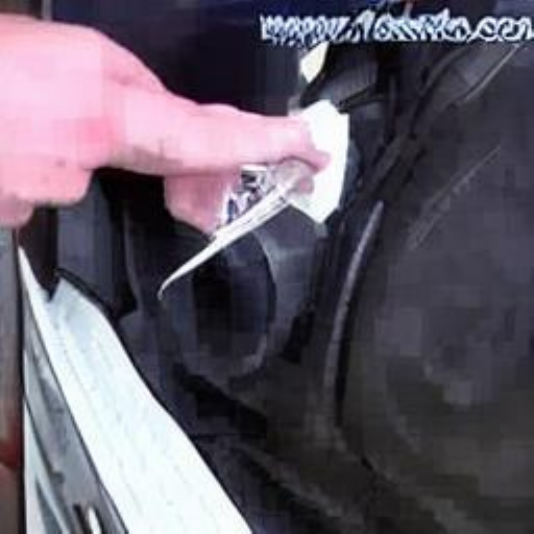} & \includegraphics[width=\linewidth, valign=m]{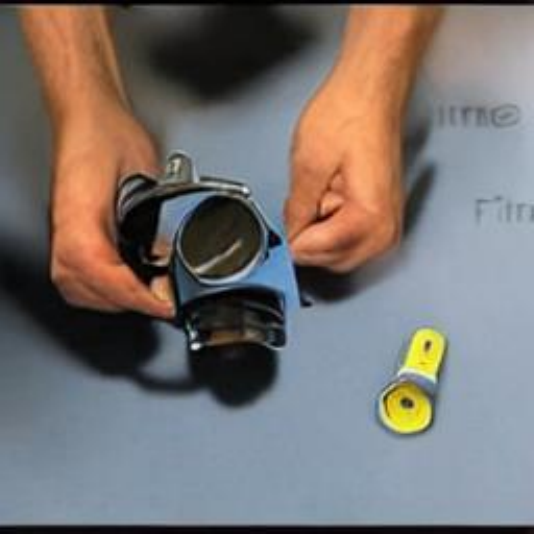} & \includegraphics[width=\linewidth, valign=m]{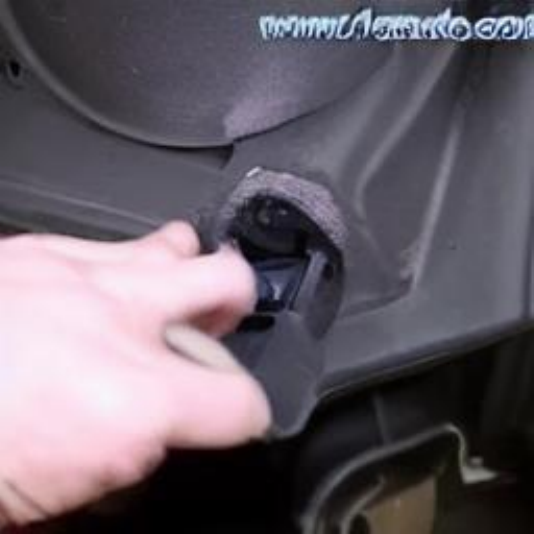} & \includegraphics[width=\linewidth, valign=m]{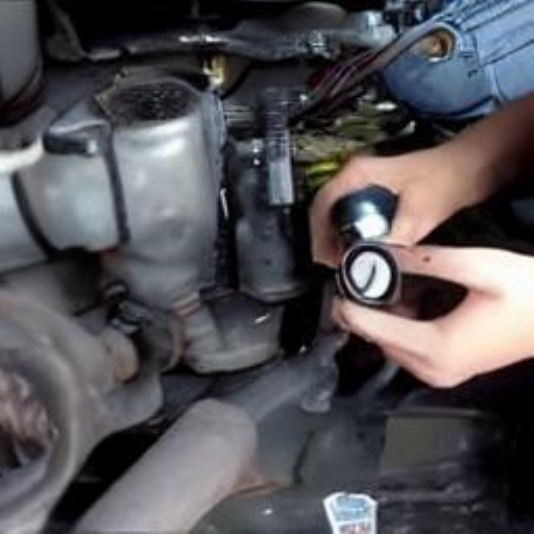} & \includegraphics[width=\linewidth, valign=m]{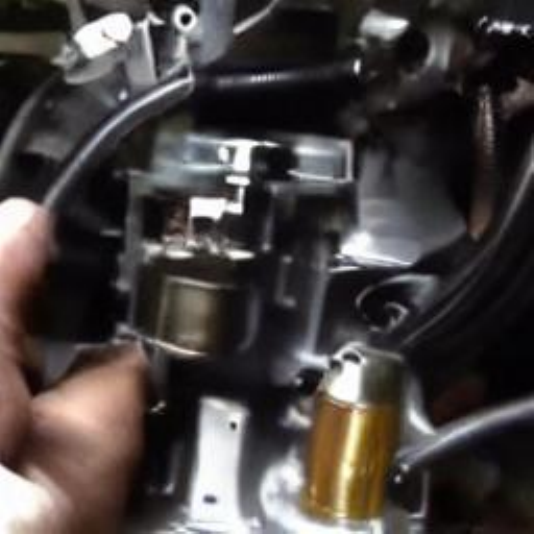} \\
\end{tabular}
  \caption{\textbf{Additional qualitative comparison on ShowHowTo.}}
  \label{fig:sup_showhow_150}
\end{figure*}

\begin{figure*}[t]
  \centering
  \setlength{\tabcolsep}{3pt}
  \renewcommand{\arraystretch}{1.2}
  \begin{tabular}{>{\centering\arraybackslash}m{1.5cm} m{0.133\linewidth} m{0.133\linewidth} m{0.133\linewidth} m{0.133\linewidth} m{0.133\linewidth} m{0.133\linewidth}}
     & \tiny\centering Season the scallops with salt and cornflour. & \tiny\centering Prepare the tofu by cutting it into 8 pieces and mixing it with corn starch solution. & \tiny\centering Place the scallops on top of the tofu and steam them for 5-6 minutes. & \tiny\centering Prepare the sauce by frying ginger and garlic in oil, then adding water and scallop. & \tiny\centering Add oyster sauce, sugar, and sesame seed oil to the sauce, and thicken it with cornstarch solution. & \tiny Pour the sauce over the scallop tofu and garnish with Chinese coriander leaf or parsley. \\
    \rotatebox[origin=c]{90}{\scriptsize Stable Diffusion} & \includegraphics[width=\linewidth, valign=m]{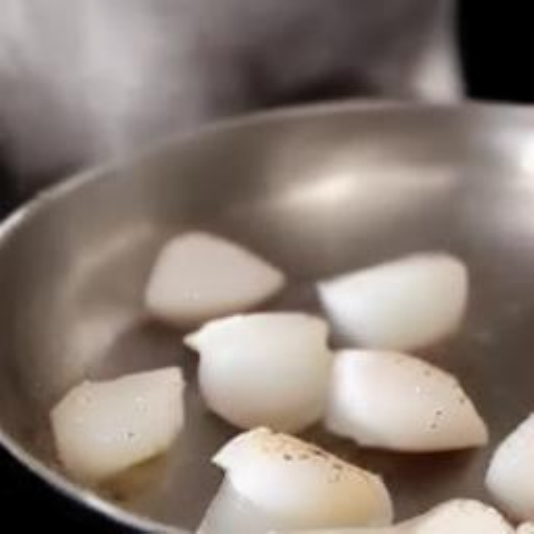} & \includegraphics[width=\linewidth, valign=m]{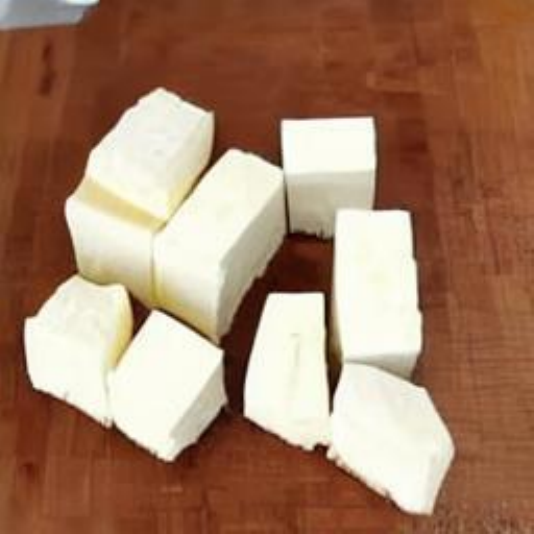} & \includegraphics[width=\linewidth, valign=m]{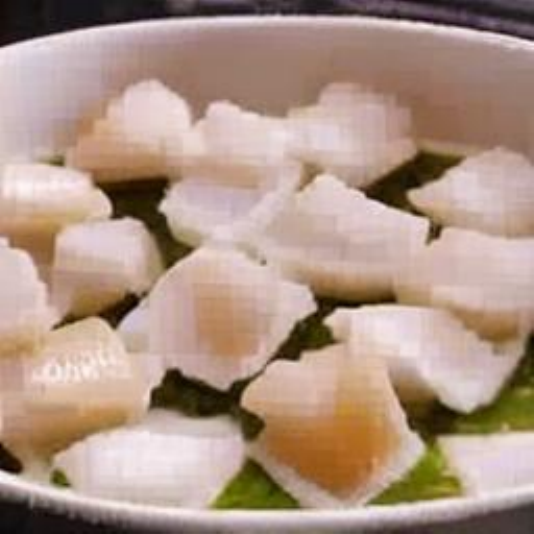} & \includegraphics[width=\linewidth, valign=m]{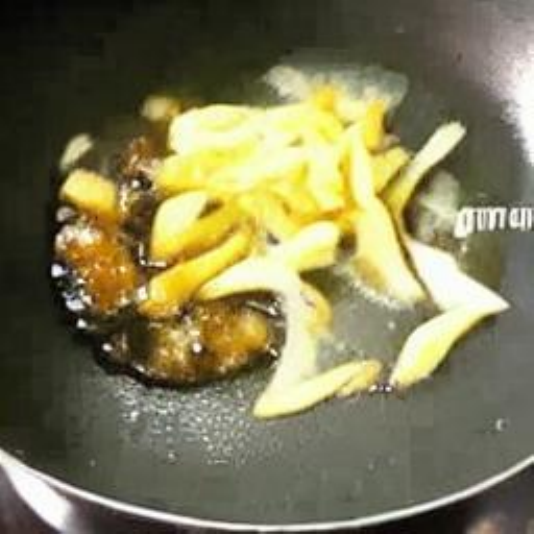} & \includegraphics[width=\linewidth, valign=m]{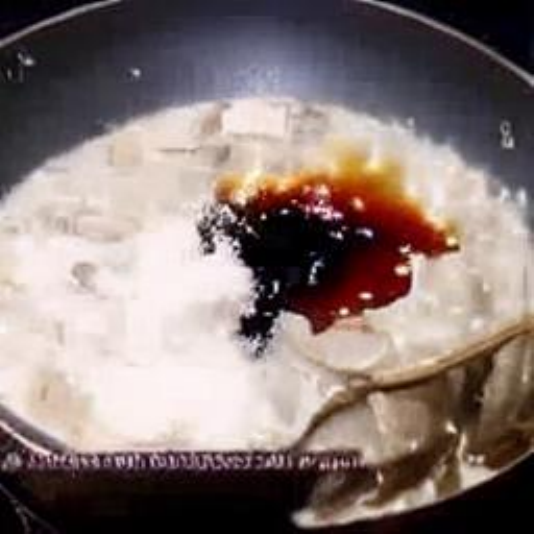} & \includegraphics[width=\linewidth, valign=m]{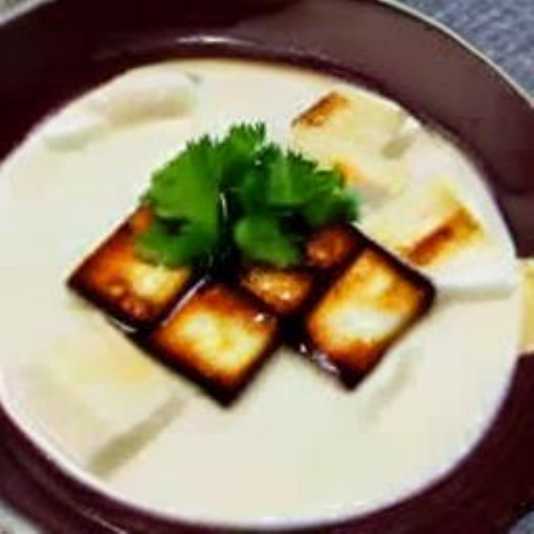} \\ \noalign{\vspace{10pt}}
    \rotatebox[origin=c]{90}{\scriptsize StackedDiffusion} & \includegraphics[width=\linewidth, valign=m]{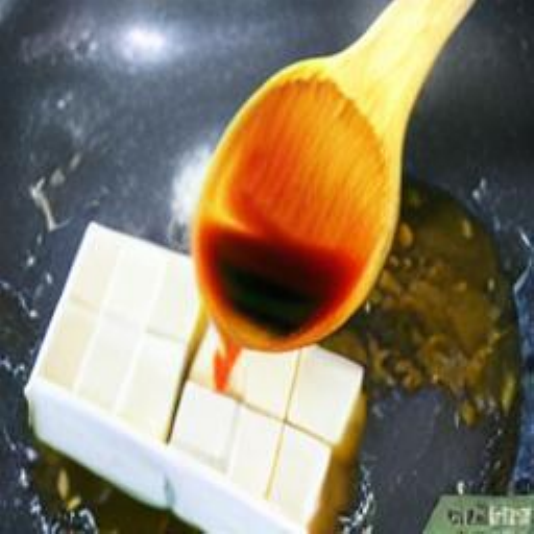} & \includegraphics[width=\linewidth, valign=m]{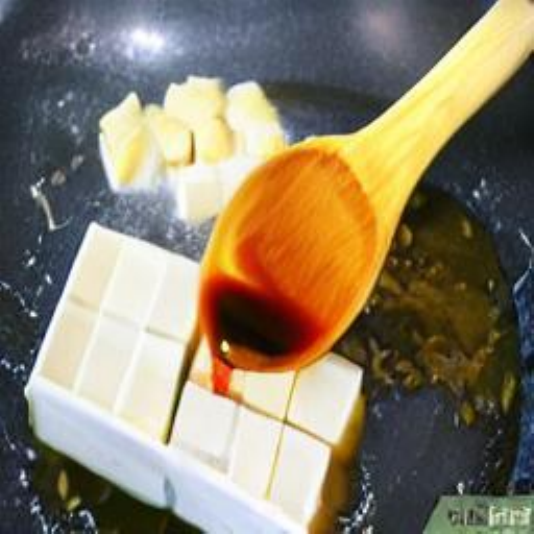} & \includegraphics[width=\linewidth, valign=m]{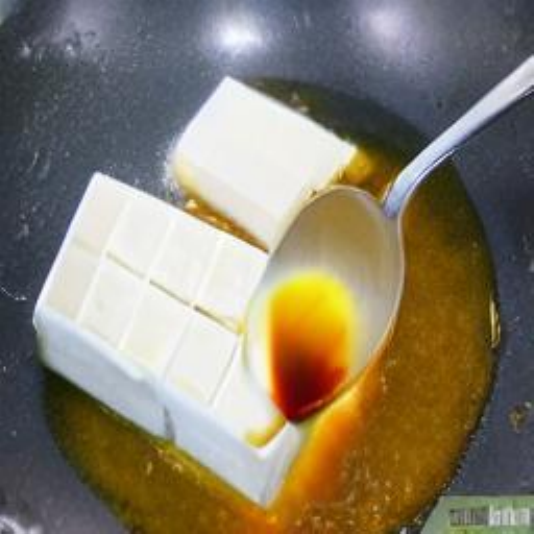} & \includegraphics[width=\linewidth, valign=m]{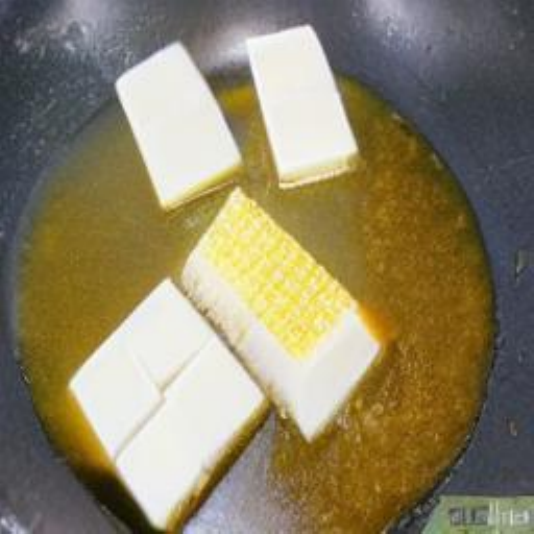} & \includegraphics[width=\linewidth, valign=m]{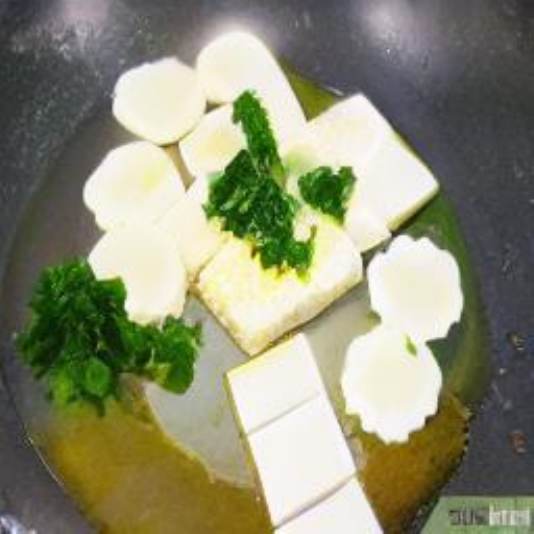} & \includegraphics[width=\linewidth, valign=m]{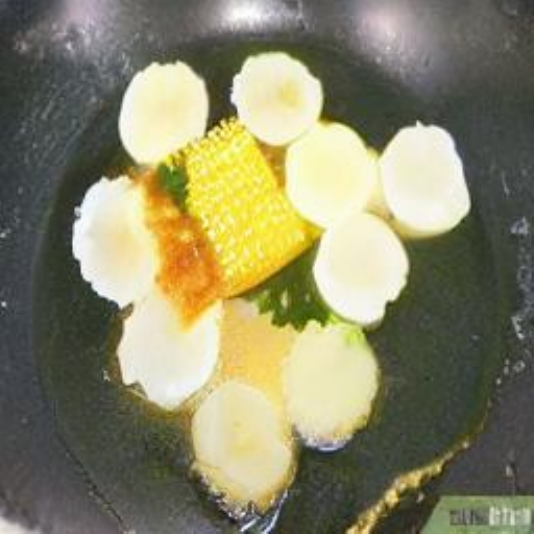} \\
\end{tabular}
  \caption{\textbf{Additional qualitative comparison on ShowHowTo.}}
  \label{fig:sup_showhow_326}
\end{figure*}

\begin{figure*}[t]
  \centering
  \setlength{\tabcolsep}{3pt}
  \renewcommand{\arraystretch}{1.2}
  \begin{tabular}{>{\centering\arraybackslash}m{1.5cm} m{0.100\linewidth} m{0.100\linewidth} m{0.100\linewidth} m{0.100\linewidth} m{0.100\linewidth} m{0.100\linewidth} m{0.100\linewidth} m{0.100\linewidth}}
     & \tiny\centering Chop 1 cup of pork into bite-sized pieces. & \tiny\centering Marinate the pork for about 10 minutes. & \tiny\centering Obtain about 2 generous cups of well-fermented kimchi and chop it into half-inch pieces. & \tiny\centering Slice half a pack of tofu into quarter-inch slices. & \tiny\centering Cut 1/2 cup worth of green onions into 1/2 inch length. & \tiny\centering Cut 1/8 cup worth of red hot pepper thinly. & \tiny\centering Fry the marinated pork with 1/2 tbsp of sesame oil in a heated pan on medium-high for about 3 minutes. & \tiny Add the chopped kimchi, one tablespoon of hot pepper powder, and half tablespoon of cooking oil. \\
    \rotatebox[origin=c]{90}{\scriptsize Stable Diffusion} & \includegraphics[width=\linewidth, valign=m]{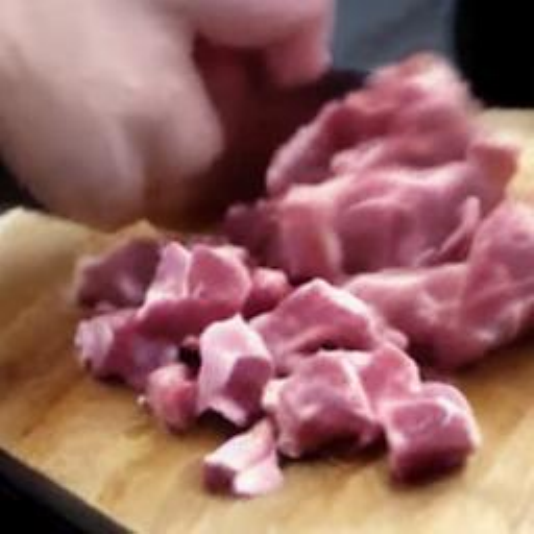} & \includegraphics[width=\linewidth, valign=m]{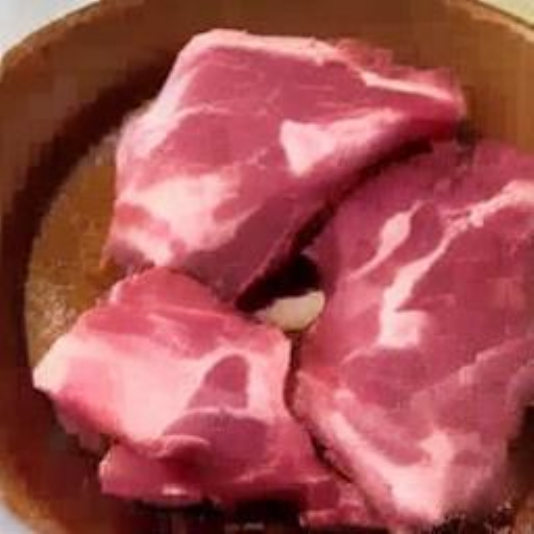} & \includegraphics[width=\linewidth, valign=m]{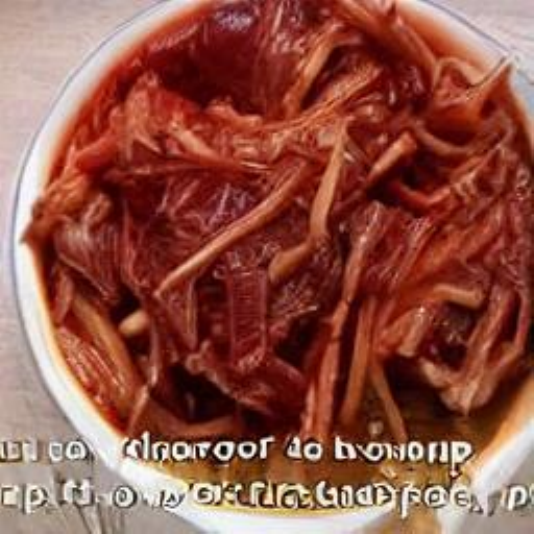} & \includegraphics[width=\linewidth, valign=m]{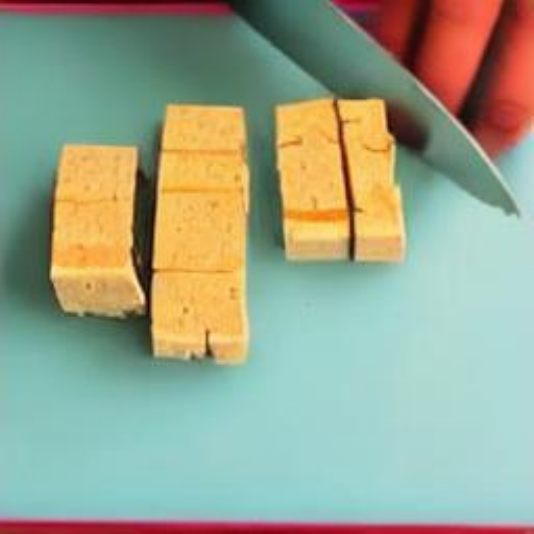} & \includegraphics[width=\linewidth, valign=m]{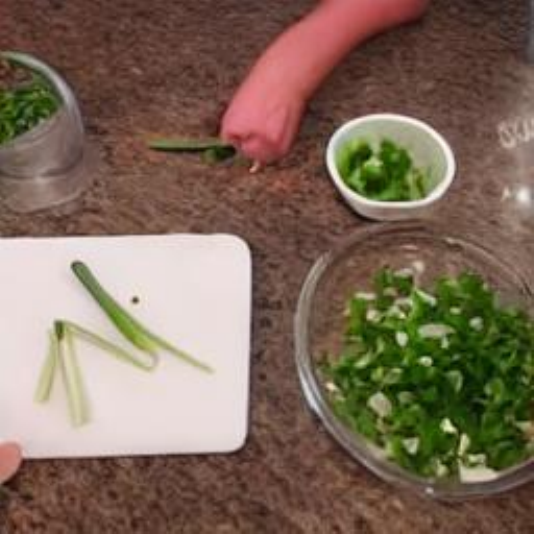} & \includegraphics[width=\linewidth, valign=m]{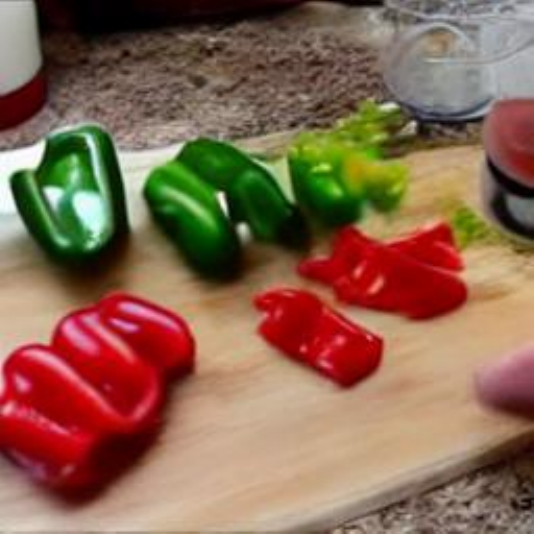} & \includegraphics[width=\linewidth, valign=m]{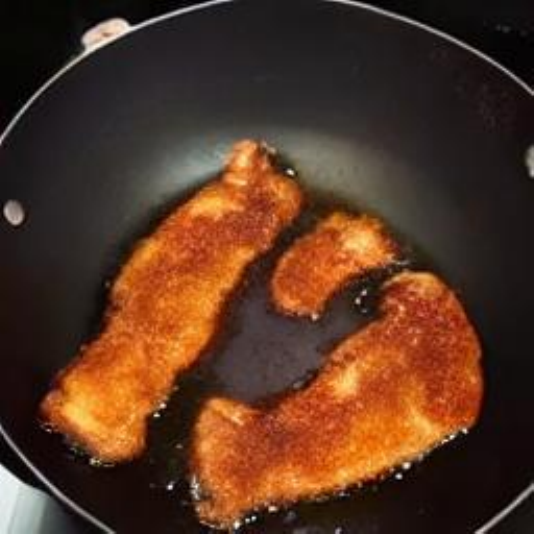} & \includegraphics[width=\linewidth, valign=m]{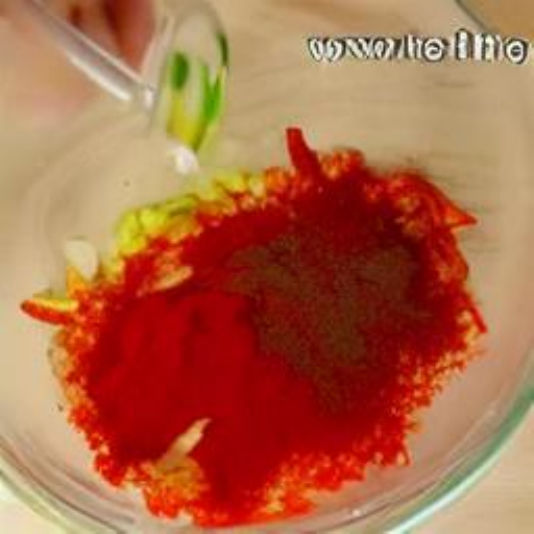} \\
\end{tabular}
  \caption{\textbf{Additional qualitative comparison on ShowHowTo.}}
  \label{fig:sup_showhow_680}
\end{figure*}

\begin{figure*}[t]
  \centering
  \setlength{\tabcolsep}{3pt}
  \renewcommand{\arraystretch}{1.2}
  \begin{tabular}{>{\centering\arraybackslash}m{1.5cm} m{0.160\linewidth} m{0.160\linewidth} m{0.160\linewidth} m{0.160\linewidth} m{0.160\linewidth}}
     & \tiny\centering Whip heavy cream for serving on top of the milkshake. & \tiny\centering Decorate the glass with peanut butter and Nutella. & \tiny\centering Add vanilla ice cream, peanut butter, and Nutella to the blender cup. & \tiny\centering Add milk to the blender cup and blend until a thick milkshake forms. & \tiny Enjoy the milkshake. \\
    \rotatebox[origin=c]{90}{\scriptsize Stable Diffusion} & \includegraphics[width=\linewidth, valign=m]{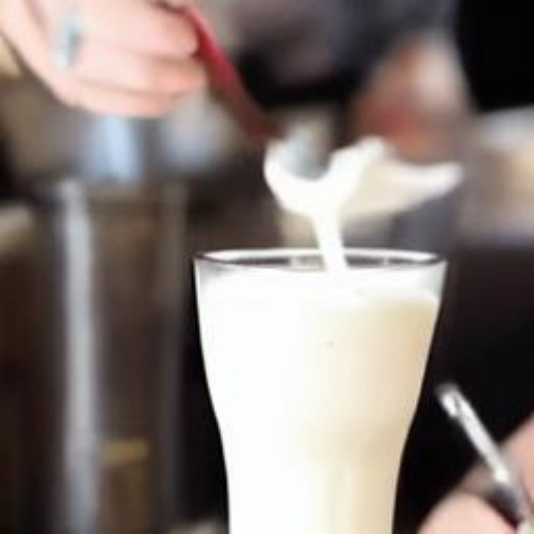} & \includegraphics[width=\linewidth, valign=m]{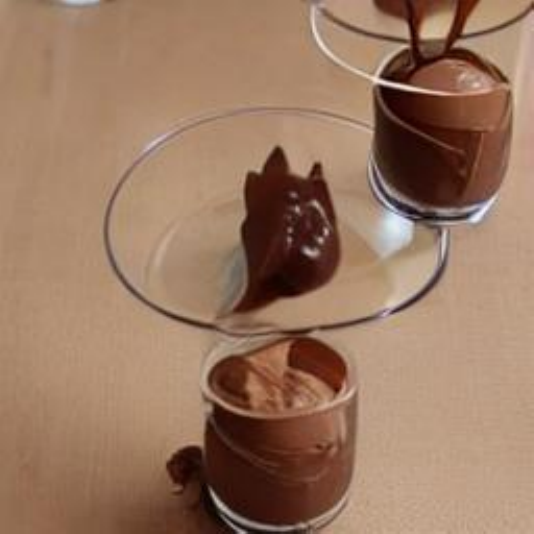} & \includegraphics[width=\linewidth, valign=m]{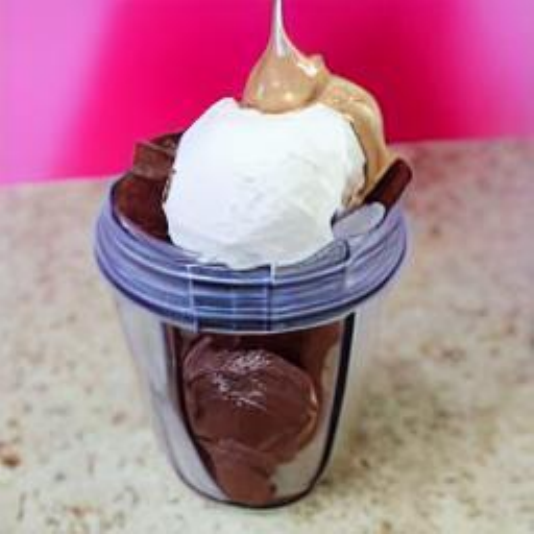} & \includegraphics[width=\linewidth, valign=m]{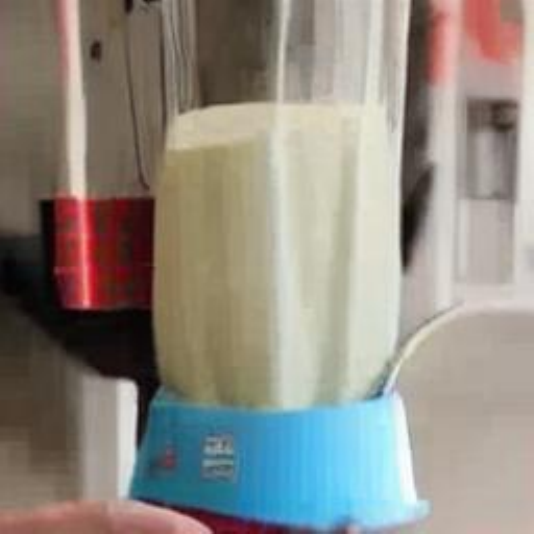} & \includegraphics[width=\linewidth, valign=m]{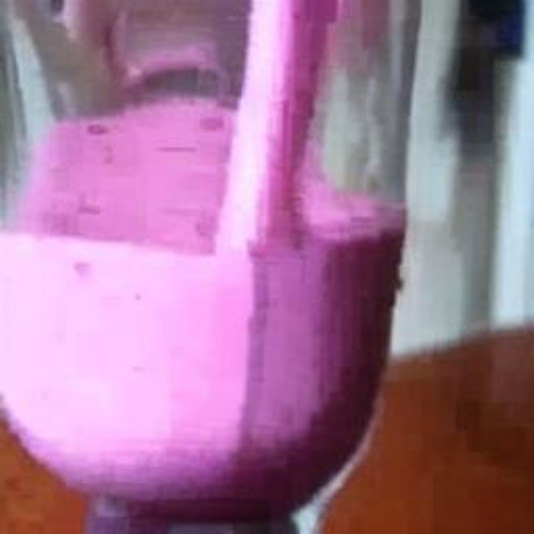} \\ \noalign{\vspace{10pt}}
    \rotatebox[origin=c]{90}{\scriptsize StackedDiffusion} & \includegraphics[width=\linewidth, valign=m]{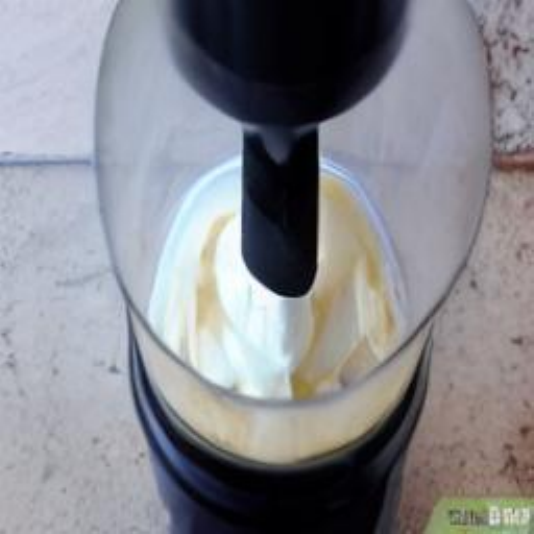} & \includegraphics[width=\linewidth, valign=m]{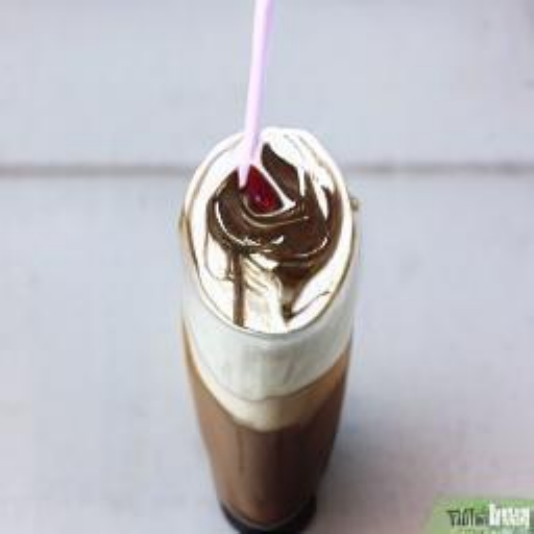} & \includegraphics[width=\linewidth, valign=m]{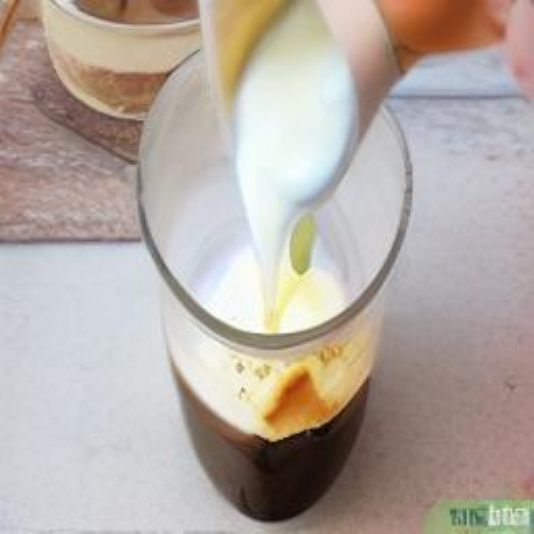} & \includegraphics[width=\linewidth, valign=m]{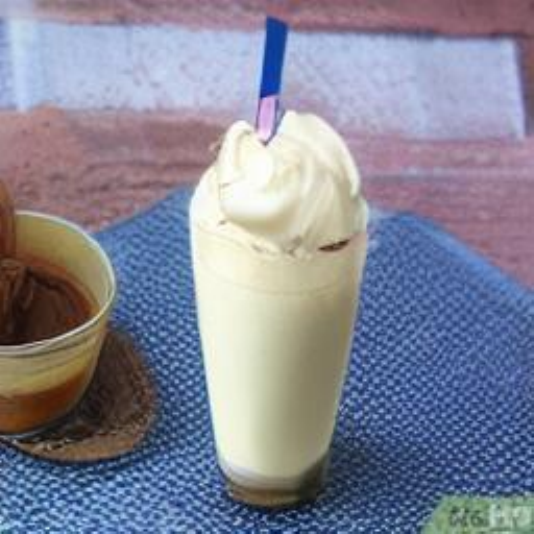} & \includegraphics[width=\linewidth, valign=m]{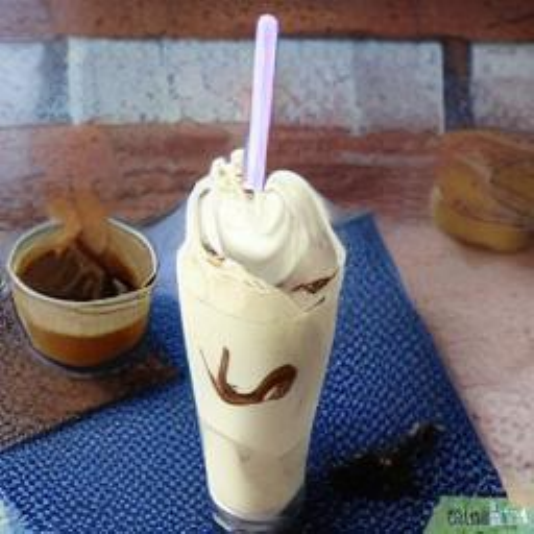} \\ \noalign{\vspace{10pt}}
    \rotatebox[origin=c]{90}{\scriptsize Bordalo et al.} & \includegraphics[width=\linewidth, valign=m]{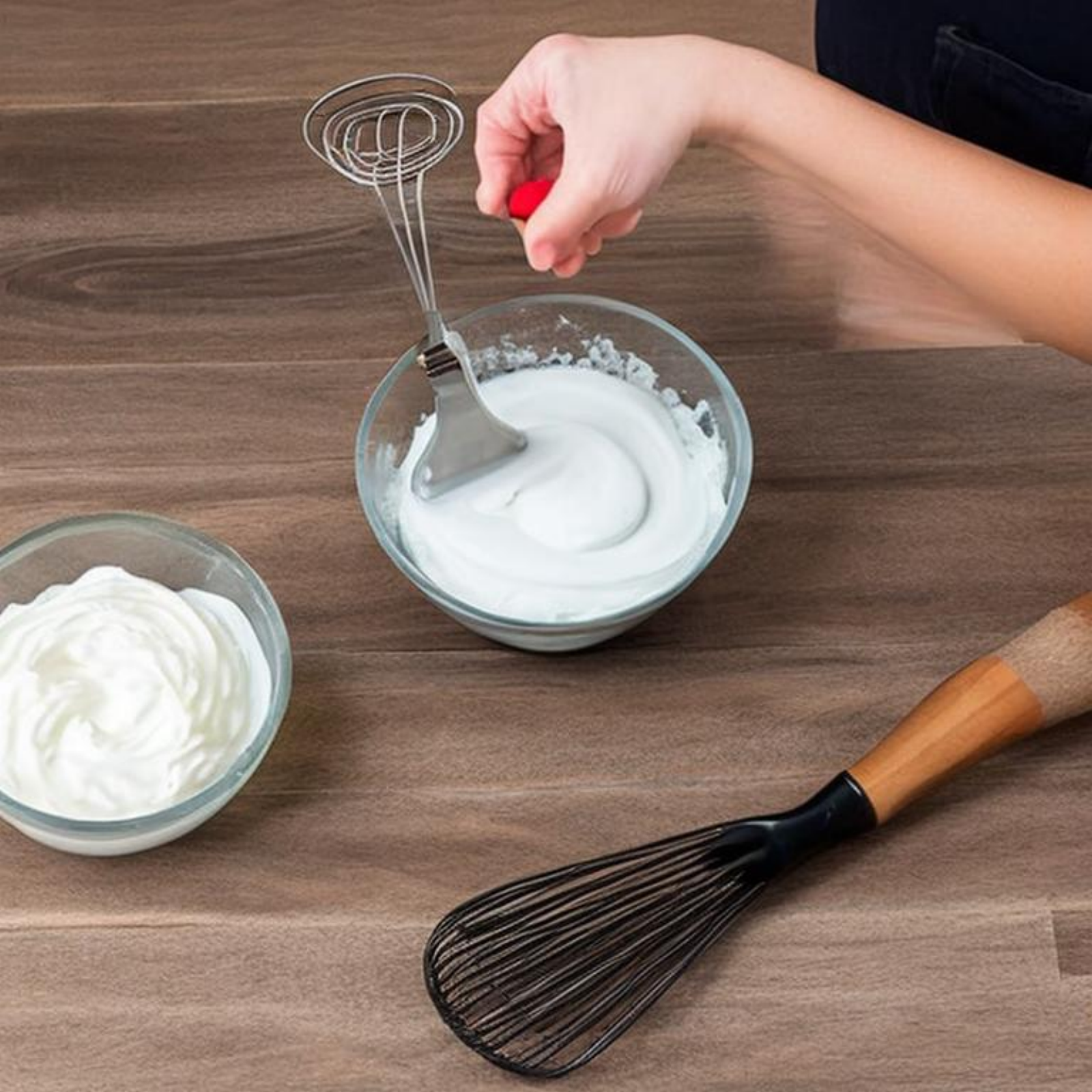} & \includegraphics[width=\linewidth, valign=m]{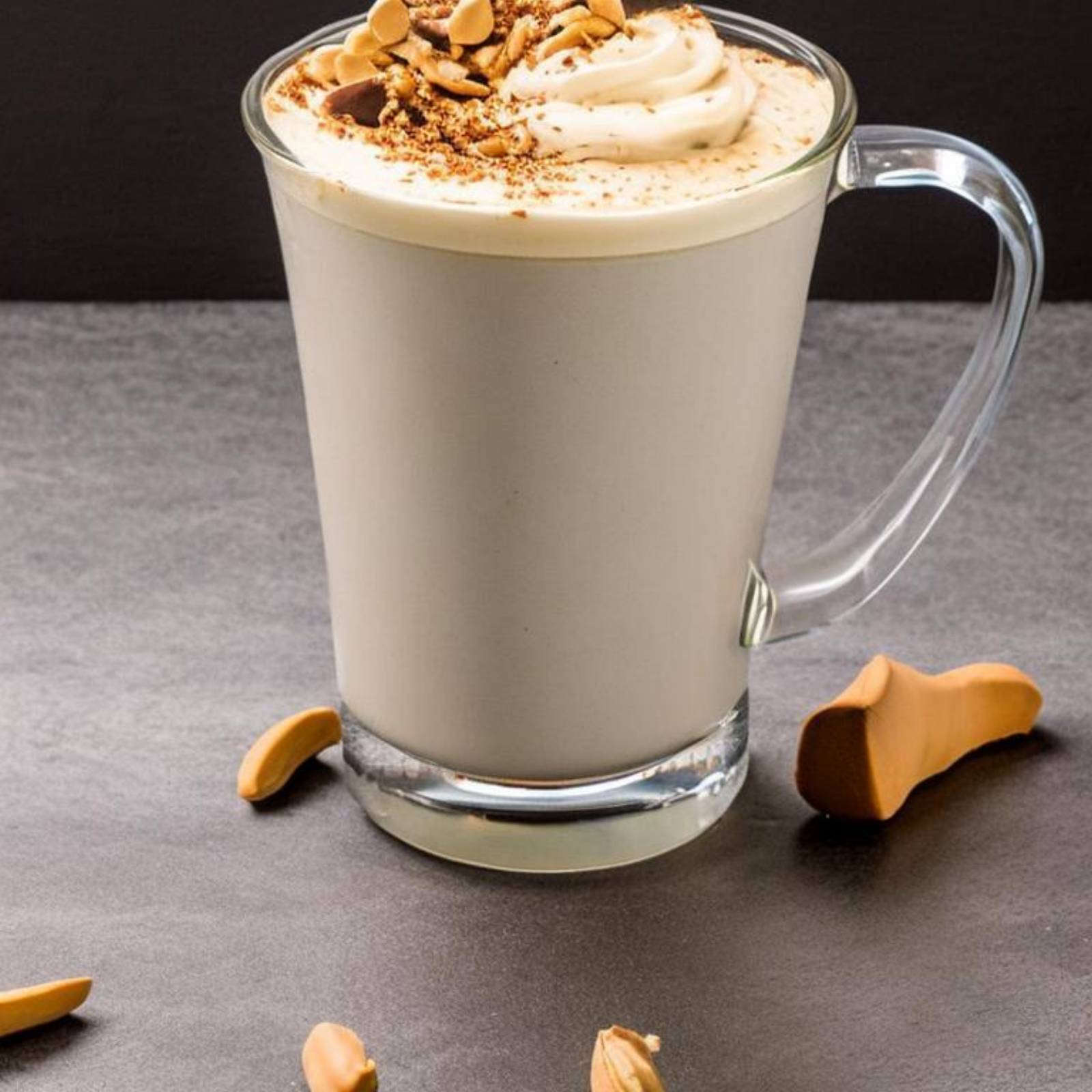} & \includegraphics[width=\linewidth, valign=m]{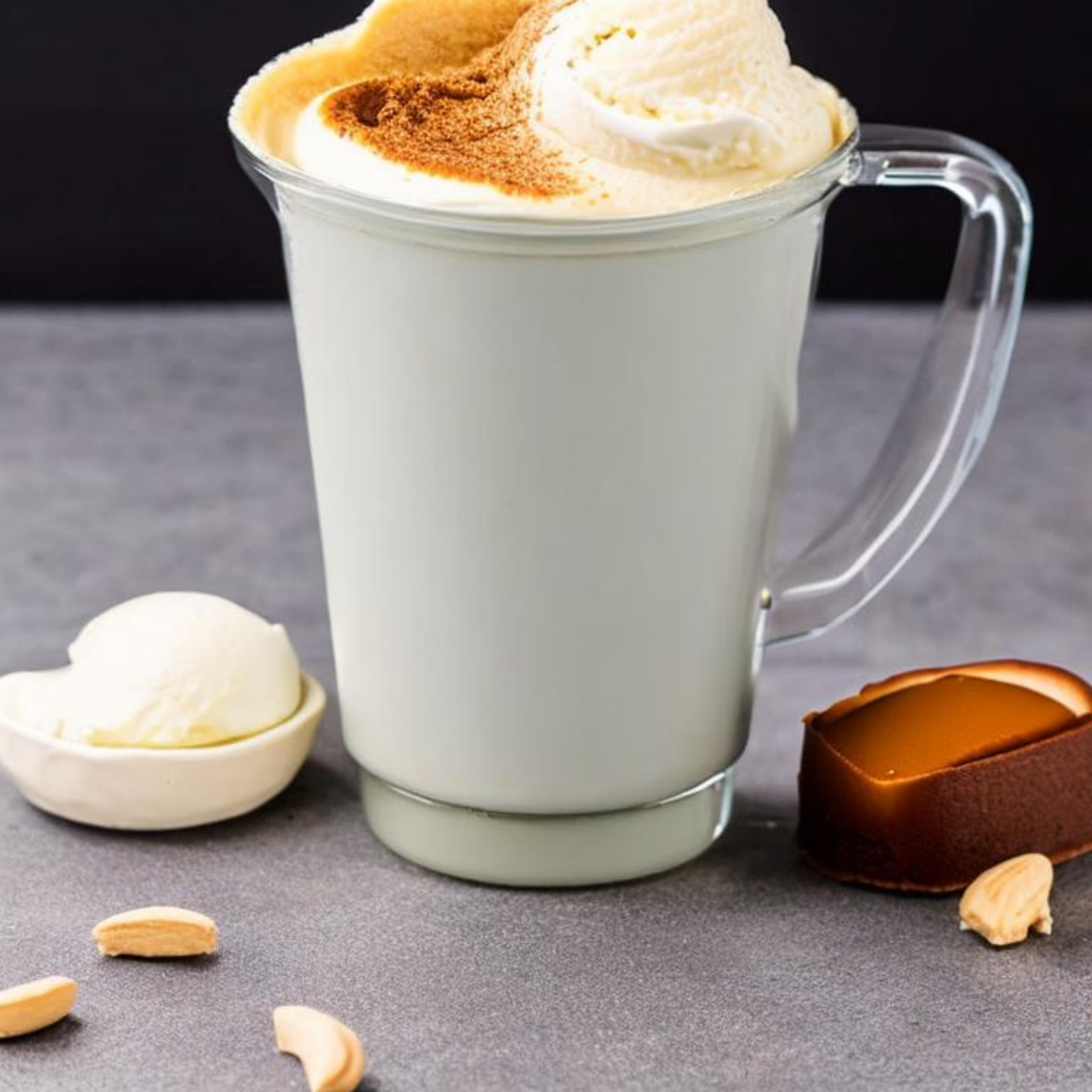} & \includegraphics[width=\linewidth, valign=m]{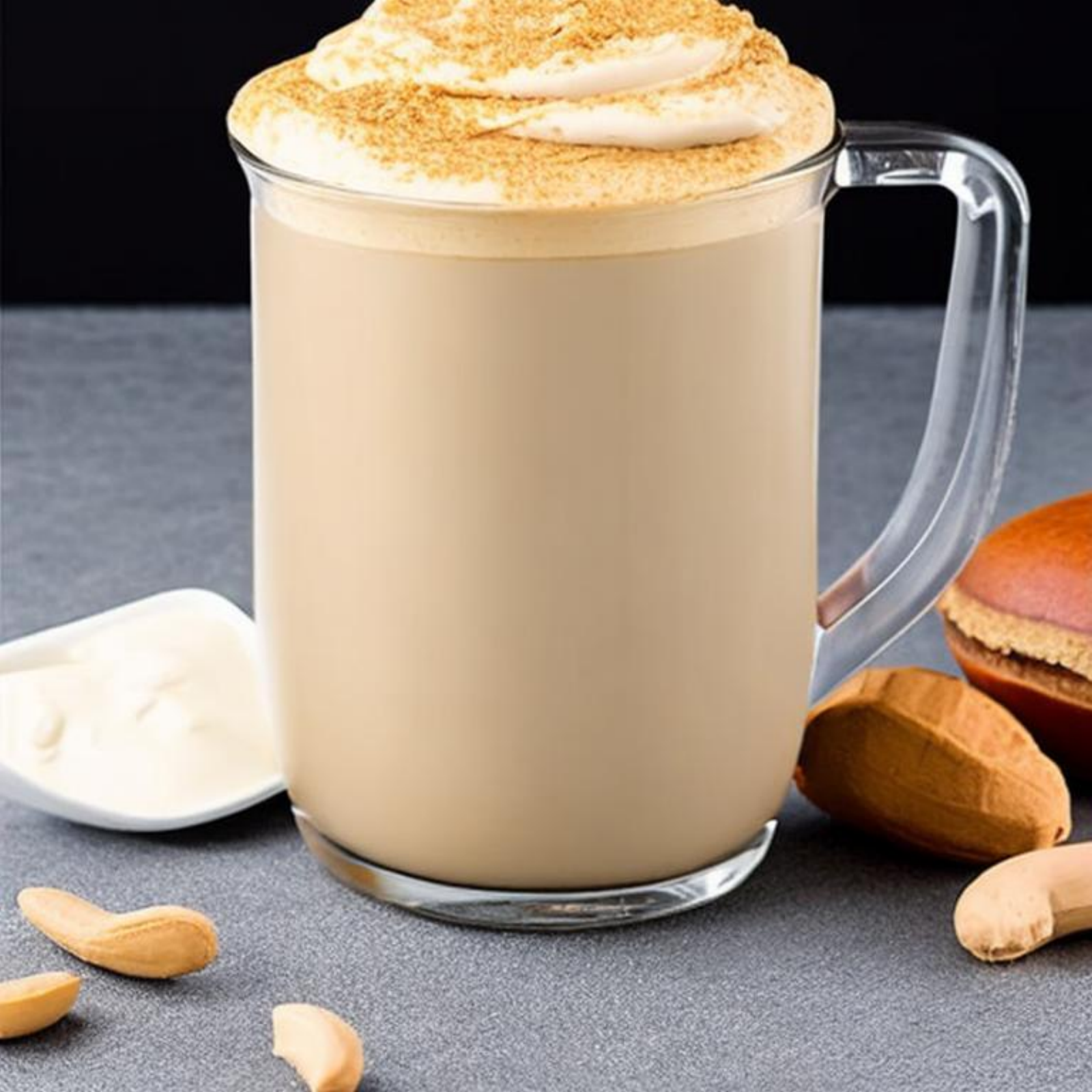} & \includegraphics[width=\linewidth, valign=m]{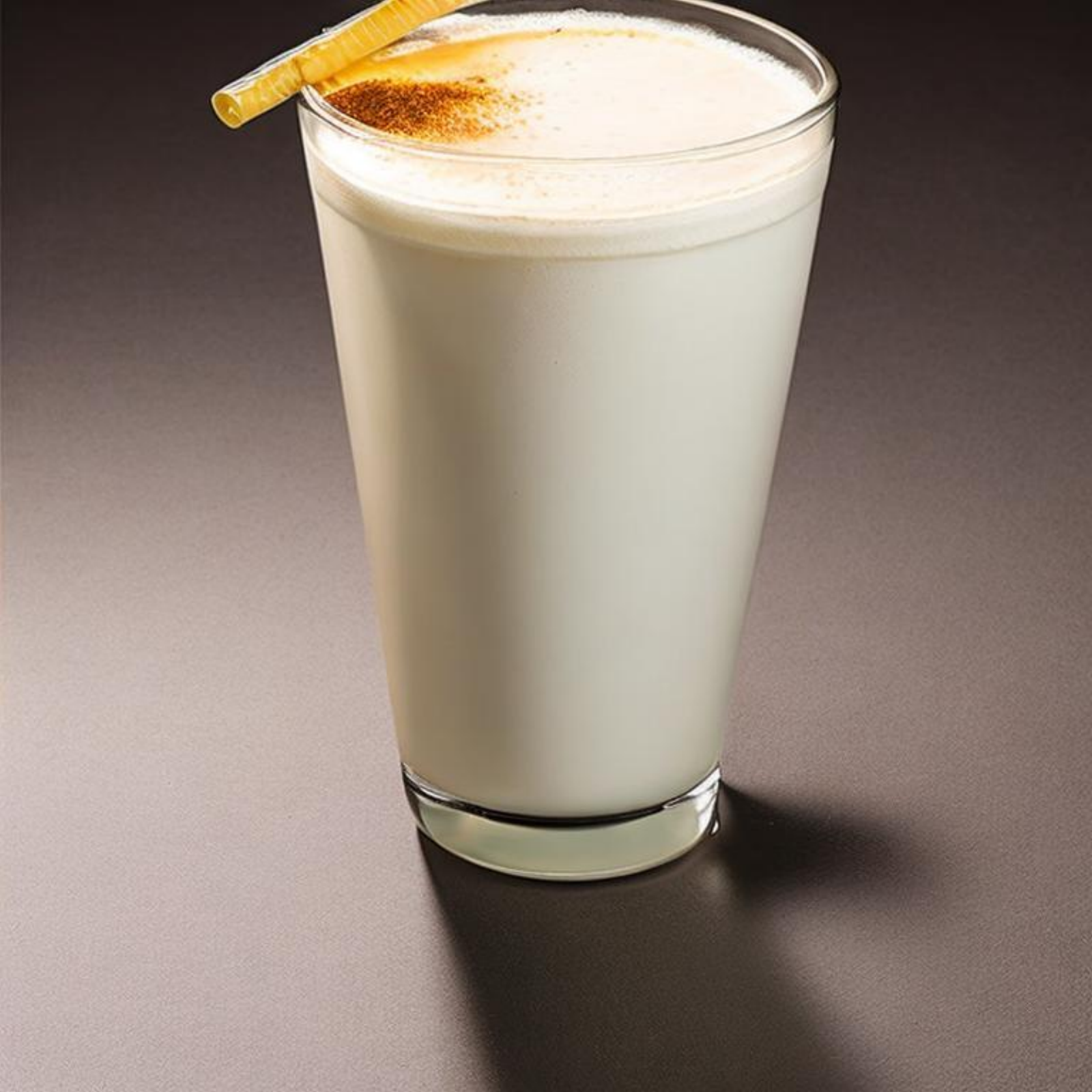} \\
\end{tabular}
  \caption{\textbf{Additional qualitative comparison on ShowHowTo.}}
  \label{fig:sup_showhow_738}
\end{figure*}

\begin{figure*}[t]
  \centering
  \setlength{\tabcolsep}{3pt}
  \renewcommand{\arraystretch}{1.2}
  \begin{tabular}{>{\centering\arraybackslash}m{1.5cm} m{0.160\linewidth} m{0.160\linewidth} m{0.160\linewidth} m{0.160\linewidth} m{0.160\linewidth}}
     & \tiny\centering Select the main firewood to use in your fireplace. & \tiny\centering Choose kindling, which are small pieces of softwood, to get the fire started. & \tiny\centering Use a few small strips of pine to get the fire started, but be aware of the potential for creosote buildup. & \tiny\centering Place the kindling inside your fireplace and surround it with larger pieces of main wood. & \tiny Light the kindling and let the fire grow. \\
    \rotatebox[origin=c]{90}{\scriptsize Stable Diffusion} & \includegraphics[width=\linewidth, valign=m]{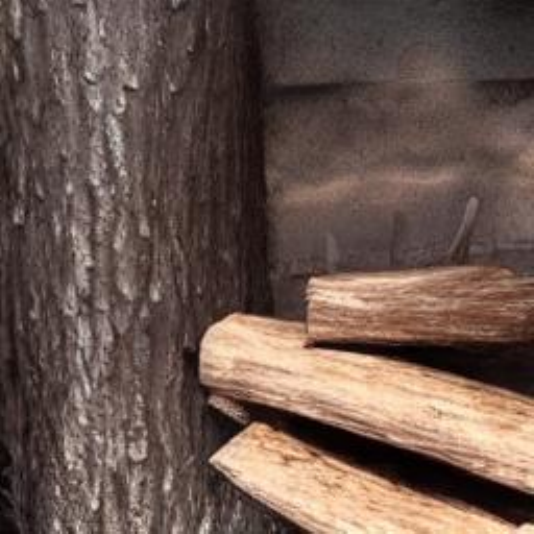} & \includegraphics[width=\linewidth, valign=m]{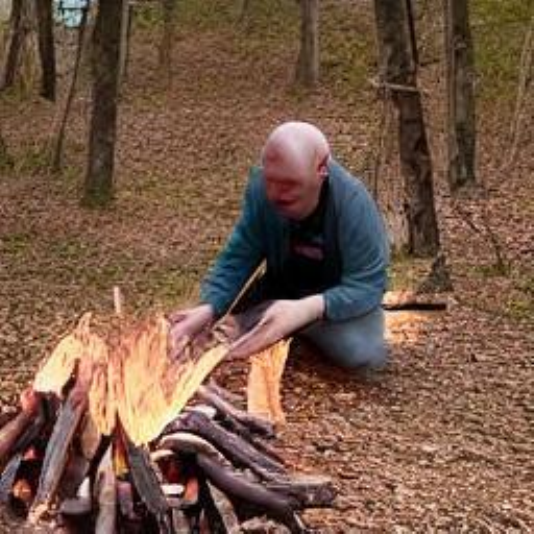} & \includegraphics[width=\linewidth, valign=m]{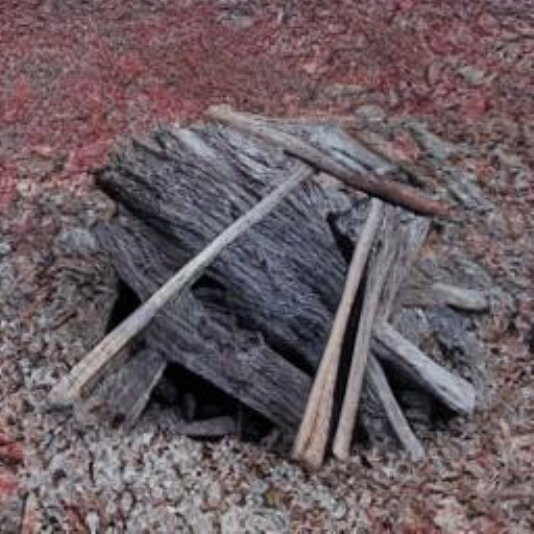} & \includegraphics[width=\linewidth, valign=m]{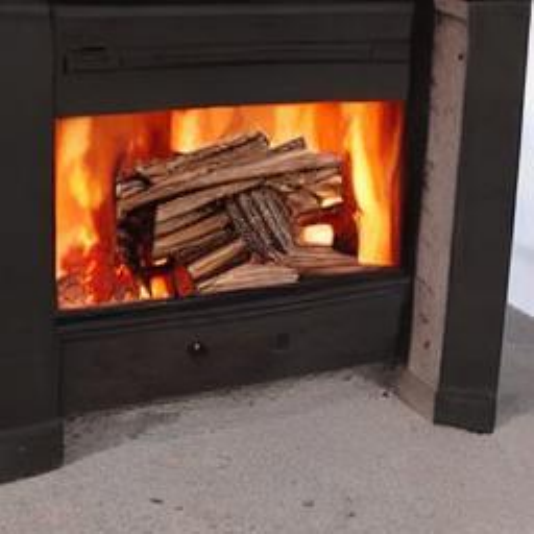} & \includegraphics[width=\linewidth, valign=m]{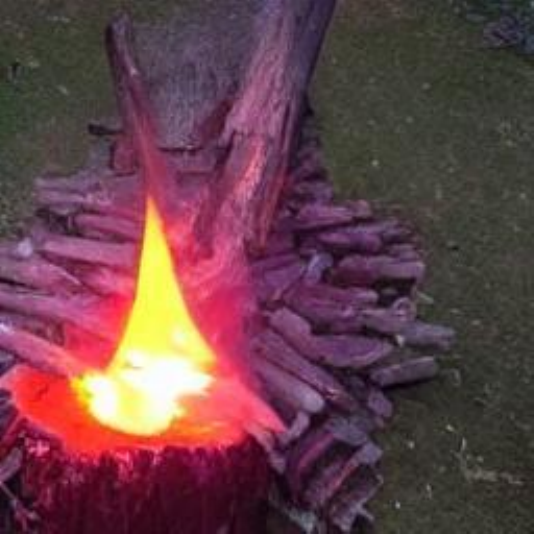} \\ \noalign{\vspace{10pt}}
    \rotatebox[origin=c]{90}{\scriptsize StackedDiffusion} & \includegraphics[width=\linewidth, valign=m]{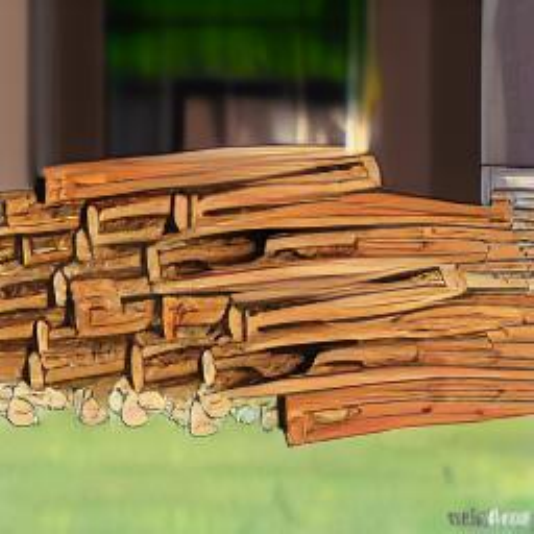} & \includegraphics[width=\linewidth, valign=m]{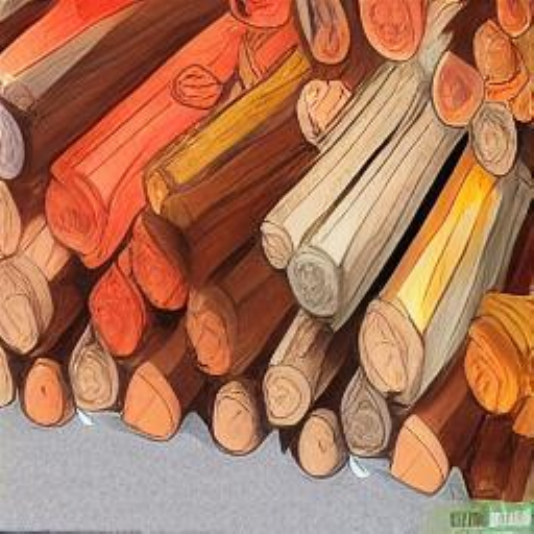} & \includegraphics[width=\linewidth, valign=m]{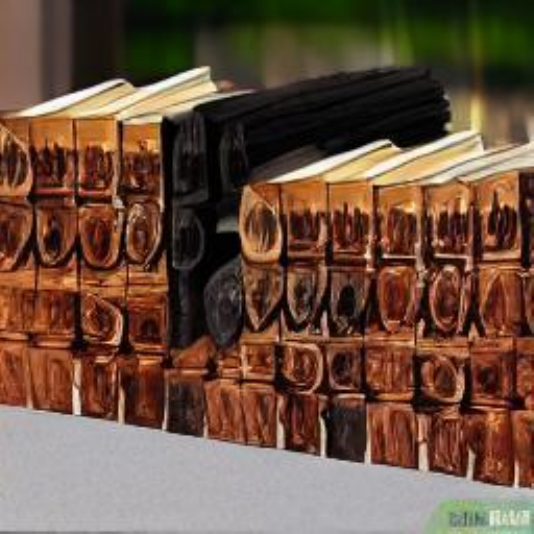} & \includegraphics[width=\linewidth, valign=m]{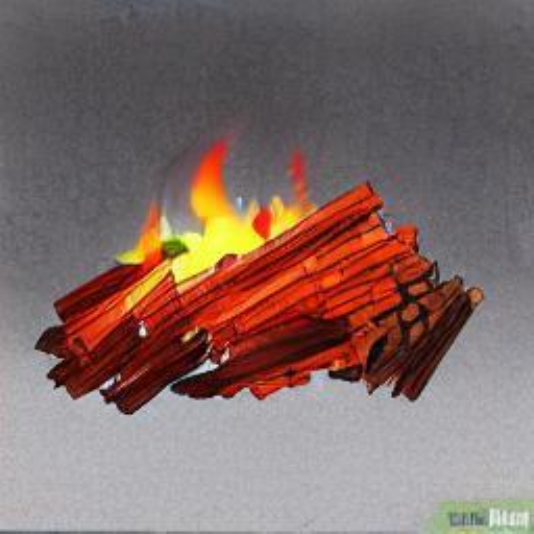} & \includegraphics[width=\linewidth, valign=m]{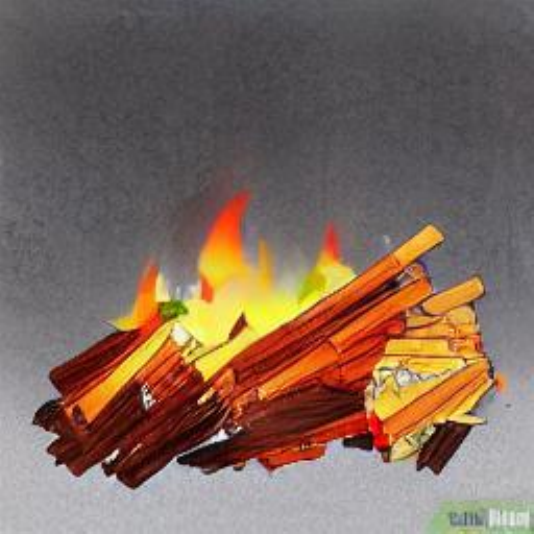} \\
\end{tabular}
  \caption{\textbf{Additional qualitative comparison on ShowHowTo.}}
  \label{fig:sup_showhow_1766}
\end{figure*}

\begin{figure}[t]
  \centering
  \setlength{\tabcolsep}{3pt}
  \renewcommand{\arraystretch}{1.2}
  \begin{tabular}{>{\centering\arraybackslash}m{1.5cm} m{0.400\linewidth} m{0.400\linewidth}}
     & \tiny\centering Buy store-bought frosting for a quick fix. & \tiny Add any embellishments onto the mini cupcakes, if desired. \\
    \rotatebox[origin=c]{90}{\scriptsize Stable Diffusion} & \includegraphics[width=\linewidth, valign=m]{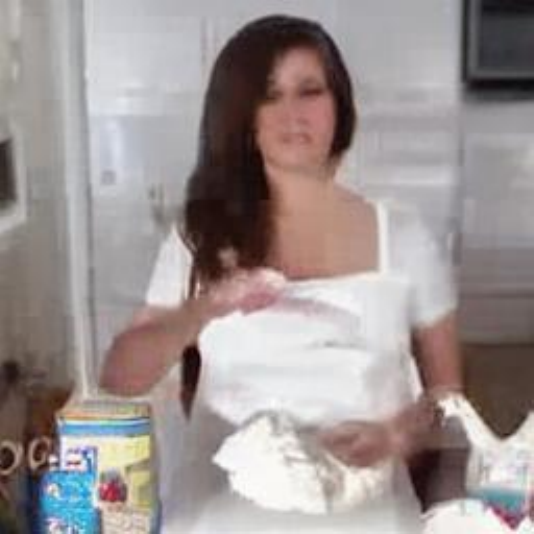} & \includegraphics[width=\linewidth, valign=m]{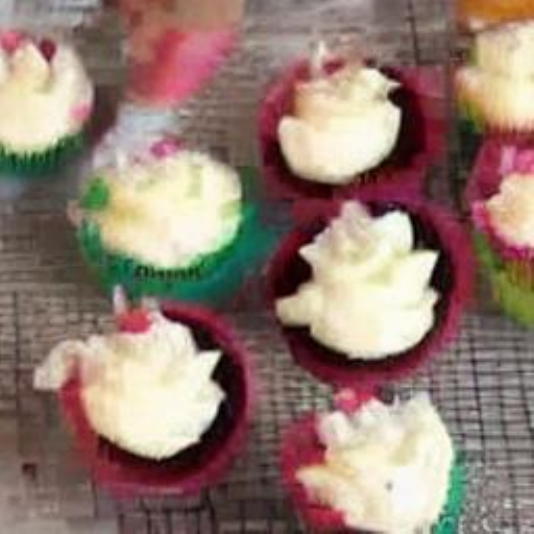} \\ \noalign{\vspace{10pt}}
    \rotatebox[origin=c]{90}{\scriptsize StackedDiffusion} & \includegraphics[width=\linewidth, valign=m]{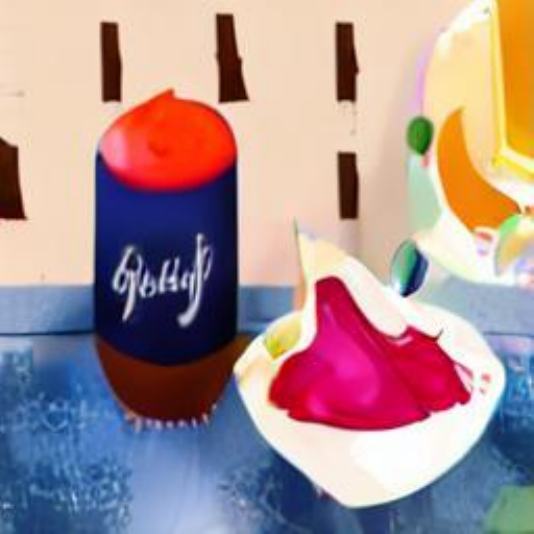} & \includegraphics[width=\linewidth, valign=m]{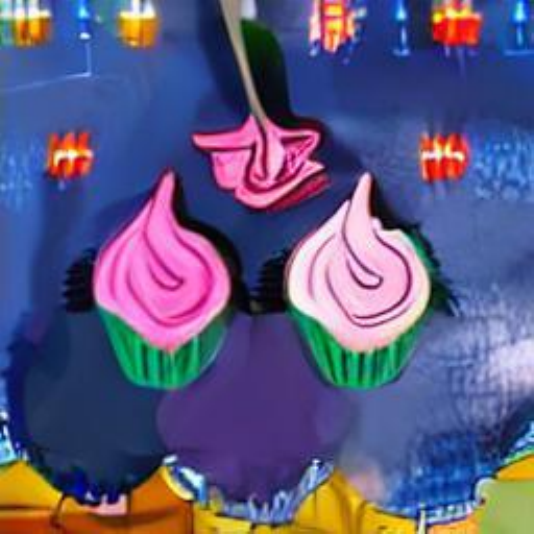} \\ \noalign{\vspace{10pt}}
    \rotatebox[origin=c]{90}{\scriptsize Bordalo et al.} & \includegraphics[width=\linewidth, valign=m]{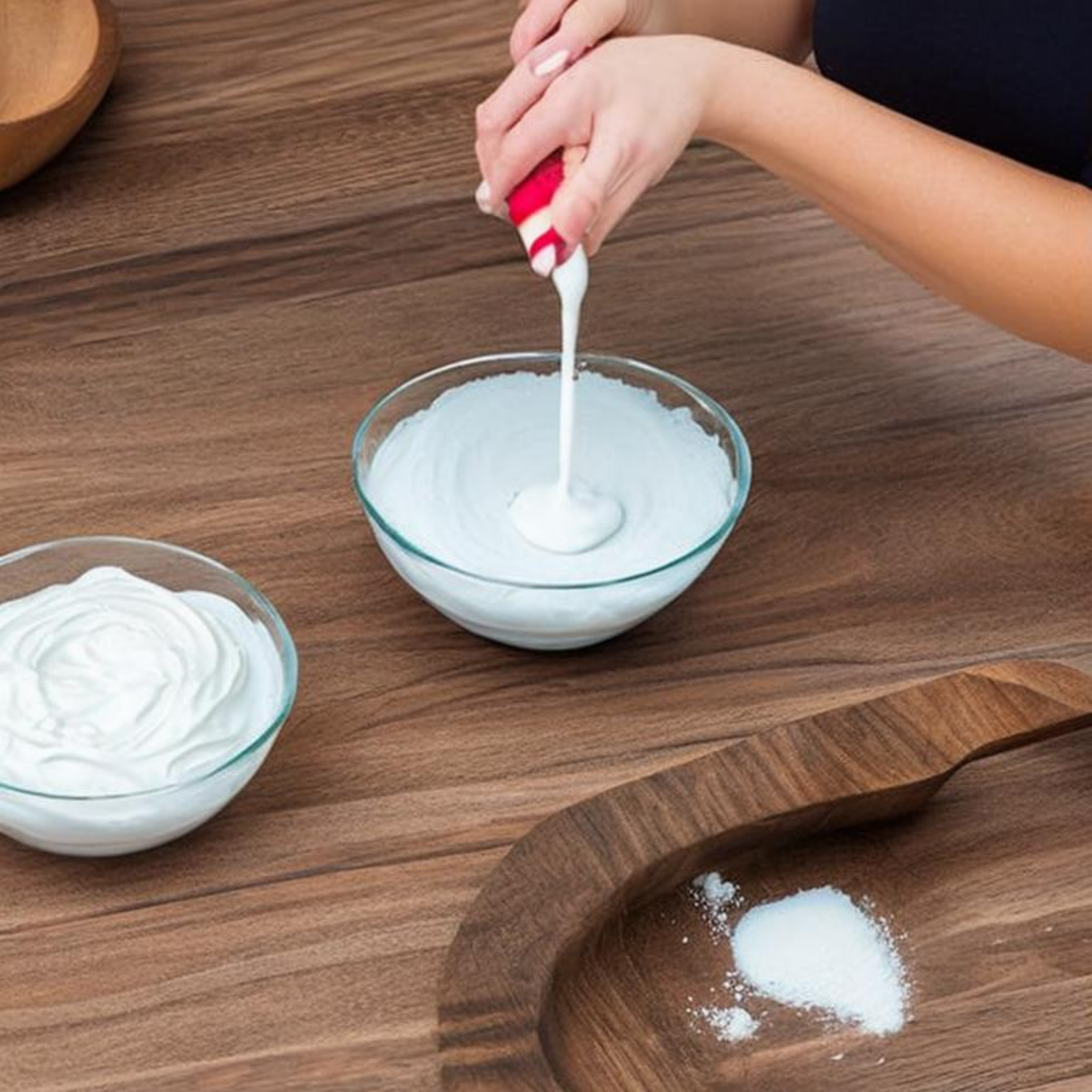} & \includegraphics[width=\linewidth, valign=m]{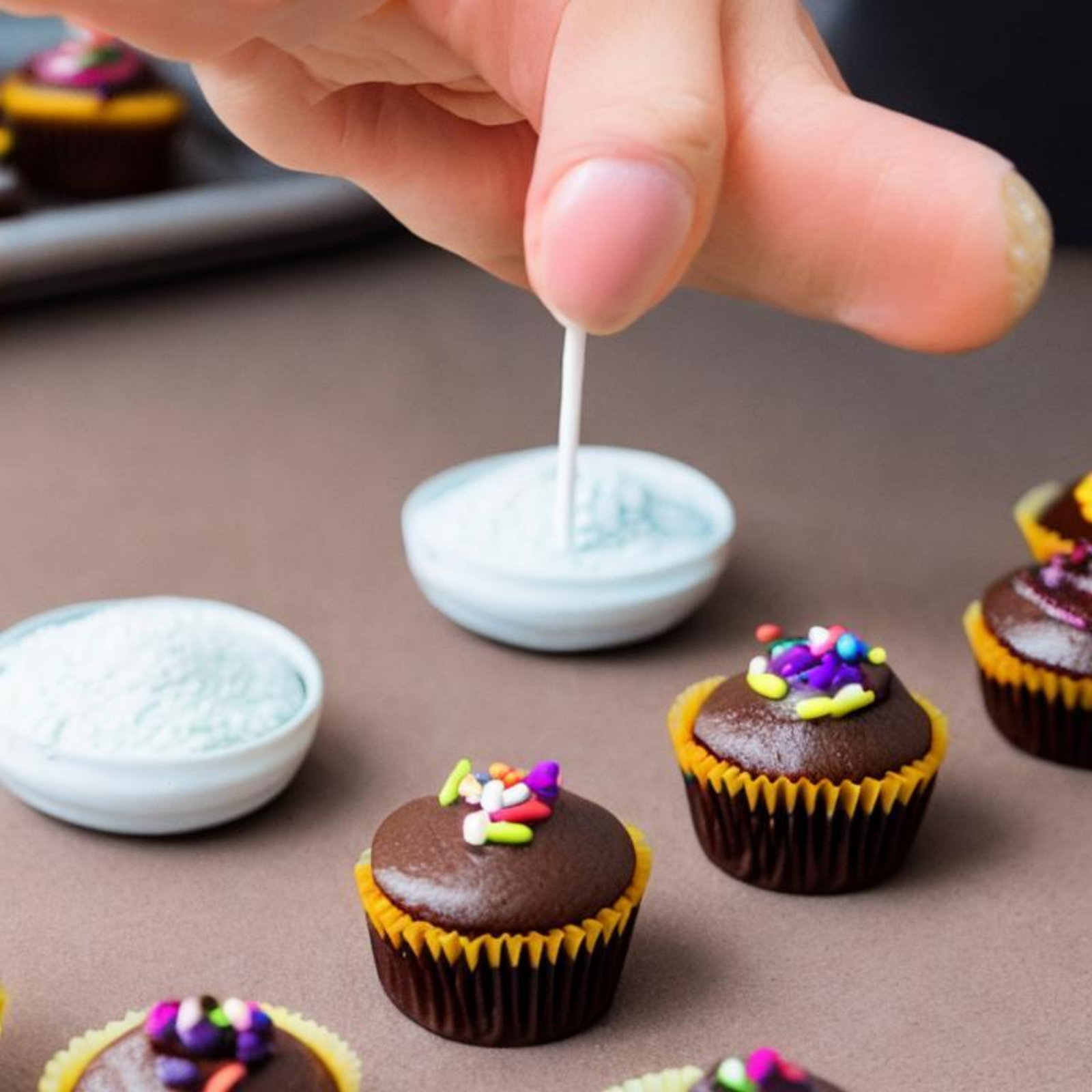} \\ \noalign{\vspace{10pt}}
    \rotatebox[origin=c]{90}{\scriptsize Baseline} & \includegraphics[width=\linewidth, valign=m]{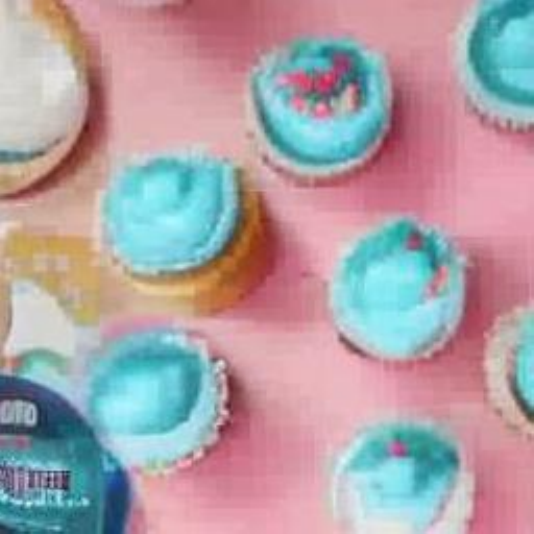} & \includegraphics[width=\linewidth, valign=m]{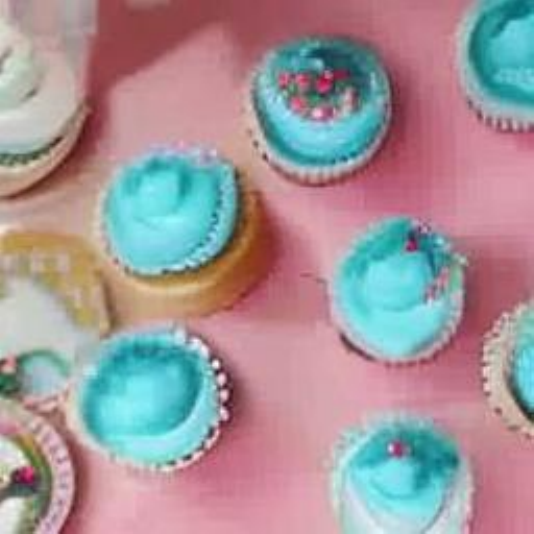} \\ \noalign{\vspace{10pt}}
    \rotatebox[origin=c]{90}{\scriptsize \textbf{InstructionCrafter}} & \includegraphics[width=\linewidth, valign=m]{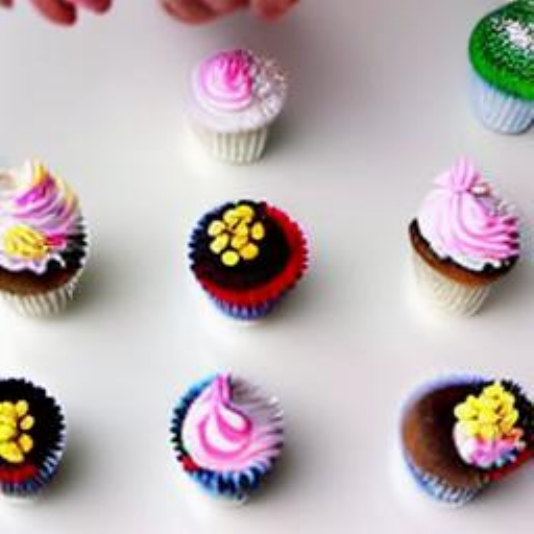} & \includegraphics[width=\linewidth, valign=m]{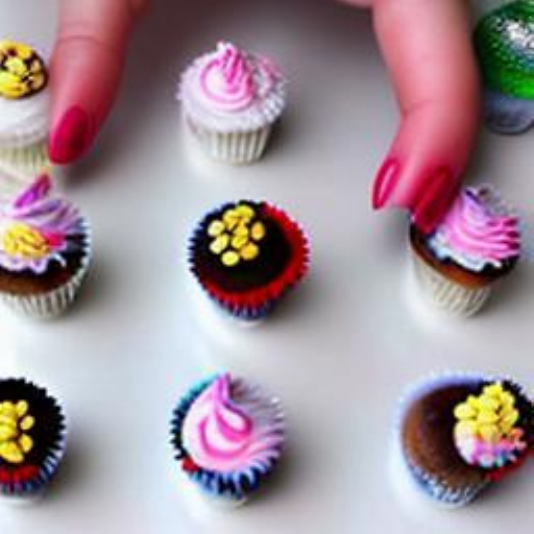} \\ \noalign{\vspace{10pt}}
  \end{tabular}
  \caption{\textbf{Additional qualitative comparison on WikiHow-VGSI.}}
  \label{fig:sup_wikihow_3}
\end{figure}

\begin{figure}[t]
  \centering
  \setlength{\tabcolsep}{3pt}
  \renewcommand{\arraystretch}{1.2}
  \begin{tabular}{>{\centering\arraybackslash}m{1.5cm} m{0.400\linewidth} m{0.400\linewidth}}
     & \tiny\centering Add a little cornstarch. & \tiny Knead the food coloring and cornstarch into the chocolate. \\
    \rotatebox[origin=c]{90}{\scriptsize Stable Diffusion} & \includegraphics[width=\linewidth, valign=m]{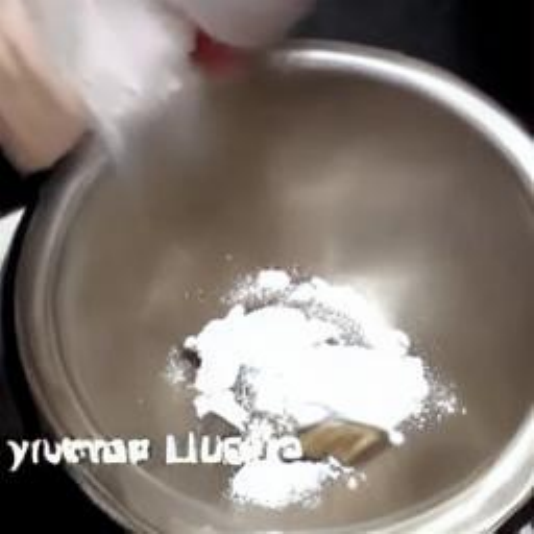} & \includegraphics[width=\linewidth, valign=m]{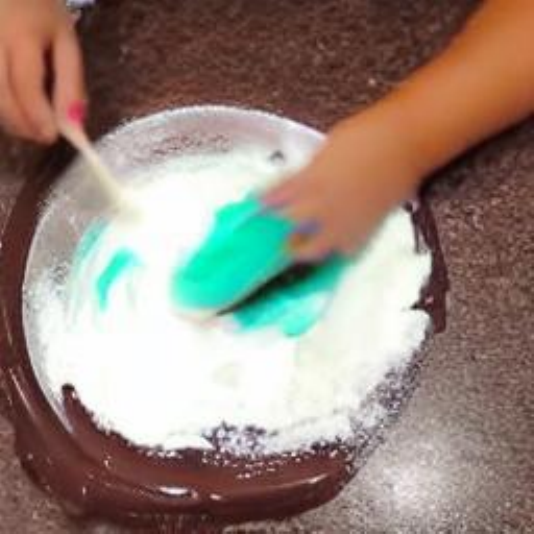} \\ \noalign{\vspace{10pt}}
    \rotatebox[origin=c]{90}{\scriptsize StackedDiffusion} & \includegraphics[width=\linewidth, valign=m]{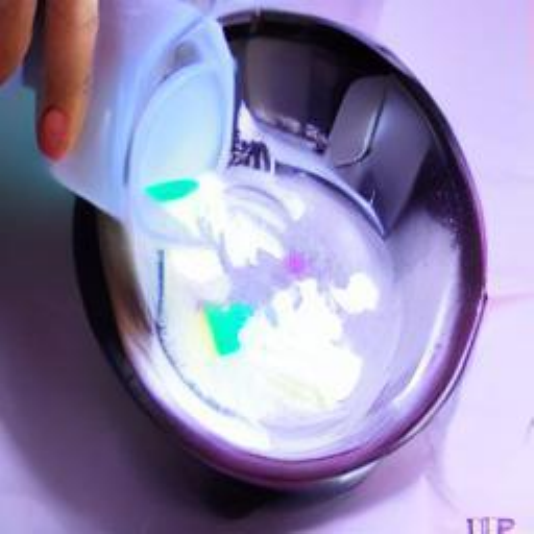} & \includegraphics[width=\linewidth, valign=m]{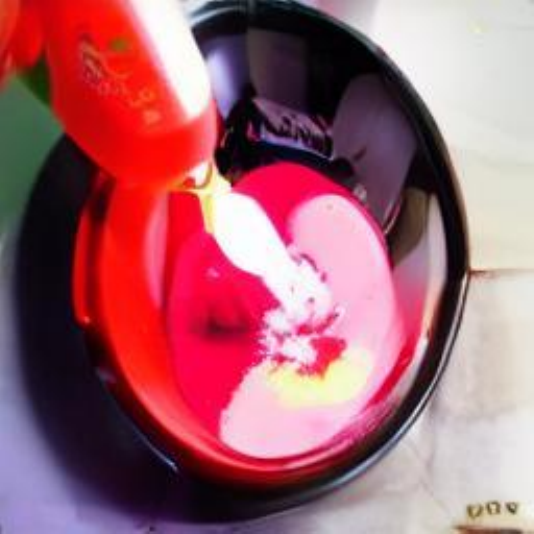} \\ \noalign{\vspace{10pt}}
    \rotatebox[origin=c]{90}{\scriptsize Bordalo et al.} & \includegraphics[width=\linewidth, valign=m]{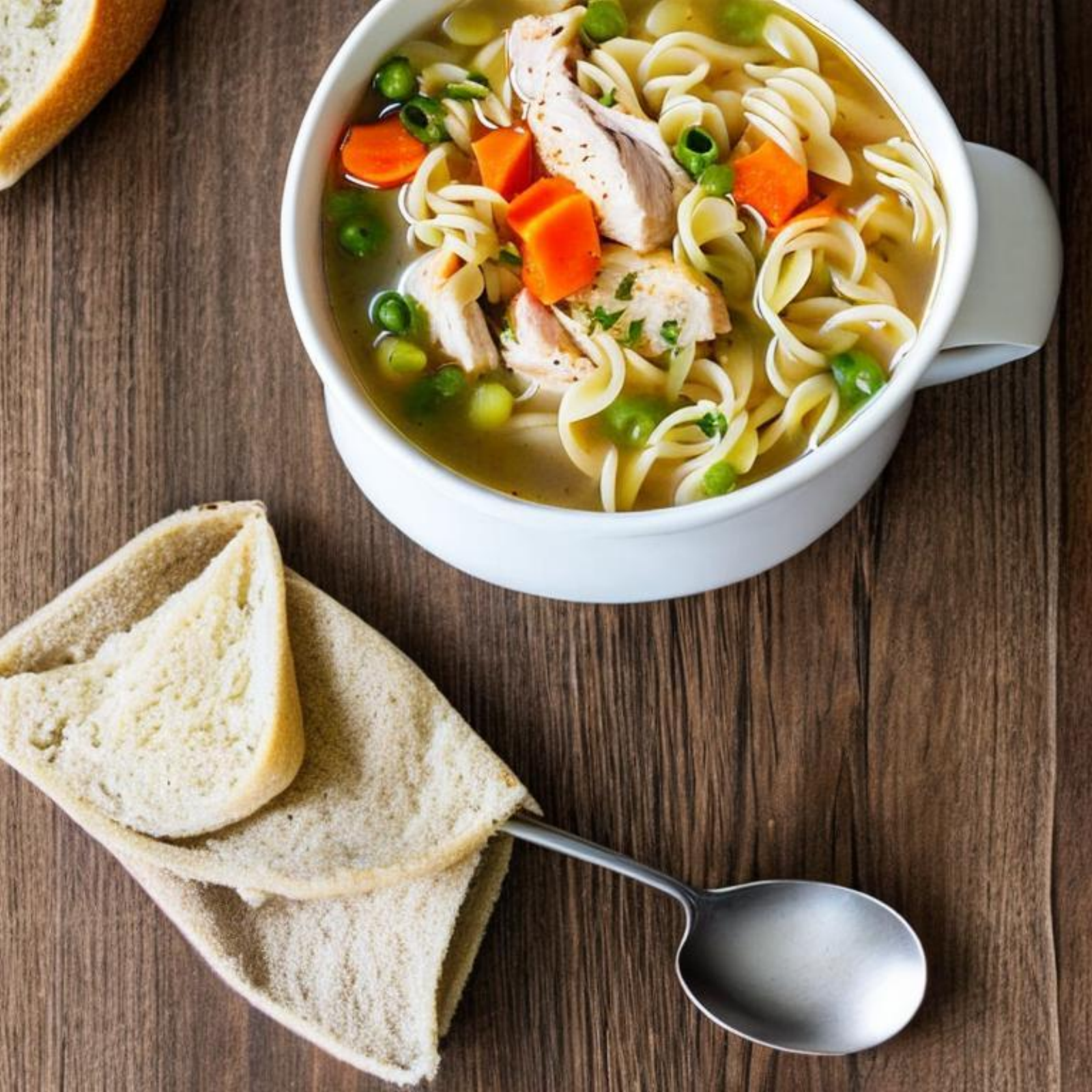} & \includegraphics[width=\linewidth, valign=m]{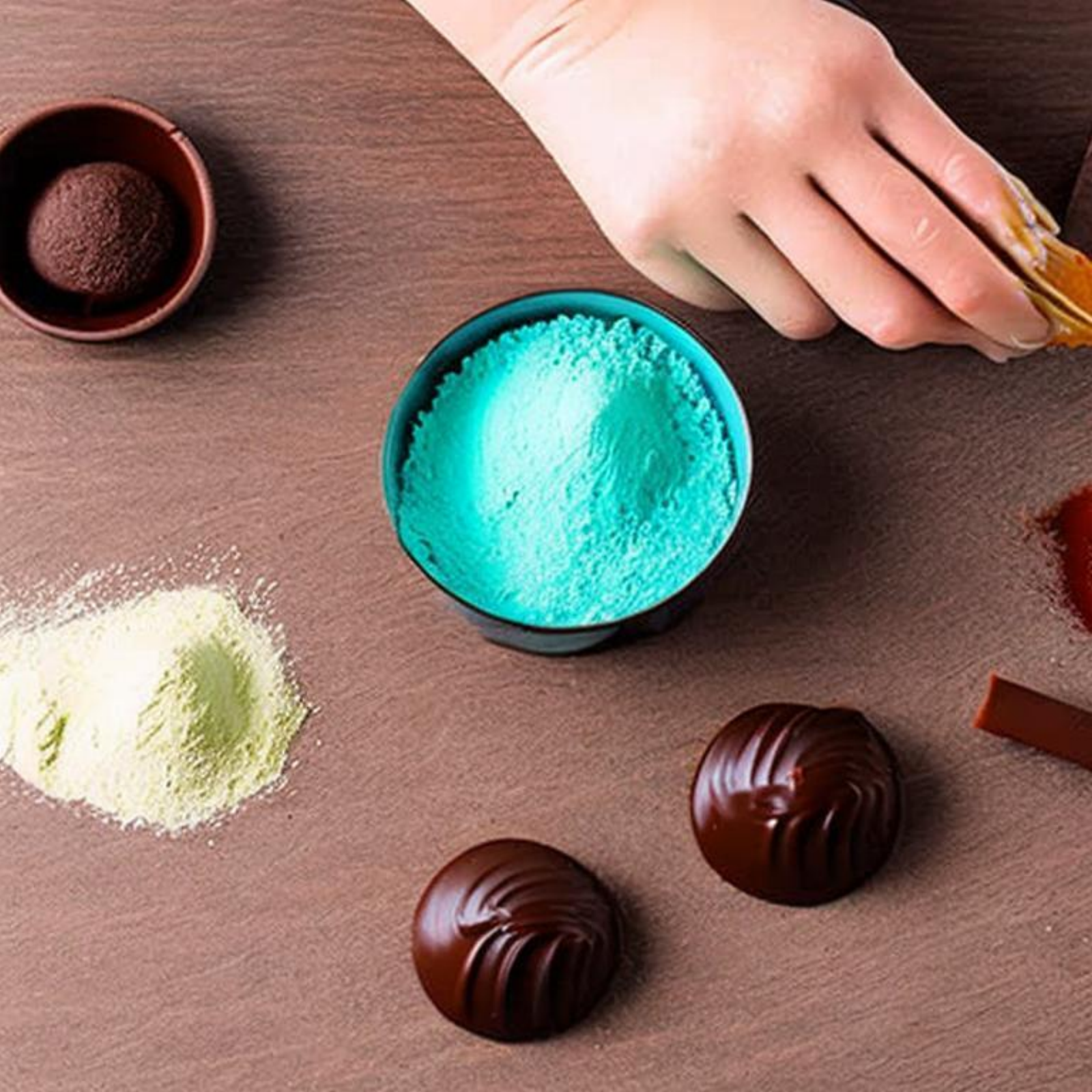} \\ \noalign{\vspace{10pt}}
    \rotatebox[origin=c]{90}{\scriptsize Baseline} & \includegraphics[width=\linewidth, valign=m]{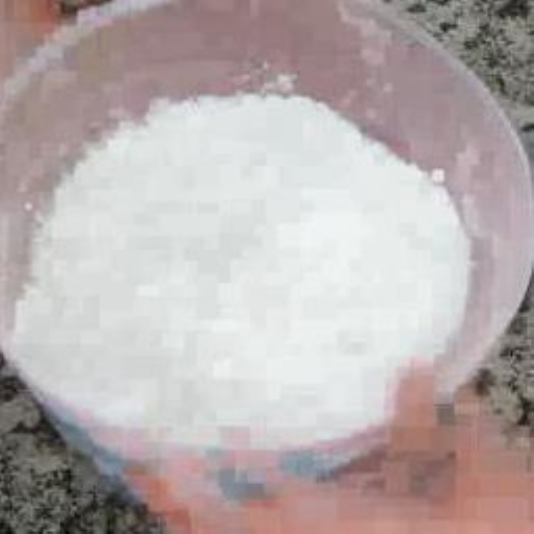} & \includegraphics[width=\linewidth, valign=m]{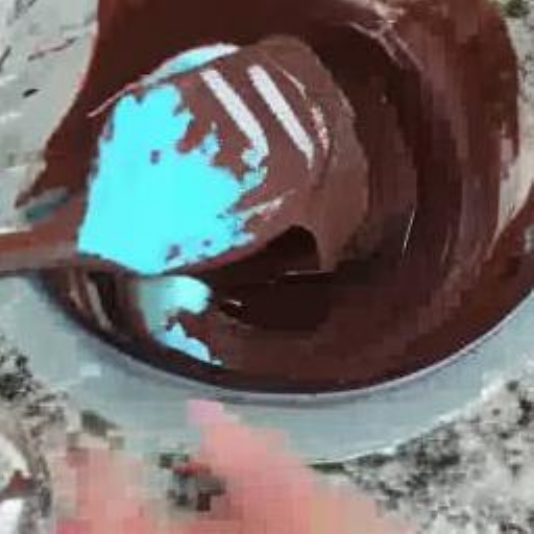} \\ \noalign{\vspace{10pt}}
    \rotatebox[origin=c]{90}{\scriptsize \textbf{InstructionCrafter}} & \includegraphics[width=\linewidth, valign=m]{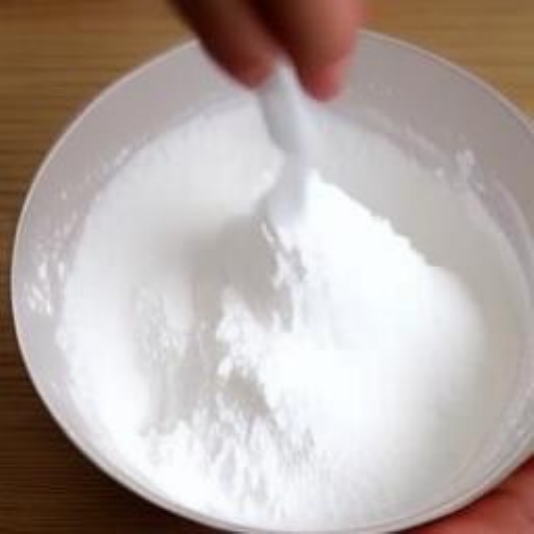} & \includegraphics[width=\linewidth, valign=m]{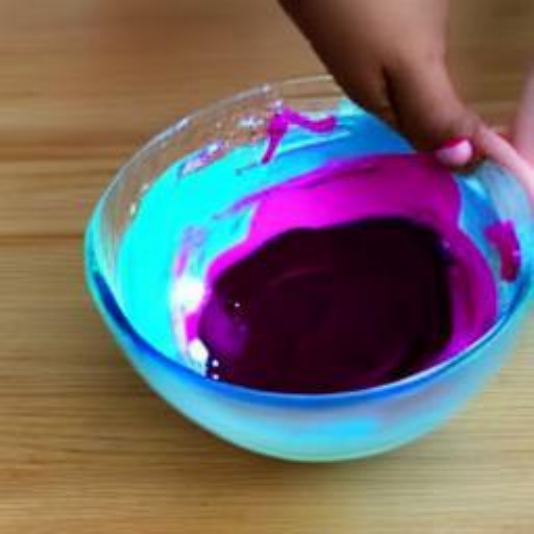} \\ \noalign{\vspace{10pt}}
  \end{tabular}
  \caption{\textbf{Additional qualitative comparison on WikiHow-VGSI.}}
  \label{fig:sup_wikihow_1505}
\end{figure}

\begin{figure*}[t]
  \centering
  \setlength{\tabcolsep}{3pt}
  \renewcommand{\arraystretch}{1.2}
  \begin{tabular}{>{\centering\arraybackslash}m{1.5cm} m{0.160\linewidth} m{0.160\linewidth} m{0.160\linewidth} m{0.160\linewidth} m{0.160\linewidth}}
     & \tiny\centering Apply primer to the slip coupling and piston end of the slip fix. & \tiny\centering Apply solvent cement over the primer on the slip coupling and slip fix. & \tiny\centering Install the coupling over the repair pipe covered with solvent cement. & \tiny\centering Repeat the previous process for the remaining repair piping. & \tiny Connect the piston end of the slip fix to the remaining piping. \\
    \rotatebox[origin=c]{90}{\scriptsize Stable Diffusion} & \includegraphics[width=\linewidth, valign=m]{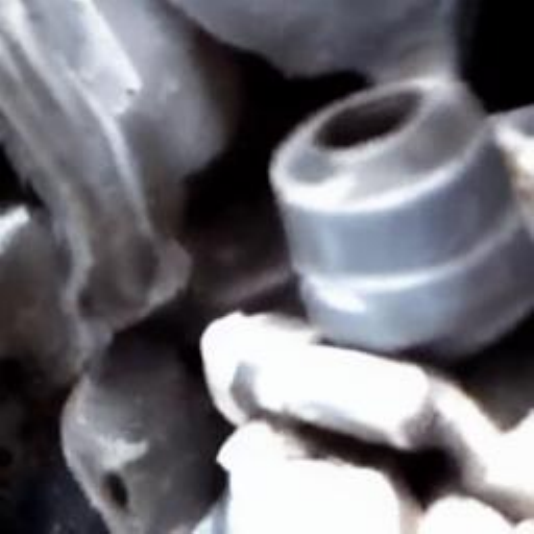} & \includegraphics[width=\linewidth, valign=m]{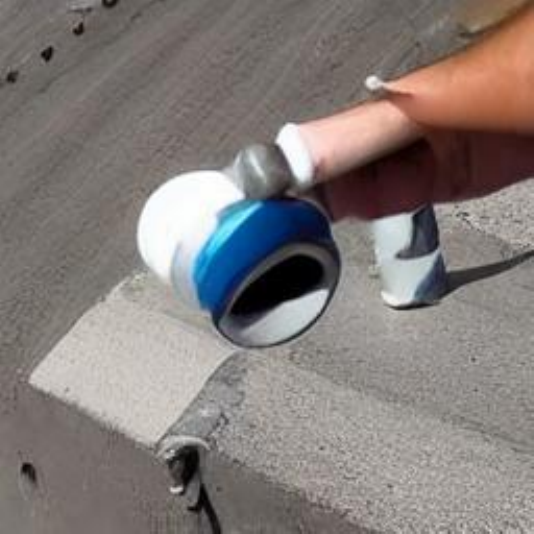} & \includegraphics[width=\linewidth, valign=m]{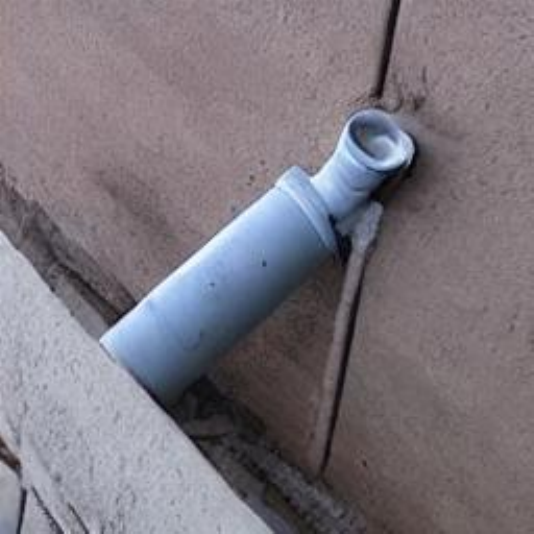} & \includegraphics[width=\linewidth, valign=m]{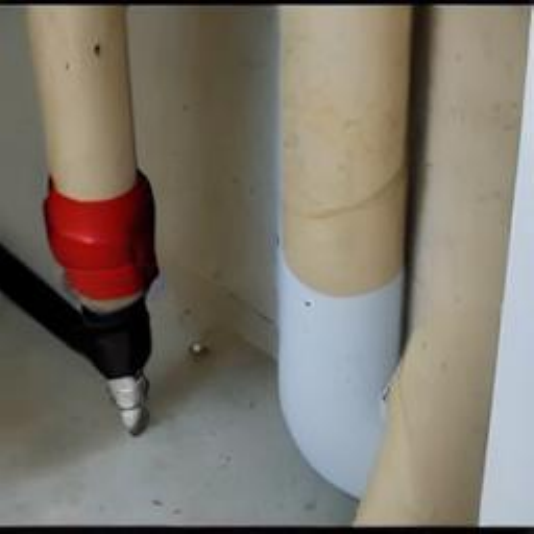} & \includegraphics[width=\linewidth, valign=m]{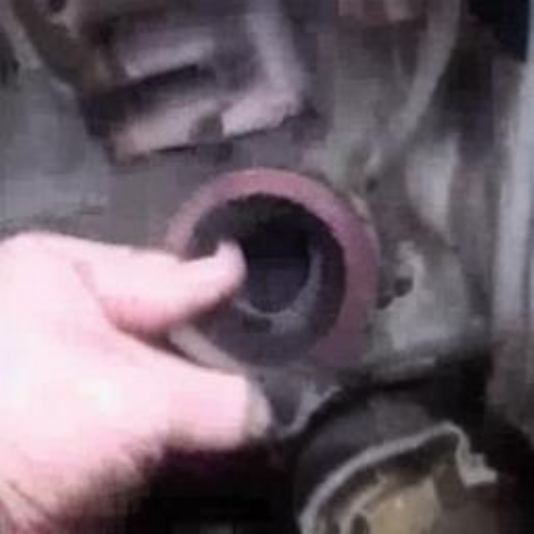} \\ \noalign{\vspace{10pt}}
    \rotatebox[origin=c]{90}{\scriptsize StackedDiffusion} & \includegraphics[width=\linewidth, valign=m]{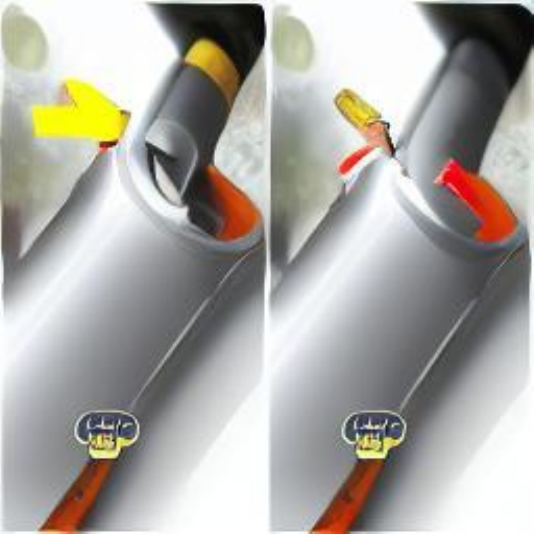} & \includegraphics[width=\linewidth, valign=m]{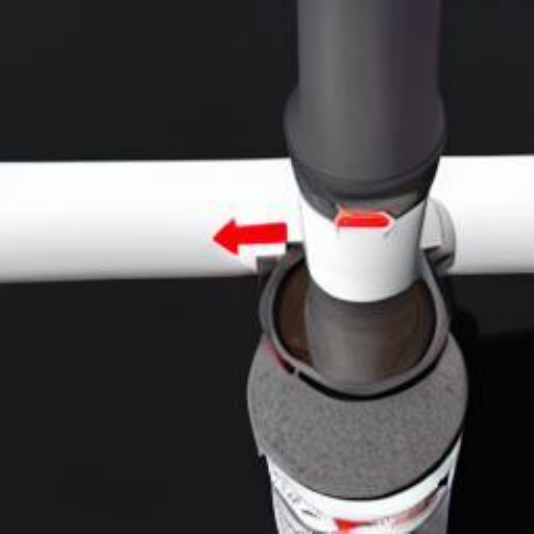} & \includegraphics[width=\linewidth, valign=m]{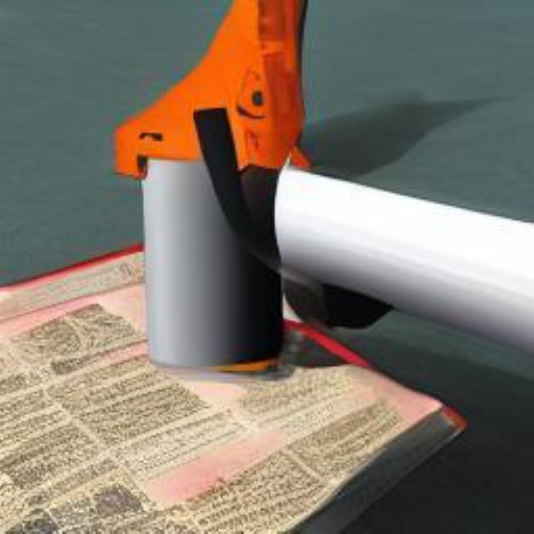} & \includegraphics[width=\linewidth, valign=m]{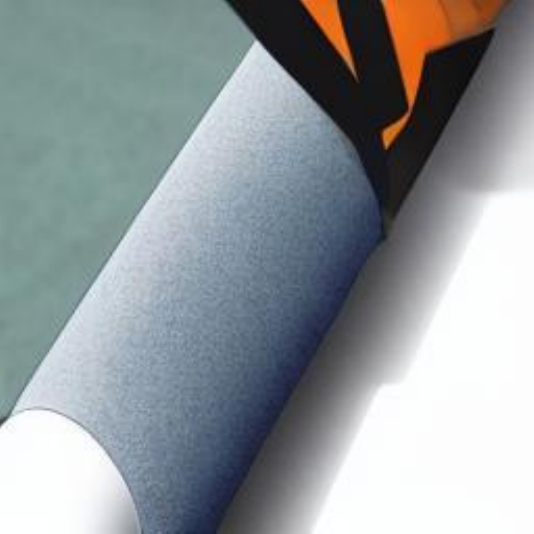} & \includegraphics[width=\linewidth, valign=m]{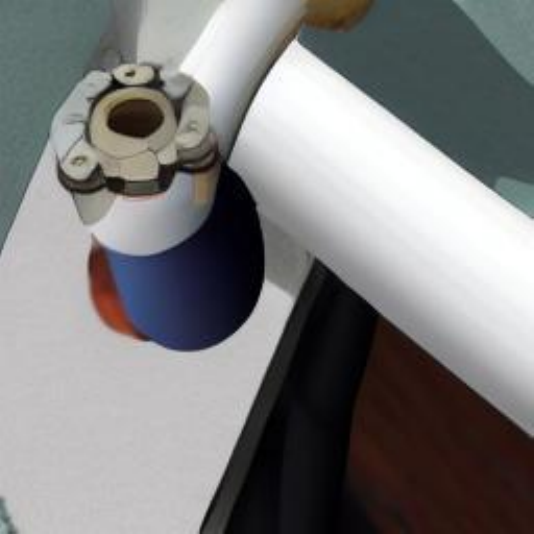} \\ \noalign{\vspace{10pt}}
    \rotatebox[origin=c]{90}{\scriptsize Bordalo et al.} & \includegraphics[width=\linewidth, valign=m]{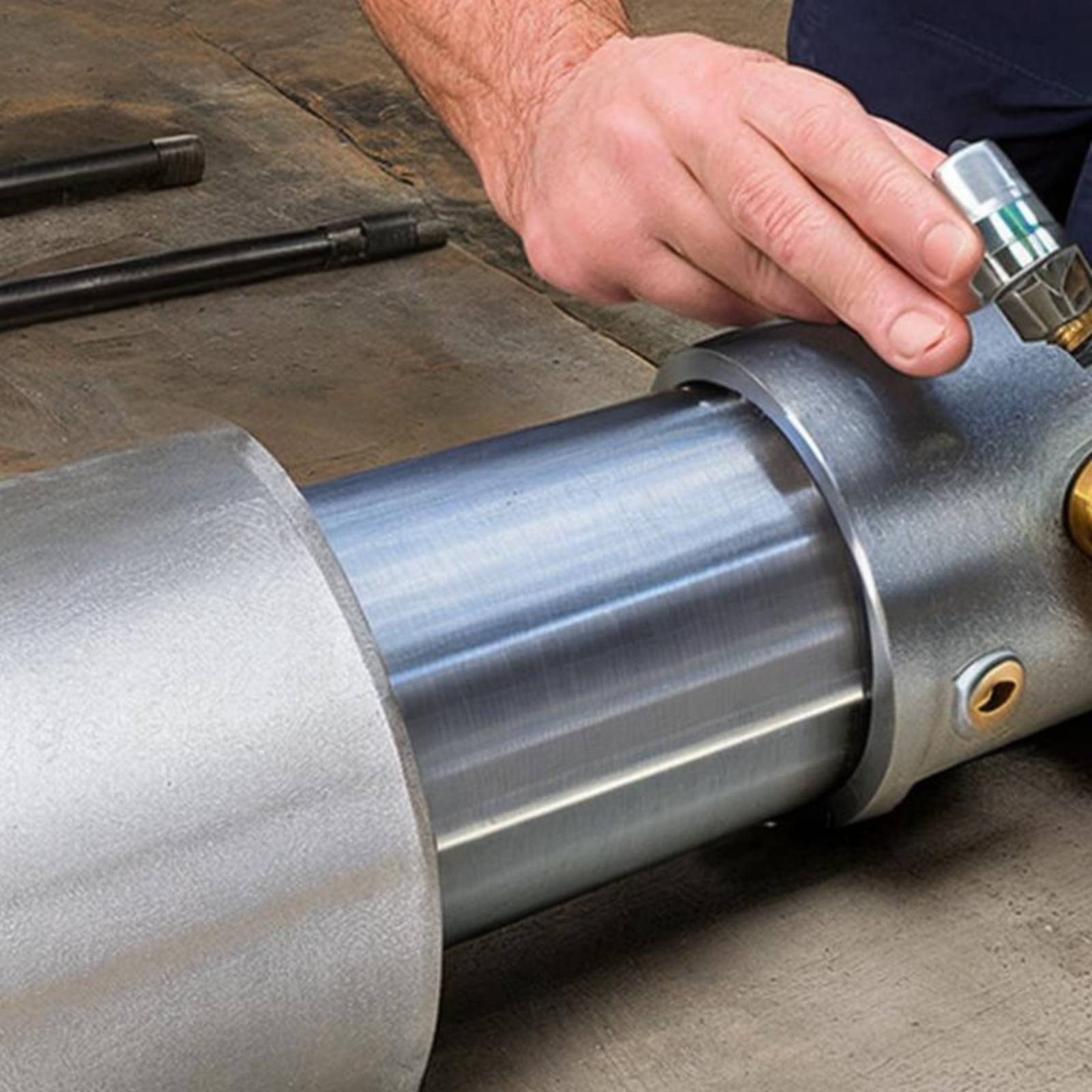} & \includegraphics[width=\linewidth, valign=m]{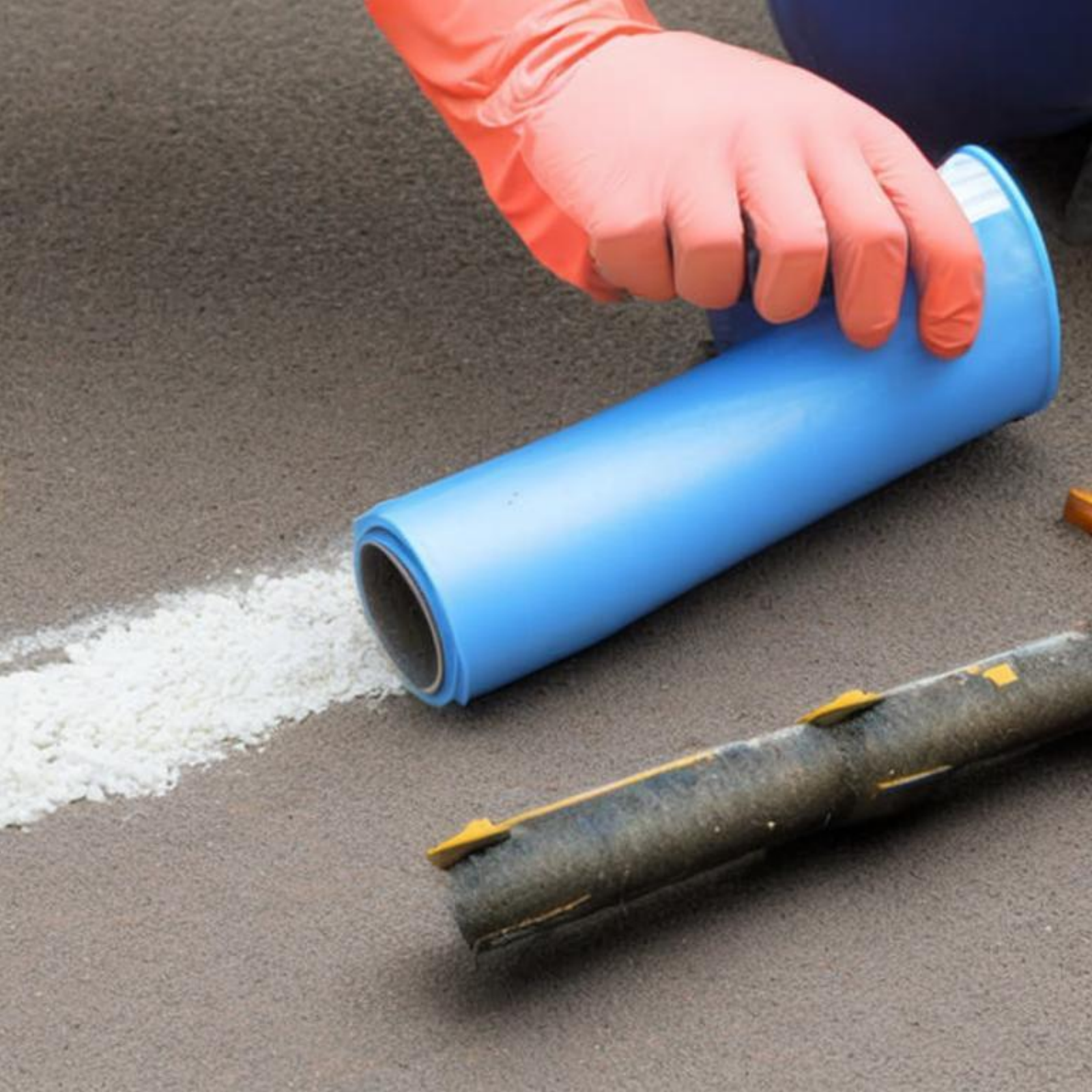} & \includegraphics[width=\linewidth, valign=m]{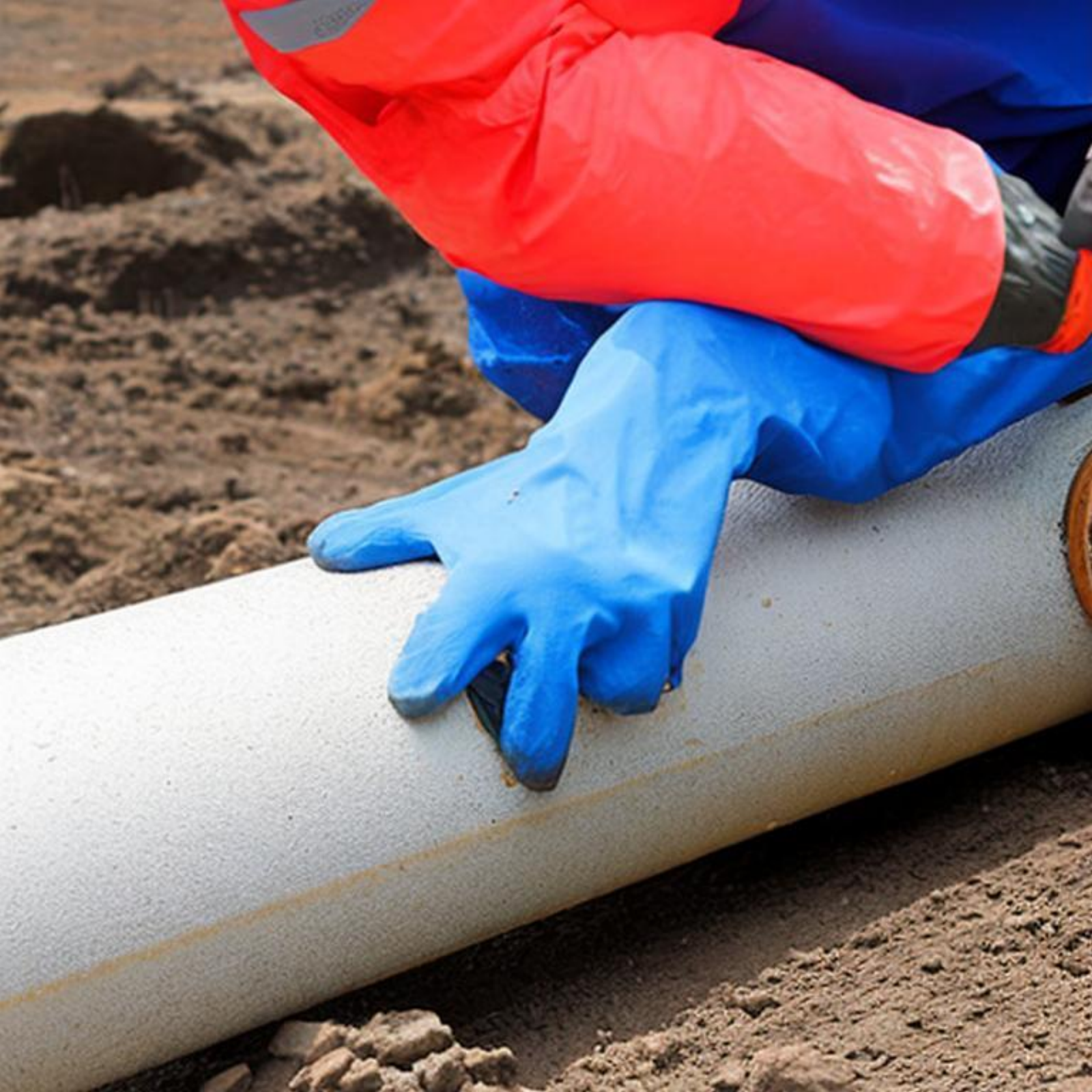} & \includegraphics[width=\linewidth, valign=m]{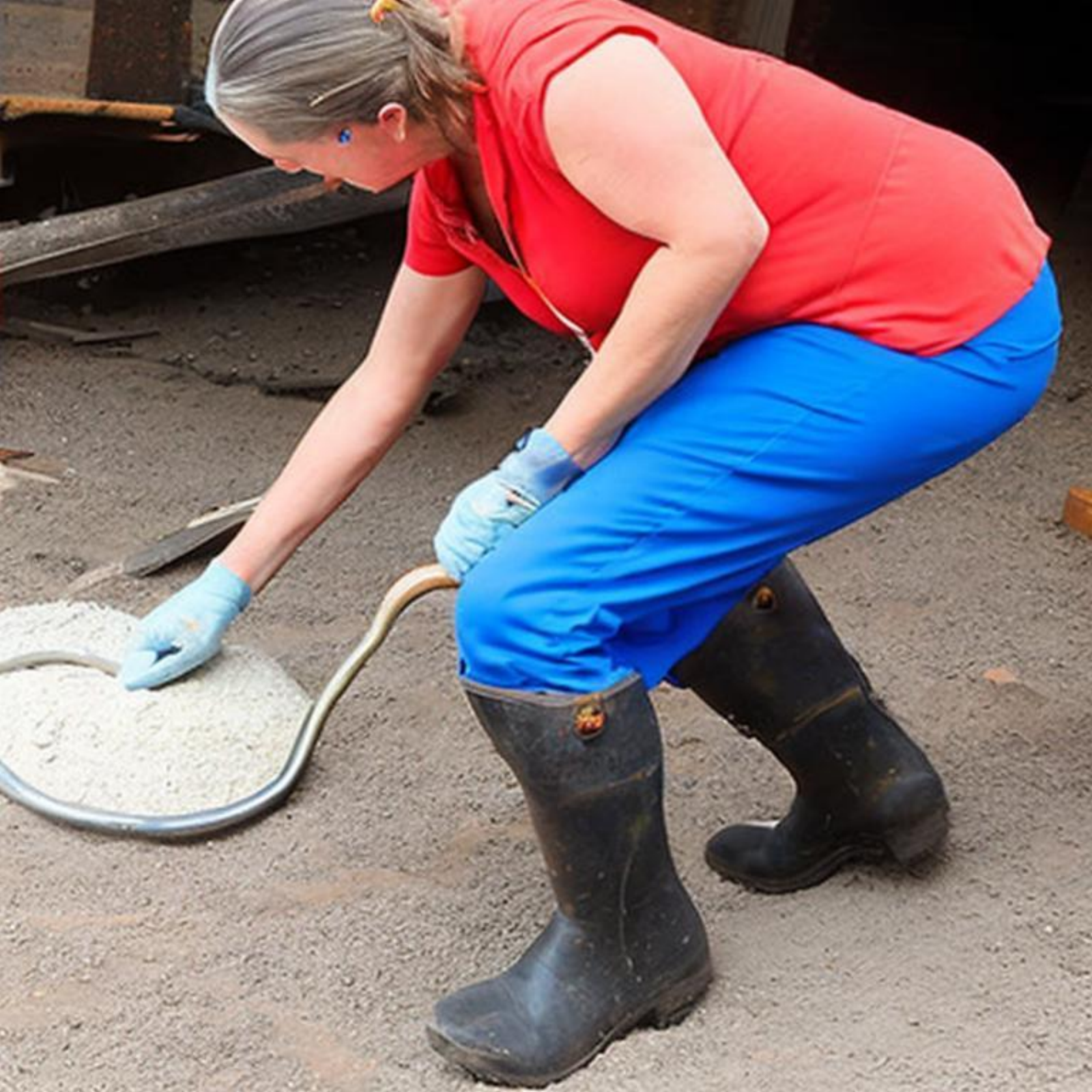} & \includegraphics[width=\linewidth, valign=m]{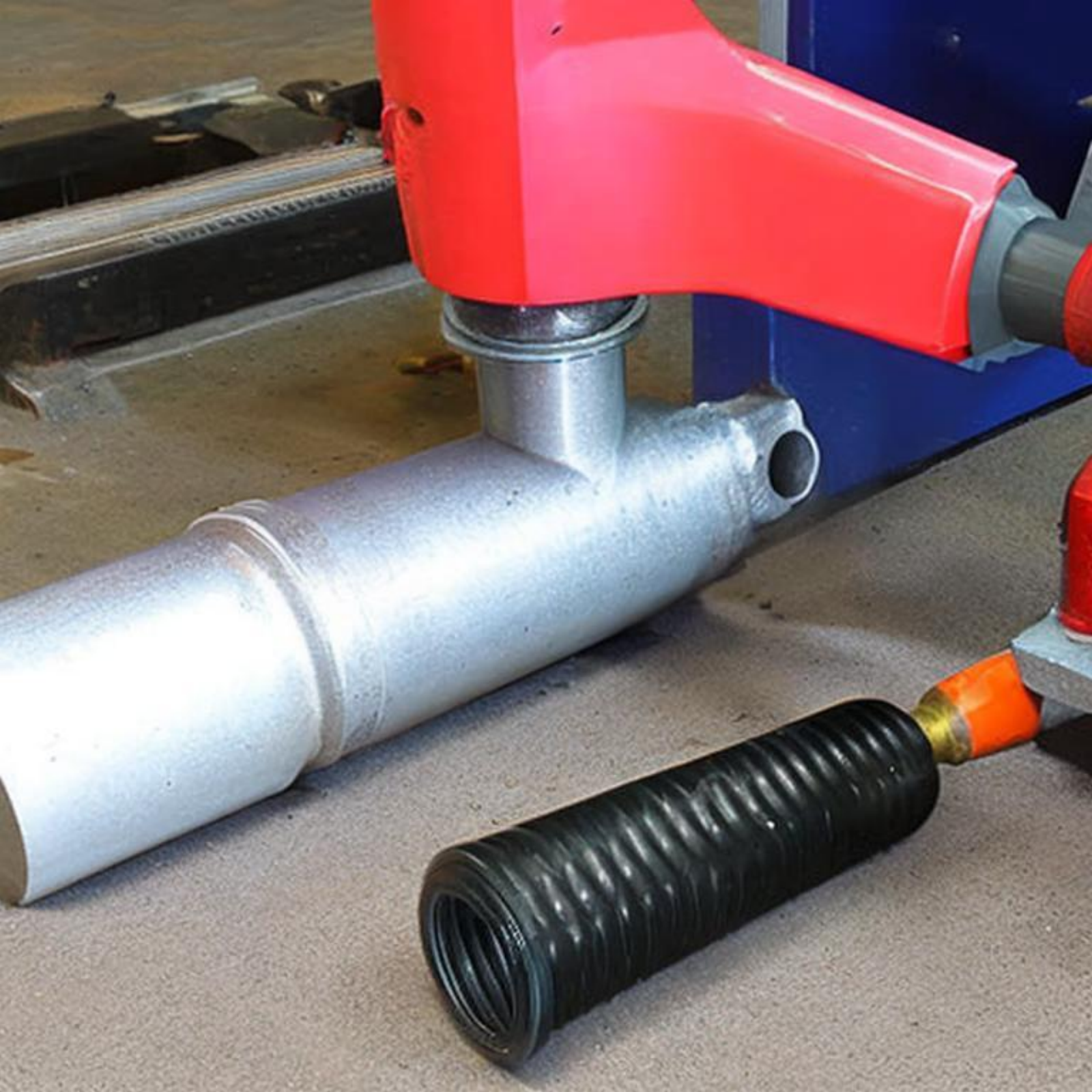} \\
\end{tabular}
  \caption{\textbf{Additional qualitative comparison on WikiHow-VGSI.}}
  \label{fig:sup_wikihow_227}
\end{figure*}

\begin{figure}[t]
  \centering
  \setlength{\tabcolsep}{3pt}
  \renewcommand{\arraystretch}{1.2}
  \begin{tabular}{>{\centering\arraybackslash}m{1.5cm} m{0.267\linewidth} m{0.267\linewidth} m{0.267\linewidth}}
     & \tiny\centering Push epoxy into deeper gouges and chips. & \tiny\centering Apply a second coat of epoxy using the kit’s brush. & \tiny Sand the repaired area with fine-grit sandpaper. \\
    \rotatebox[origin=c]{90}{\scriptsize Stable Diffusion} & \includegraphics[width=\linewidth, valign=m]{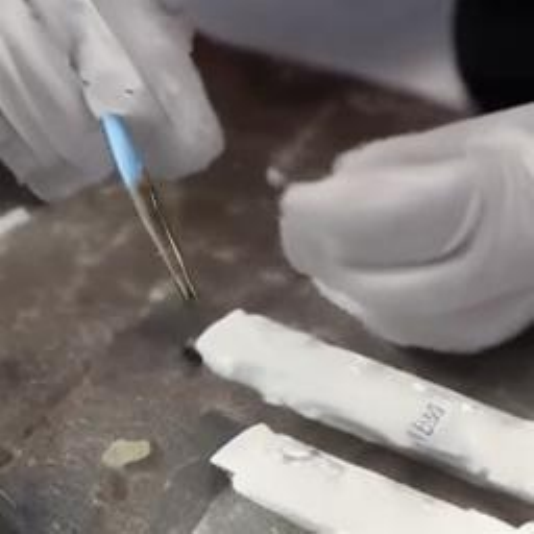} & \includegraphics[width=\linewidth, valign=m]{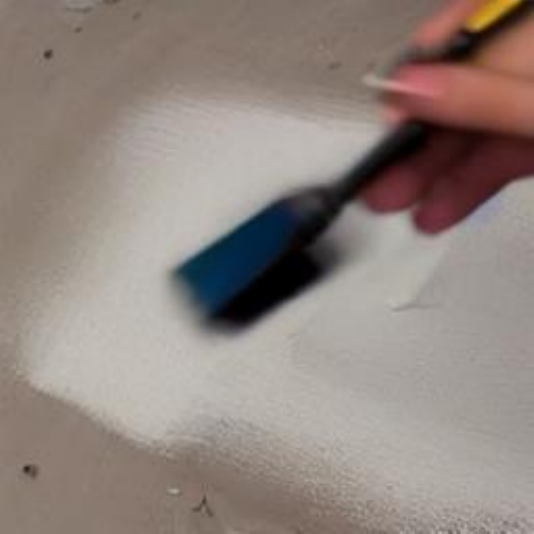} & \includegraphics[width=\linewidth, valign=m]{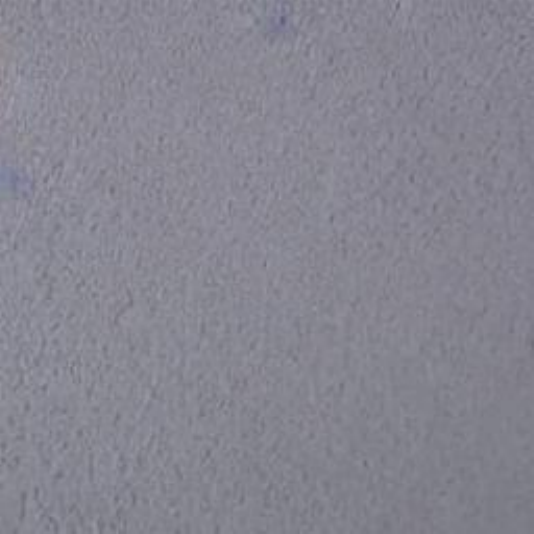} \\ \noalign{\vspace{10pt}}
    \rotatebox[origin=c]{90}{\scriptsize StackedDiffusion} & \includegraphics[width=\linewidth, valign=m]{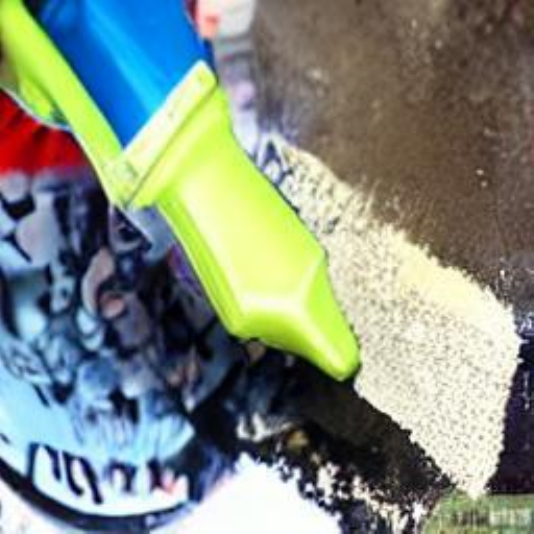} & \includegraphics[width=\linewidth, valign=m]{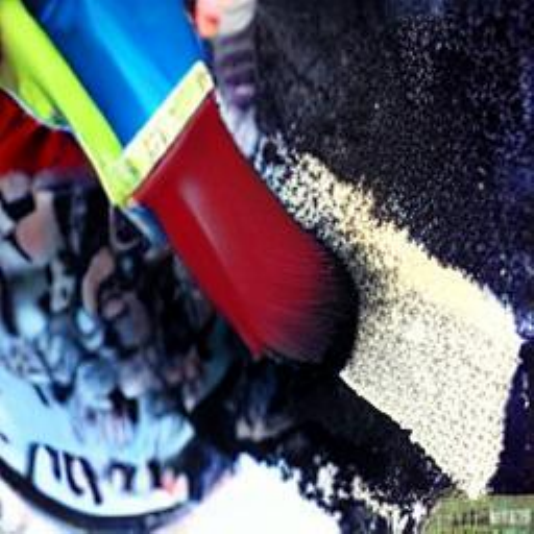} & \includegraphics[width=\linewidth, valign=m]{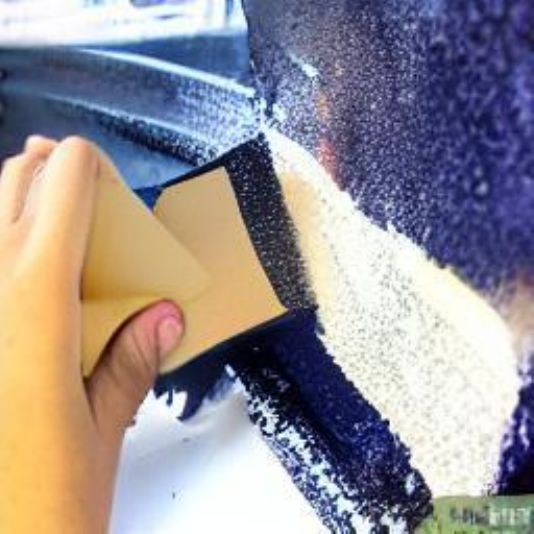} \\ \noalign{\vspace{10pt}}
    \rotatebox[origin=c]{90}{\scriptsize Bordalo et al.} & \includegraphics[width=\linewidth, valign=m]{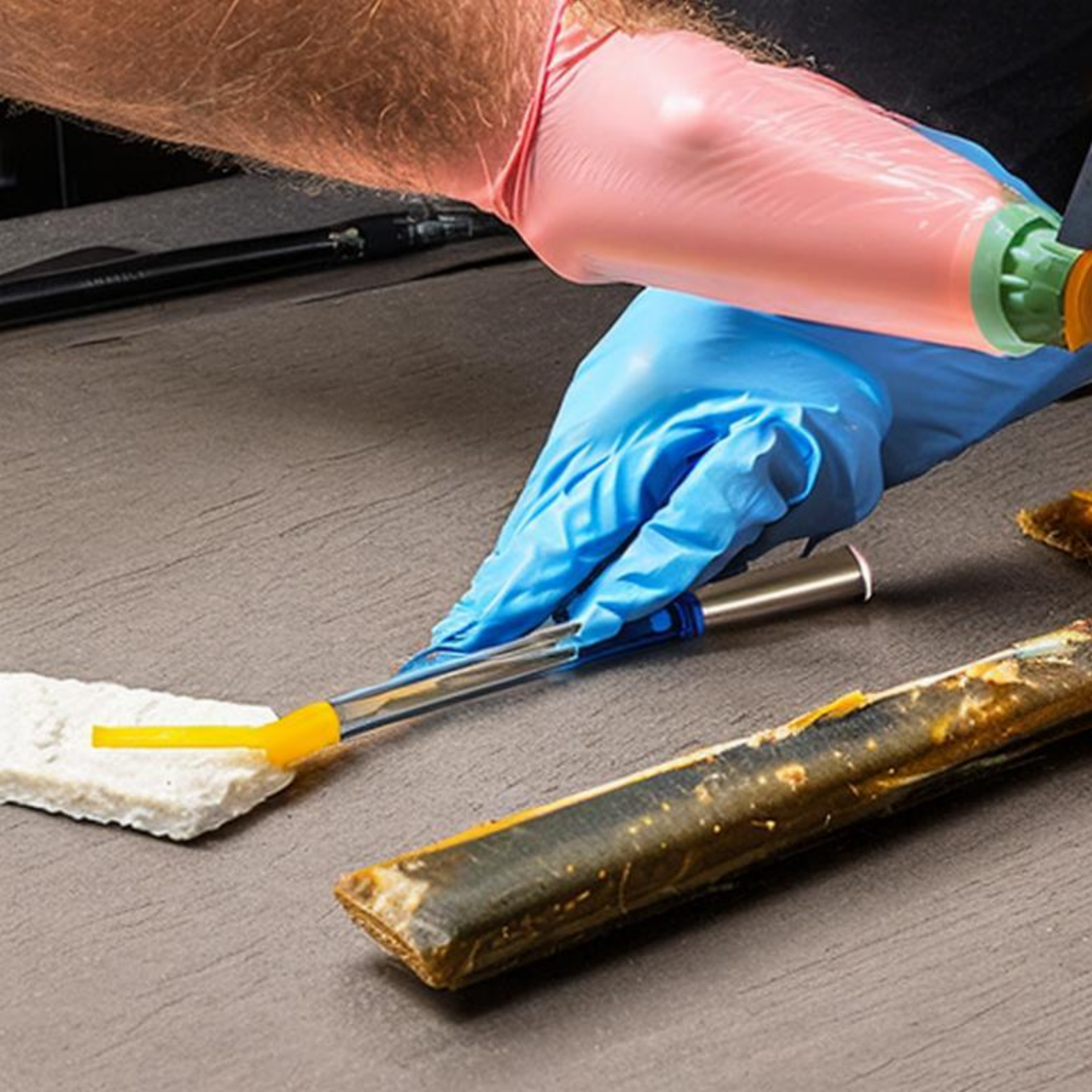} & \includegraphics[width=\linewidth, valign=m]{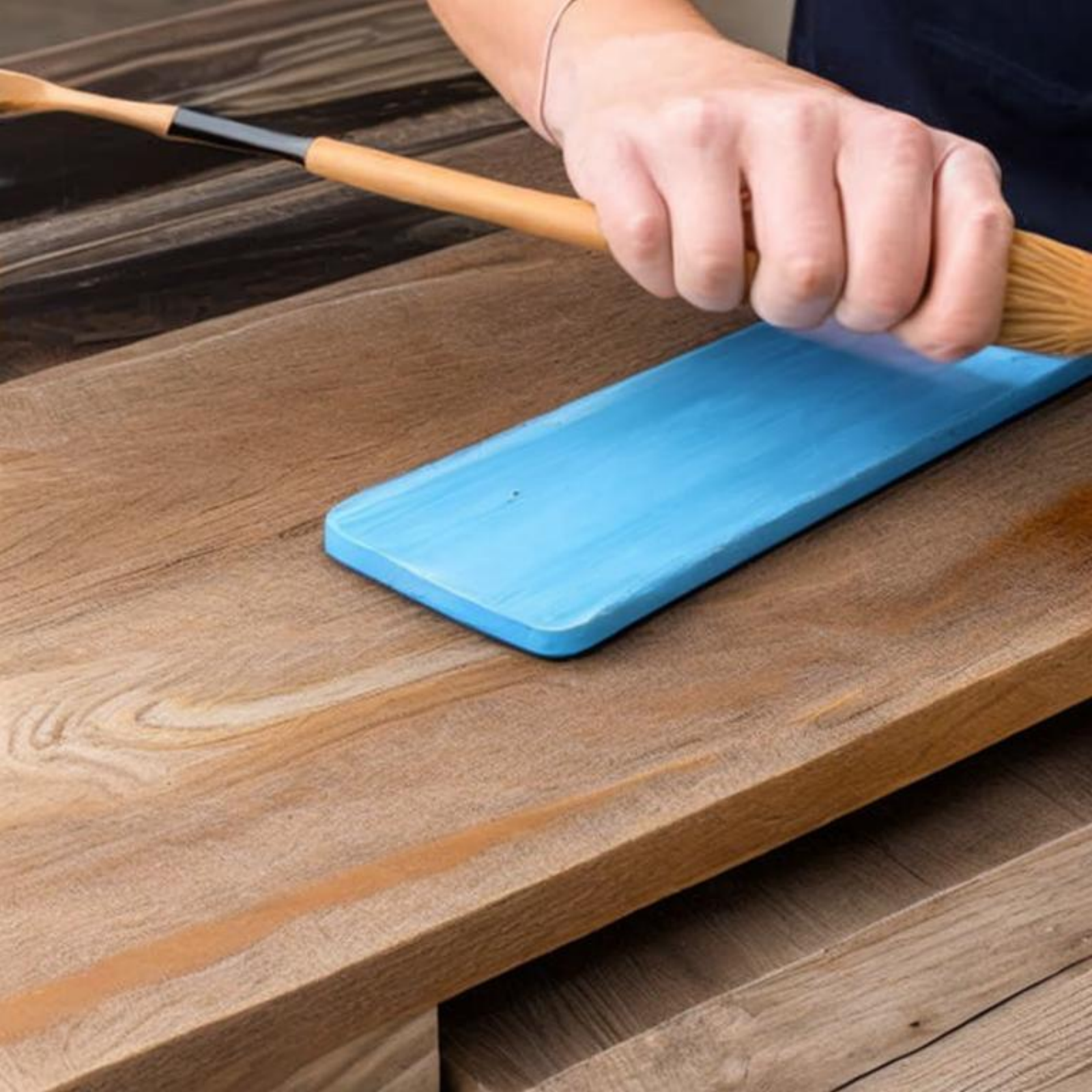} & \includegraphics[width=\linewidth, valign=m]{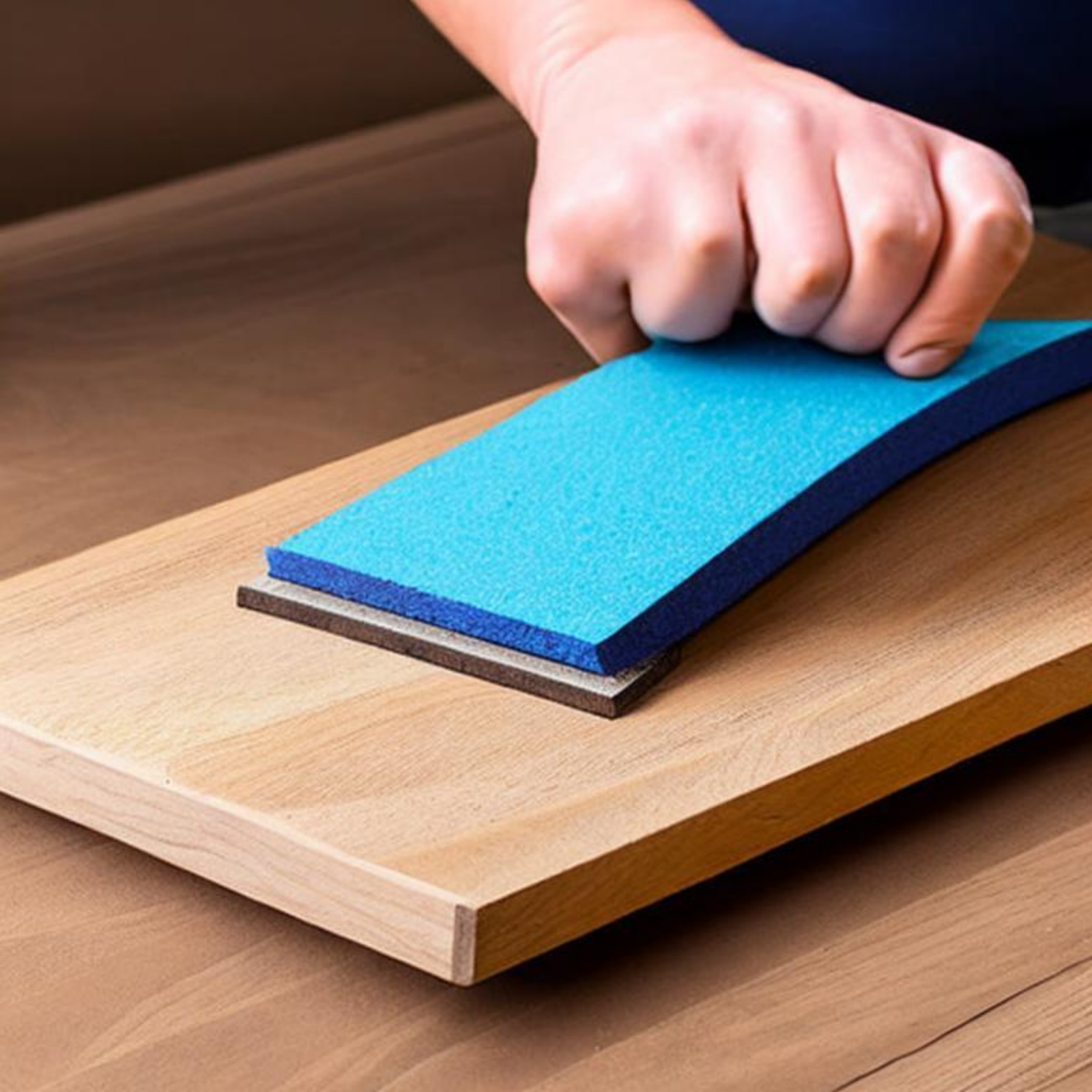} \\ \noalign{\vspace{10pt}}
    \rotatebox[origin=c]{90}{\scriptsize Baseline} & \includegraphics[width=\linewidth, valign=m]{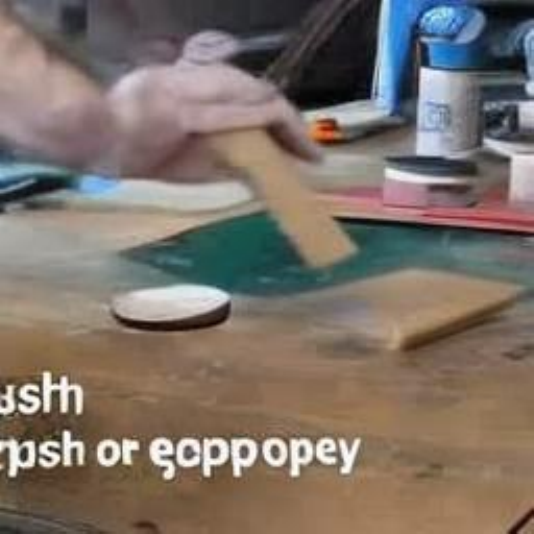} & \includegraphics[width=\linewidth, valign=m]{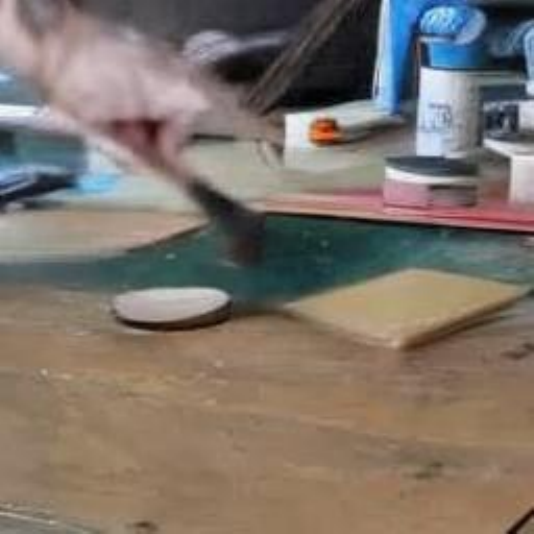} & \includegraphics[width=\linewidth, valign=m]{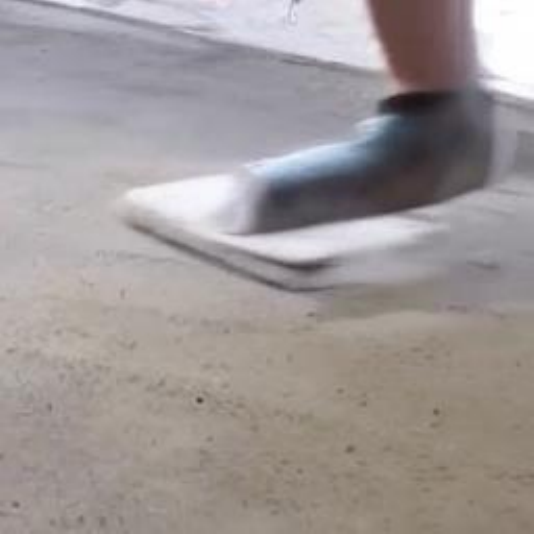} \\ \noalign{\vspace{10pt}}
    \rotatebox[origin=c]{90}{\scriptsize \textbf{InstructionCrafter}} & \includegraphics[width=\linewidth, valign=m]{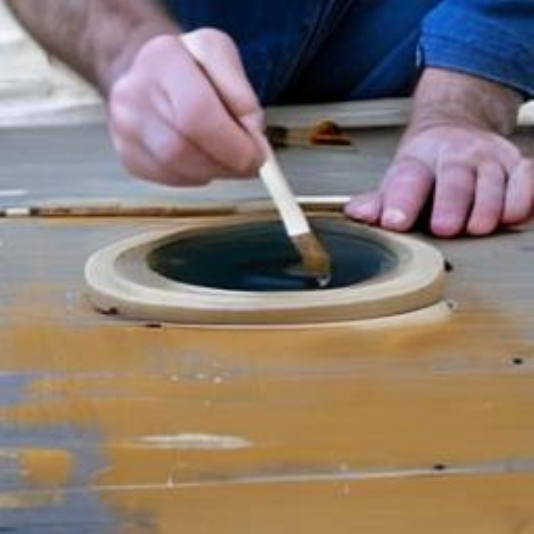} & \includegraphics[width=\linewidth, valign=m]{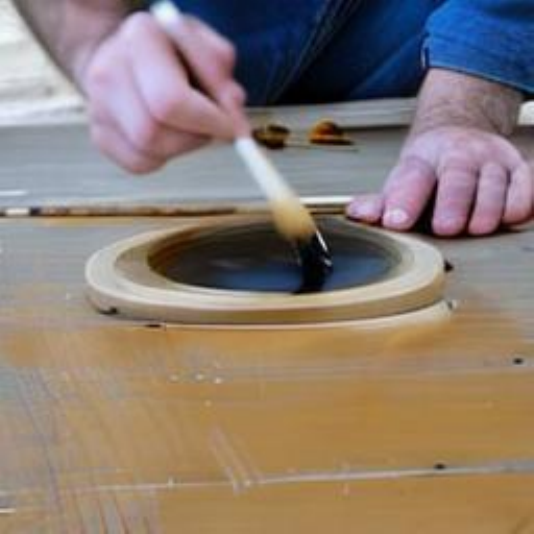} & \includegraphics[width=\linewidth, valign=m]{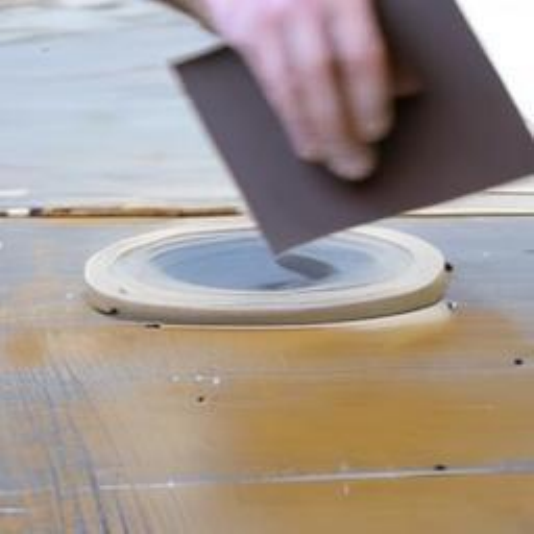} \\
\end{tabular}
  \caption{\textbf{Additional qualitative comparison on WikiHow-VGSI.}}
  \label{fig:sup_wikihow_452}
\end{figure}

\begin{figure}[t]
  \centering
  \setlength{\tabcolsep}{3pt}
  \renewcommand{\arraystretch}{1.2}
  \begin{tabular}{>{\centering\arraybackslash}m{1.5cm} m{0.267\linewidth} m{0.267\linewidth} m{0.267\linewidth}}
     & \tiny\centering Wash the chicken legs. & \tiny\centering Coat the chicken in the butter mixture. & \tiny Serve the chicken. \\
    \rotatebox[origin=c]{90}{\scriptsize Stable Diffusion} & \includegraphics[width=\linewidth, valign=m]{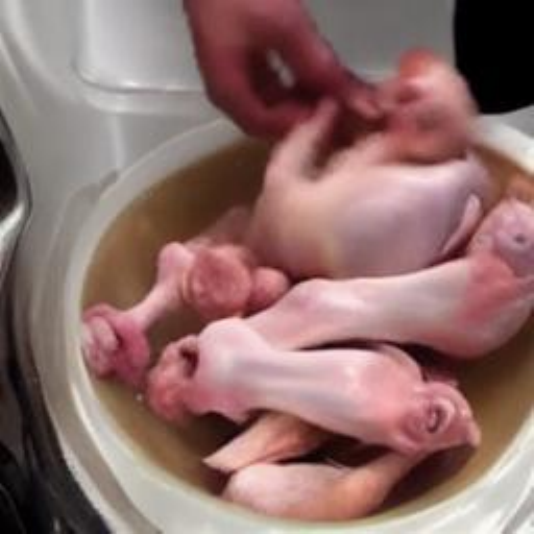} & \includegraphics[width=\linewidth, valign=m]{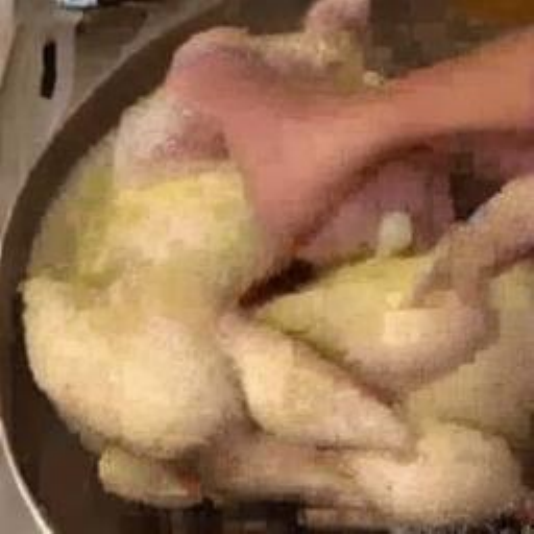} & \includegraphics[width=\linewidth, valign=m]{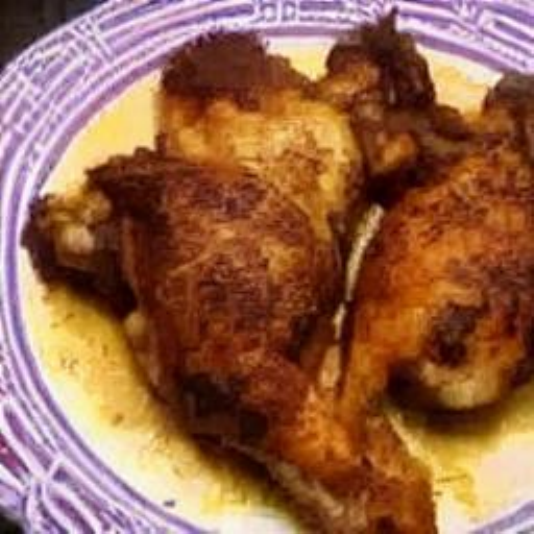} \\ \noalign{\vspace{10pt}}
    \rotatebox[origin=c]{90}{\scriptsize StackedDiffusion} & \includegraphics[width=\linewidth, valign=m]{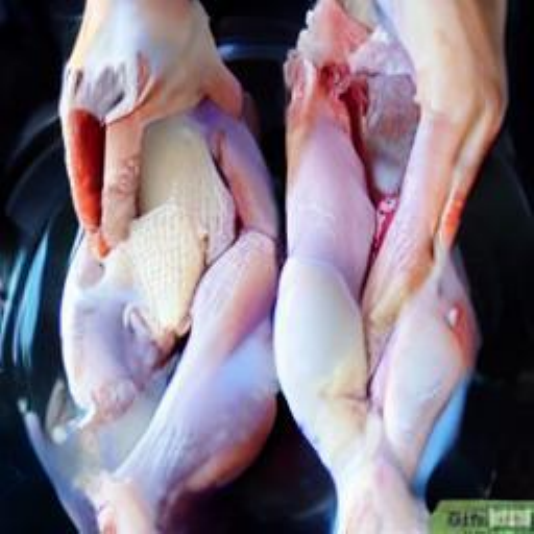} & \includegraphics[width=\linewidth, valign=m]{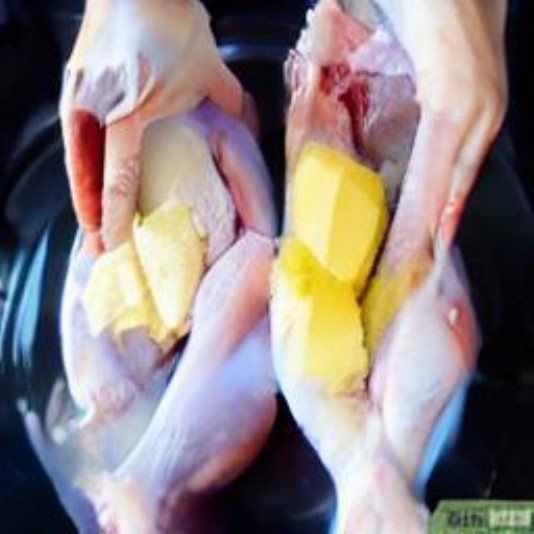} & \includegraphics[width=\linewidth, valign=m]{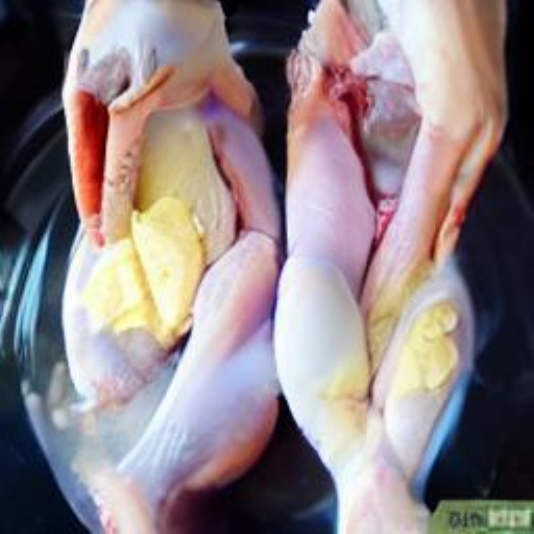} \\ \noalign{\vspace{10pt}}
    \rotatebox[origin=c]{90}{\scriptsize Bordalo et al.} & \includegraphics[width=\linewidth, valign=m]{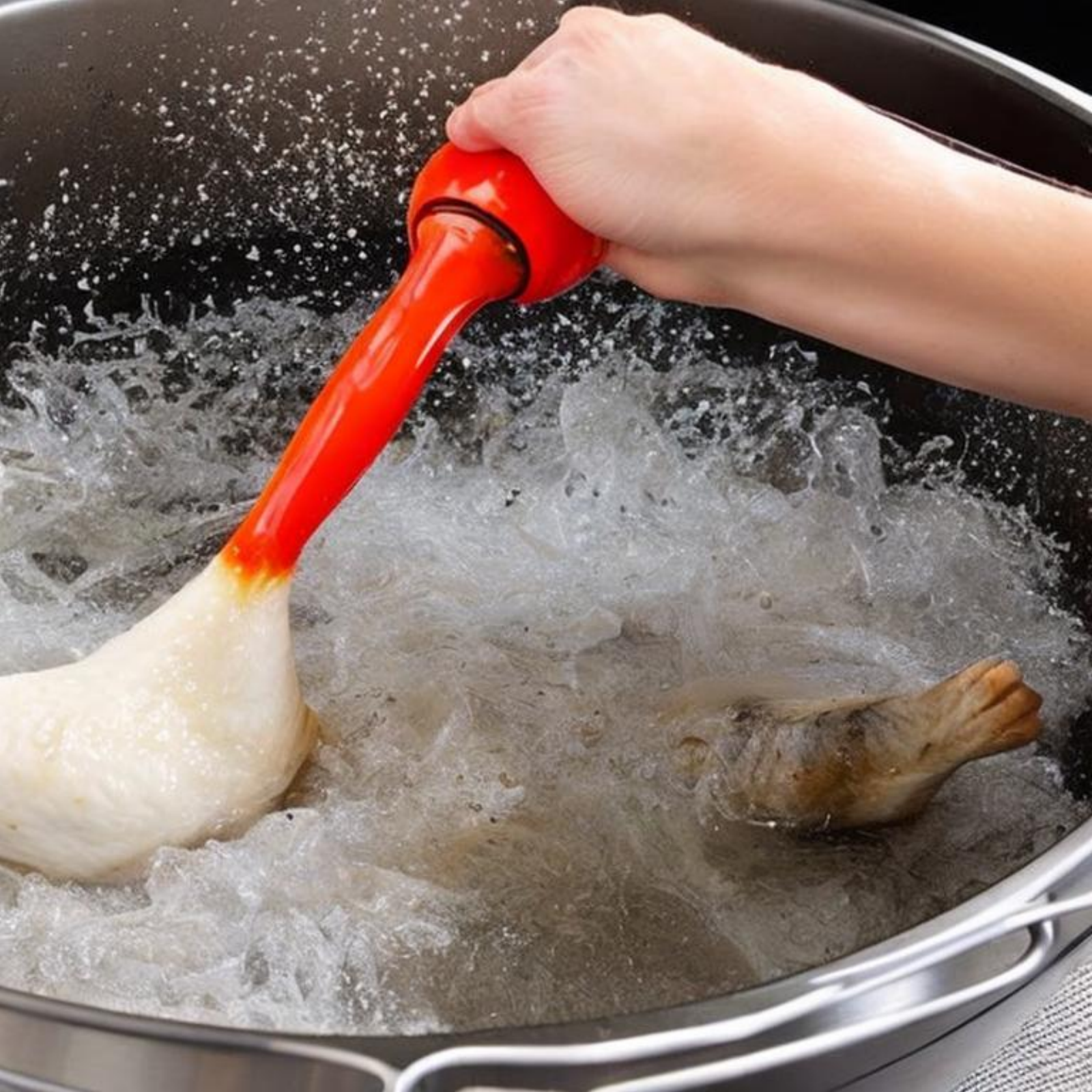} & \includegraphics[width=\linewidth, valign=m]{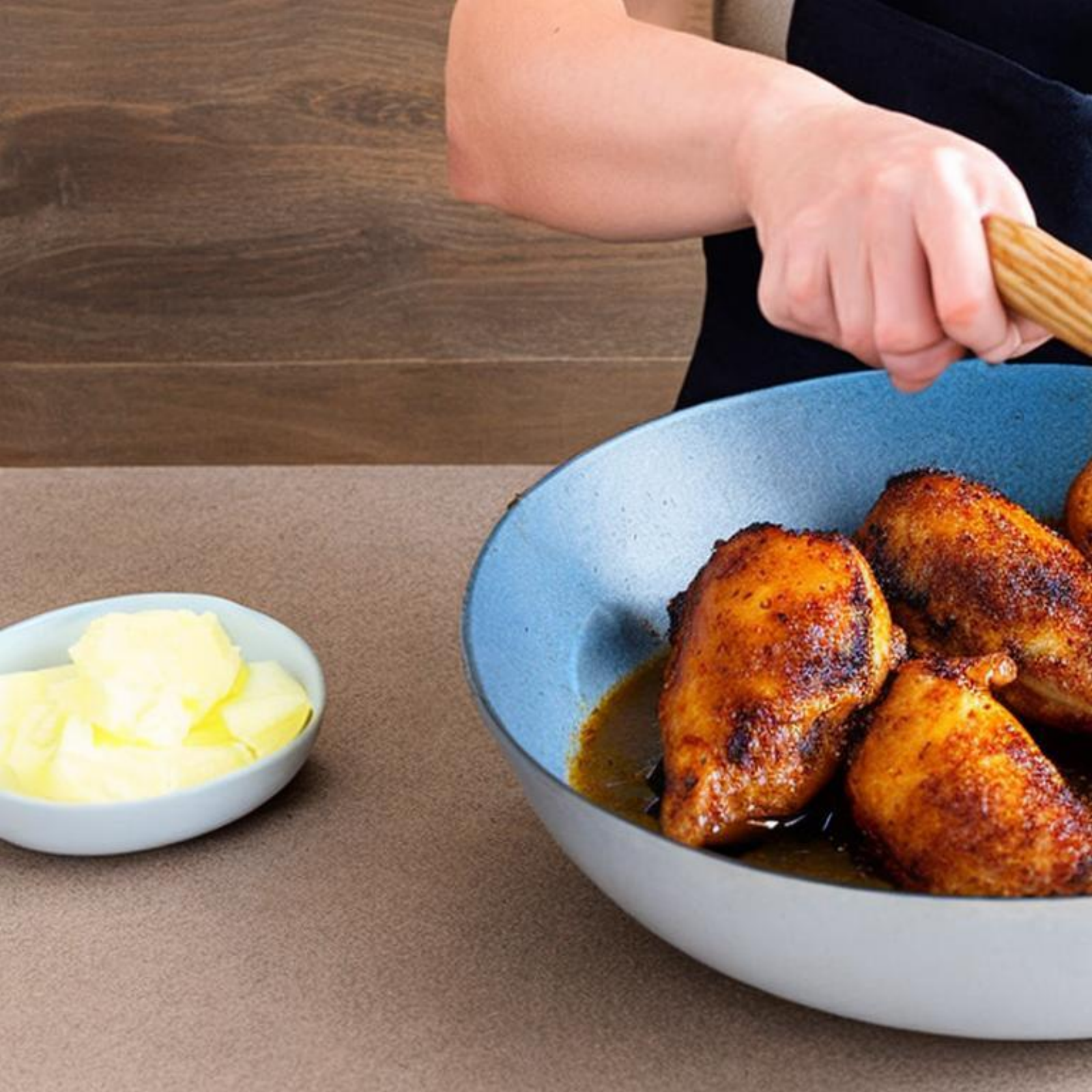} & \includegraphics[width=\linewidth, valign=m]{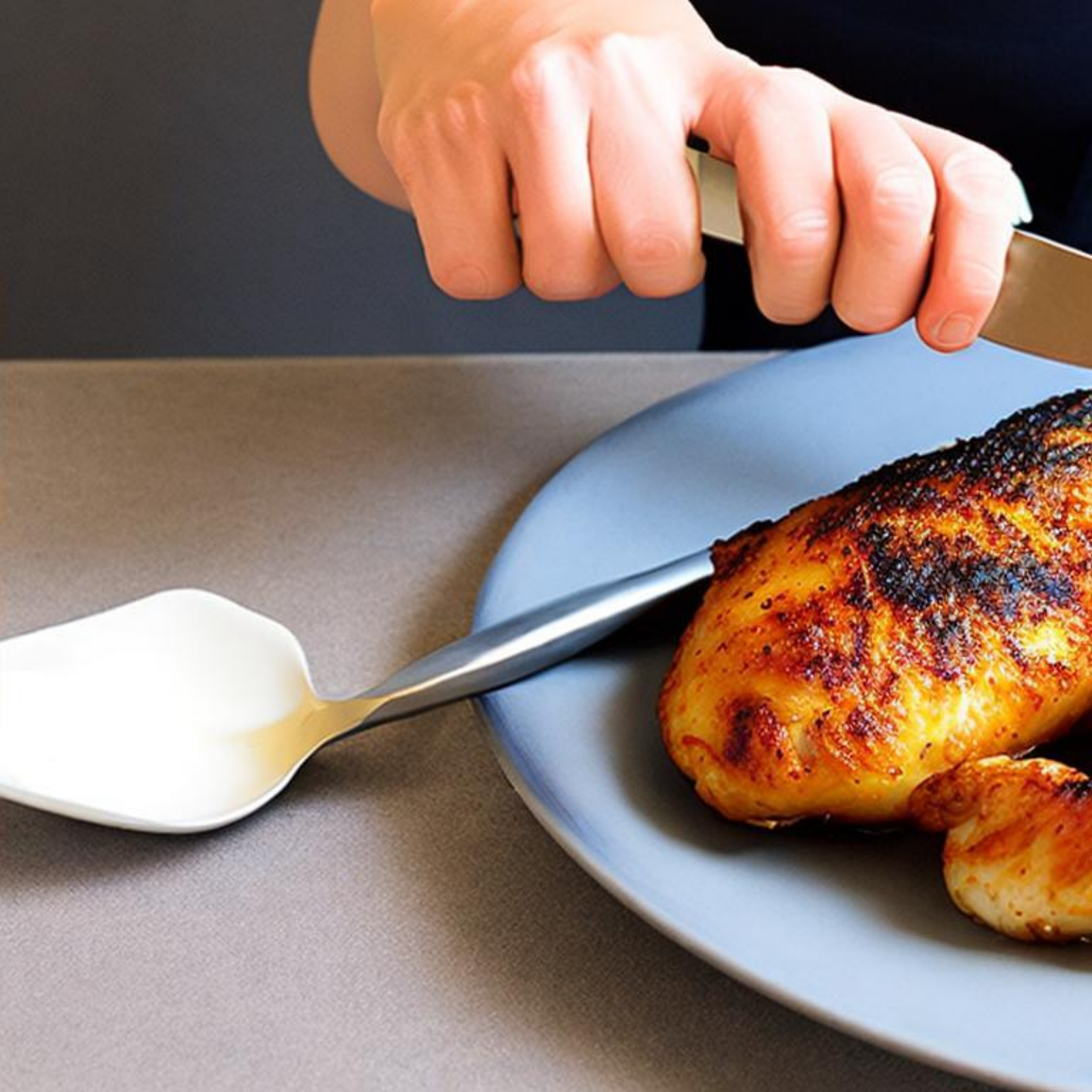} \\ \noalign{\vspace{10pt}}
    \rotatebox[origin=c]{90}{\scriptsize Baseline} & \includegraphics[width=\linewidth, valign=m]{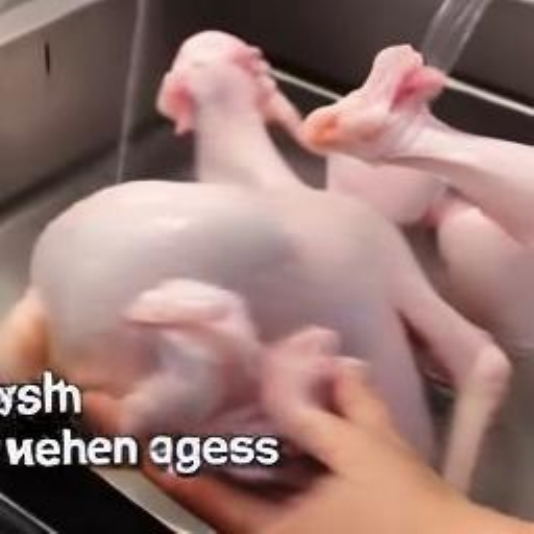} & \includegraphics[width=\linewidth, valign=m]{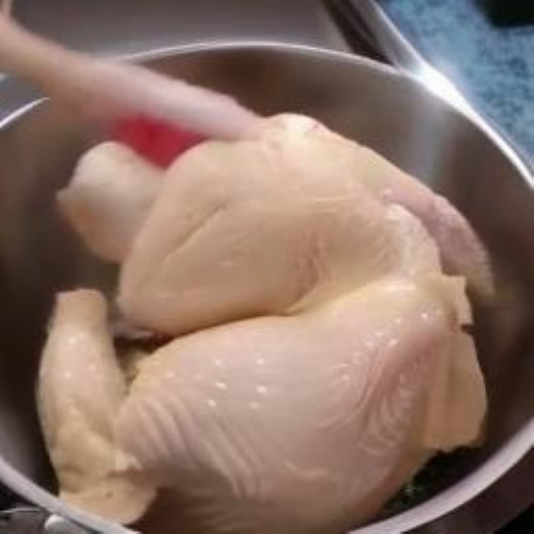} & \includegraphics[width=\linewidth, valign=m]{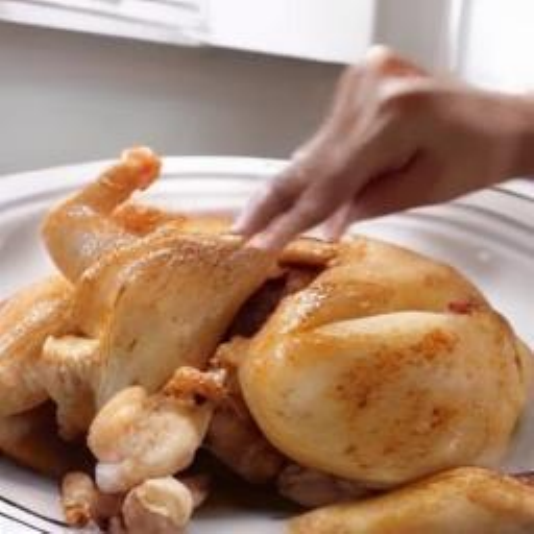} \\ \noalign{\vspace{10pt}}
    \rotatebox[origin=c]{90}{\scriptsize \textbf{InstructionCrafter}} & \includegraphics[width=\linewidth, valign=m]{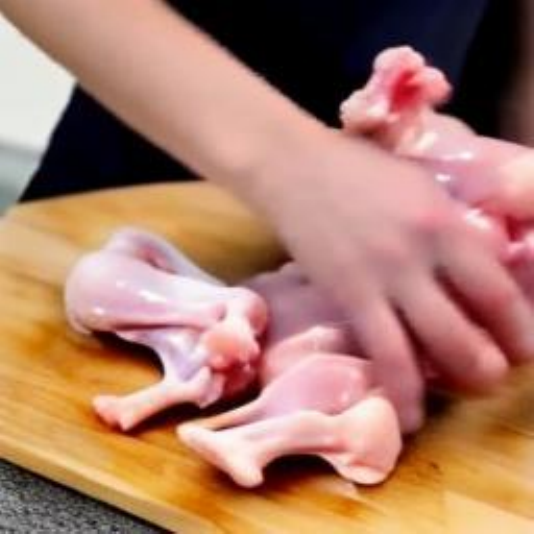} & \includegraphics[width=\linewidth, valign=m]{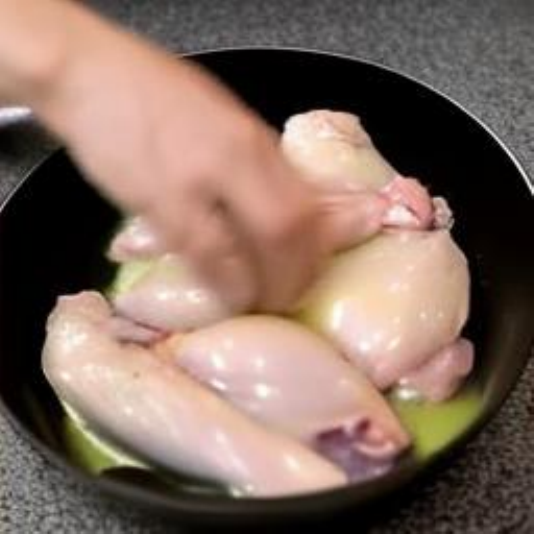} & \includegraphics[width=\linewidth, valign=m]{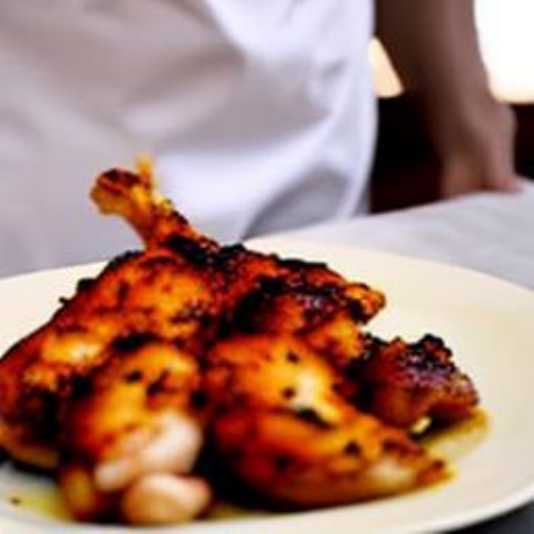} \\ \noalign{\vspace{10pt}}
  \end{tabular}
  \caption{\textbf{Additional qualitative comparison on WikiHow-VGSI.}}
  \label{fig:sup_wikihow_759}
\end{figure}

\begin{figure*}[t]
  \centering
  \setlength{\tabcolsep}{3pt}
  \renewcommand{\arraystretch}{1.2}
  \begin{tabular}{>{\centering\arraybackslash}m{1.5cm} m{0.18\linewidth} m{0.18\linewidth} m{0.18\linewidth}}
     & \tiny\centering Steep tea with other marinade ingredients. & \tiny\centering Continue marinating at room temperature or in the fridge. & \tiny Serve in the shell. \\
    \rotatebox[origin=c]{90}{\scriptsize Stable Diffusion} & \includegraphics[width=\linewidth, valign=m]{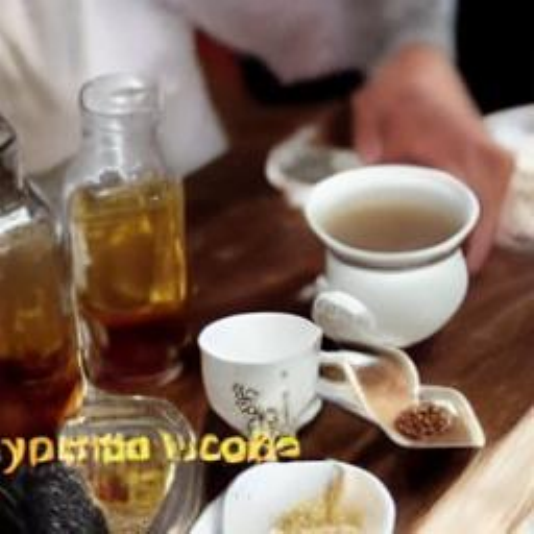} & \includegraphics[width=\linewidth, valign=m]{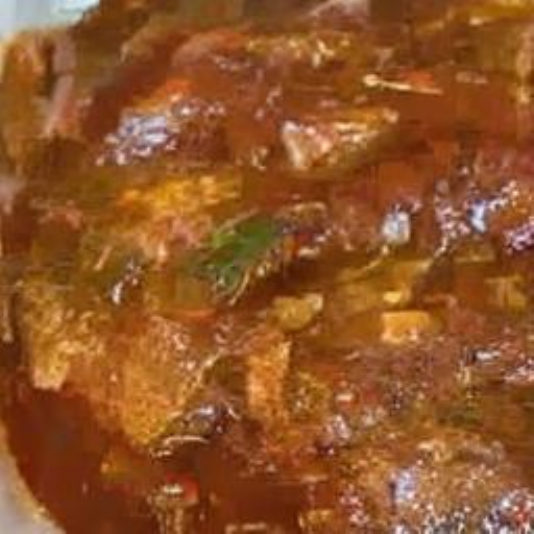} & \includegraphics[width=\linewidth, valign=m]{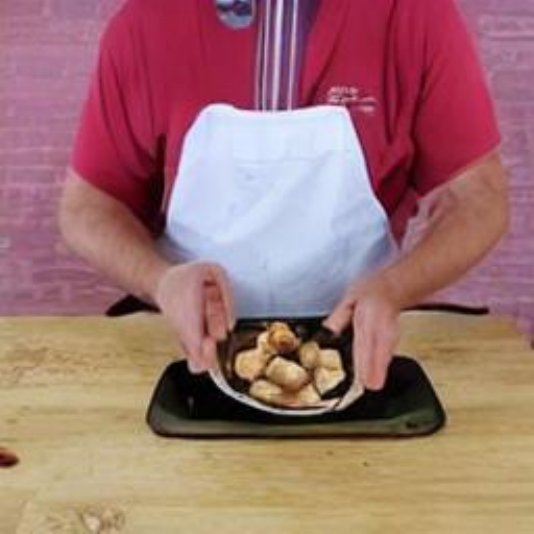} \\ \noalign{\vspace{10pt}}
    \rotatebox[origin=c]{90}{\scriptsize StackedDiffusion} & \includegraphics[width=\linewidth, valign=m]{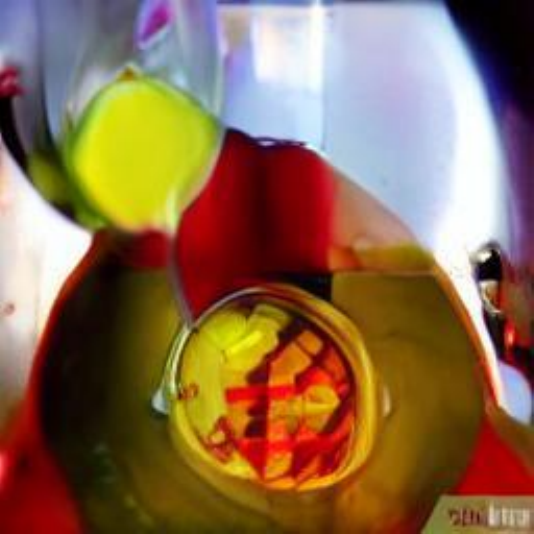} & \includegraphics[width=\linewidth, valign=m]{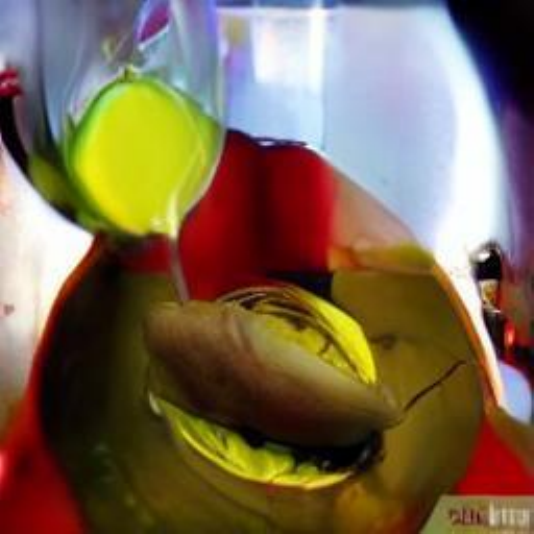} & \includegraphics[width=\linewidth, valign=m]{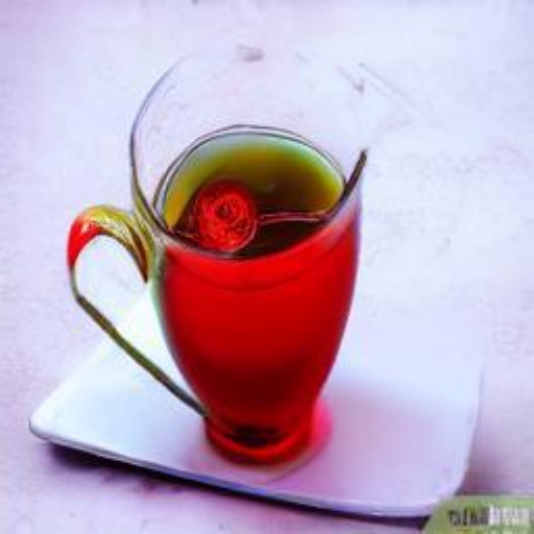} \\ \noalign{\vspace{10pt}}
    \rotatebox[origin=c]{90}{\scriptsize Bordalo et al.} & \includegraphics[width=\linewidth, valign=m]{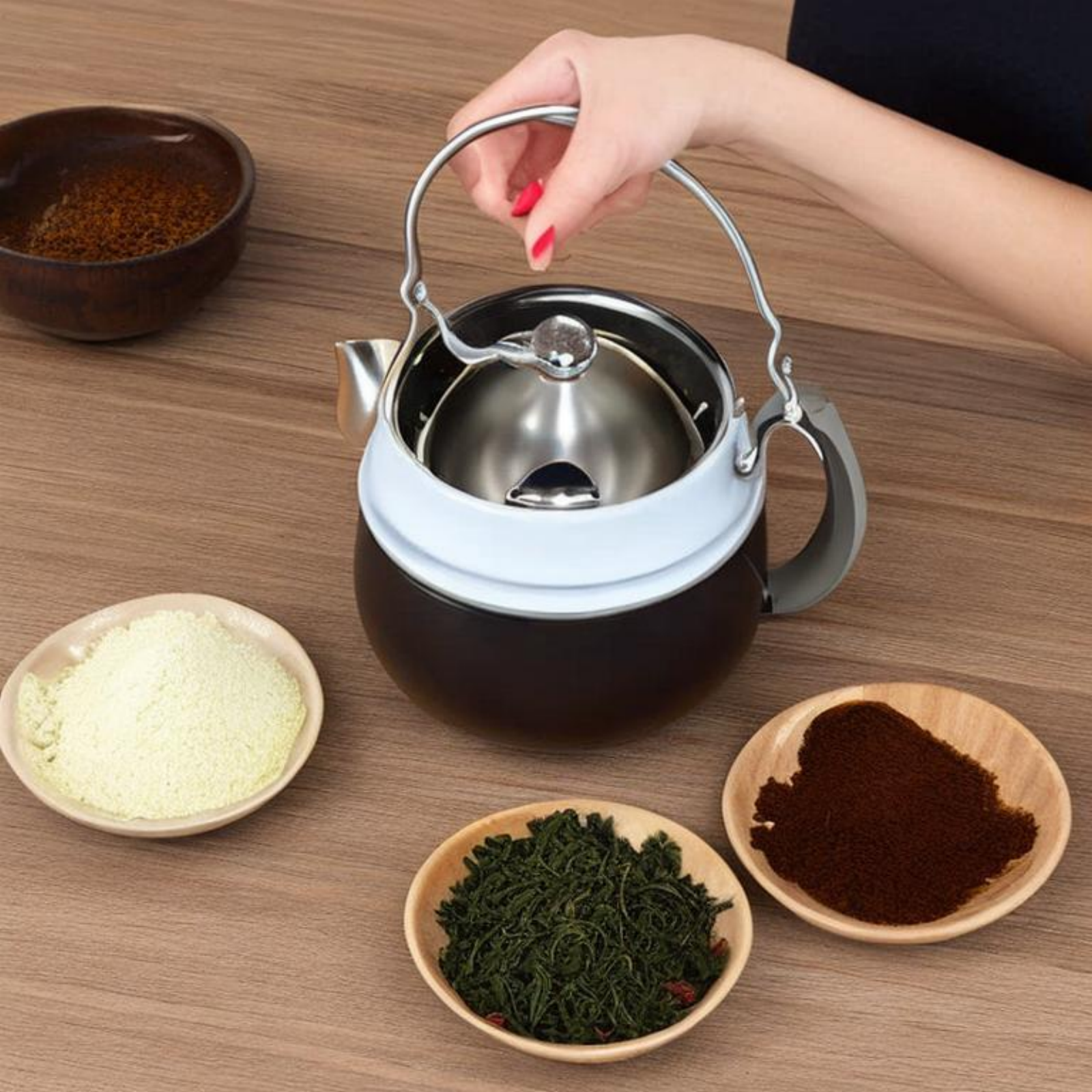} & \includegraphics[width=\linewidth, valign=m]{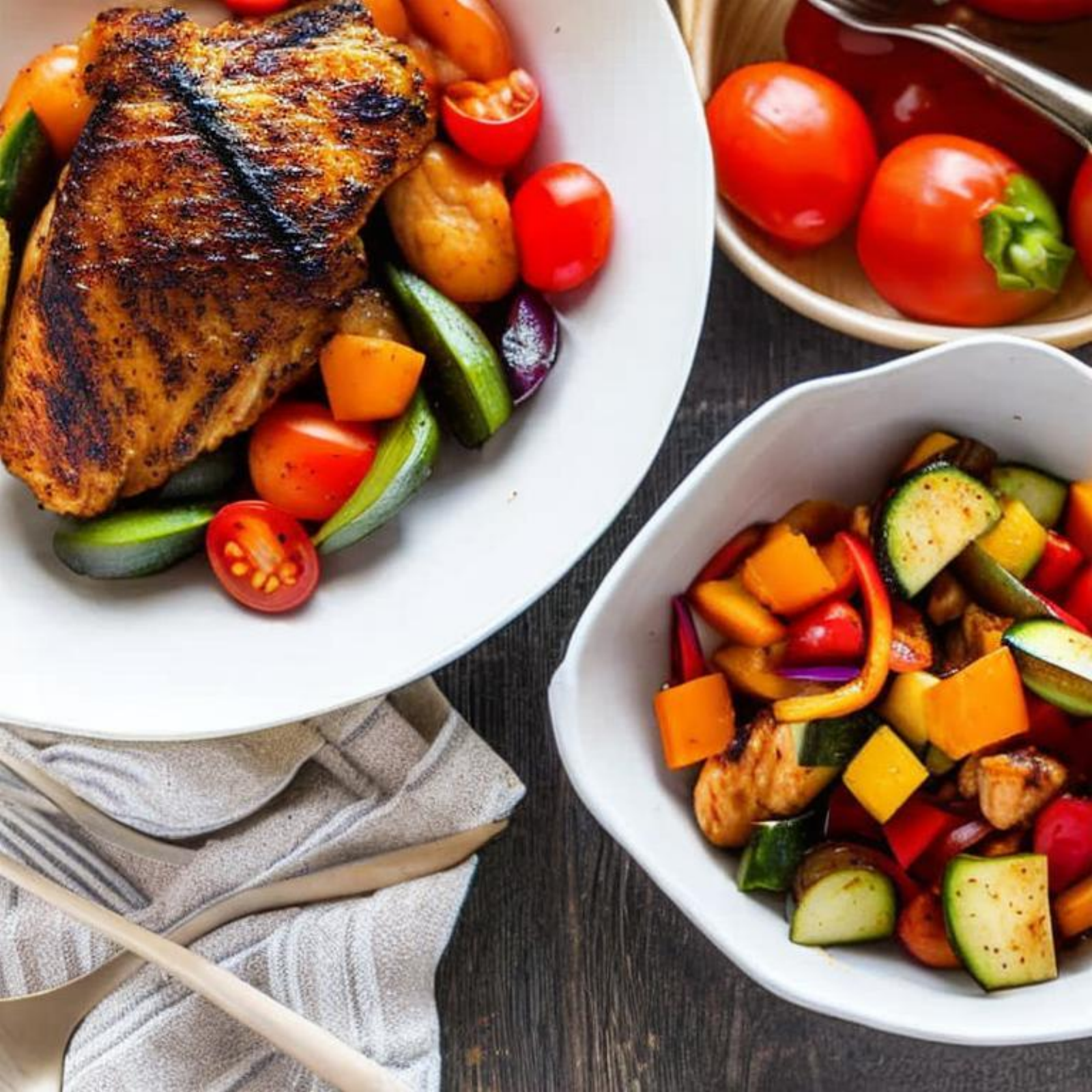} & \includegraphics[width=\linewidth, valign=m]{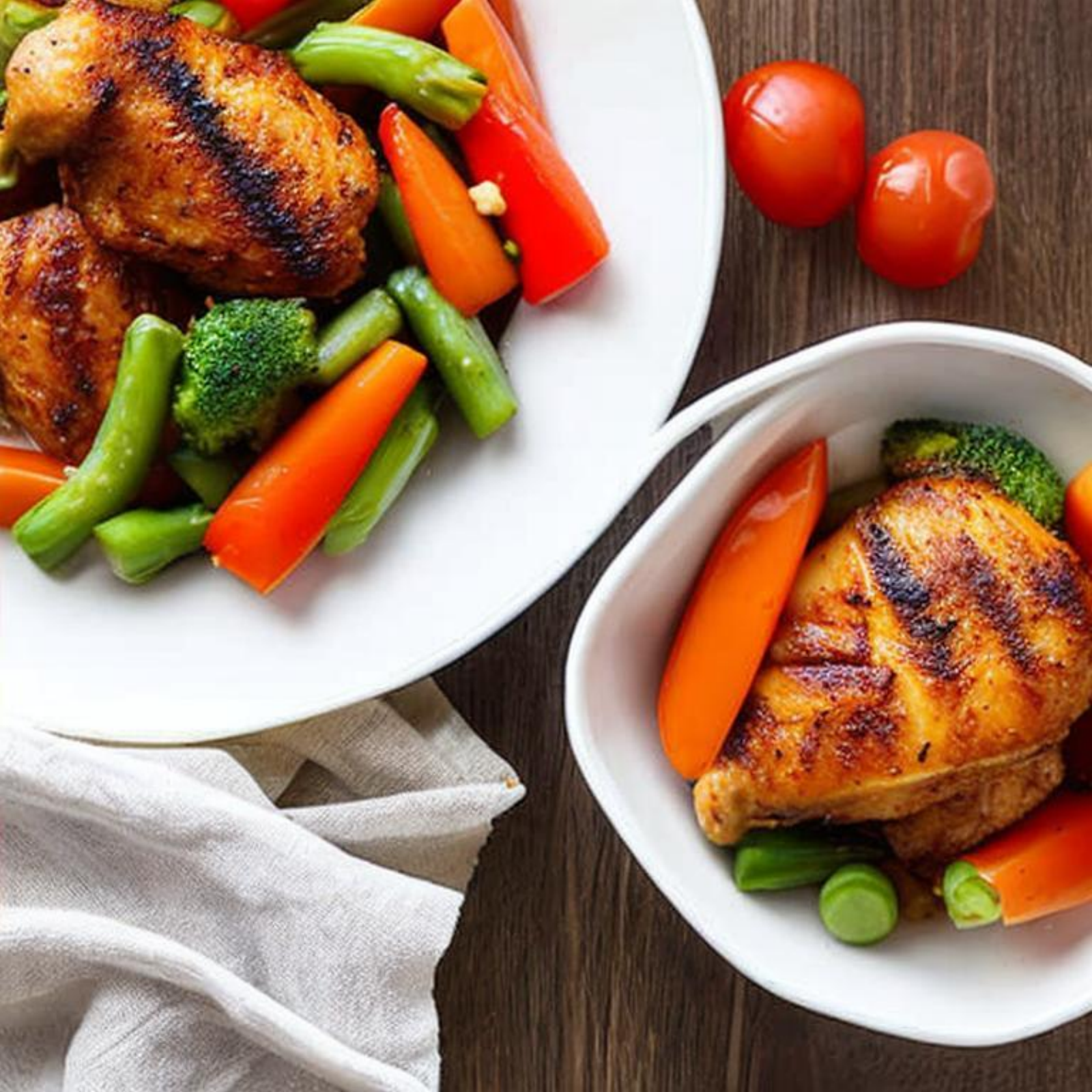} \\ \noalign{\vspace{10pt}}
    \rotatebox[origin=c]{90}{\scriptsize Baseline} & \includegraphics[width=\linewidth, valign=m]{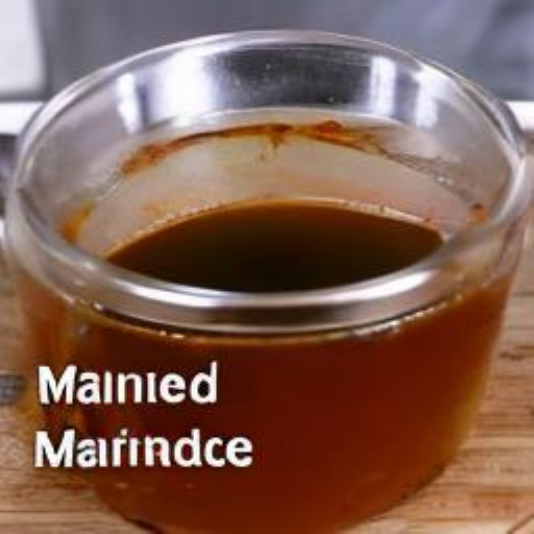} & \includegraphics[width=\linewidth, valign=m]{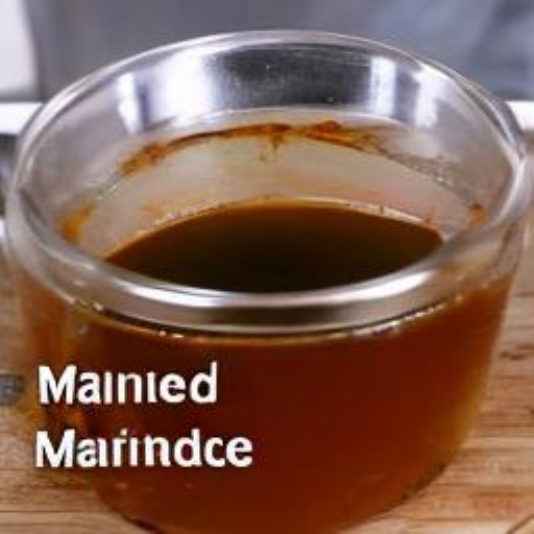} & \includegraphics[width=\linewidth, valign=m]{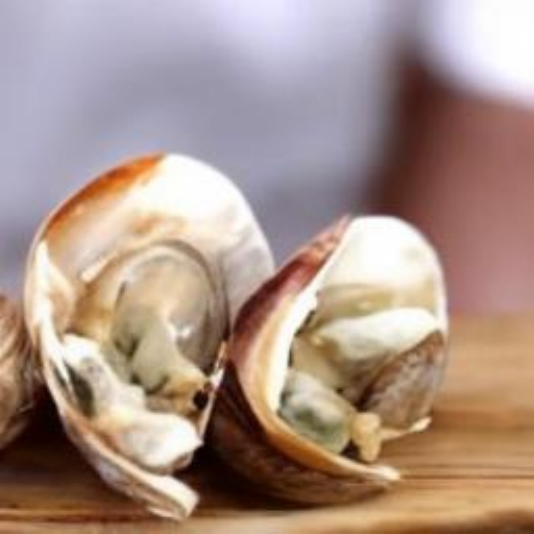} \\ \noalign{\vspace{10pt}}
    \rotatebox[origin=c]{90}{\scriptsize \textbf{InstructionCrafter}} & \includegraphics[width=\linewidth, valign=m]{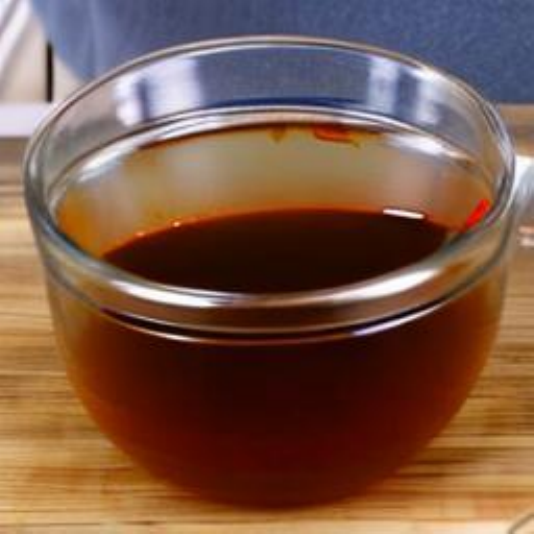} & \includegraphics[width=\linewidth, valign=m]{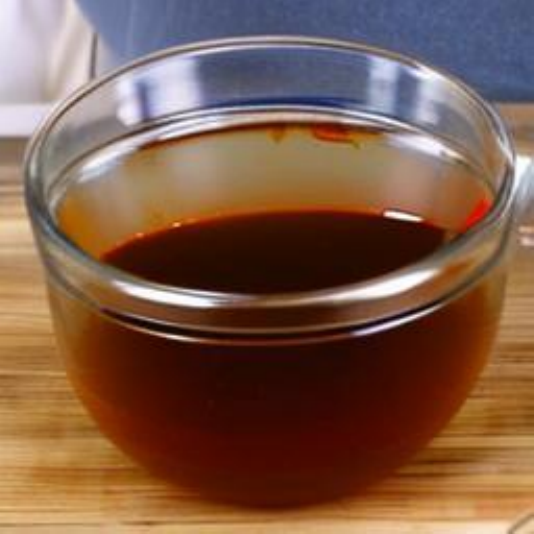} & \includegraphics[width=\linewidth, valign=m]{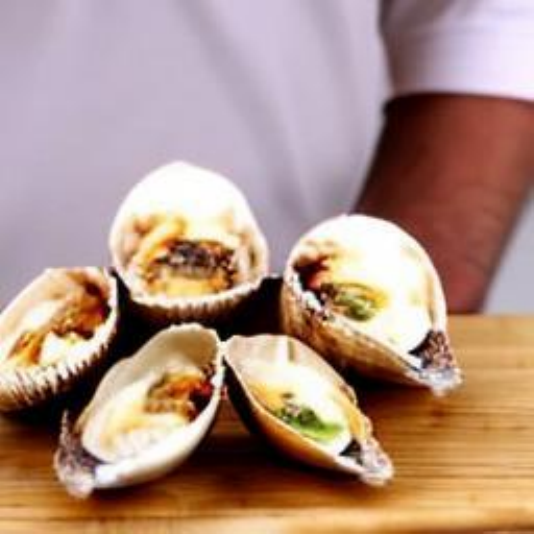} \\ \noalign{\vspace{10pt}}
  \end{tabular}
  \caption{\textbf{Additional qualitative comparison on WikiHow-VGSI.}}
  \label{fig:sup_wikihow_840}
\end{figure*}

\begin{figure*}[t]
  \centering
  \setlength{\tabcolsep}{3pt}
  \renewcommand{\arraystretch}{1.2}
  \begin{tabular}{>{\centering\arraybackslash}m{1.5cm} m{0.200\linewidth} m{0.200\linewidth} m{0.200\linewidth} m{0.200\linewidth}}
     & \tiny\centering Buff with wet 200 grit abrasive. & \tiny\centering Rinse all materials. & \tiny\centering Repeat with 400 grit abrasive. & \tiny Rinse again. \\
    \rotatebox[origin=c]{90}{\scriptsize Stable Diffusion} & \includegraphics[width=\linewidth, valign=m]{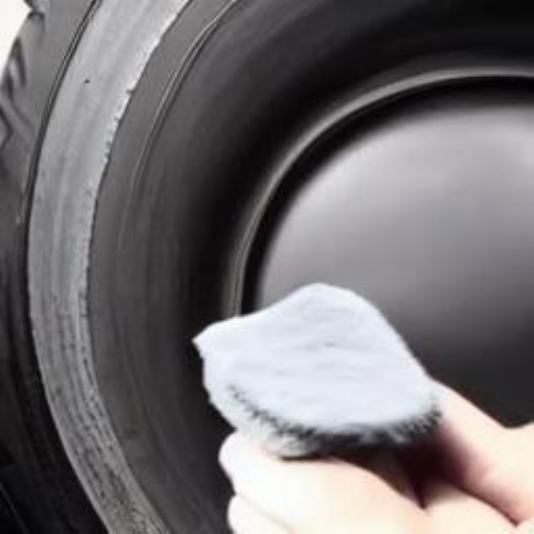} & \includegraphics[width=\linewidth, valign=m]{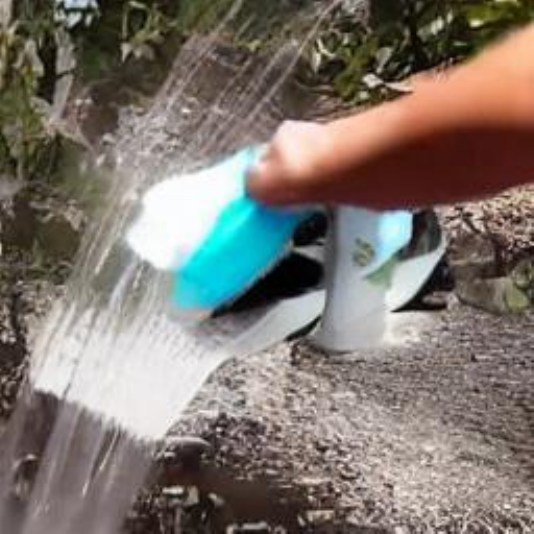} & \includegraphics[width=\linewidth, valign=m]{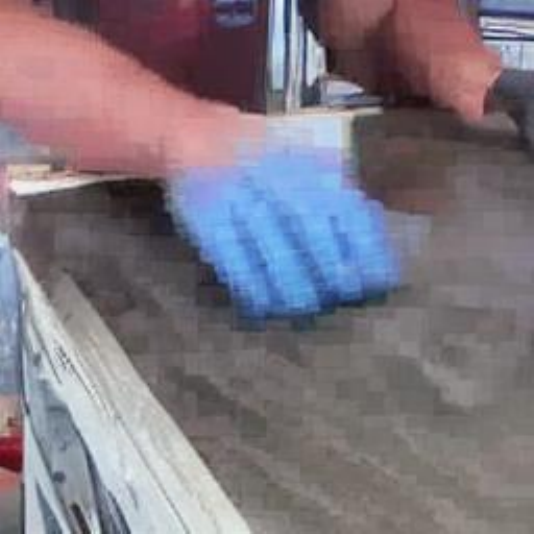} & \includegraphics[width=\linewidth, valign=m]{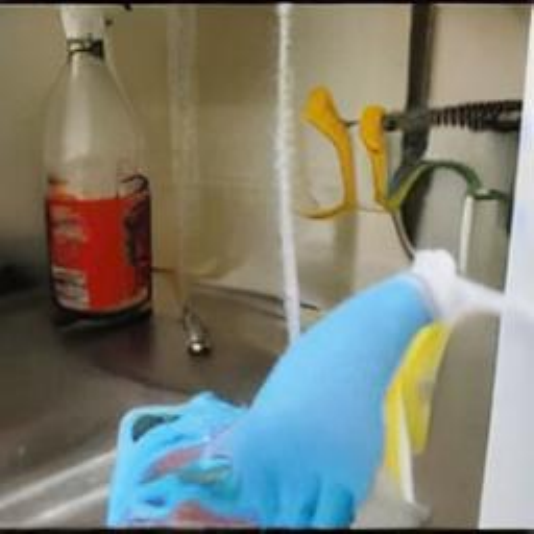} \\ \noalign{\vspace{10pt}}
    \rotatebox[origin=c]{90}{\scriptsize StackedDiffusion} & \includegraphics[width=\linewidth, valign=m]{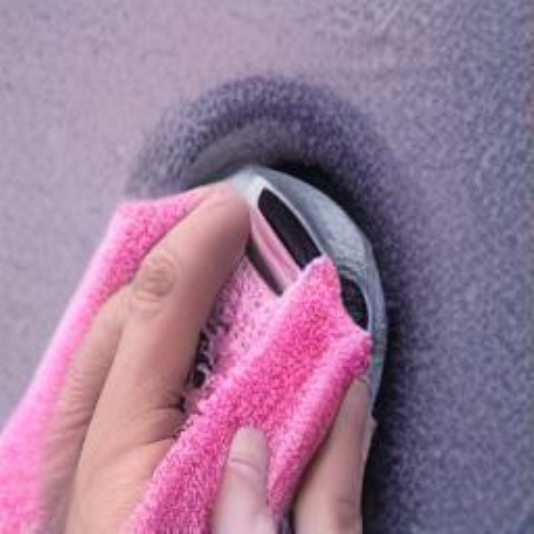} & \includegraphics[width=\linewidth, valign=m]{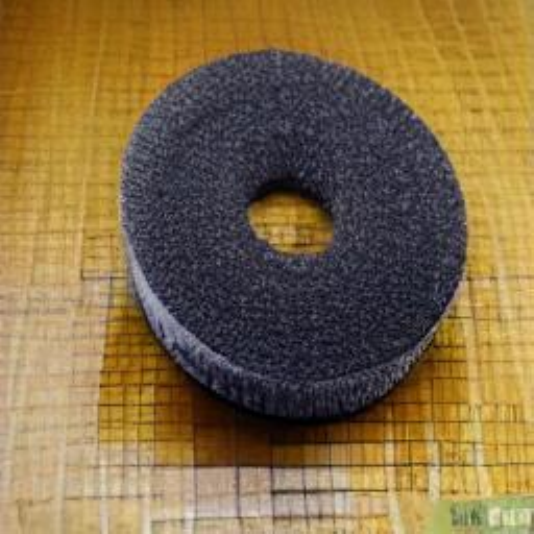} & \includegraphics[width=\linewidth, valign=m]{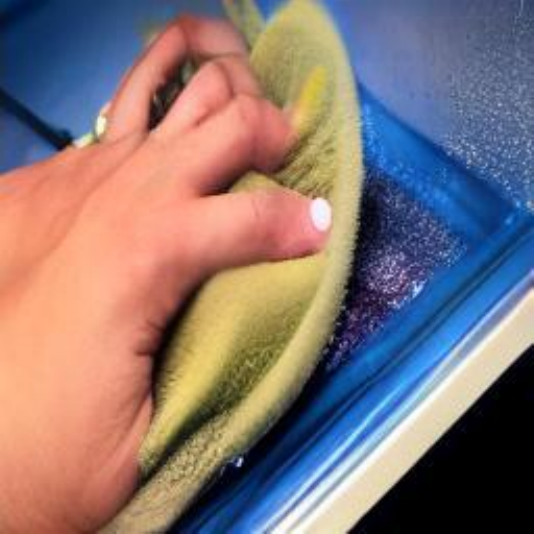} & \includegraphics[width=\linewidth, valign=m]{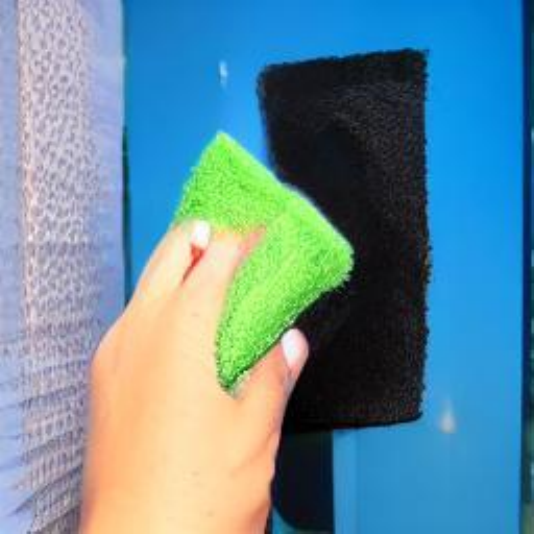} \\ \noalign{\vspace{10pt}}
    \rotatebox[origin=c]{90}{\scriptsize Bordalo et al.} & \includegraphics[width=\linewidth, valign=m]{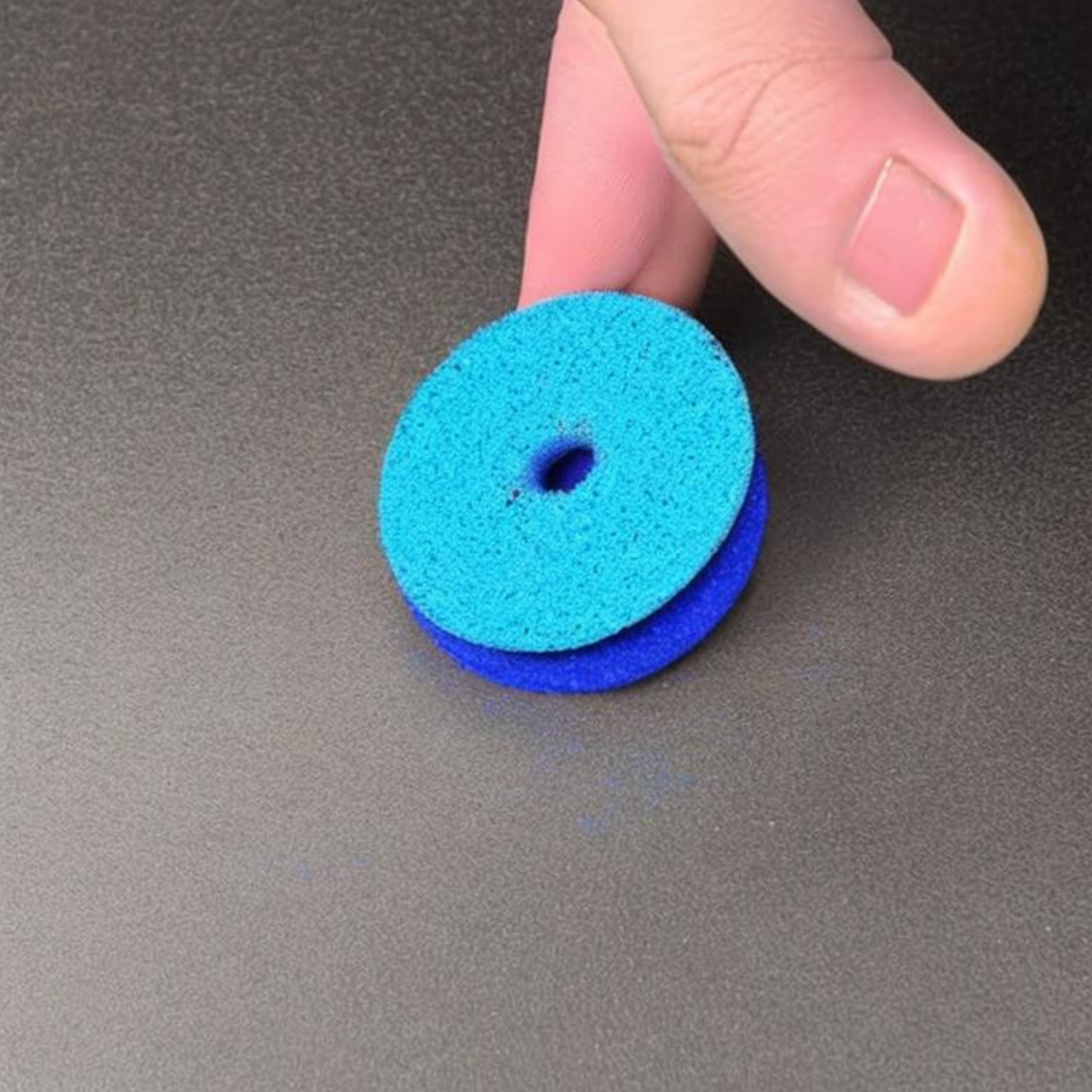} & \includegraphics[width=\linewidth, valign=m]{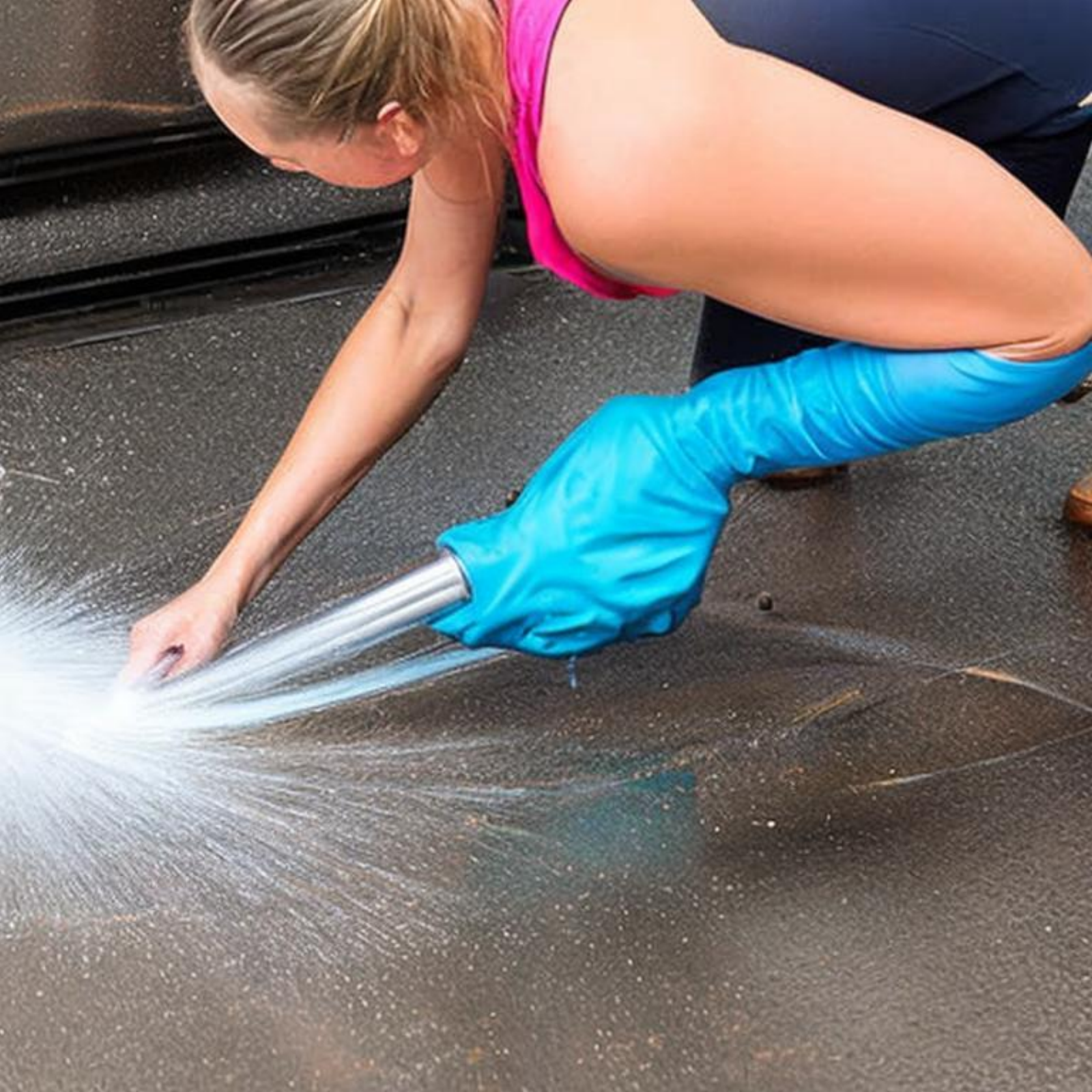} & \includegraphics[width=\linewidth, valign=m]{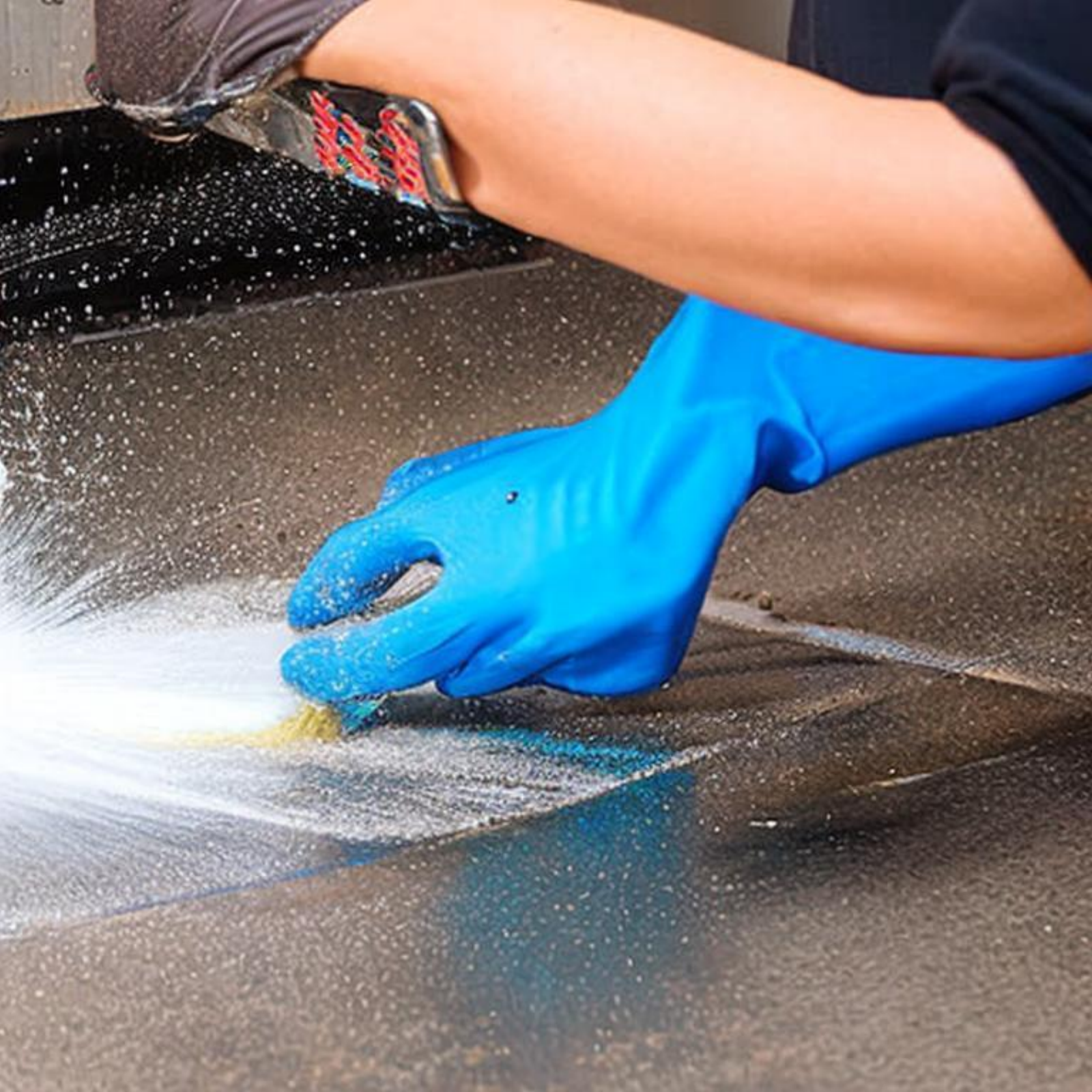} & \includegraphics[width=\linewidth, valign=m]{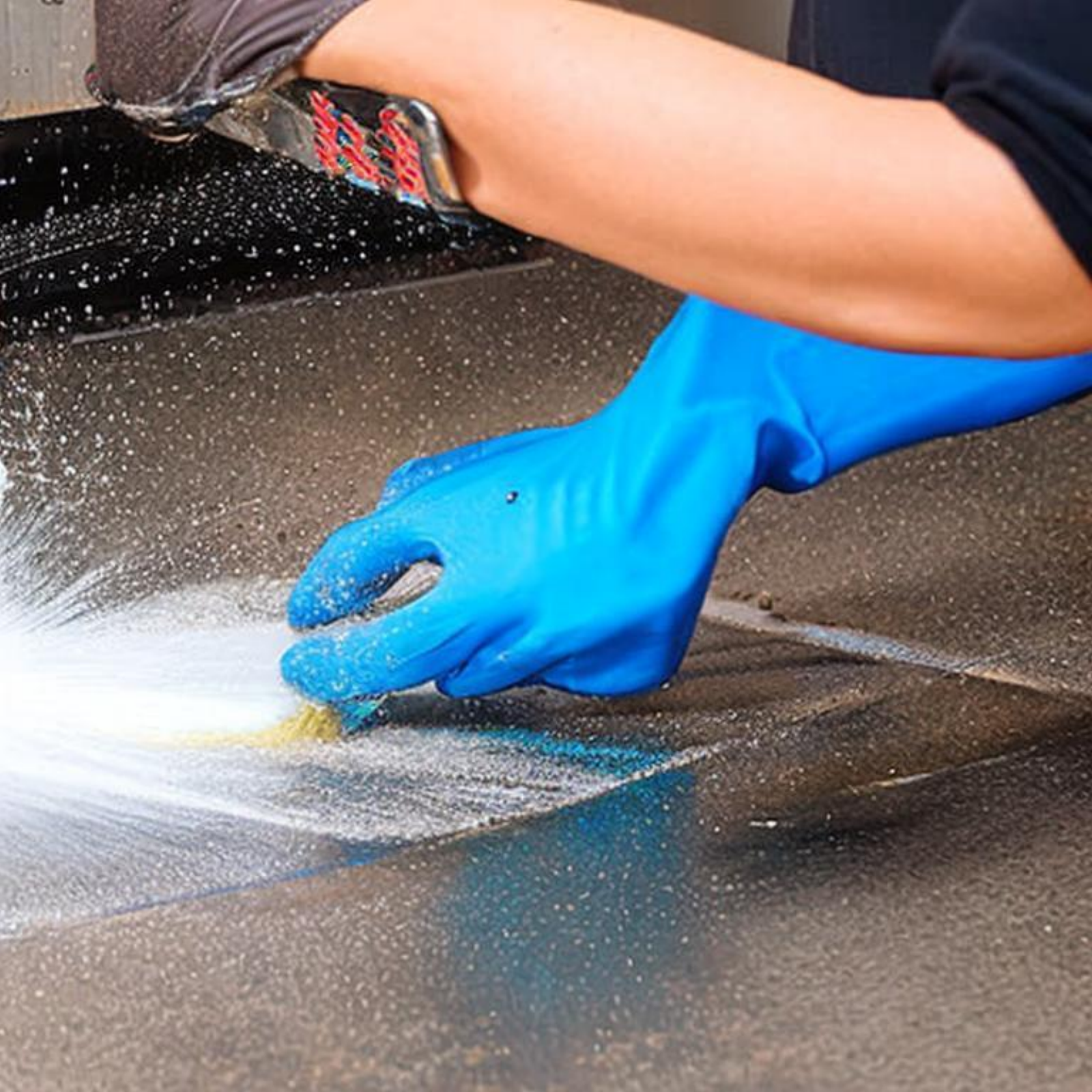} \\ \noalign{\vspace{10pt}}
    \rotatebox[origin=c]{90}{\scriptsize Baseline} & \includegraphics[width=\linewidth, valign=m]{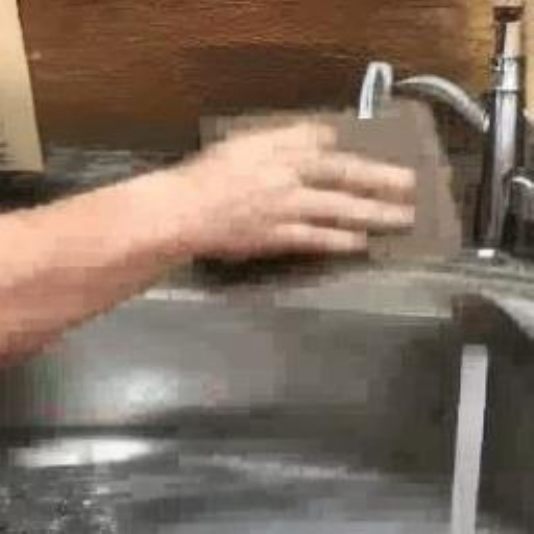} & \includegraphics[width=\linewidth, valign=m]{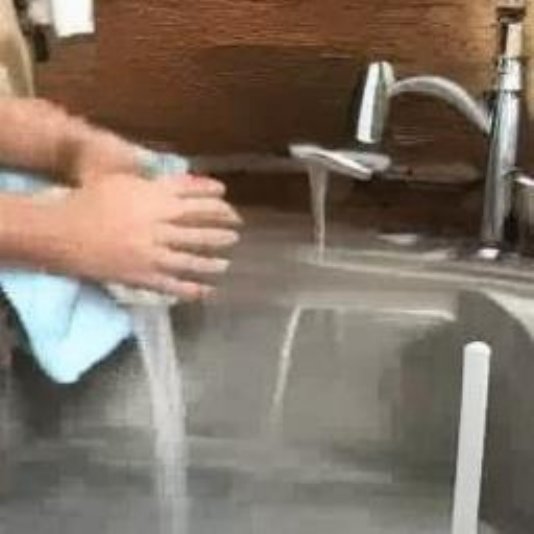} & \includegraphics[width=\linewidth, valign=m]{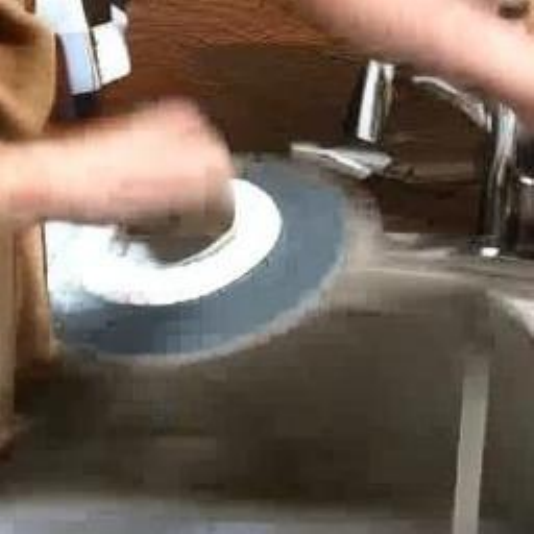} & \includegraphics[width=\linewidth, valign=m]{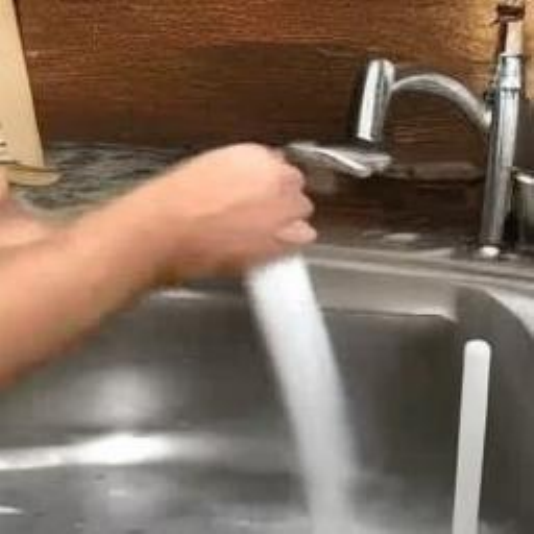} \\
\end{tabular}
  \caption{\textbf{Additional qualitative comparison on WikiHow-VGSI.}}
  \label{fig:sup_wikihow_893}
\end{figure*}

\begin{figure}[t]
  \centering
  \setlength{\tabcolsep}{3pt}
  \renewcommand{\arraystretch}{1.2}
  \begin{tabular}{>{\centering\arraybackslash}m{1.5cm} m{0.267\linewidth} m{0.267\linewidth} m{0.267\linewidth}}
     & \tiny\centering Combine the paste ingredients and grind. & \tiny\centering Cut the meat into bite sized pieces. & \tiny Garnish and serve. \\
    \rotatebox[origin=c]{90}{\scriptsize Stable Diffusion} & \includegraphics[width=\linewidth, valign=m]{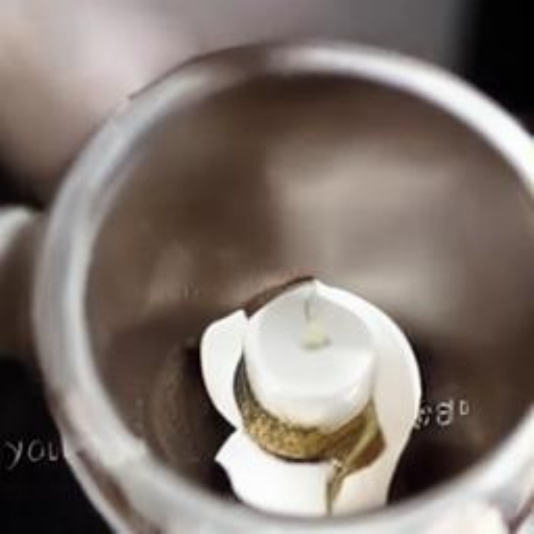} & \includegraphics[width=\linewidth, valign=m]{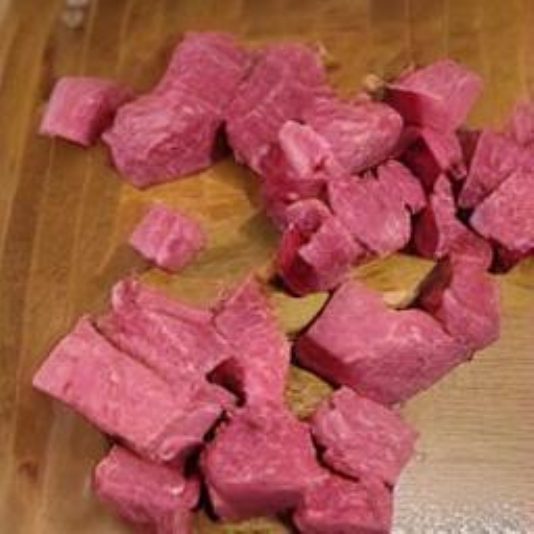} & \includegraphics[width=\linewidth, valign=m]{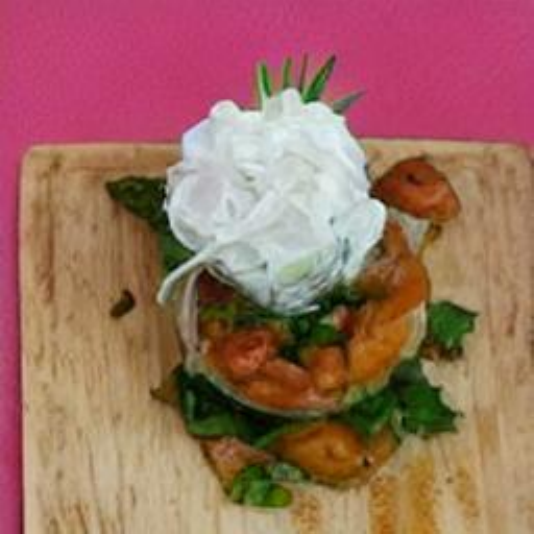} \\ \noalign{\vspace{10pt}}
    \rotatebox[origin=c]{90}{\scriptsize StackedDiffusion} & \includegraphics[width=\linewidth, valign=m]{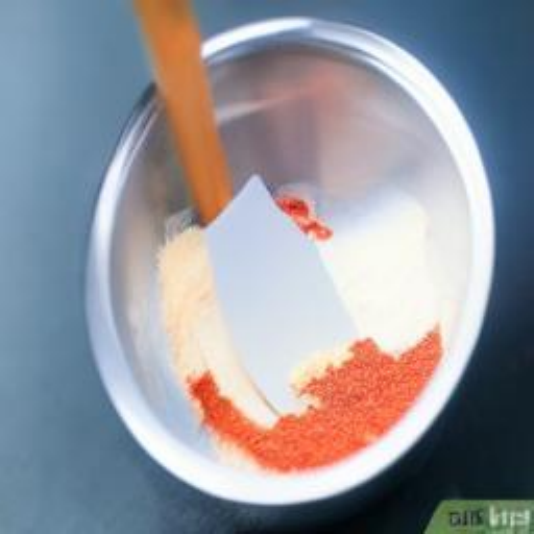} & \includegraphics[width=\linewidth, valign=m]{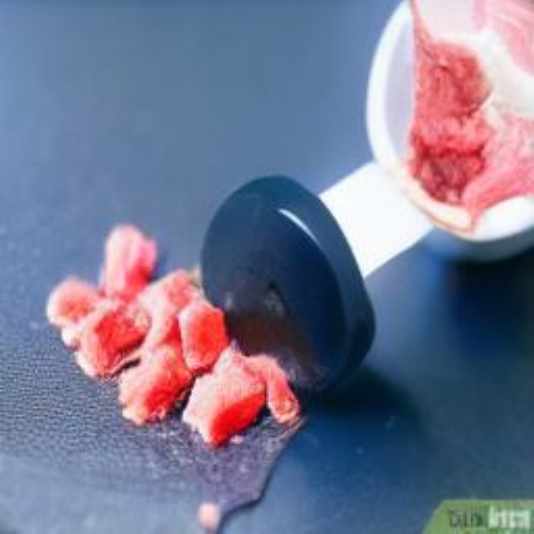} & \includegraphics[width=\linewidth, valign=m]{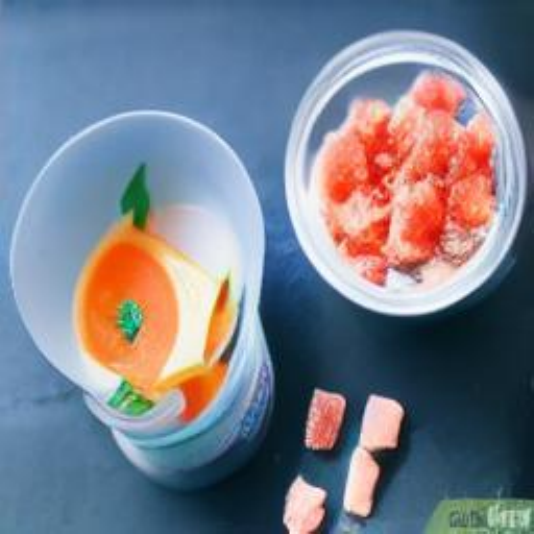} \\ \noalign{\vspace{10pt}}
    \rotatebox[origin=c]{90}{\scriptsize Bordalo et al.} & \includegraphics[width=\linewidth, valign=m]{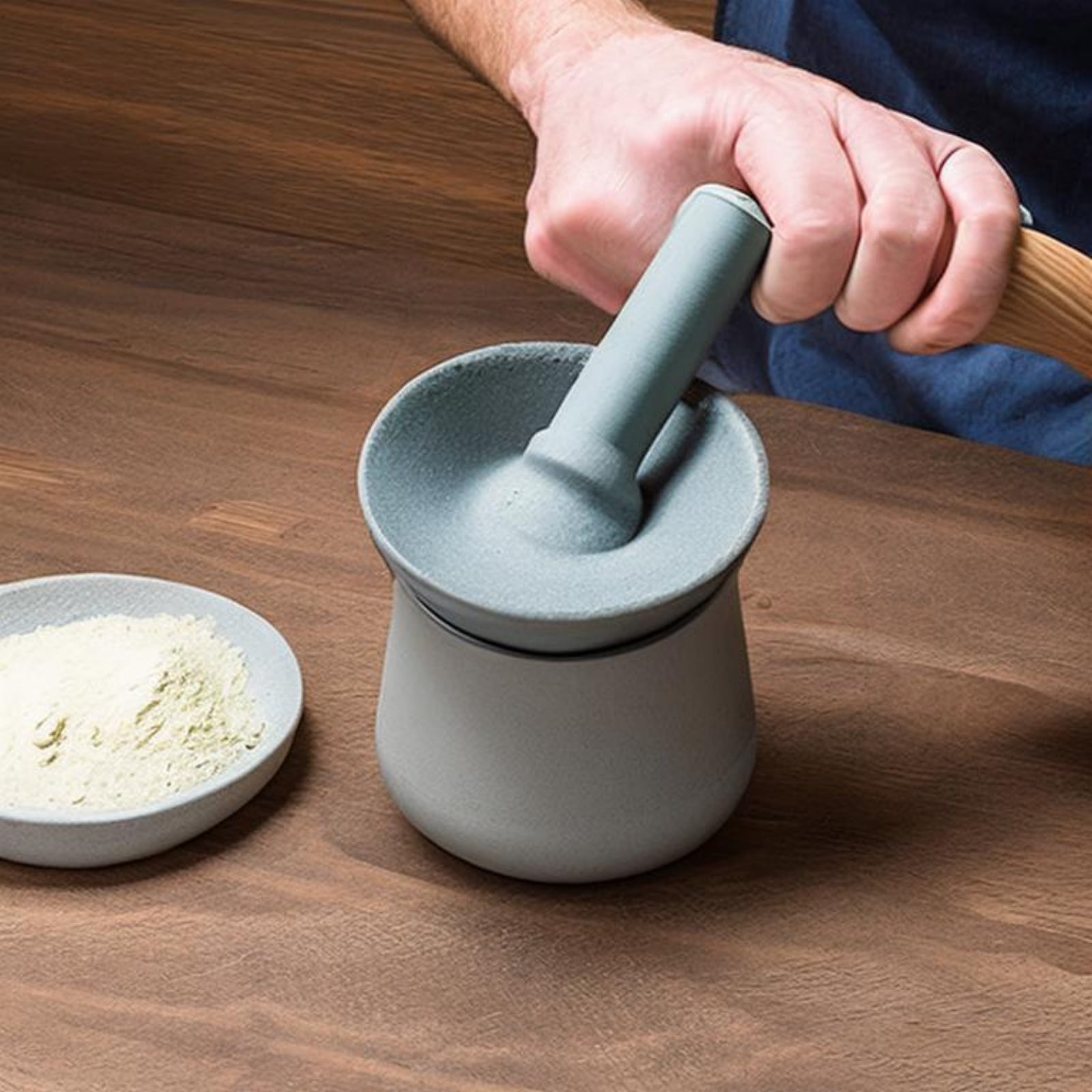} & \includegraphics[width=\linewidth, valign=m]{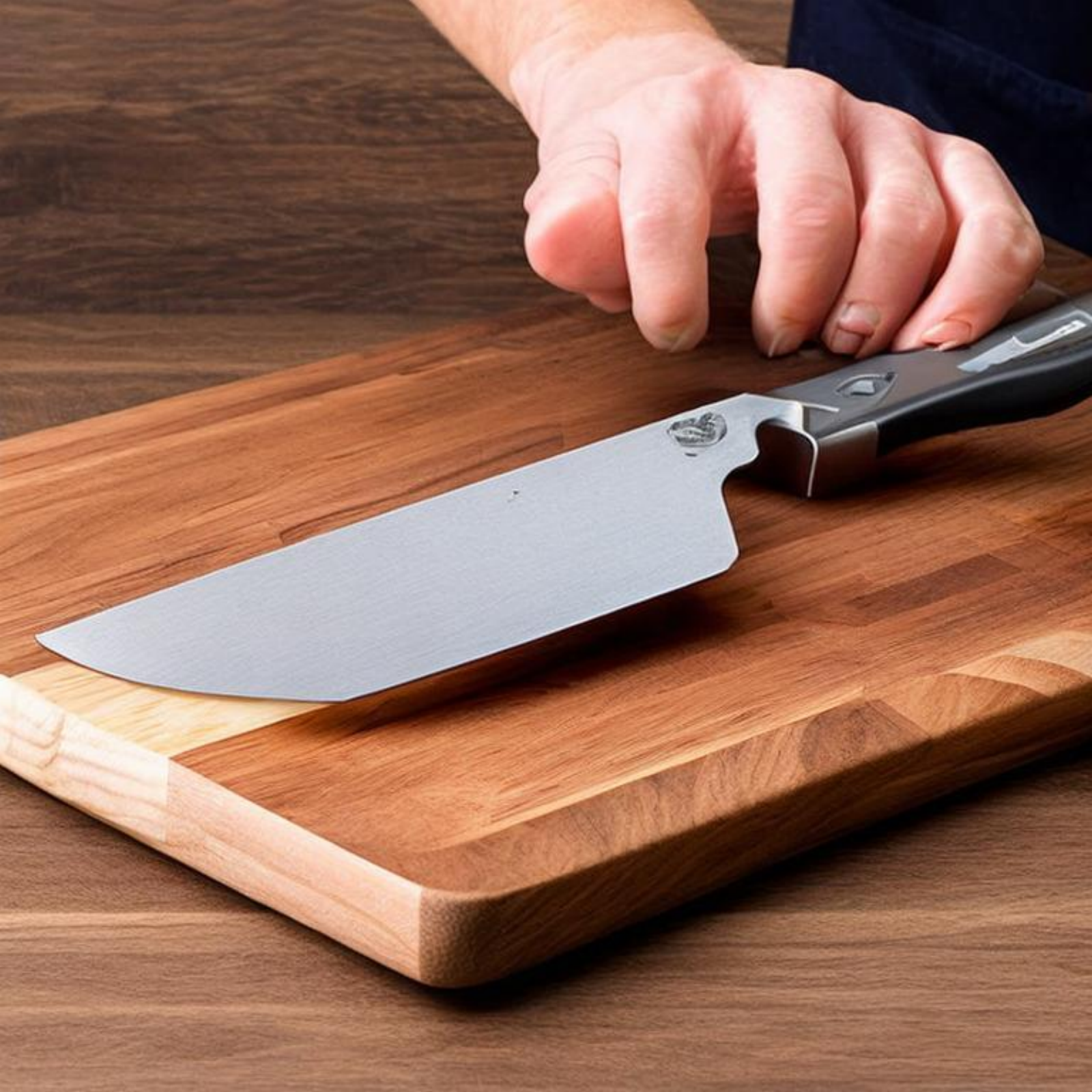} & \includegraphics[width=\linewidth, valign=m]{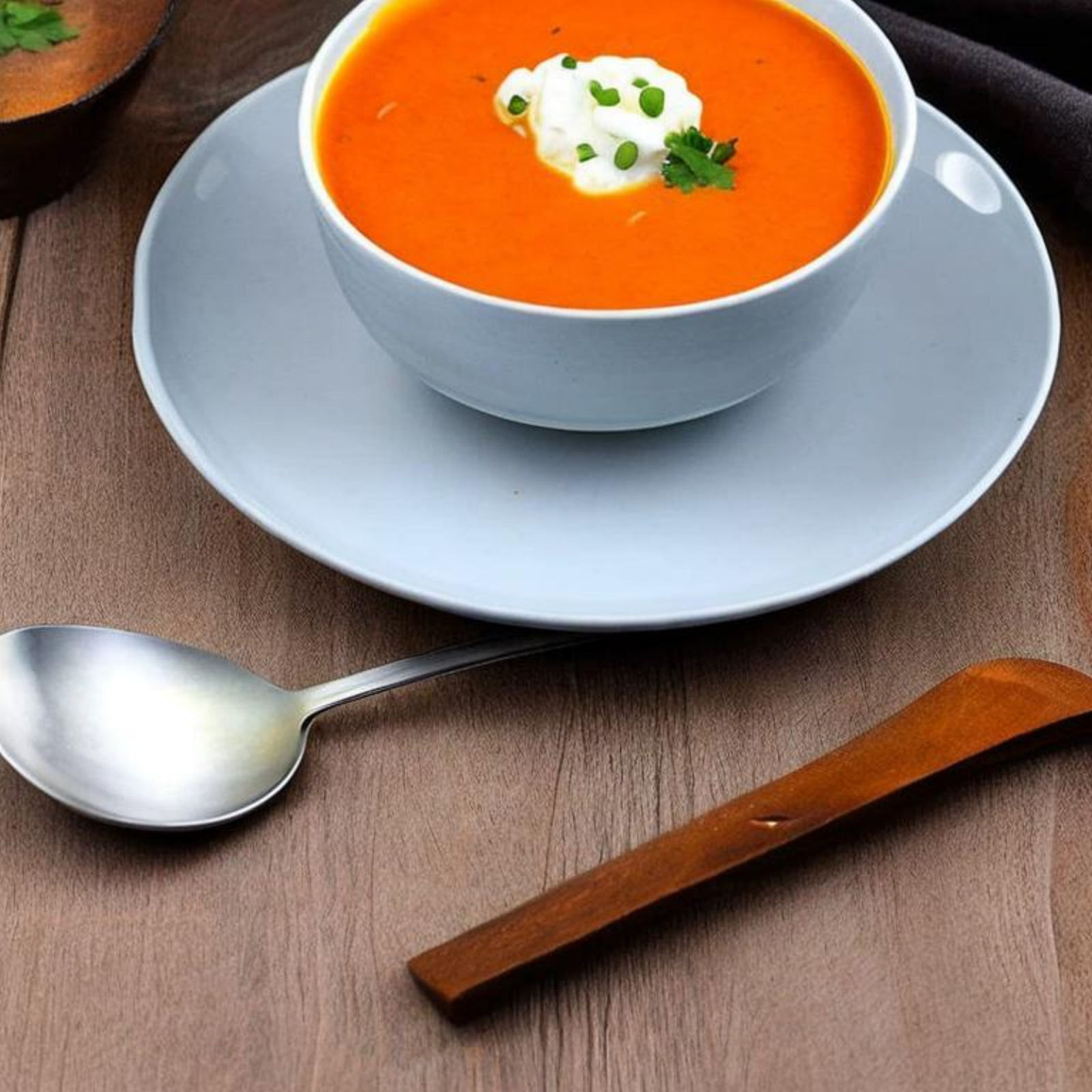} \\ \noalign{\vspace{10pt}}
    \rotatebox[origin=c]{90}{\scriptsize Baseline} & \includegraphics[width=\linewidth, valign=m]{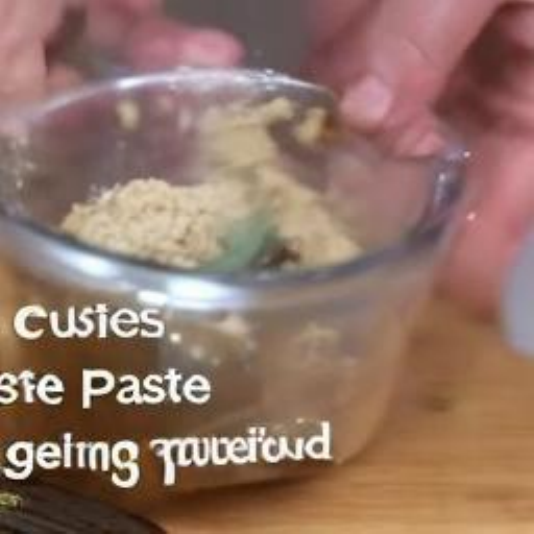} & \includegraphics[width=\linewidth, valign=m]{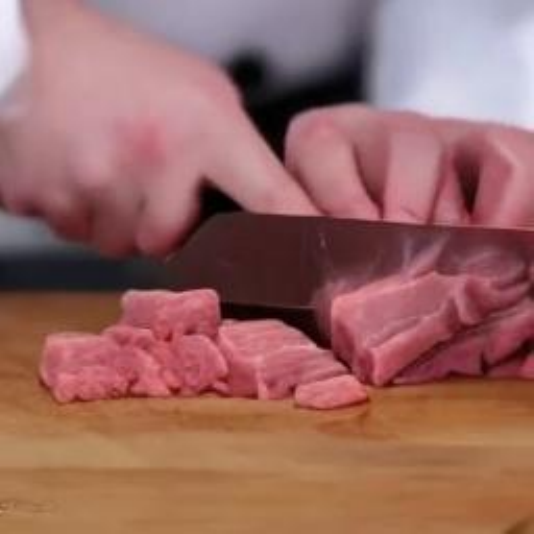} & \includegraphics[width=\linewidth, valign=m]{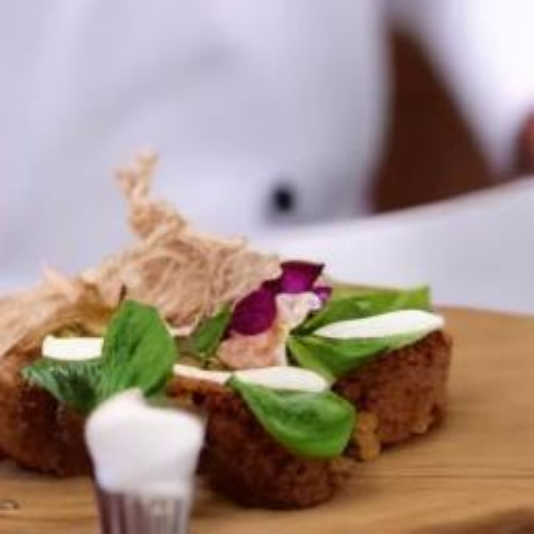} \\ \noalign{\vspace{10pt}}
    \rotatebox[origin=c]{90}{\scriptsize \textbf{InstructionCrafter}} & \includegraphics[width=\linewidth, valign=m]{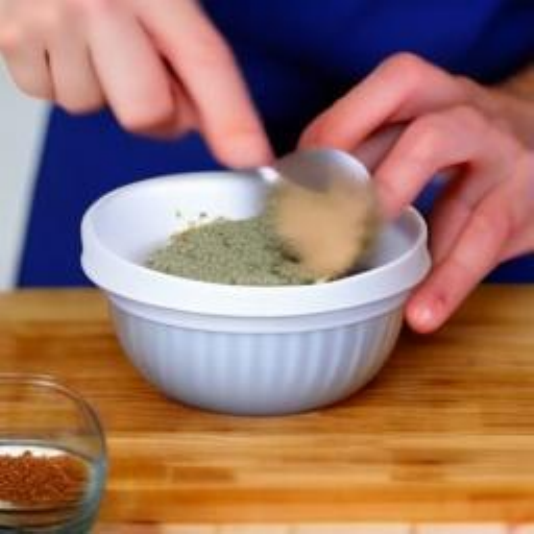} & \includegraphics[width=\linewidth, valign=m]{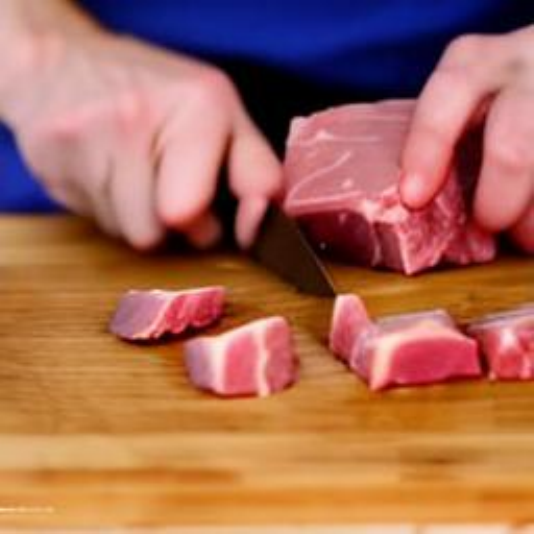} & \includegraphics[width=\linewidth, valign=m]{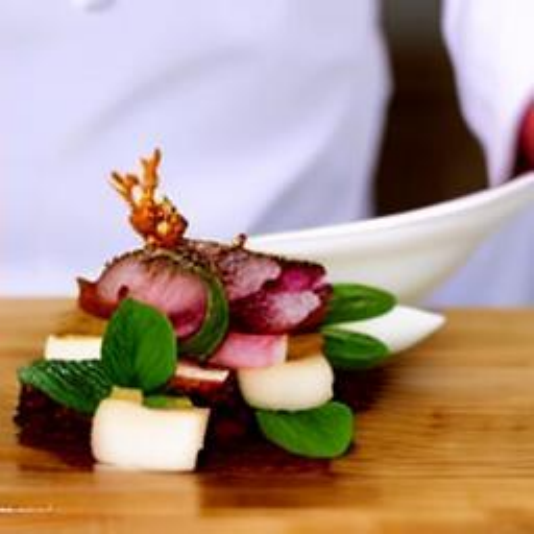} \\ \noalign{\vspace{10pt}}
  \end{tabular}
  \caption{\textbf{Additional qualitative comparison on WikiHow-VGSI.}}
  \label{fig:sup_wikihow_1174}
\end{figure}

\begin{figure}[t]
  \centering
  \setlength{\tabcolsep}{3pt}
  \renewcommand{\arraystretch}{1.2}
  \begin{tabular}{>{\centering\arraybackslash}m{1.5cm} m{0.267\linewidth} m{0.267\linewidth} m{0.267\linewidth}}
     & \tiny\centering Soak the prunes. & \tiny\centering Heat all the ingredients. & \tiny Blend and cool before using. \\
    \rotatebox[origin=c]{90}{\scriptsize Stable Diffusion} & \includegraphics[width=\linewidth, valign=m]{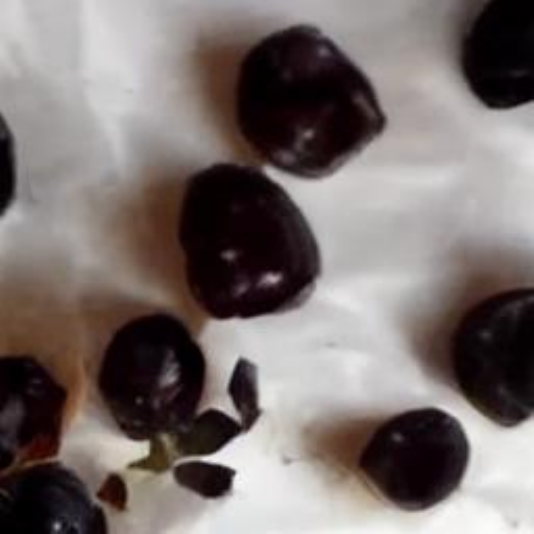} & \includegraphics[width=\linewidth, valign=m]{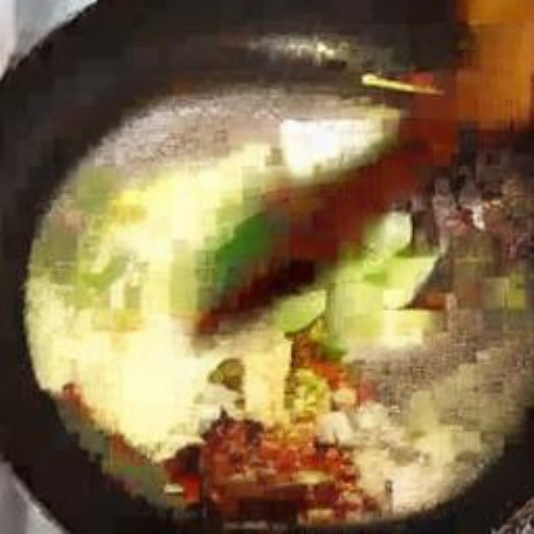} & \includegraphics[width=\linewidth, valign=m]{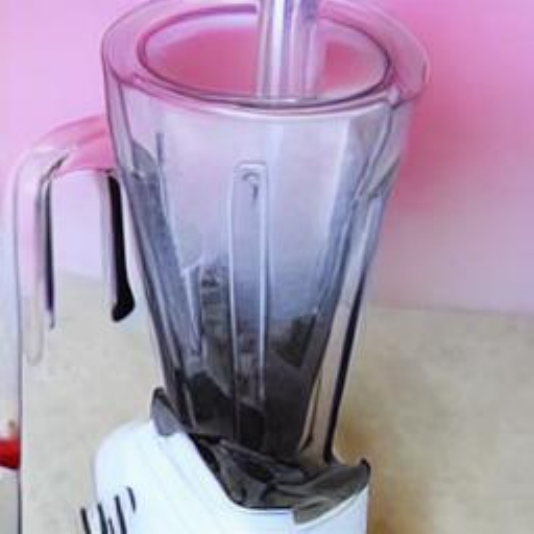} \\ \noalign{\vspace{10pt}}
    \rotatebox[origin=c]{90}{\scriptsize StackedDiffusion} & \includegraphics[width=\linewidth, valign=m]{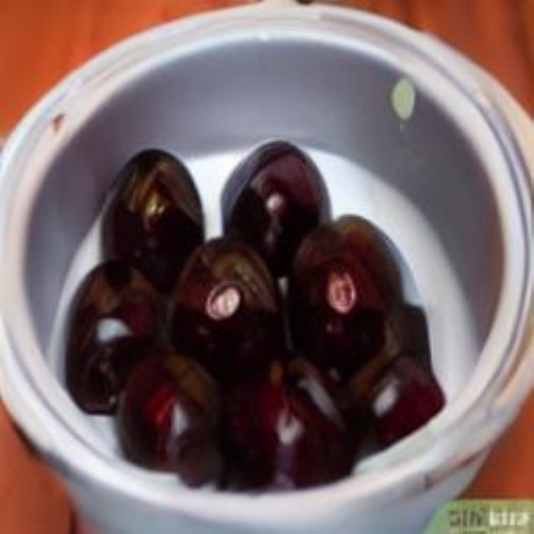} & \includegraphics[width=\linewidth, valign=m]{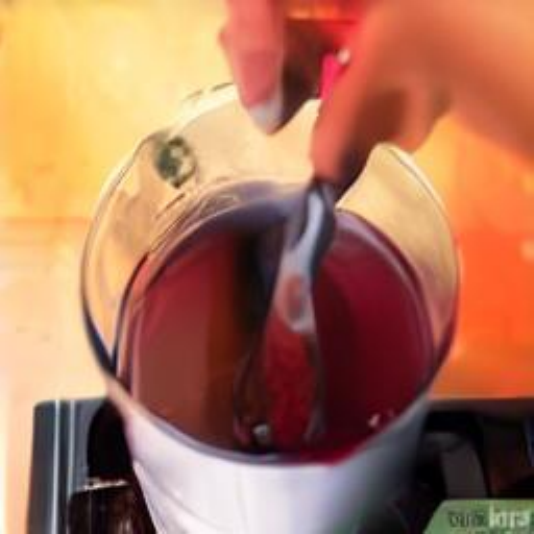} & \includegraphics[width=\linewidth, valign=m]{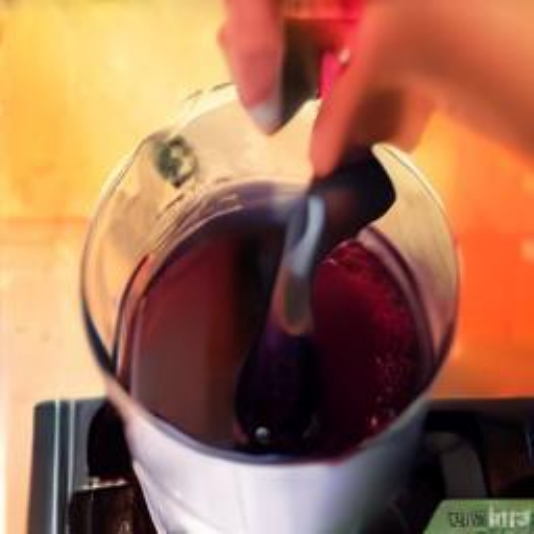} \\ \noalign{\vspace{10pt}}
    \rotatebox[origin=c]{90}{\scriptsize Bordalo et al.} & \includegraphics[width=\linewidth, valign=m]{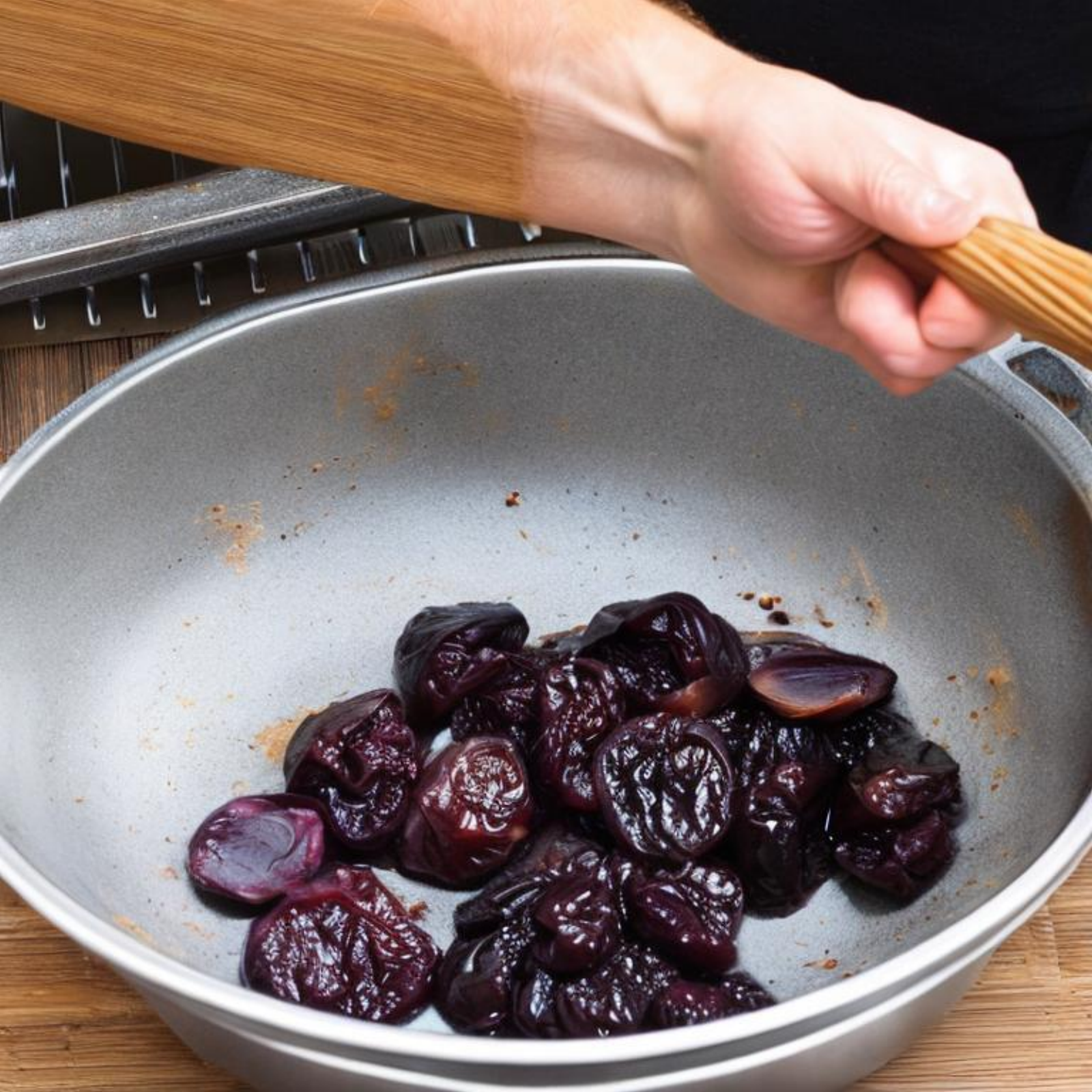} & \includegraphics[width=\linewidth, valign=m]{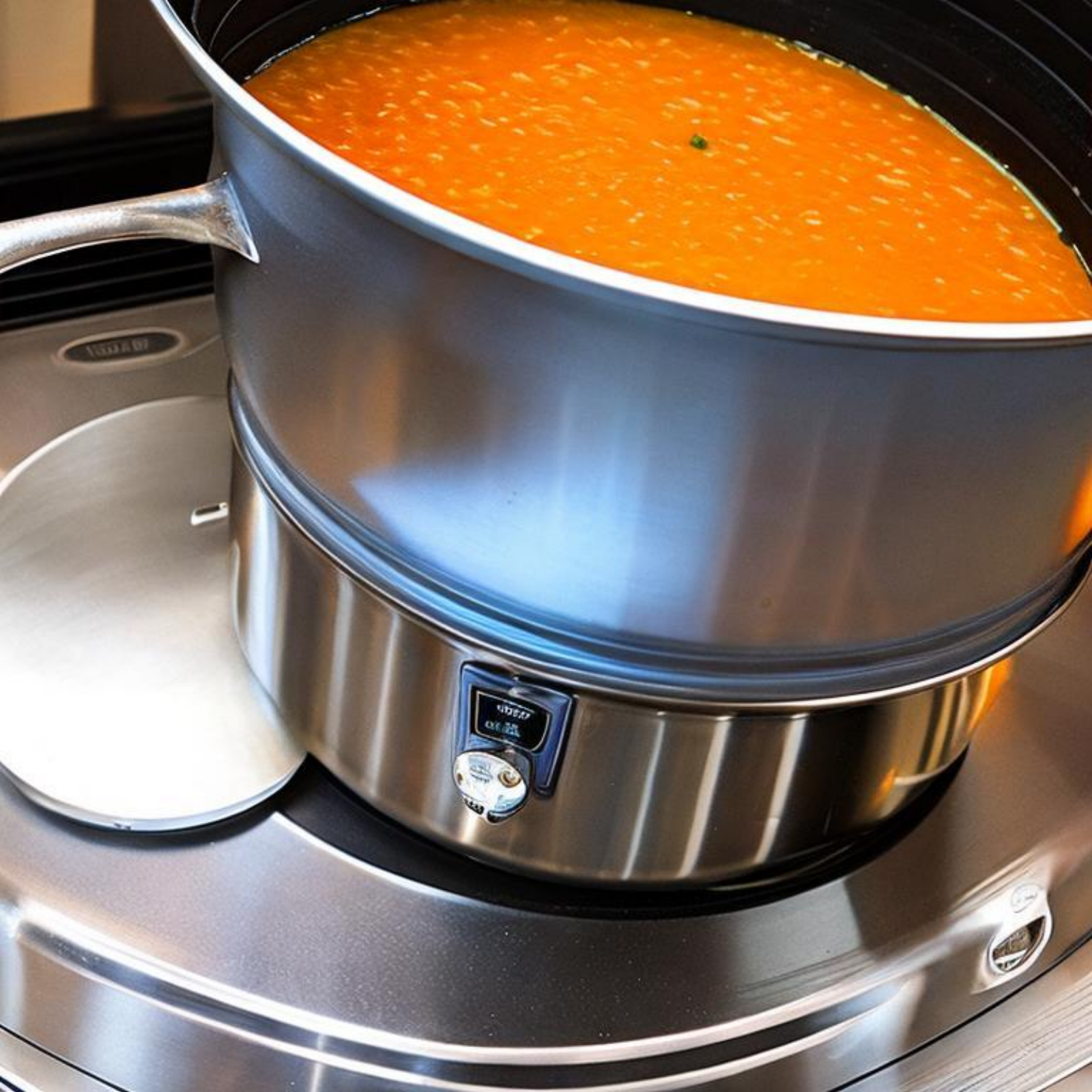} & \includegraphics[width=\linewidth, valign=m]{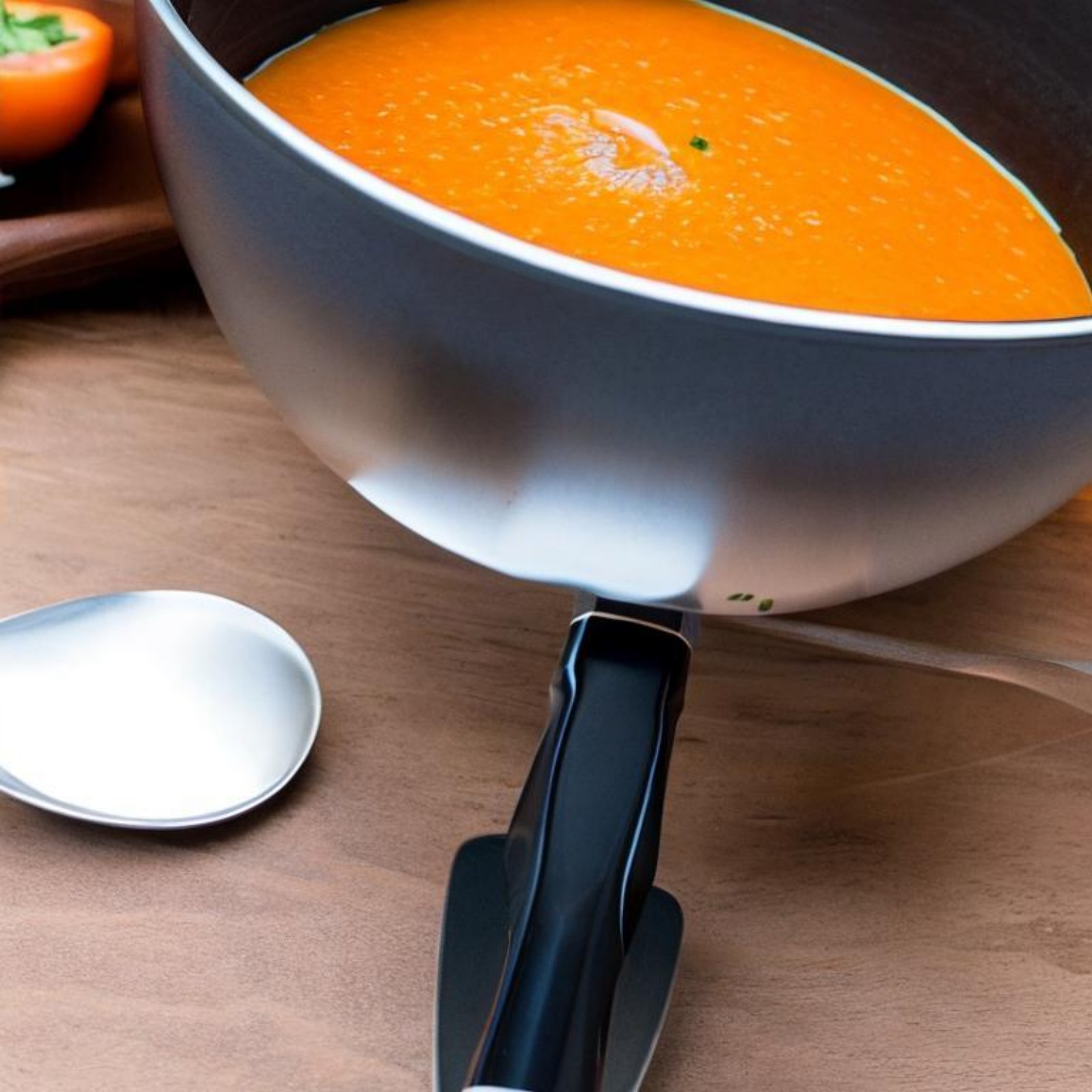} \\ \noalign{\vspace{10pt}}
    \rotatebox[origin=c]{90}{\scriptsize Baseline} & \includegraphics[width=\linewidth, valign=m]{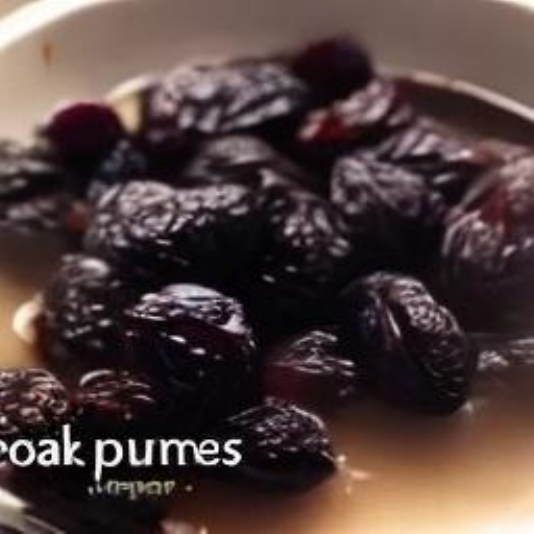} & \includegraphics[width=\linewidth, valign=m]{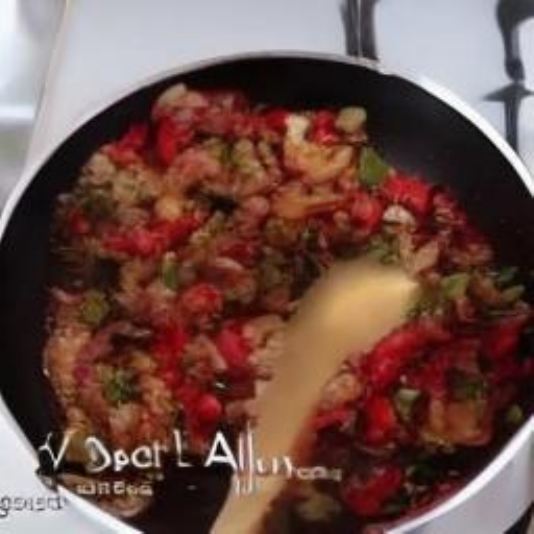} & \includegraphics[width=\linewidth, valign=m]{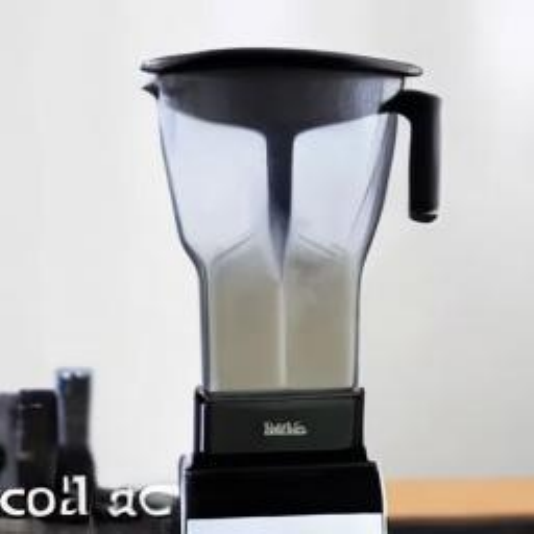} \\ \noalign{\vspace{10pt}}
    \rotatebox[origin=c]{90}{\scriptsize \textbf{InstructionCrafter}} & \includegraphics[width=\linewidth, valign=m]{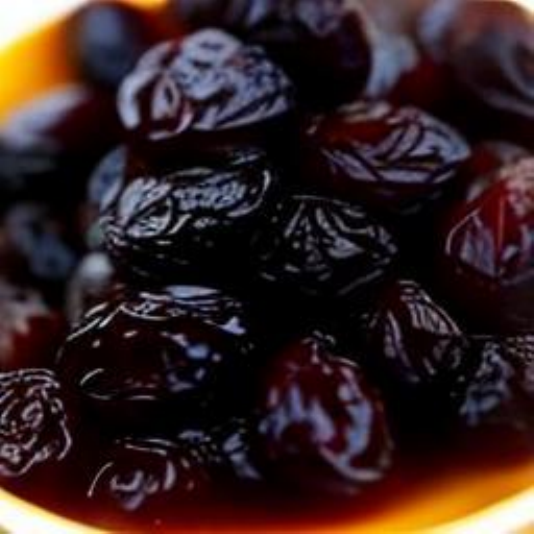} & \includegraphics[width=\linewidth, valign=m]{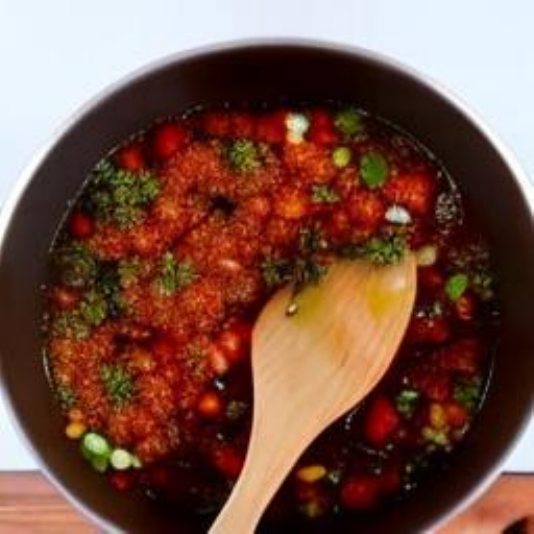} & \includegraphics[width=\linewidth, valign=m]{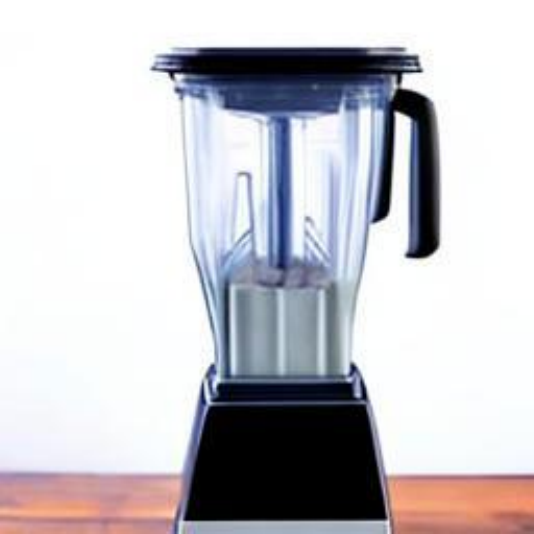} \\ \noalign{\vspace{10pt}}
  \end{tabular}
  \caption{\textbf{Additional qualitative comparison on WikiHow-VGSI.}}
  \label{fig:sup_wikihow_1632}
\end{figure}

\begin{figure*}[t]
  \centering
  \setlength{\tabcolsep}{3pt}
  \renewcommand{\arraystretch}{1.2}
  \begin{tabular}{>{\centering\arraybackslash}m{1.5cm} m{0.200\linewidth} m{0.200\linewidth} m{0.200\linewidth} m{0.200\linewidth}}
     & \tiny\centering Mix the white layer first. & \tiny\centering Create the orange layer. & \tiny\centering Make the final layer--yellow. & \tiny Finished. \\
    \rotatebox[origin=c]{90}{\scriptsize Stable Diffusion} & \includegraphics[width=\linewidth, valign=m]{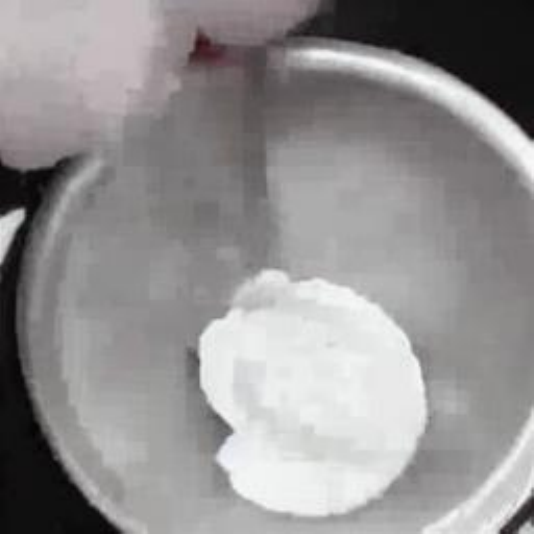} & \includegraphics[width=\linewidth, valign=m]{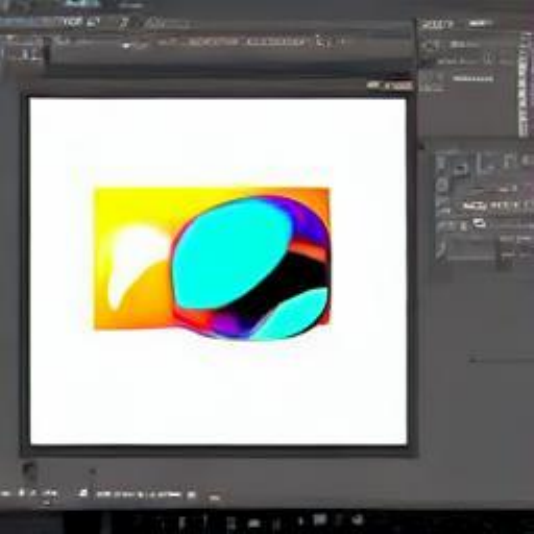} & \includegraphics[width=\linewidth, valign=m]{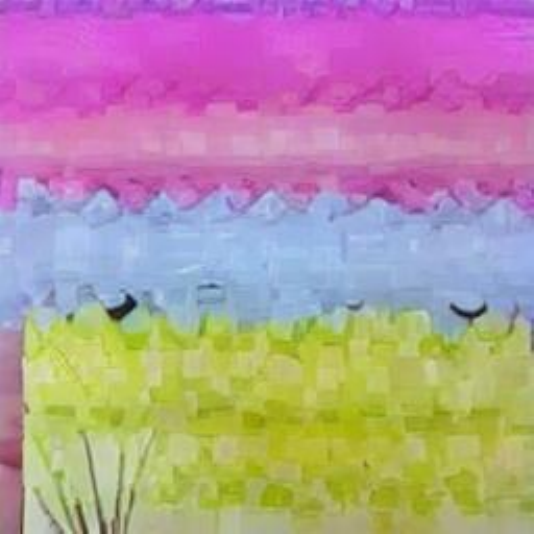} & \includegraphics[width=\linewidth, valign=m]{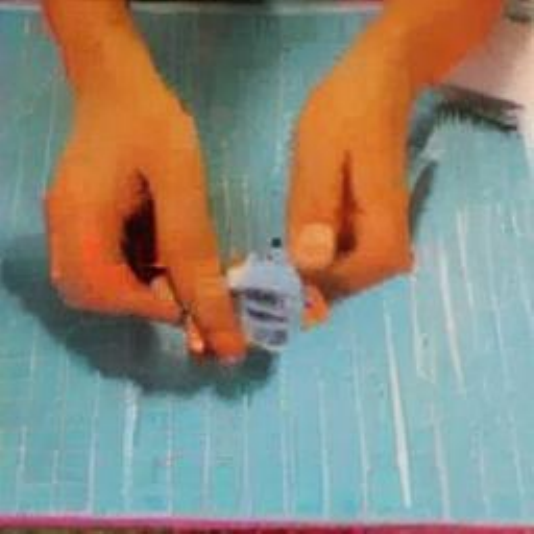} \\ \noalign{\vspace{10pt}}
    \rotatebox[origin=c]{90}{\scriptsize StackedDiffusion} & \includegraphics[width=\linewidth, valign=m]{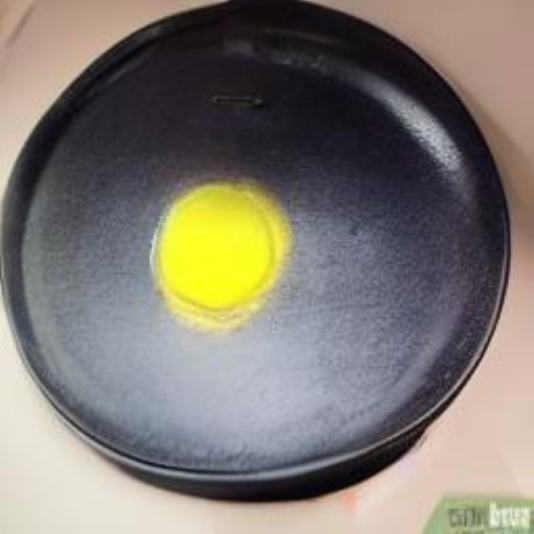} & \includegraphics[width=\linewidth, valign=m]{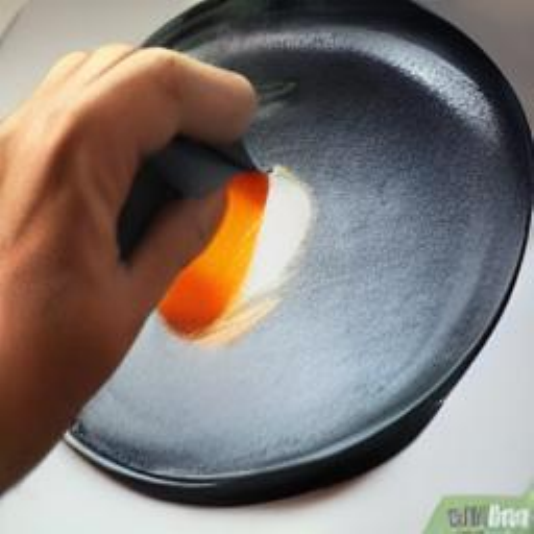} & \includegraphics[width=\linewidth, valign=m]{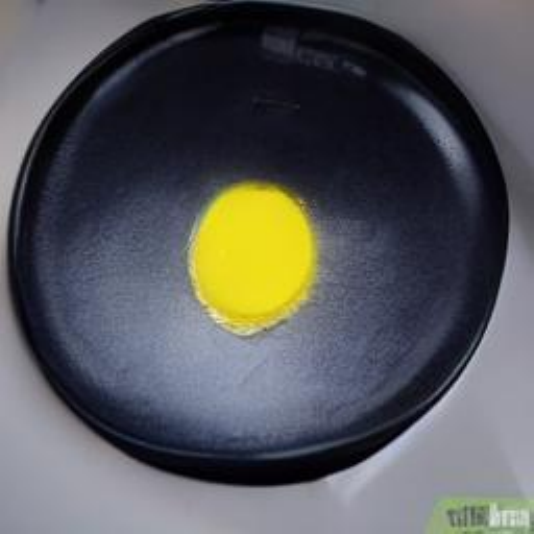} & \includegraphics[width=\linewidth, valign=m]{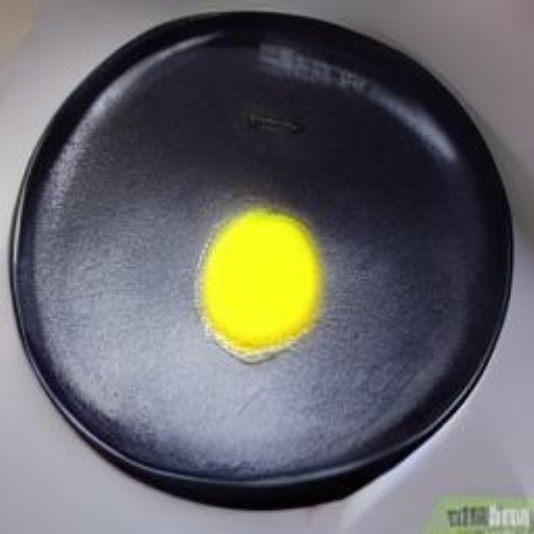} \\ \noalign{\vspace{10pt}}
    \rotatebox[origin=c]{90}{\scriptsize Bordalo et al.} & \includegraphics[width=\linewidth, valign=m]{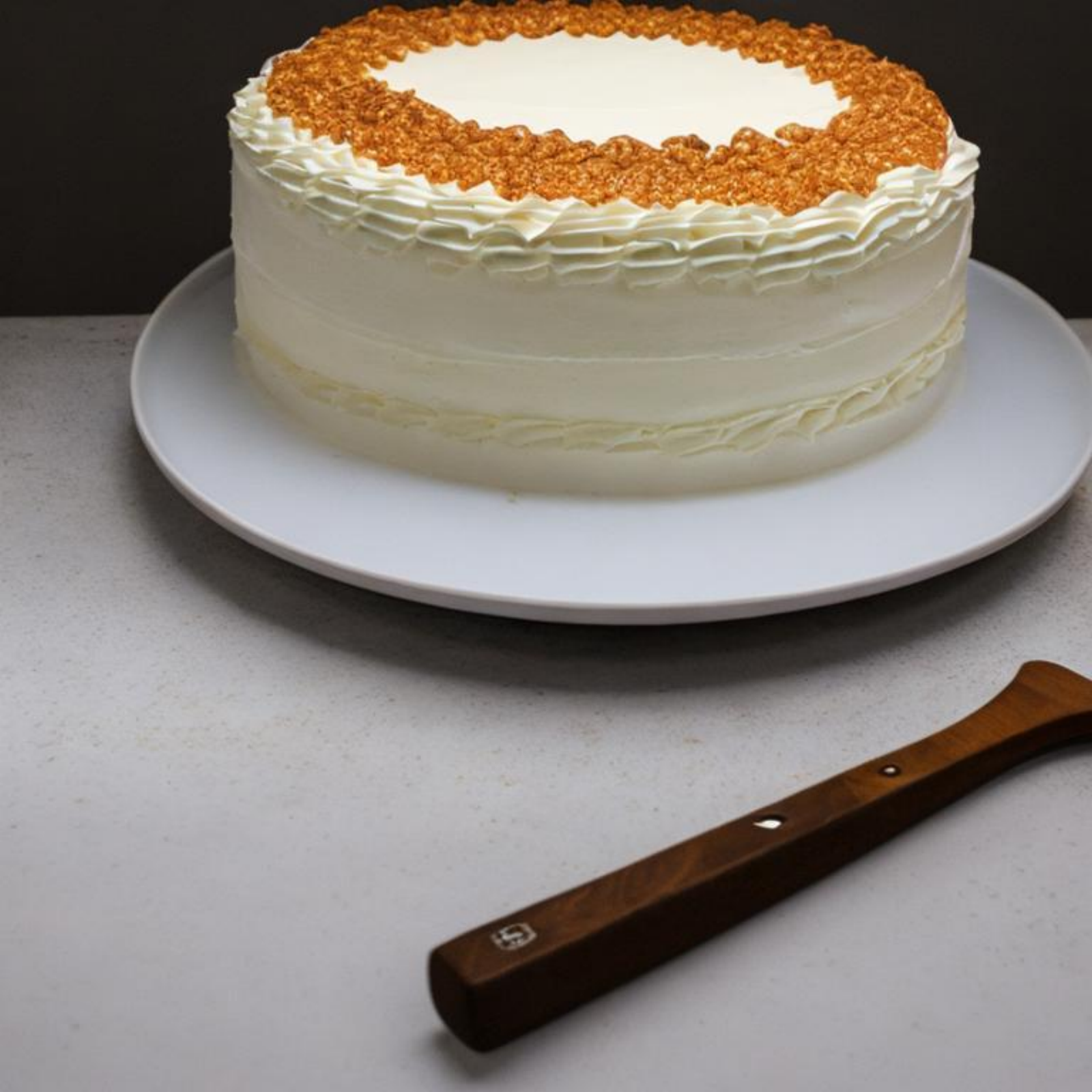} & \includegraphics[width=\linewidth, valign=m]{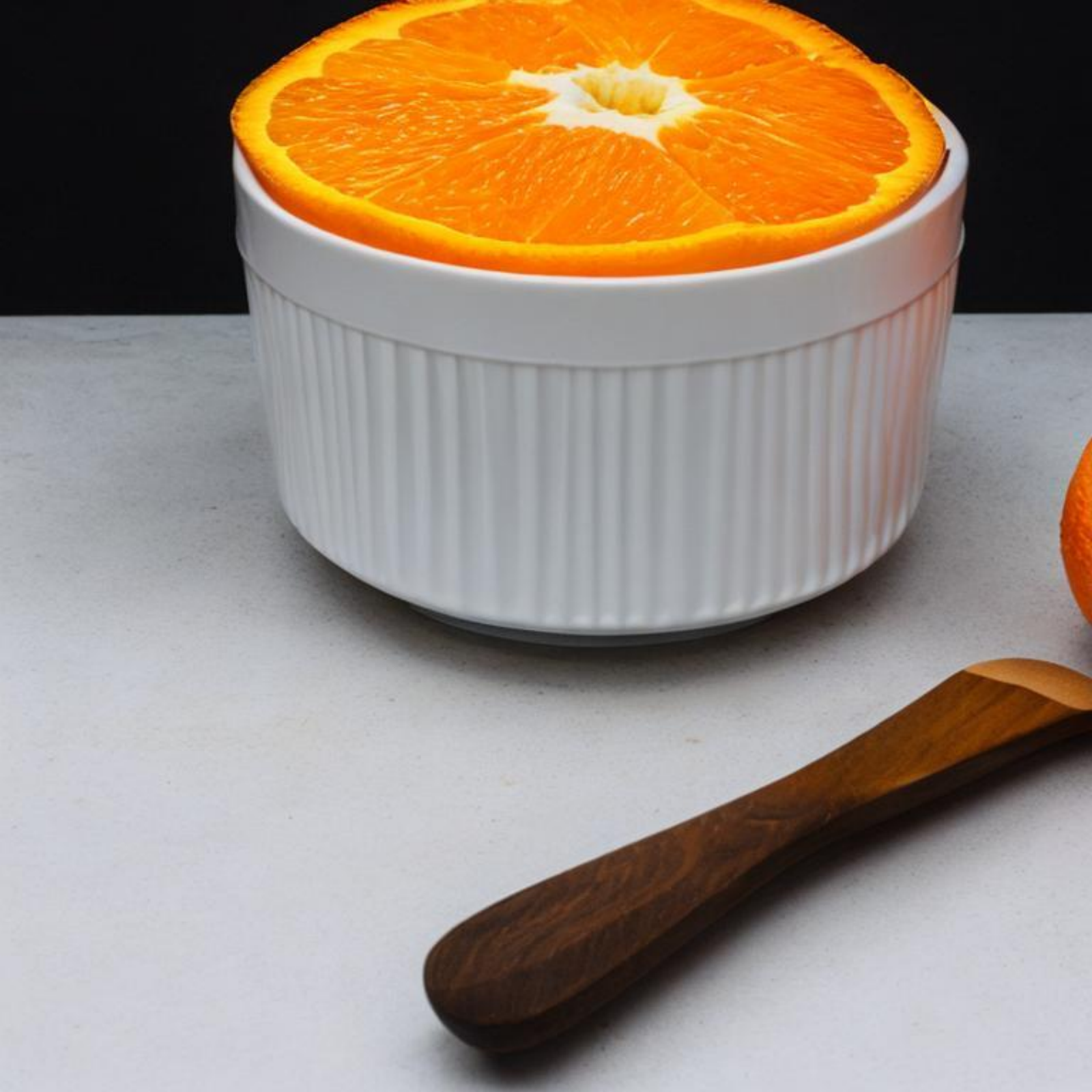} & \includegraphics[width=\linewidth, valign=m]{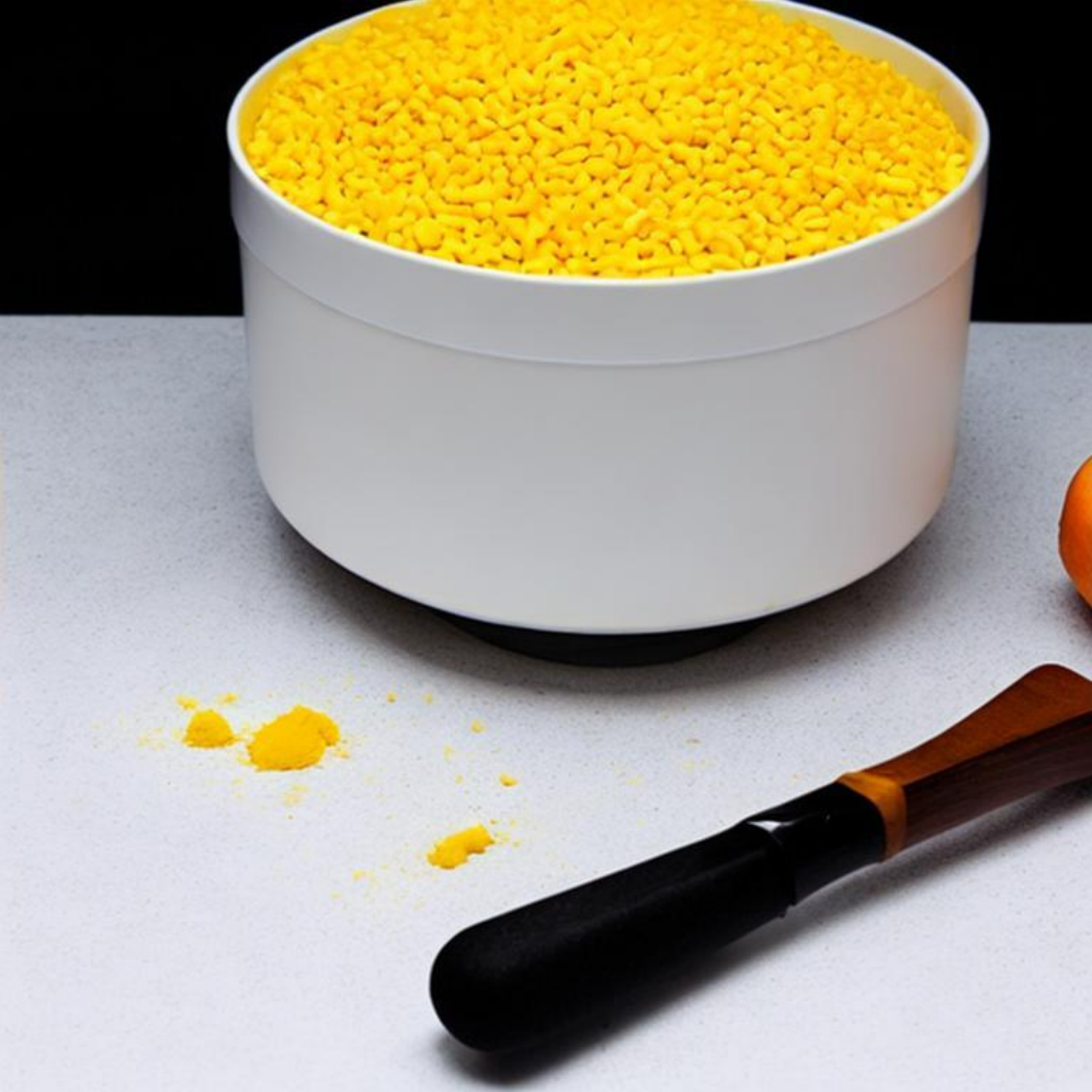} & \includegraphics[width=\linewidth, valign=m]{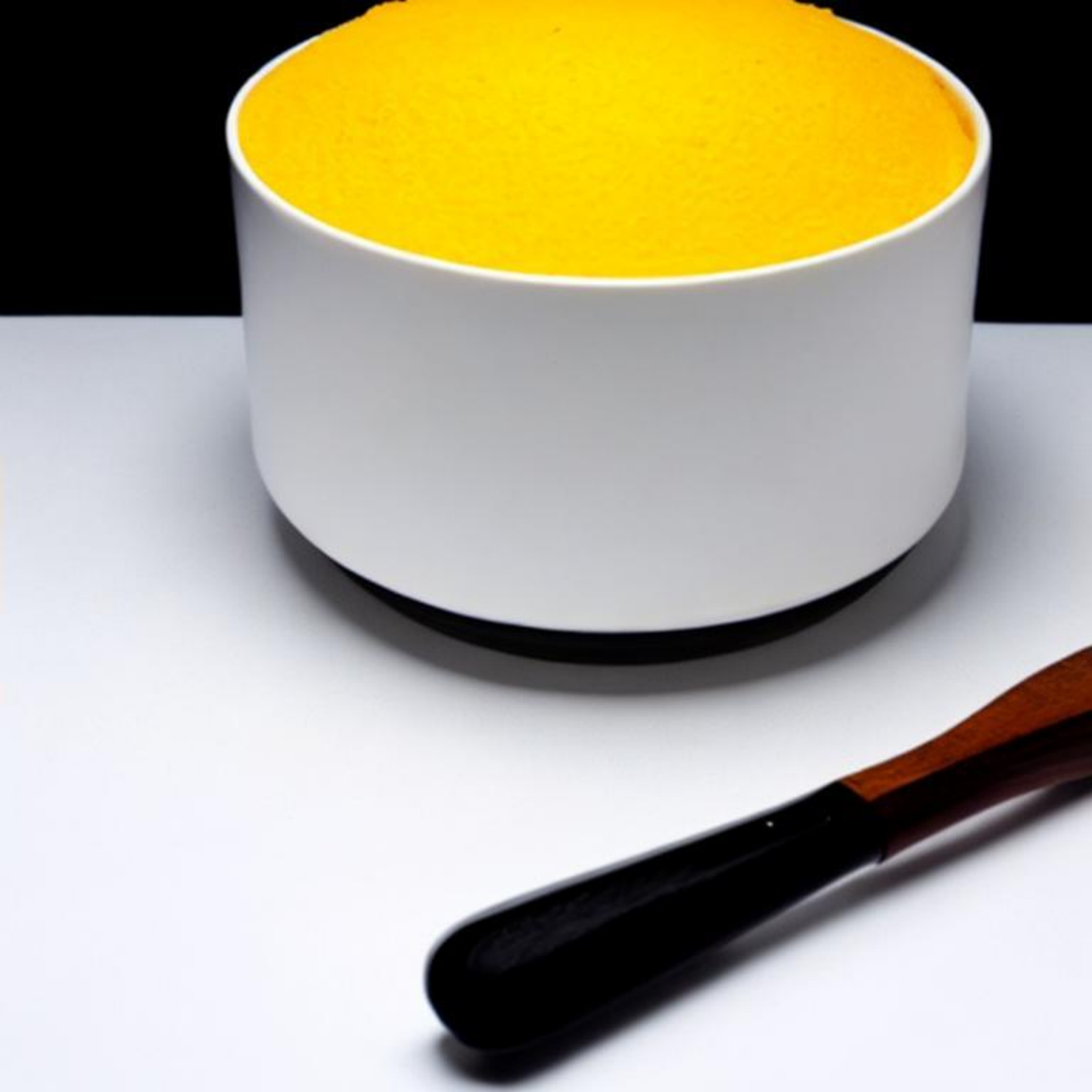} \\ \noalign{\vspace{10pt}}
    \rotatebox[origin=c]{90}{\scriptsize Baseline} & \includegraphics[width=\linewidth, valign=m]{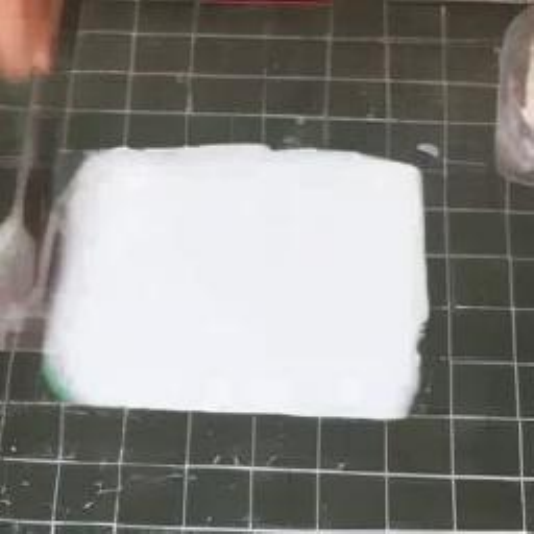} & \includegraphics[width=\linewidth, valign=m]{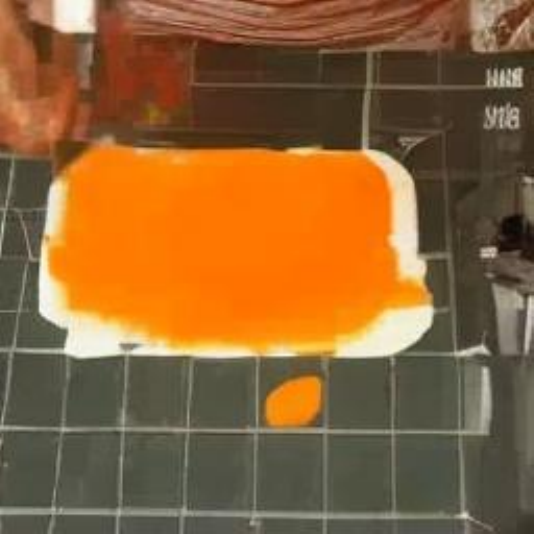} & \includegraphics[width=\linewidth, valign=m]{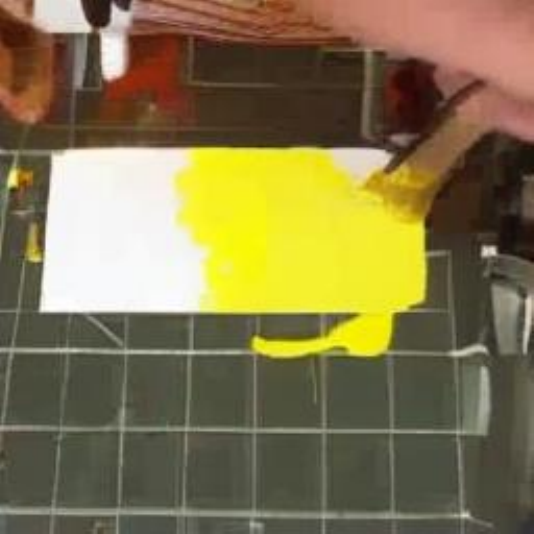} & \includegraphics[width=\linewidth, valign=m]{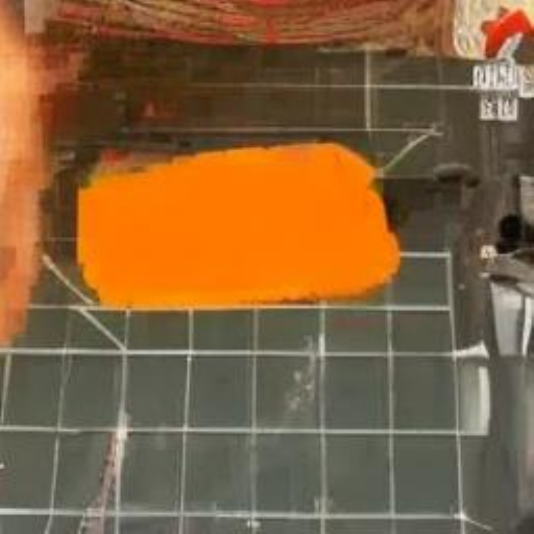} \\
\end{tabular}
  \caption{\textbf{Additional qualitative comparison on WikiHow-VGSI.}}
  \label{fig:sup_wikihow_1580}
\end{figure*}

We show additional qualitative comparisons with prior methods in~\cref{fig:sup_showhow_3951} to~\cref{fig:sup_showhow_1766} for ShowHowTo~\cite{souvcek2025showhowto} and~\cref{fig:sup_wikihow_3} to~\cref{fig:sup_wikihow_1580} for WikiHow-VGSI~\cite{yang2021visual}. The results further demonstrate that existing methods often fail to generate high-quality, consistent, and step-faithful images, while our InstructionCrafter produces high-quality images that are consistent across steps and faithful to the instructions. In particular, large watermarks and blurriness significantly degrade the clarity of visual instructions, which limits their real-world applications. Our method effectively mitigates these issues and generates clear images.

\subsection{Additional Qualitative Examples}
\begin{figure*}
    \centering
    \includegraphics[width=\textwidth]{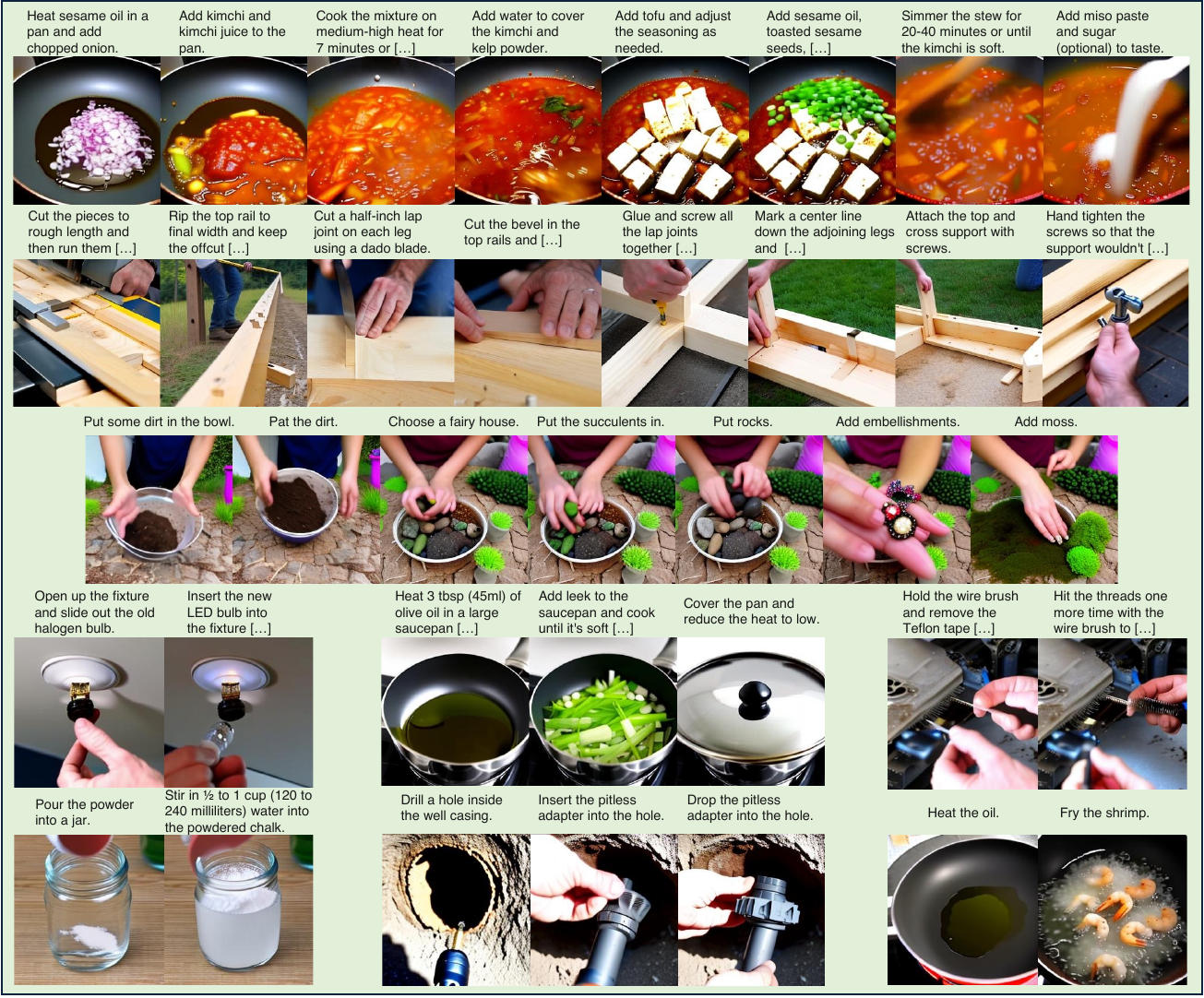}
    \caption{\textbf{Additional qualitative examples of our method on ShowHowTo~\cite{souvcek2025showhowto} and WikiHow-VGSI~\cite{yang2021visual}.} Our InstructionCrafter generates diverse, high-quality, consistent, and step-faithful images across various instructional scenarios.}
    \label{fig:sup_ours}
\end{figure*}

We show more qualitative examples of our method in~\cref{fig:sup_ours}. The results further demonstrate that our InstructionCrafter generates diverse, high-quality, consistent, and step-faithful images across various instructional scenarios.

\subsection{Failure Cases}
\label{sec:failure_cases}
\begin{figure*}[t]
  \centering
  \setlength{\tabcolsep}{3pt}
  \renewcommand{\arraystretch}{1.2}
  \begin{tabular}{>{\centering\arraybackslash}m{1.5cm} m{0.160\linewidth} m{0.160\linewidth} m{0.160\linewidth} m{0.160\linewidth} m{0.160\linewidth}}
     & \tiny\centering Cut the rosemary bonsai to maintain its shape and size. & \tiny\centering Cut the bonsai to the bud, following the natural growth habit. & \tiny\centering Maintain the bonsai by cutting it to the original outline. & \tiny\centering Water the bonsai regularly to keep it alive. & \tiny Learn to do light pruning to maintain the bonsai's shape and size. \\
    \rotatebox[origin=c]{90}{\scriptsize Stable Diffusion} & \includegraphics[width=\linewidth, valign=m]{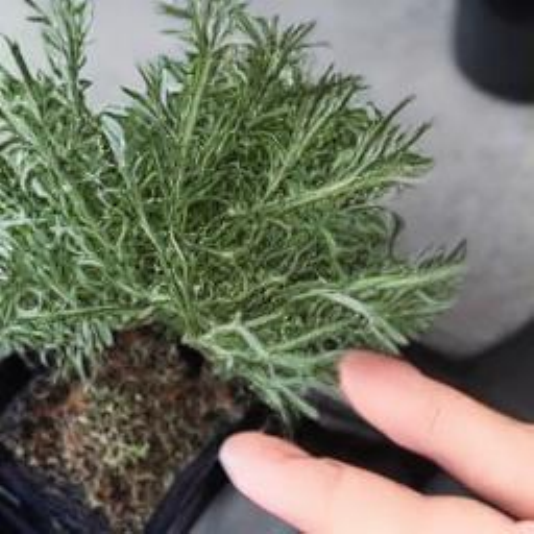} & \includegraphics[width=\linewidth, valign=m]{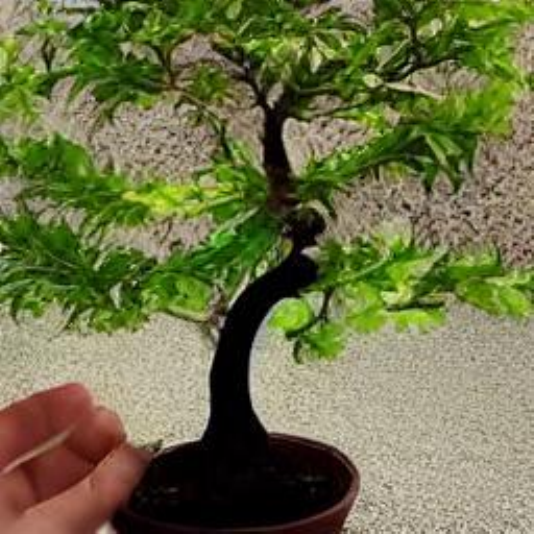} & \includegraphics[width=\linewidth, valign=m]{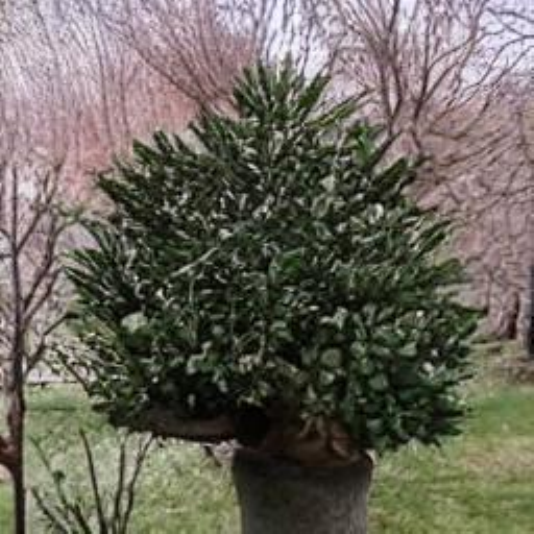} & \includegraphics[width=\linewidth, valign=m]{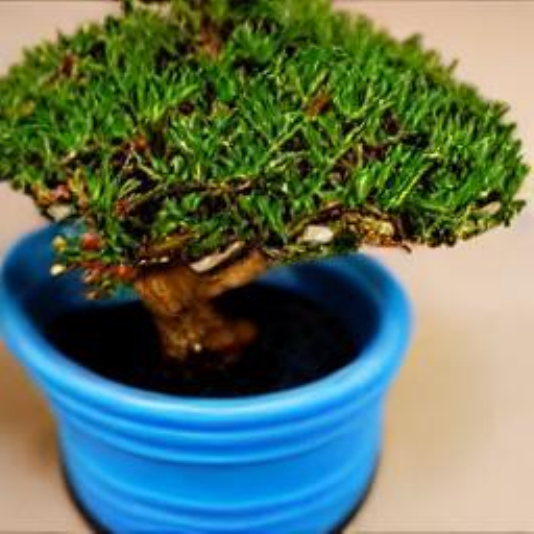} & \includegraphics[width=\linewidth, valign=m]{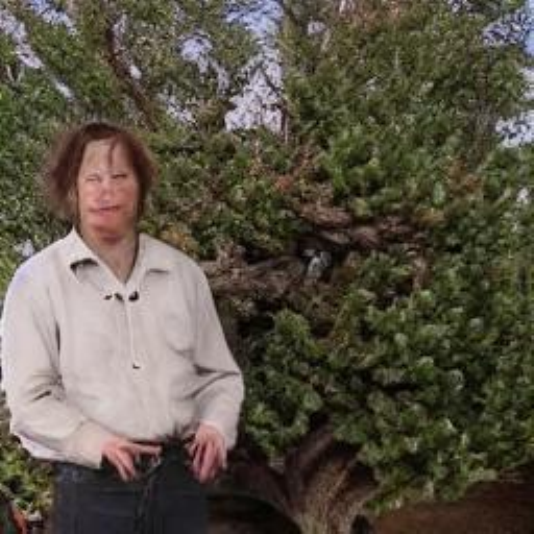} \\ \noalign{\vspace{10pt}}
    \rotatebox[origin=c]{90}{\scriptsize StackedDiffusion} & \includegraphics[width=\linewidth, valign=m]{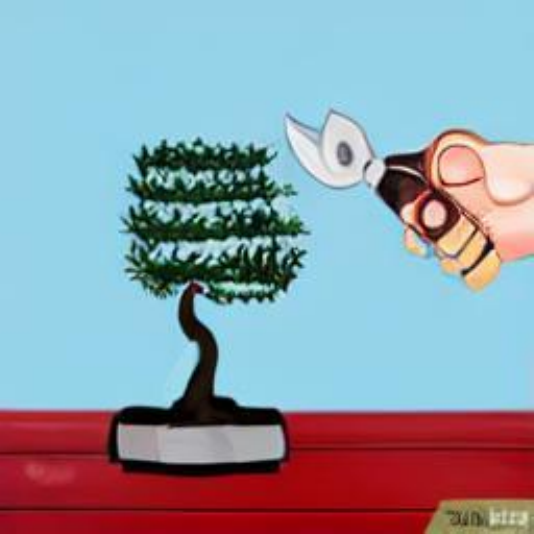} & \includegraphics[width=\linewidth, valign=m]{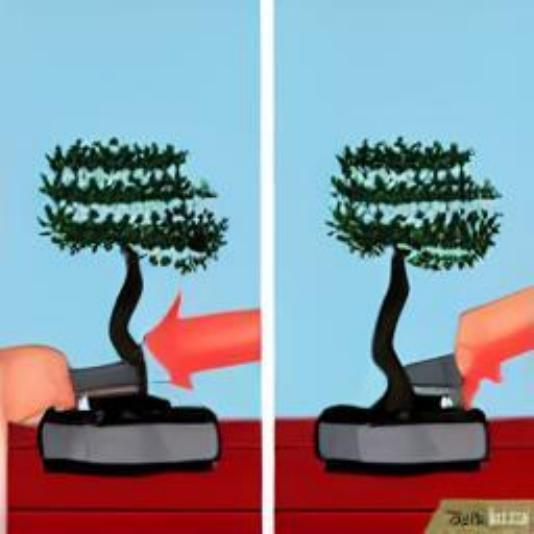} & \includegraphics[width=\linewidth, valign=m]{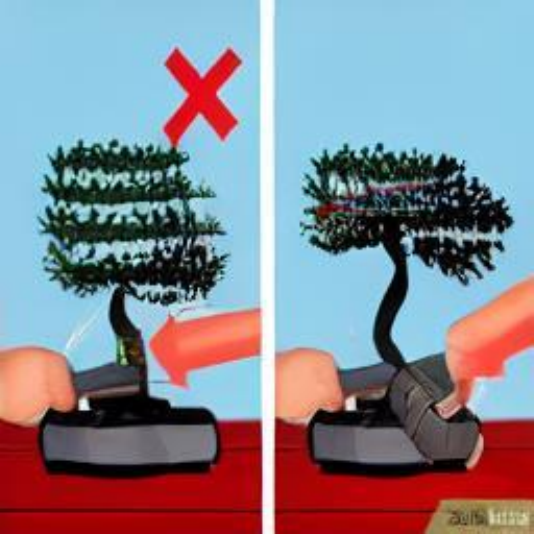} & \includegraphics[width=\linewidth, valign=m]{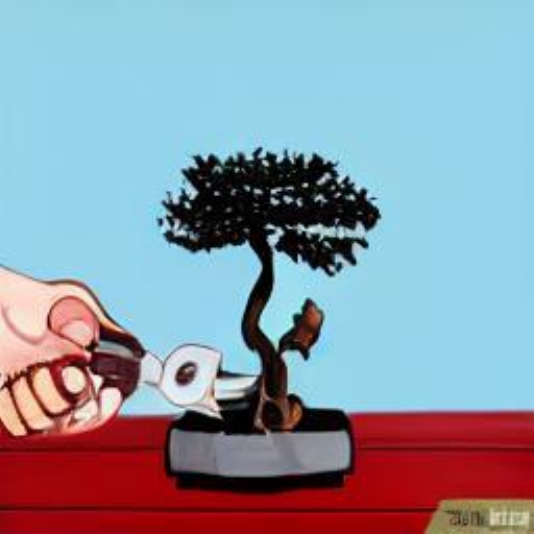} & \includegraphics[width=\linewidth, valign=m]{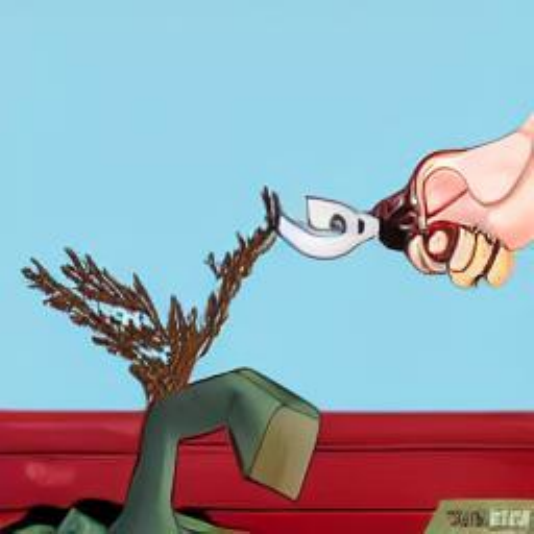} \\ \noalign{\vspace{10pt}}
    \rotatebox[origin=c]{90}{\scriptsize Bordalo et al.} & \includegraphics[width=\linewidth, valign=m]{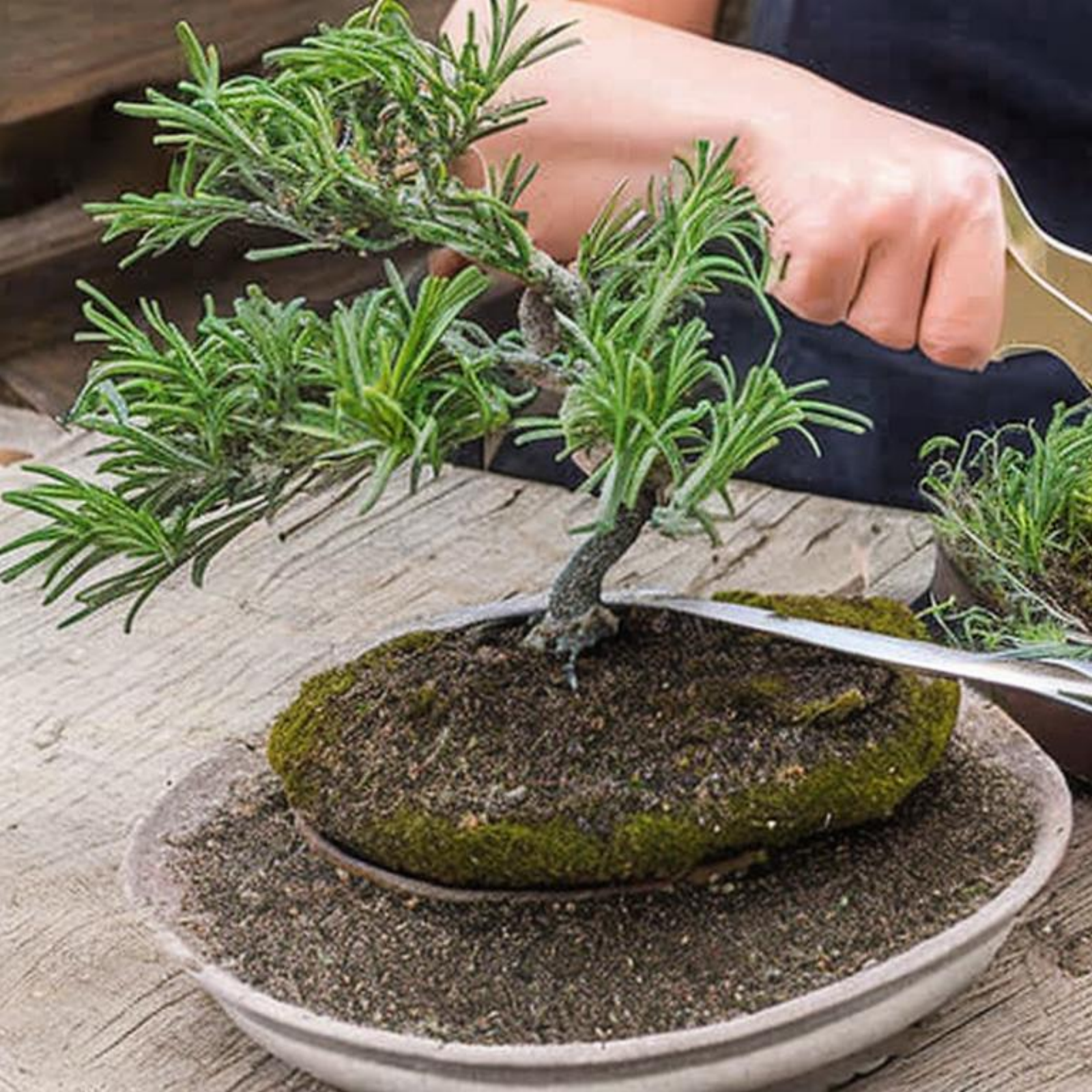} & \includegraphics[width=\linewidth, valign=m]{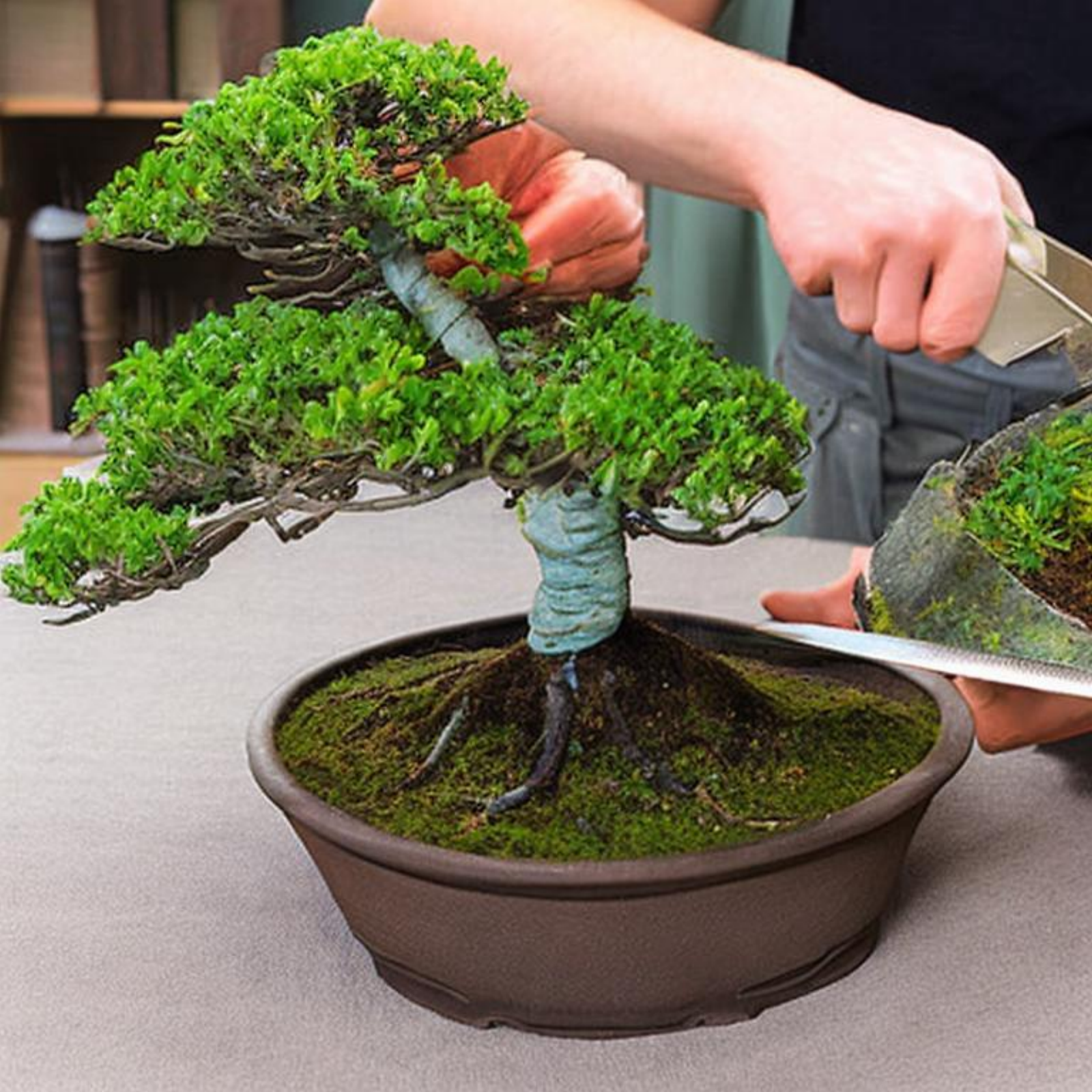} & \includegraphics[width=\linewidth, valign=m]{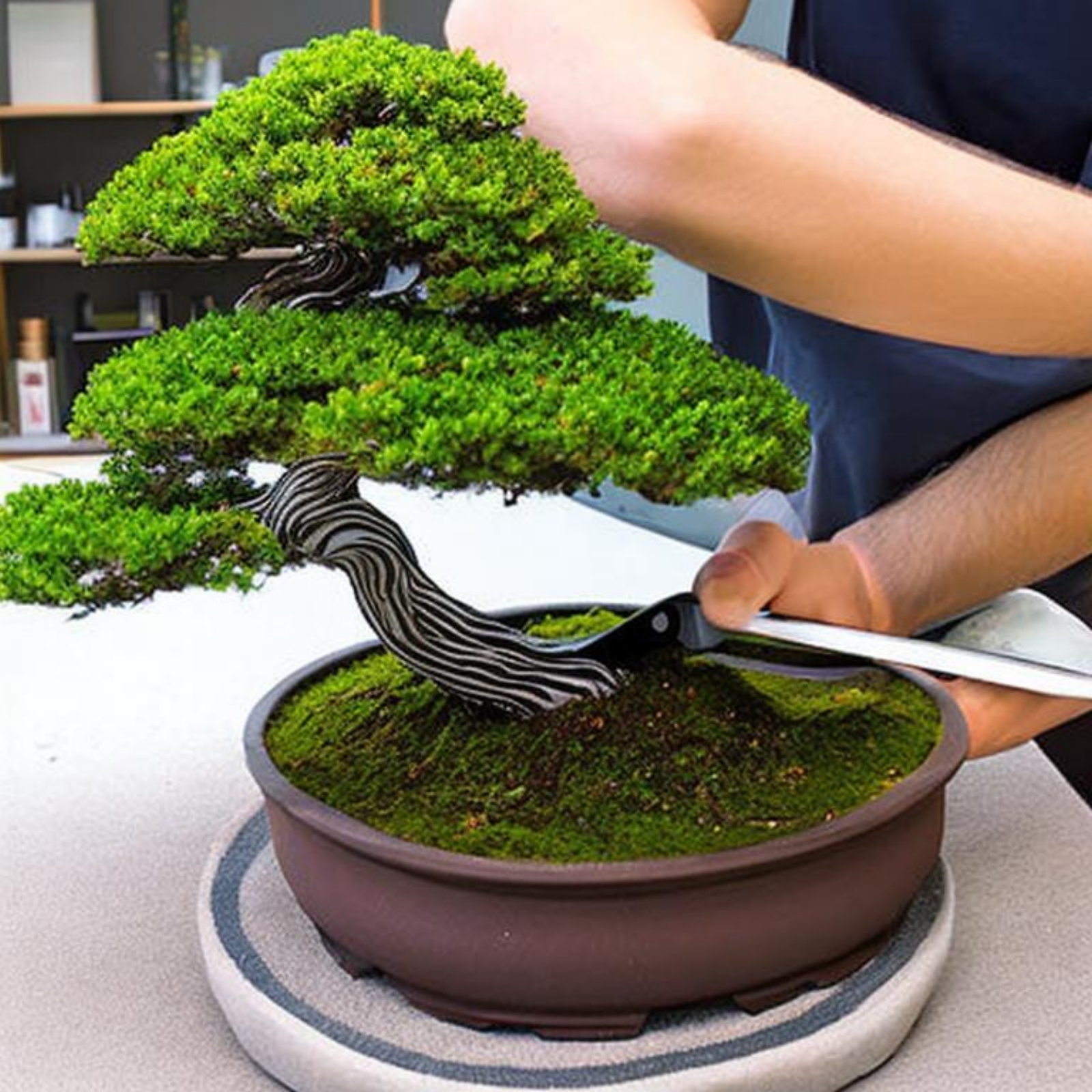} & \includegraphics[width=\linewidth, valign=m]{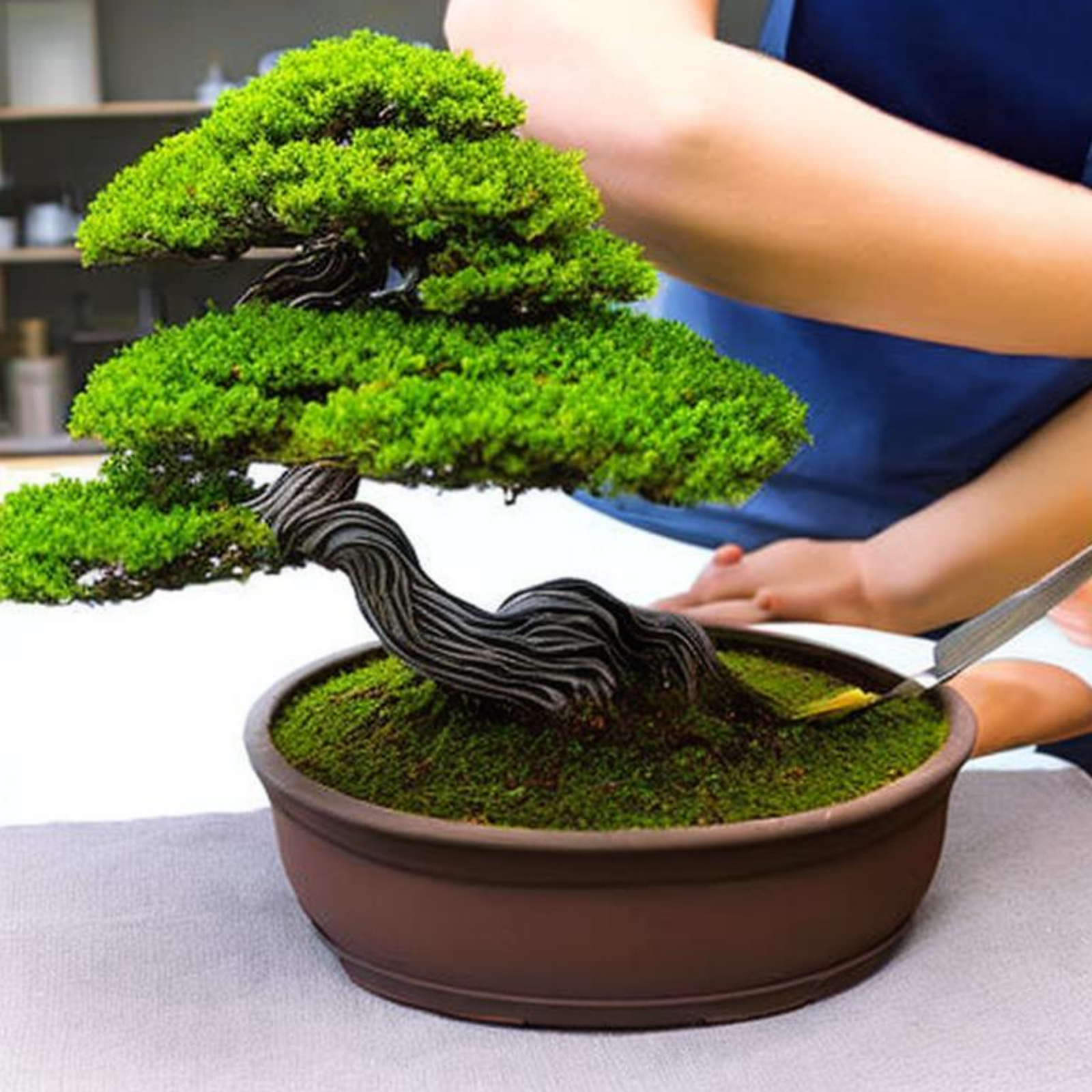} & \includegraphics[width=\linewidth, valign=m]{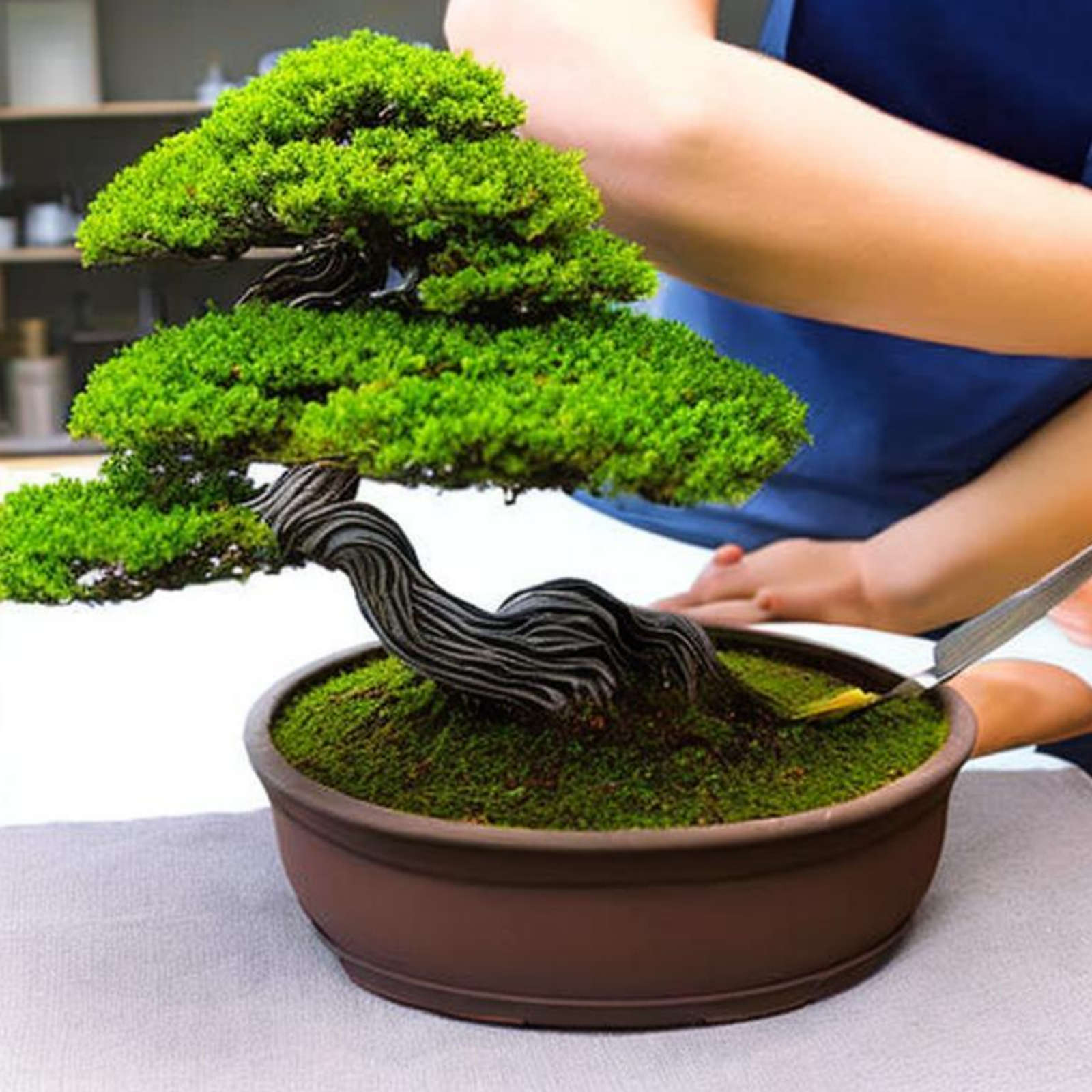} \\ \noalign{\vspace{10pt}}
    \rotatebox[origin=c]{90}{\scriptsize Baseline} & \includegraphics[width=\linewidth, valign=m]{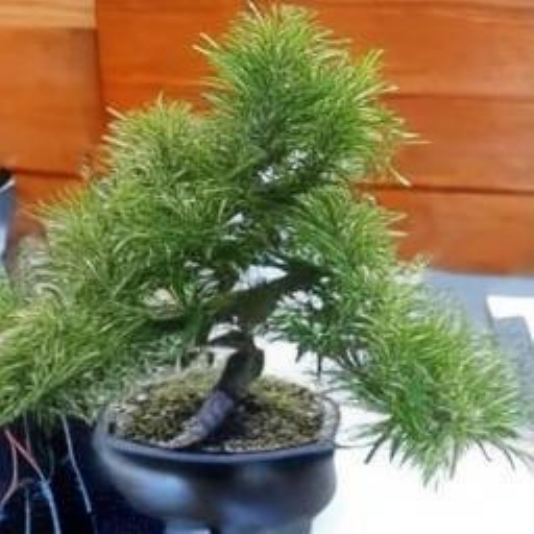} & \includegraphics[width=\linewidth, valign=m]{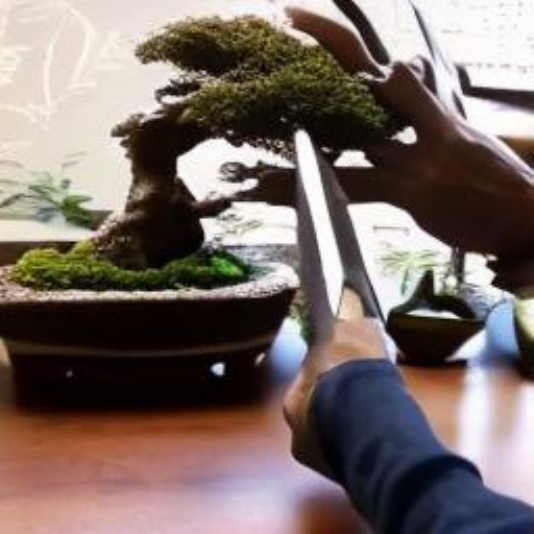} & \includegraphics[width=\linewidth, valign=m]{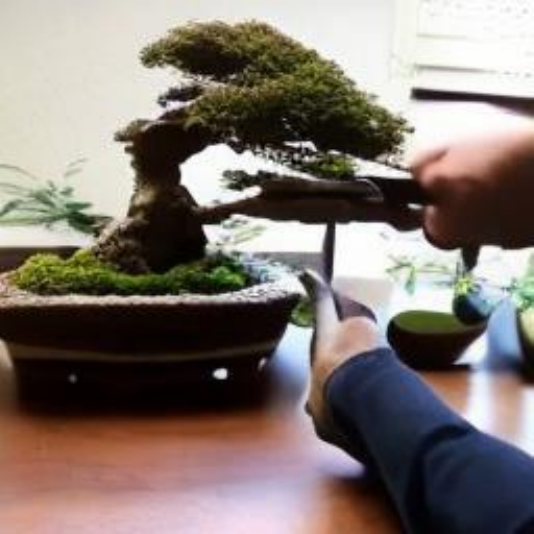} & \includegraphics[width=\linewidth, valign=m]{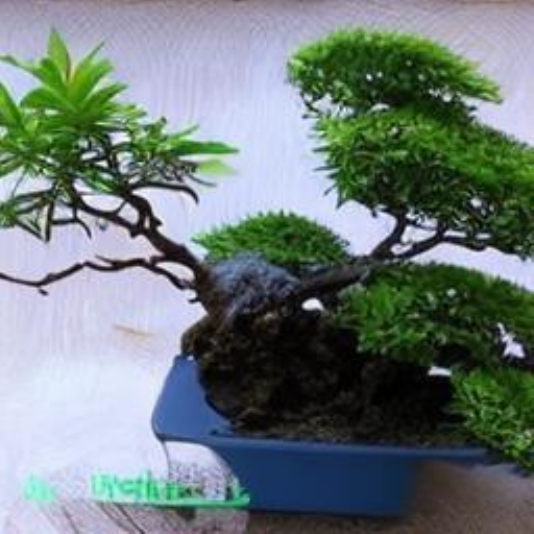} & \includegraphics[width=\linewidth, valign=m]{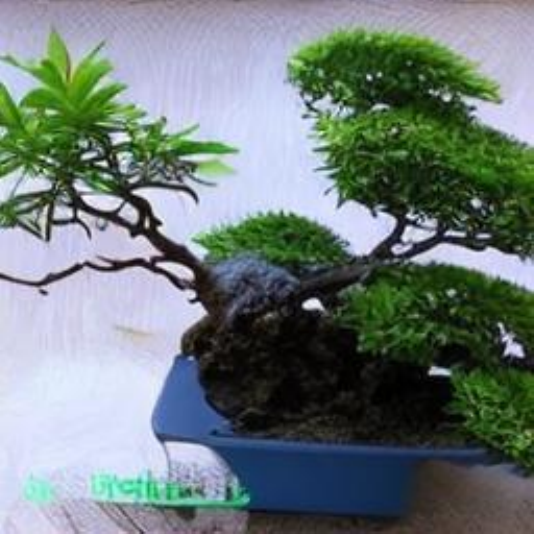} \\ \noalign{\vspace{10pt}}
    \rotatebox[origin=c]{90}{\scriptsize \textbf{InstructionCrafter}} & \includegraphics[width=\linewidth, valign=m]{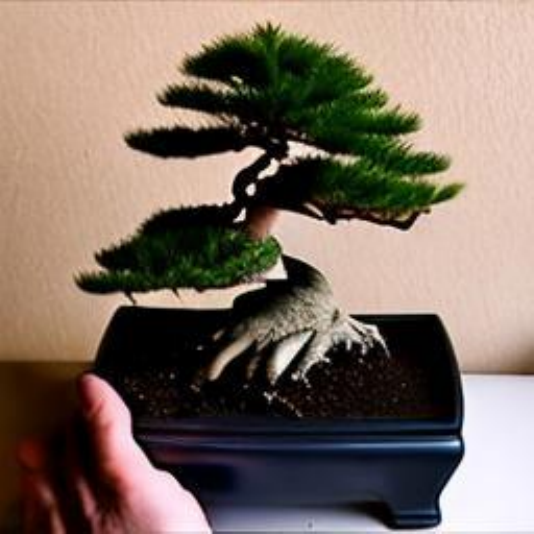} & \includegraphics[width=\linewidth, valign=m]{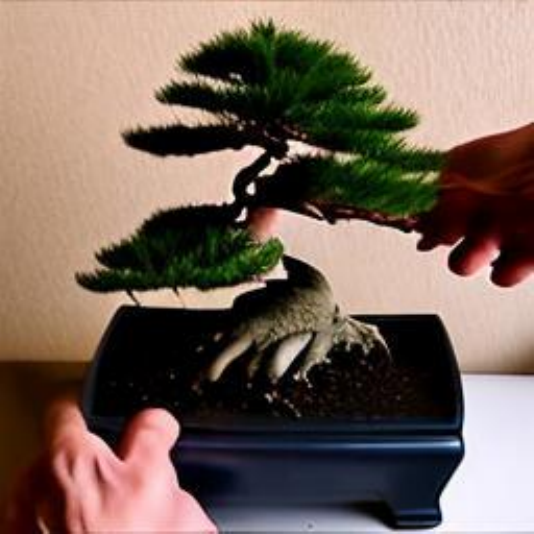} & \includegraphics[width=\linewidth, valign=m]{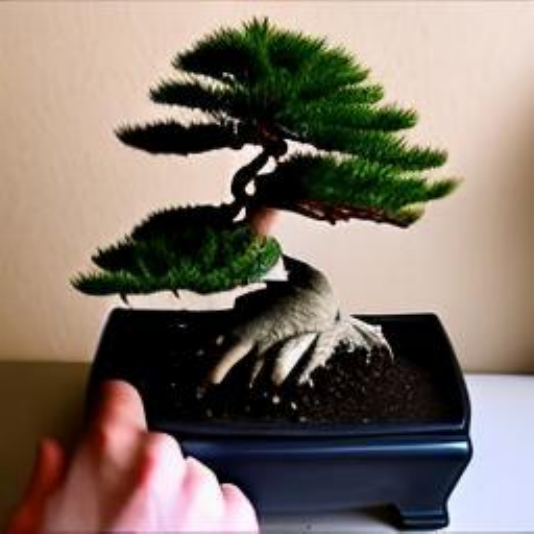} & \includegraphics[width=\linewidth, valign=m]{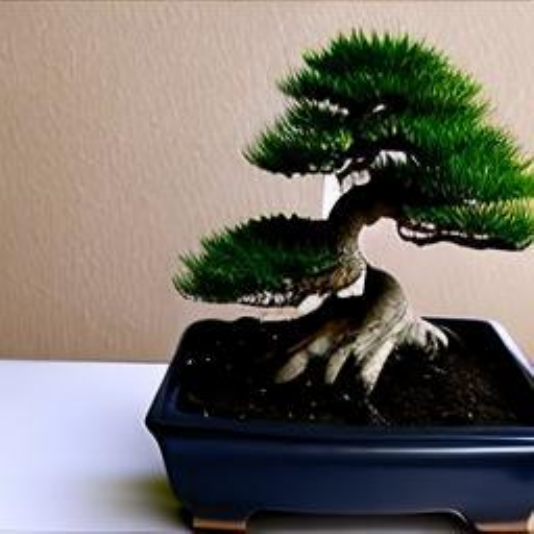} & \includegraphics[width=\linewidth, valign=m]{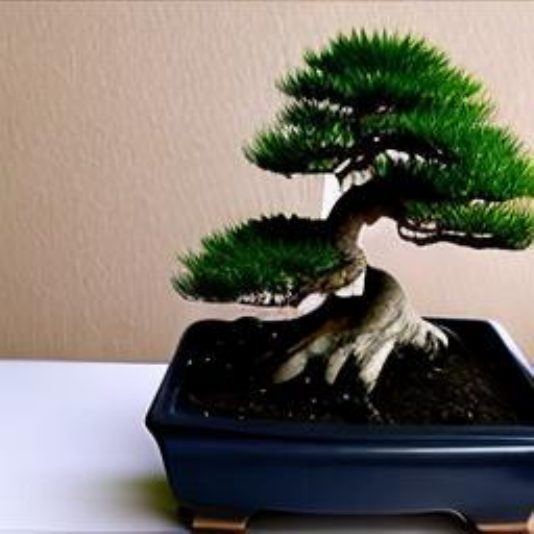} \\ \noalign{\vspace{10pt}}
  \end{tabular}
  \caption{\textbf{Failure cases where none of the methods could generate satisfactory images.}}
  \label{fig:fail1}
\end{figure*}

\begin{figure*}[t]
  \centering
  \setlength{\tabcolsep}{3pt}
  \renewcommand{\arraystretch}{1.2}
  \begin{tabular}{>{\centering\arraybackslash}m{1.5cm} m{0.100\linewidth} m{0.100\linewidth} m{0.100\linewidth} m{0.100\linewidth} m{0.100\linewidth} m{0.100\linewidth} m{0.100\linewidth} m{0.100\linewidth}}
     & \tiny\centering Build a basic bandpass filter using a capacitor, inductor, and resistor. & \tiny\centering Set up the oscilloscope's frequency response analysis application. & \tiny\centering Hook up the waveform generator to the input of the device. & \tiny\centering Probe channel one on the input of the device. & \tiny\centering Probe channel two on the output of the device. & \tiny\centering Configure the frequency sweep settings. & \tiny\centering Run the frequency response analysis test. & \tiny View the Bode plot and analyze the results. \\
    \rotatebox[origin=c]{90}{\scriptsize Stable Diffusion} & \includegraphics[width=\linewidth, valign=m]{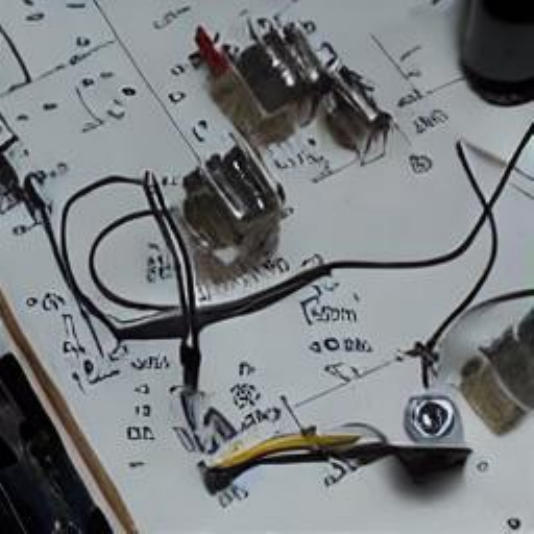} & \includegraphics[width=\linewidth, valign=m]{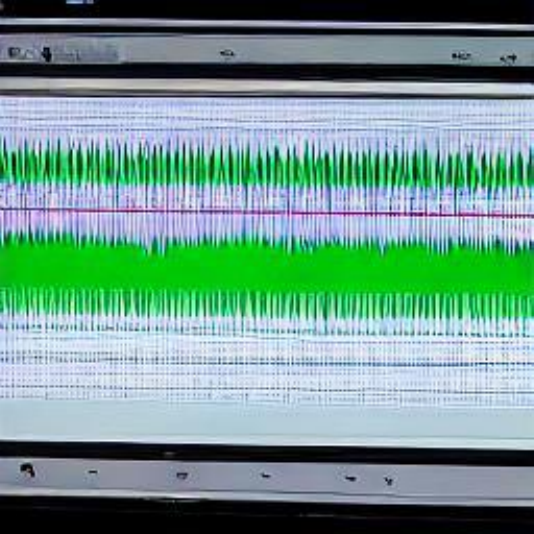} & \includegraphics[width=\linewidth, valign=m]{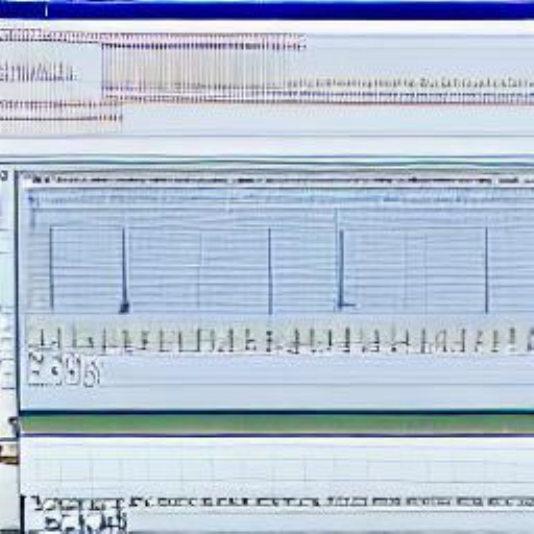} & \includegraphics[width=\linewidth, valign=m]{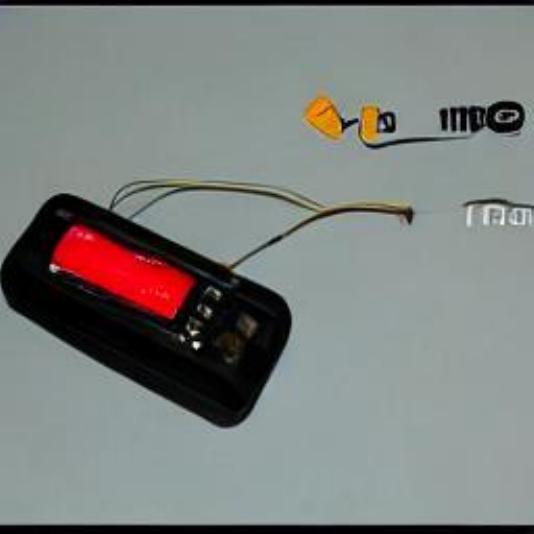} & \includegraphics[width=\linewidth, valign=m]{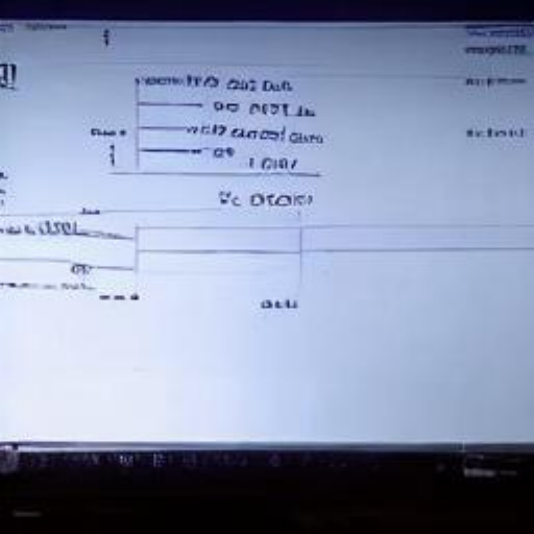} & \includegraphics[width=\linewidth, valign=m]{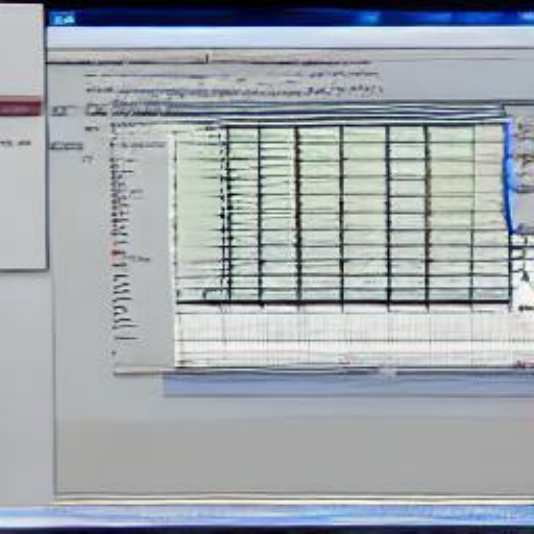} & \includegraphics[width=\linewidth, valign=m]{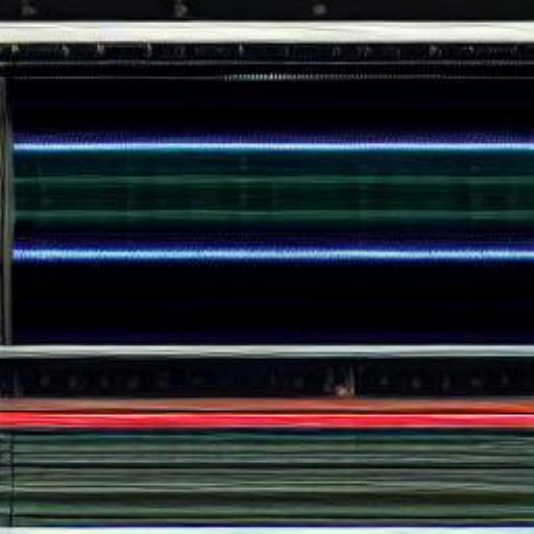} & \includegraphics[width=\linewidth, valign=m]{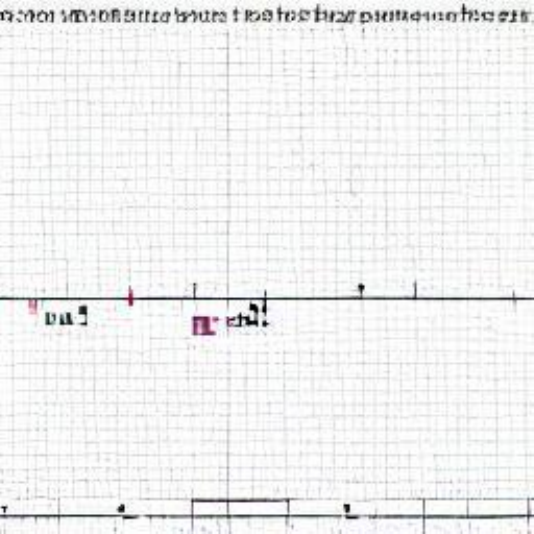} \\ \noalign{\vspace{10pt}}
    \rotatebox[origin=c]{90}{\scriptsize StackedDiffusion} & \includegraphics[width=\linewidth, valign=m]{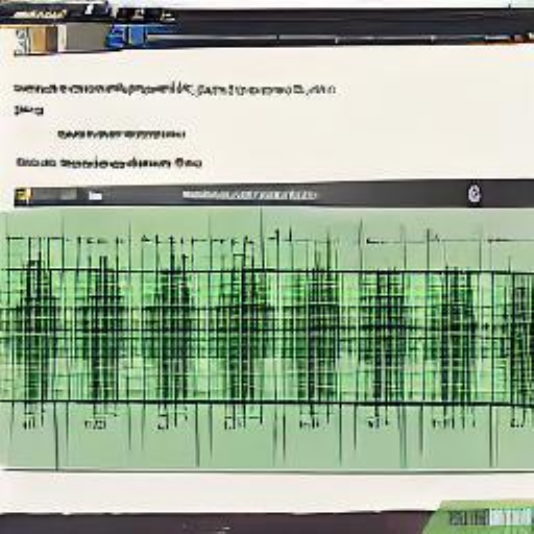} & \includegraphics[width=\linewidth, valign=m]{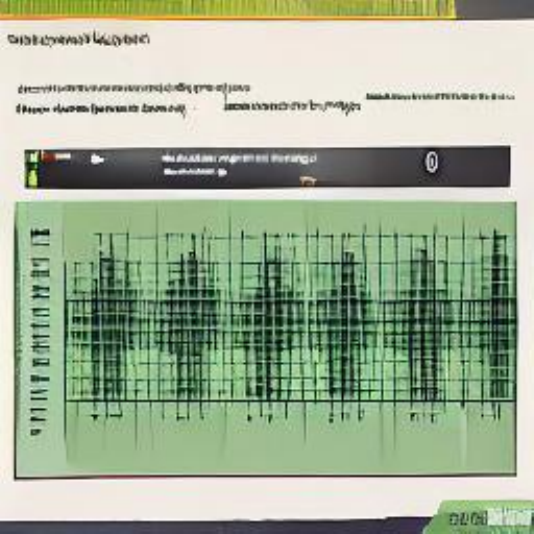} & \includegraphics[width=\linewidth, valign=m]{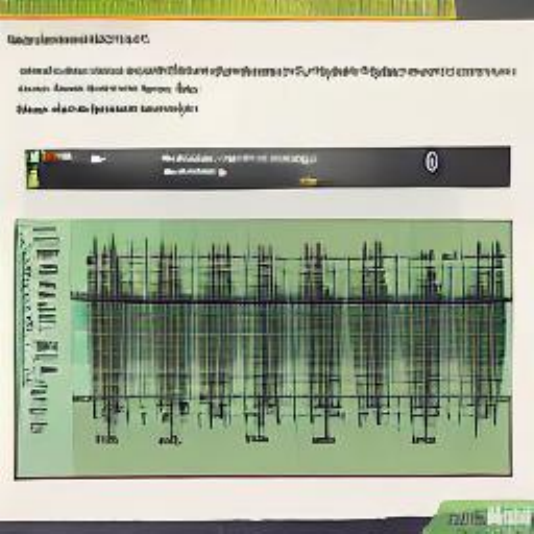} & \includegraphics[width=\linewidth, valign=m]{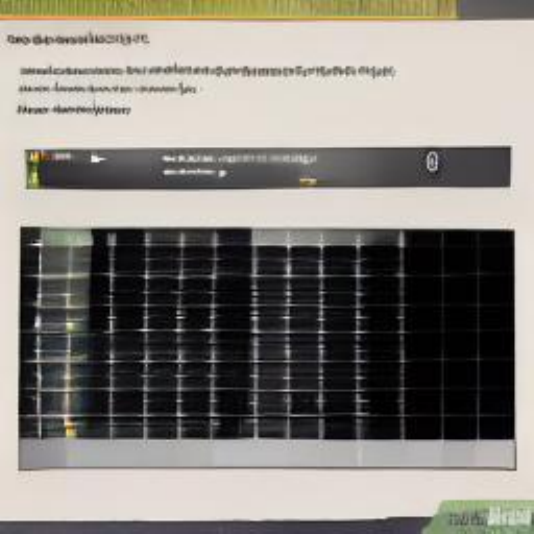} & \includegraphics[width=\linewidth, valign=m]{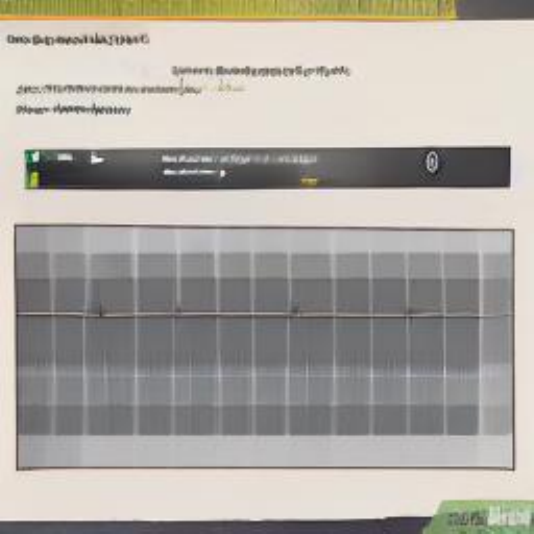} & \includegraphics[width=\linewidth, valign=m]{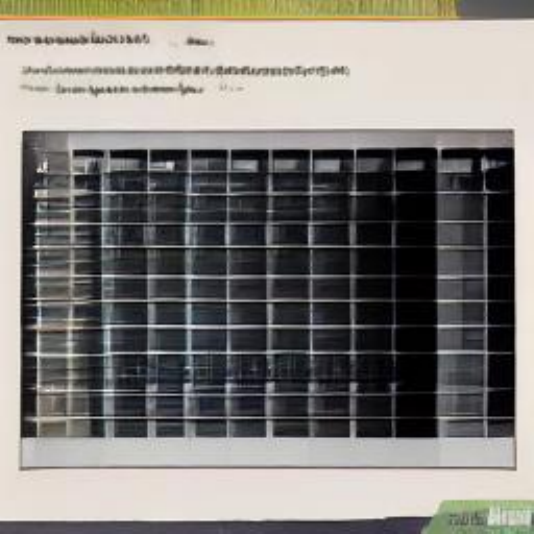} & \includegraphics[width=\linewidth, valign=m]{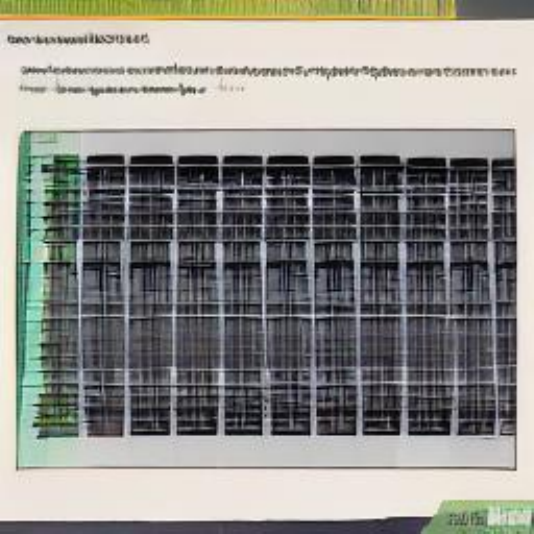} & \includegraphics[width=\linewidth, valign=m]{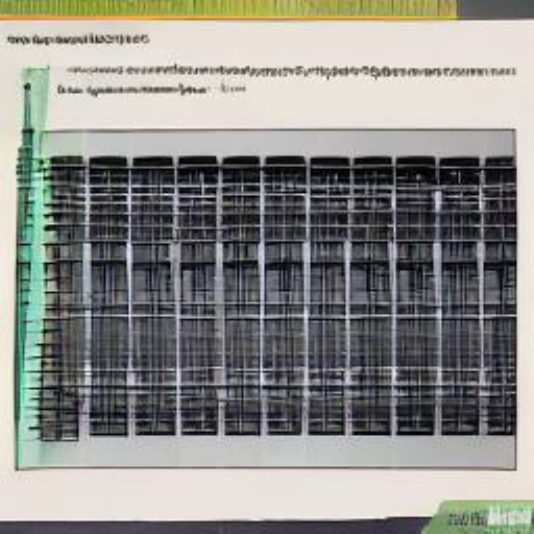} \\ \noalign{\vspace{10pt}}
    \rotatebox[origin=c]{90}{\scriptsize Bordalo et al.} & \includegraphics[width=\linewidth, valign=m]{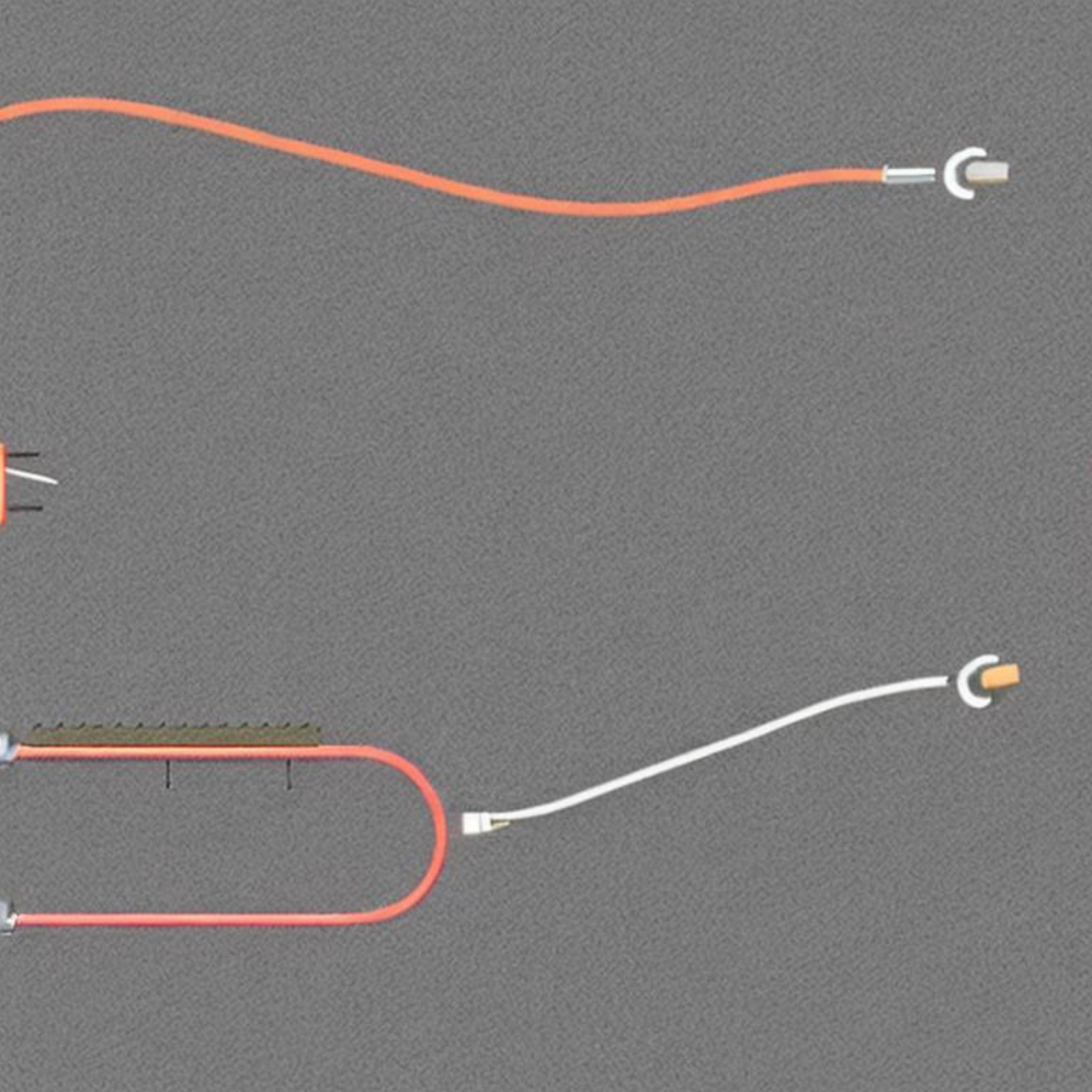} & \includegraphics[width=\linewidth, valign=m]{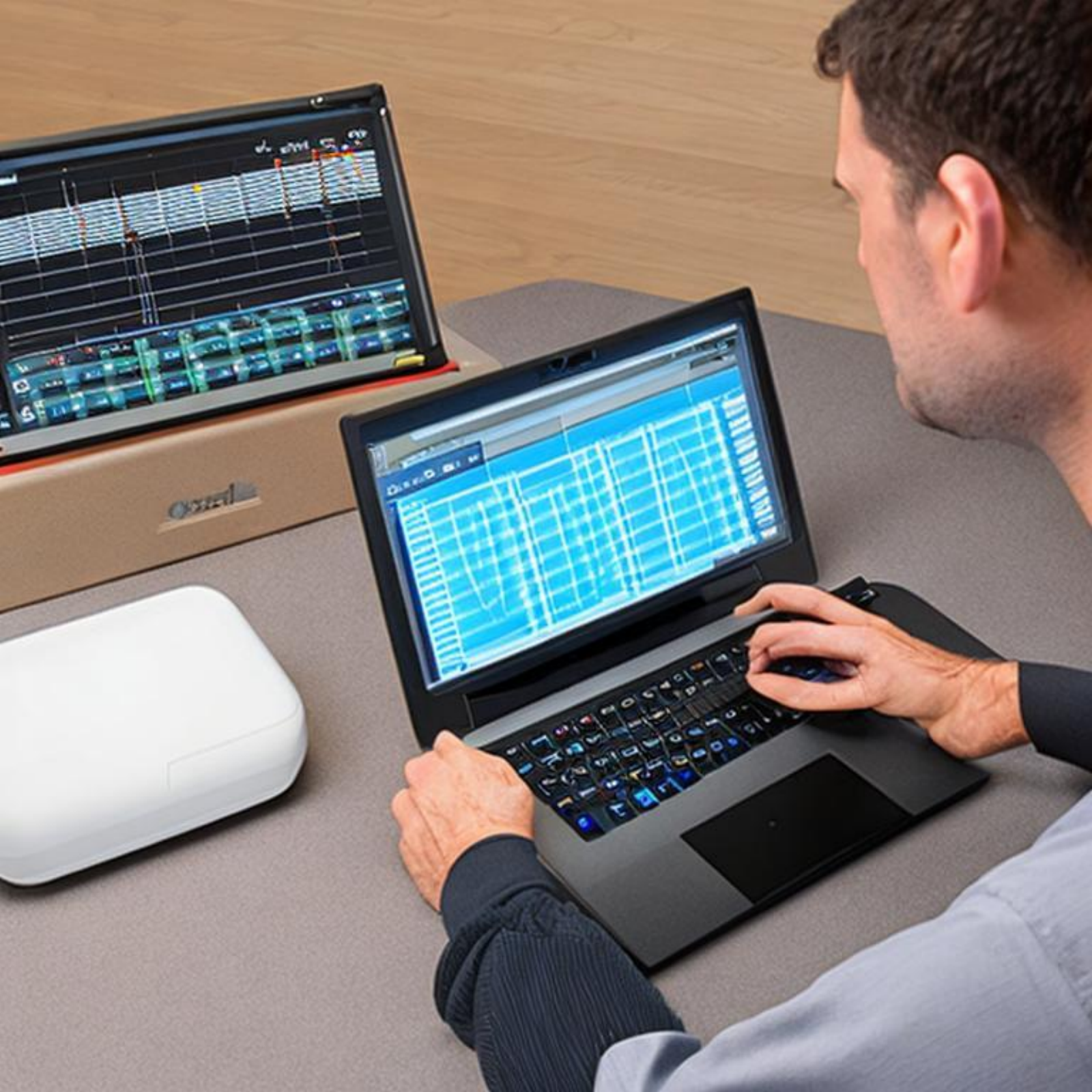} & \includegraphics[width=\linewidth, valign=m]{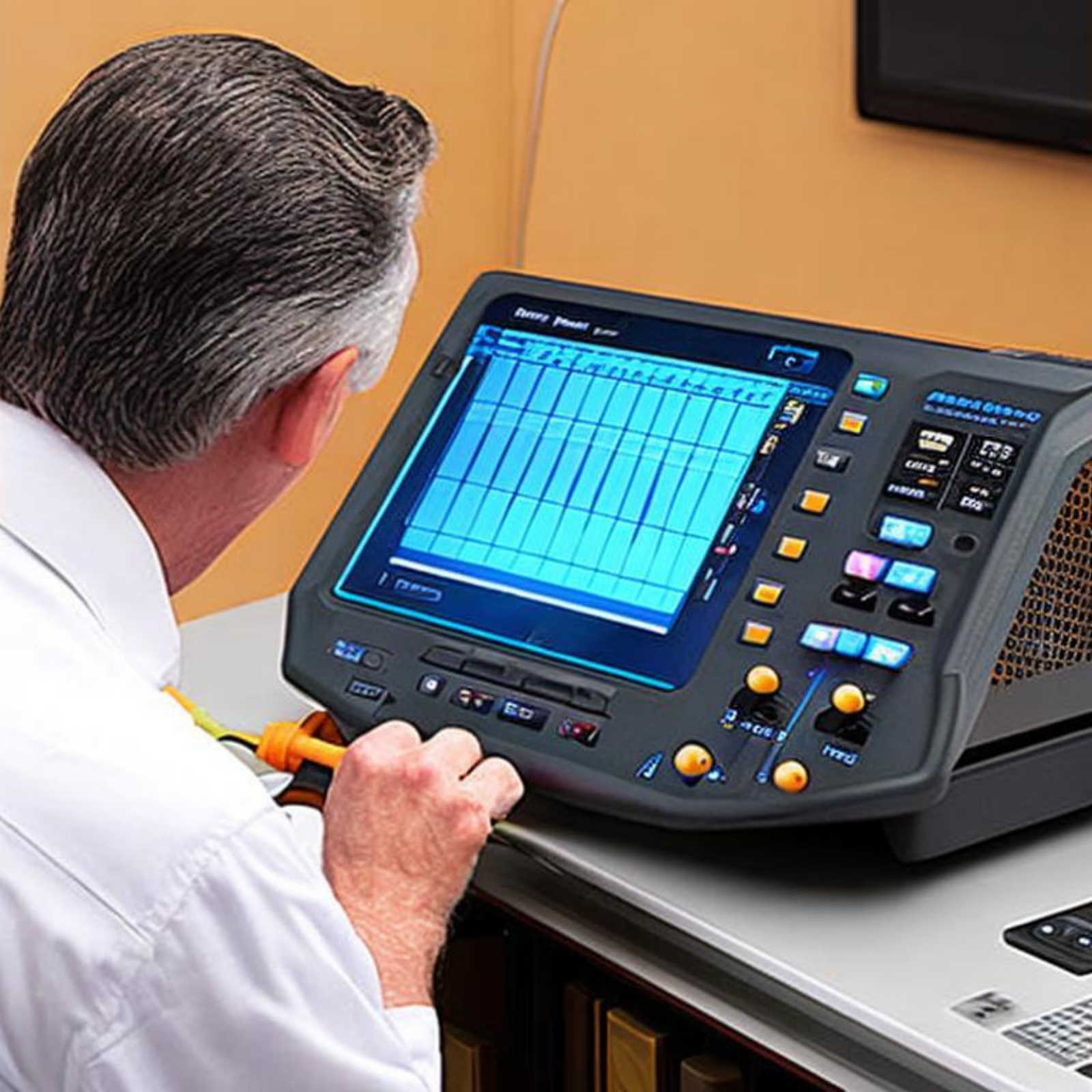} & \includegraphics[width=\linewidth, valign=m]{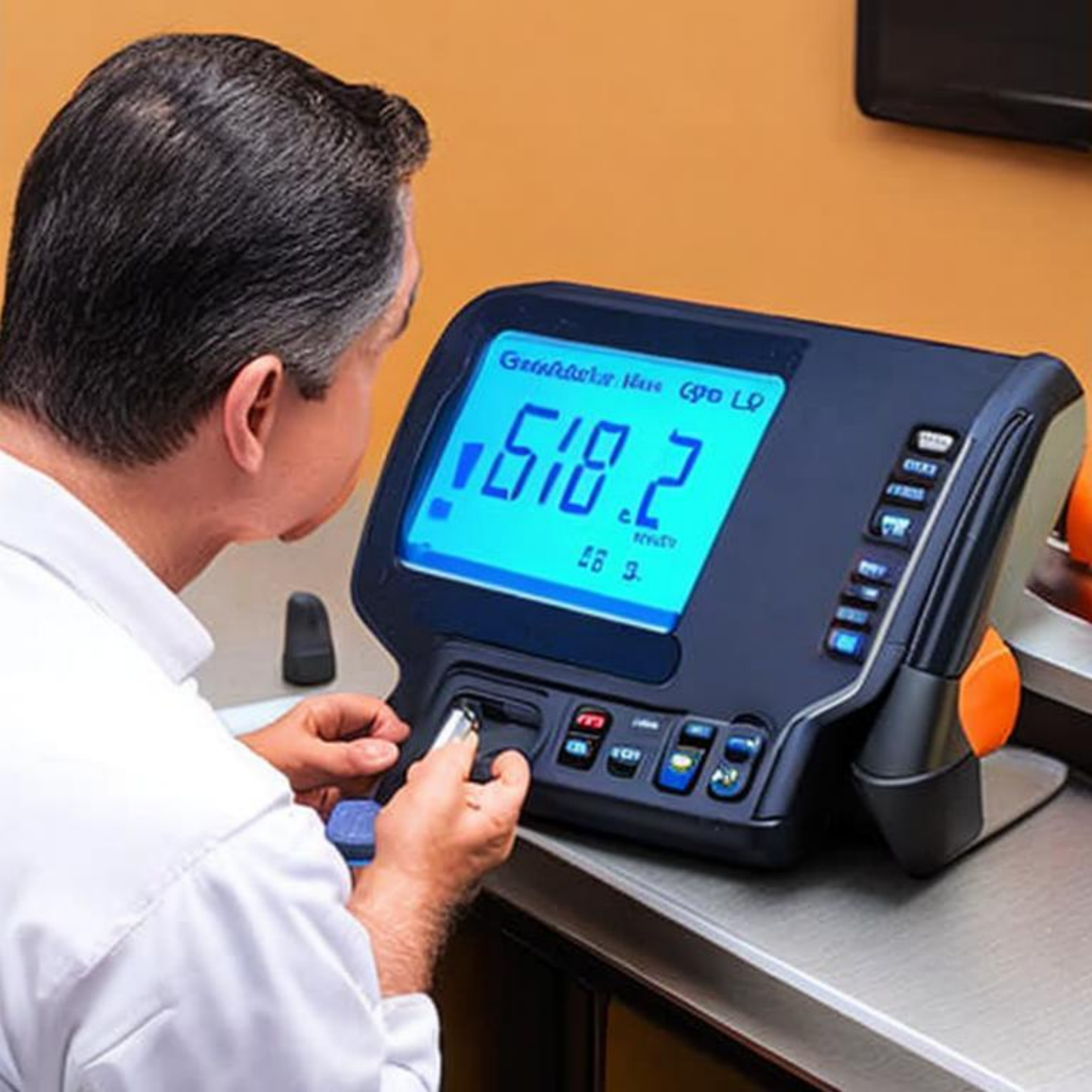} & \includegraphics[width=\linewidth, valign=m]{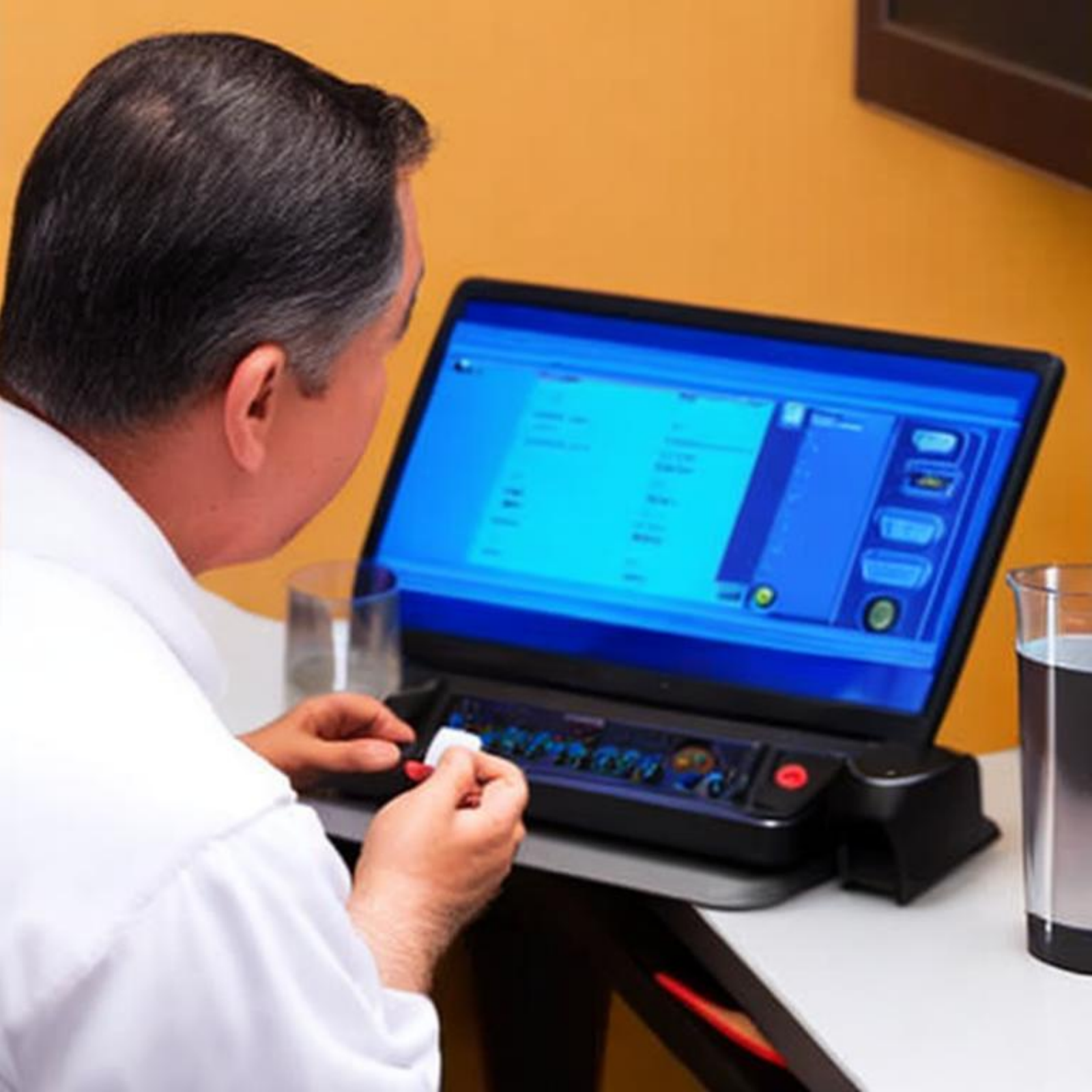} & \includegraphics[width=\linewidth, valign=m]{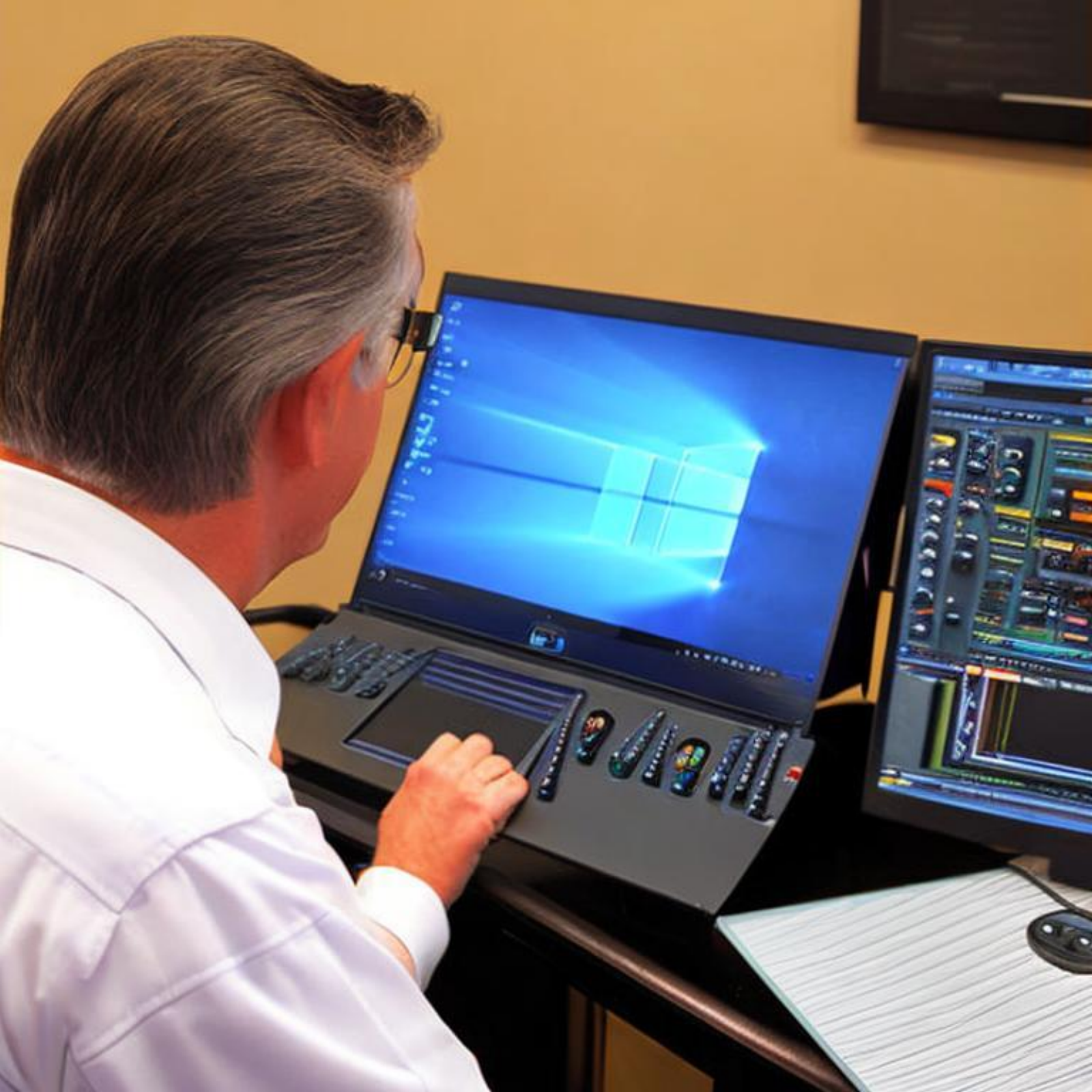} & \includegraphics[width=\linewidth, valign=m]{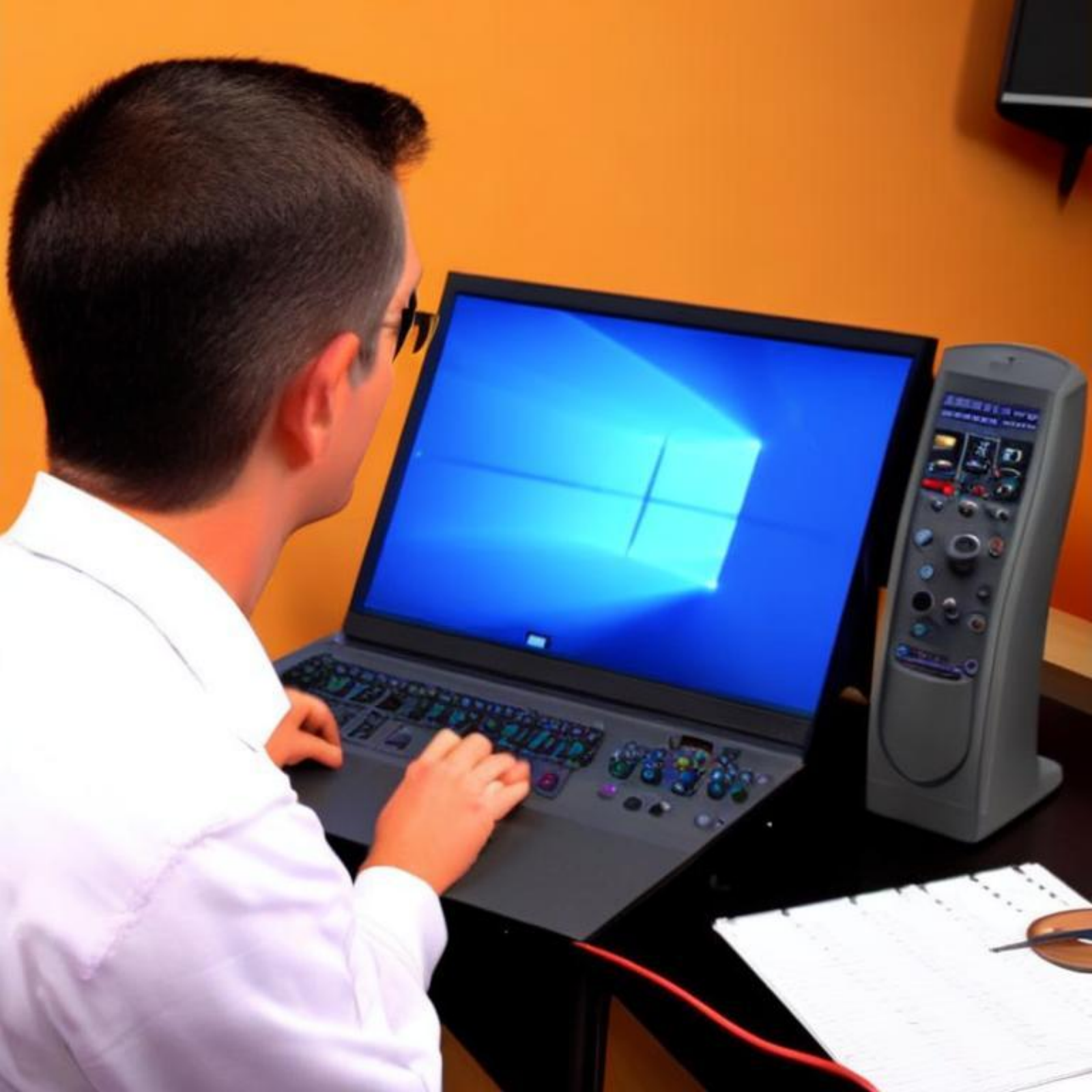} & \includegraphics[width=\linewidth, valign=m]{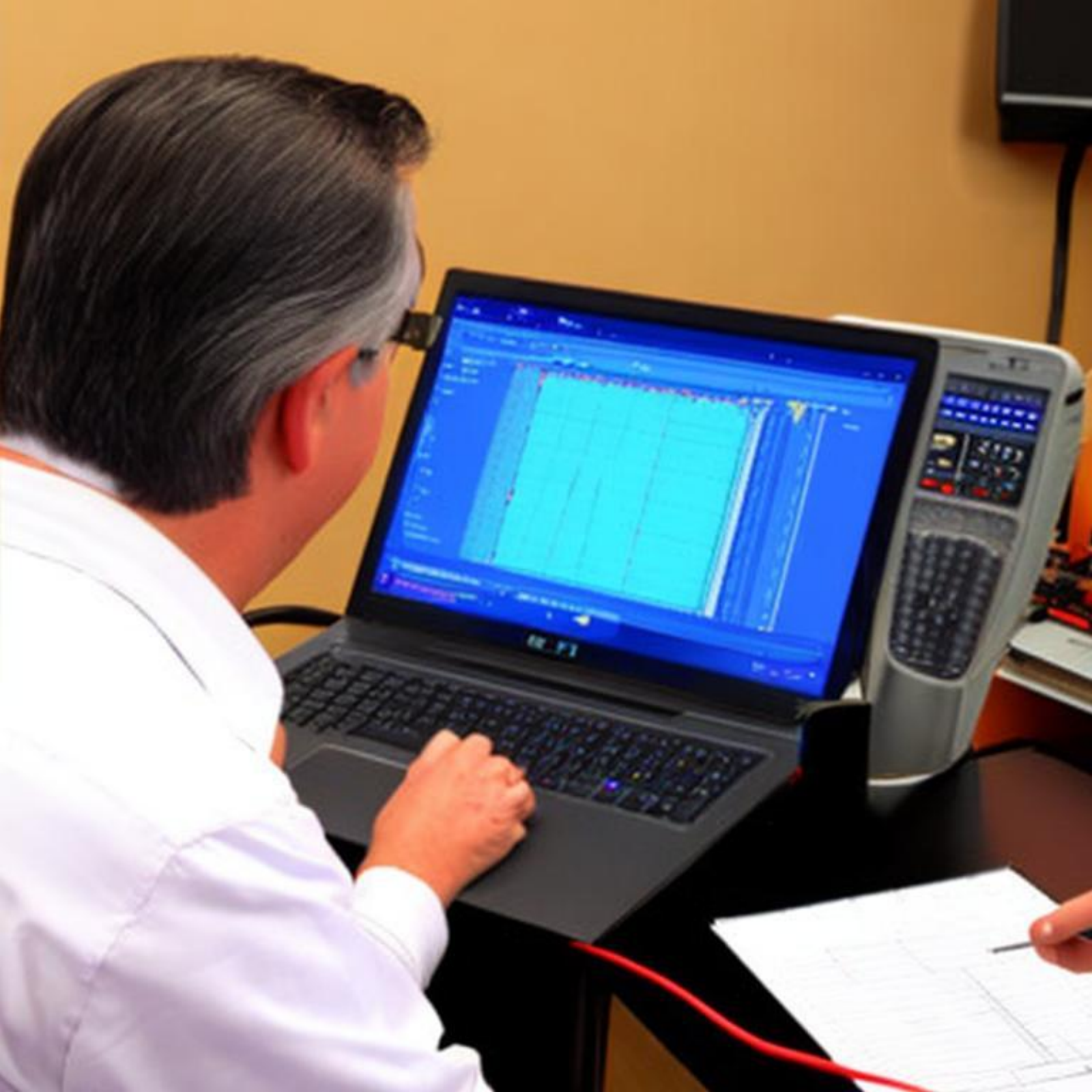} \\ \noalign{\vspace{10pt}}
    \rotatebox[origin=c]{90}{\scriptsize Baseline} & \includegraphics[width=\linewidth, valign=m]{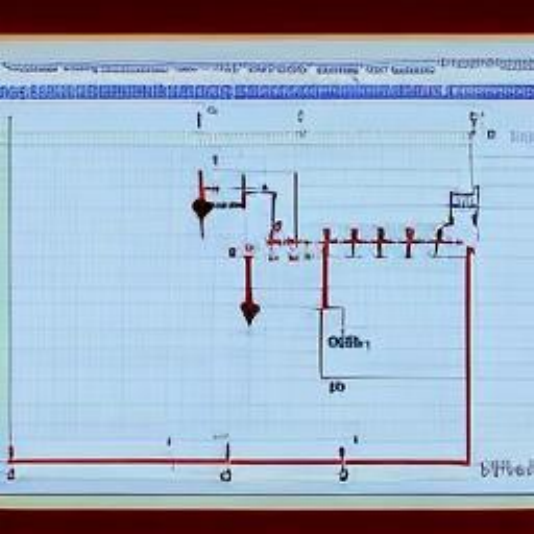} & \includegraphics[width=\linewidth, valign=m]{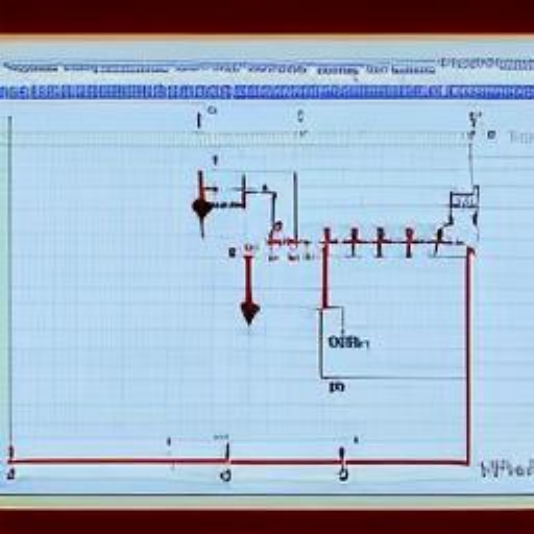} & \includegraphics[width=\linewidth, valign=m]{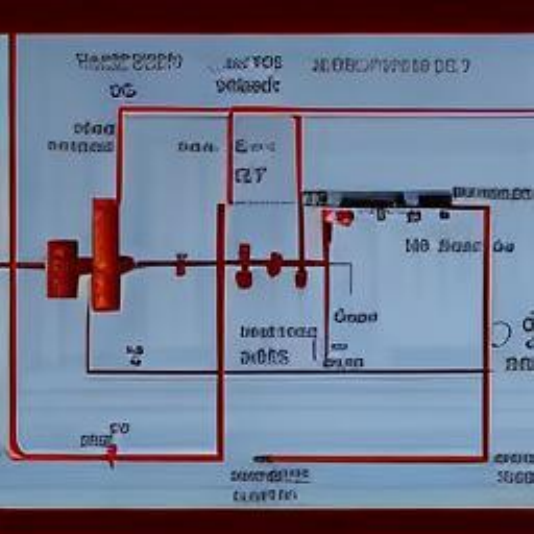} & \includegraphics[width=\linewidth, valign=m]{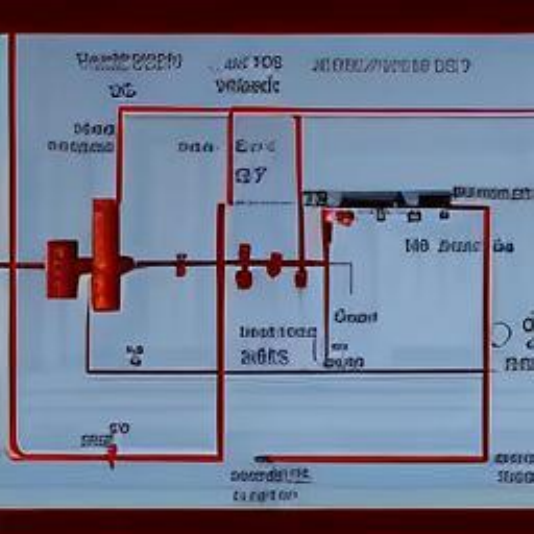} & \includegraphics[width=\linewidth, valign=m]{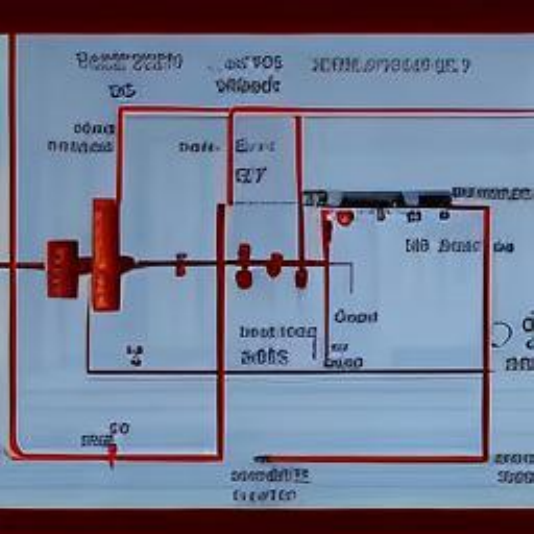} & \includegraphics[width=\linewidth, valign=m]{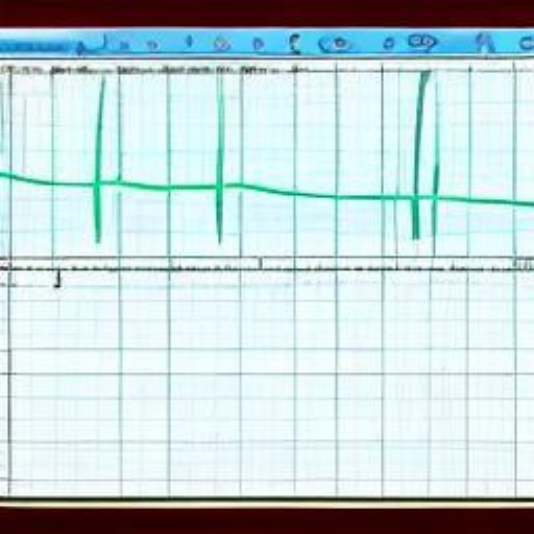} & \includegraphics[width=\linewidth, valign=m]{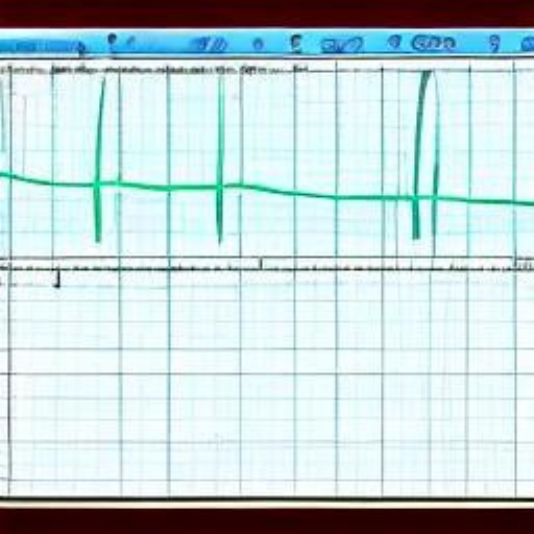} & \includegraphics[width=\linewidth, valign=m]{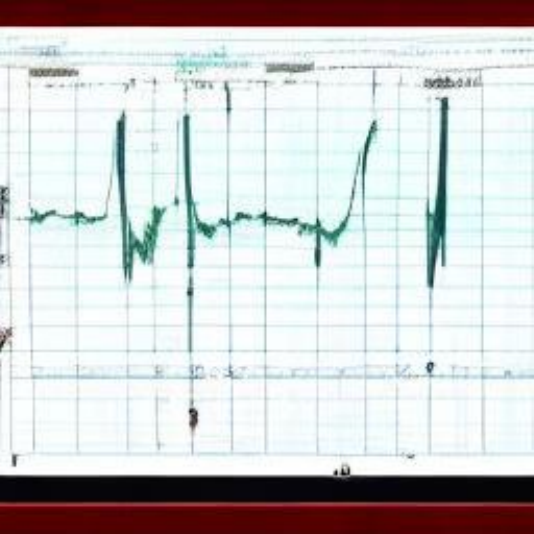} \\ \noalign{\vspace{10pt}}
    \rotatebox[origin=c]{90}{\scriptsize \textbf{InstructionCrafter}} & \includegraphics[width=\linewidth, valign=m]{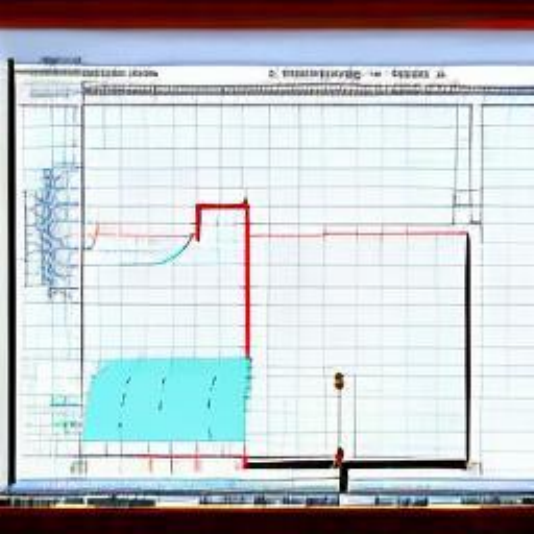} & \includegraphics[width=\linewidth, valign=m]{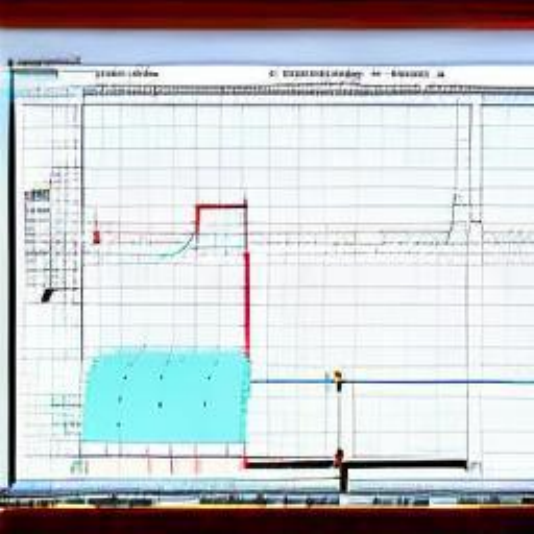} & \includegraphics[width=\linewidth, valign=m]{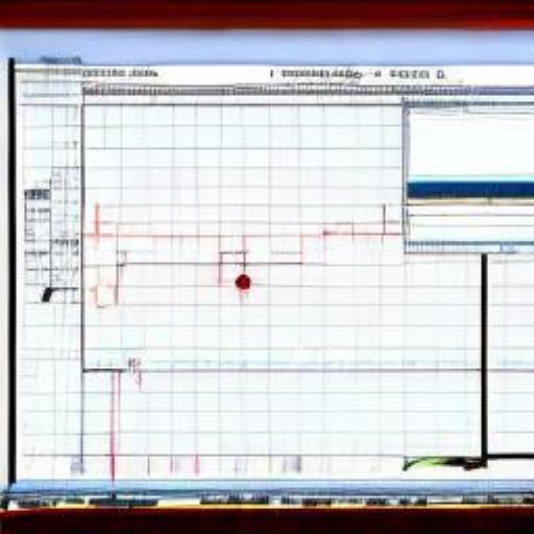} & \includegraphics[width=\linewidth, valign=m]{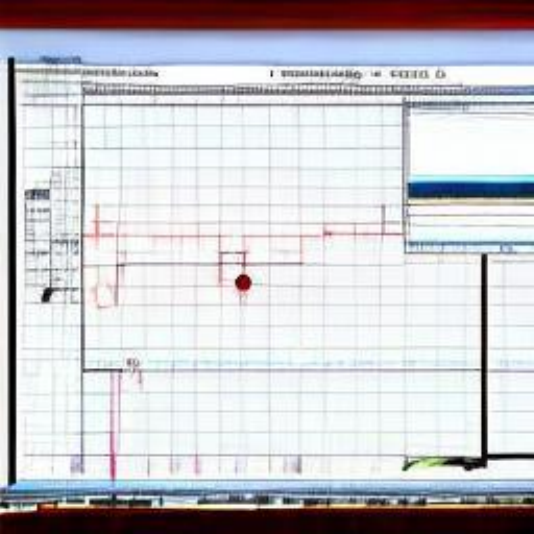} & \includegraphics[width=\linewidth, valign=m]{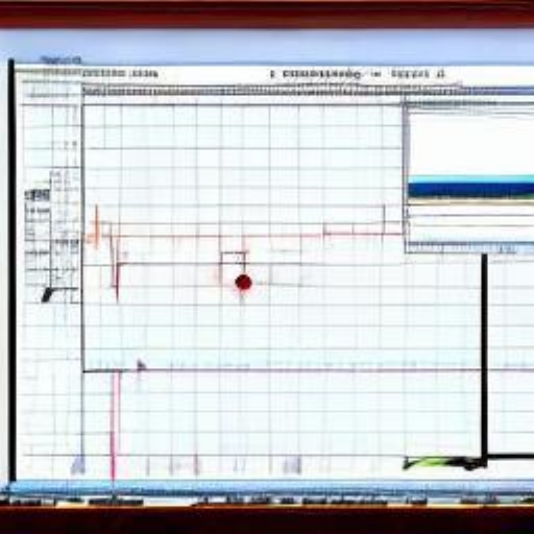} & \includegraphics[width=\linewidth, valign=m]{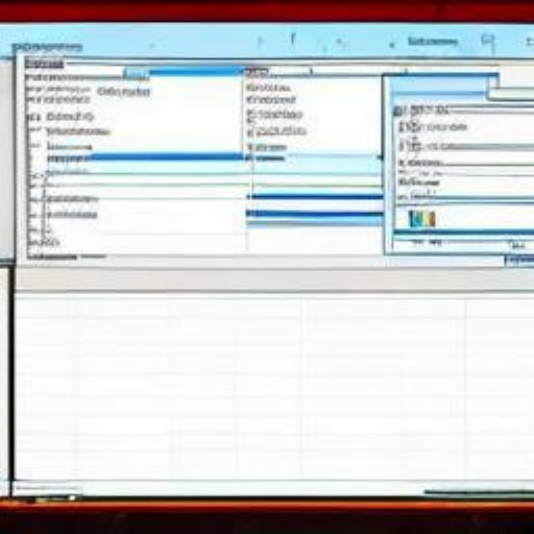} & \includegraphics[width=\linewidth, valign=m]{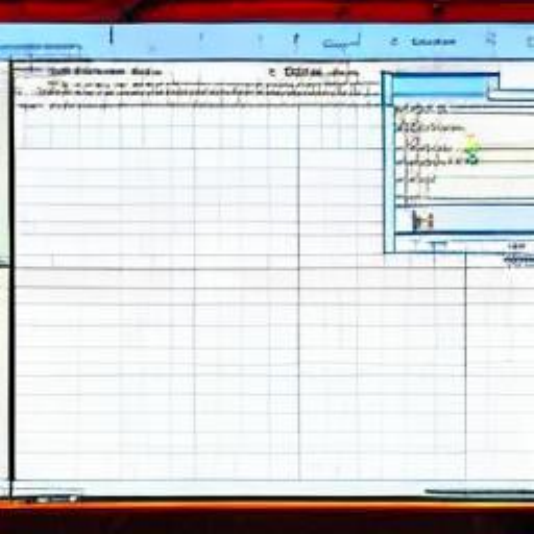} & \includegraphics[width=\linewidth, valign=m]{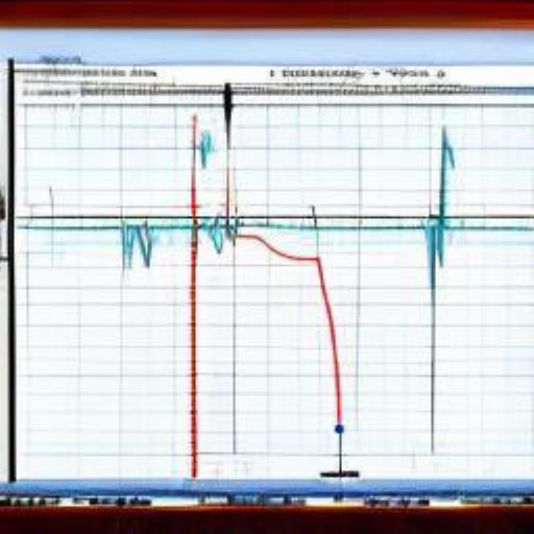} \\ \noalign{\vspace{10pt}}
  \end{tabular}
  \caption{\textbf{Failure cases where none of the methods could generate satisfactory images.}}
  \label{fig:fail2}
\end{figure*}

As discussed in the limitations of the main paper, we show several failure cases in~\cref{fig:fail1} and~\cref{fig:fail2}. Our model sometimes fails to understand abstract instructions such as ``maintain'', ``keep'' and ``learn'' which require high-level reasoning beyond visual understanding as shown in~\cref{fig:fail1}. Additionally, our model may struggle to generate highly complex scenes that require advanced expertise, such as frequency response testing in~\cref{fig:fail2}. We note that all compared methods also fail in these cases where understanding such instructions and generating corresponding images is challenging. These issues could be alleviated by leveraging stronger diffusion backbones and more diverse and high-quality training datasets in future work.

% Additional qualitative comparisons

% Additional qualitative examples

% Failure cases

{
    \clearpage
    \small
    \bibliographystyle{bmvc2k}
    \bibliography{egbib}
}